\documentclass{article} 
\usepackage{iclr2026_conference,times}

\usepackage{lineno}

\usepackage{pifont}
\usepackage{amssymb}
\usepackage{wrapfig}
\usepackage{graphicx}
\usepackage{printlen}
\usepackage{multirow}
\usepackage{color, colortbl}
\usepackage{makecell}
\usepackage{longtable}
\usepackage[export]{adjustbox}

\usepackage{ltcaption}

\newcommand{\blfootnote}[1]{%
  \begingroup
  \renewcommand\thefootnote{}\footnote{#1}%
  \addtocounter{footnote}{-1}%
  \endgroup
}

\usepackage{enumitem}
\newcommand\mybullet[1]{{%
    \fboxsep=0pt
    \fbox{\rule{0pt}{1em}%
\rule{.5em}{#1em}}}}

\usepackage{ifmtarg}
\makeatletter
\newcommand\bitem[1][]{%
  \@ifmtarg{#1}{
\item[$\bullet$]}
  {
\item[\mybullet{#1}]}}
\makeatother

\usepackage{arydshln}
\makeatletter
\def\adl@drawiv#1#2#3{%
  \hskip.5\tabcolsep
  \xleaders#3{#2.5\@tempdimb #1{1}#2.5\@tempdimb}%
  #2\z@ plus1fil minus1fil\relax
\hskip.5\tabcolsep}
\newcommand{\cdashlinelr}[1]{%
  \noalign{\vskip\aboverulesep
    \global\let\@dashdrawstore\adl@draw
  \global\let\adl@draw\adl@drawiv}
  \cdashline{#1}
  \noalign{\global\let\adl@draw\@dashdrawstore
\vskip\belowrulesep}}
\makeatother

\usepackage[utf8]{inputenc} 
\usepackage[T1]{fontenc}    
\usepackage{url}            
\usepackage{booktabs}       
\usepackage{amsfonts}       
\usepackage{nicefrac}       
\usepackage{microtype}      
\usepackage{xcolor}         
\usepackage{arydshln}

\usepackage{algorithm}
\usepackage{algpseudocode}
\usepackage{etoolbox}

\usepackage{graphicx}
\usepackage{subcaption}
\usepackage{tcolorbox}
\usepackage[misc]{ifsym} 

\DeclareCaptionSubType{algorithm}

\definecolor{lightyellow}{RGB}{255,255,204}

\usepackage{amsmath,amsfonts,bm}

\def\eqref#1{equation~\ref{#1}}

\def\1{\bm{1}}

\DeclareMathAlphabet{\mathsfit}{\encodingdefault}{\sfdefault}{m}{sl}
\SetMathAlphabet{\mathsfit}{bold}{\encodingdefault}{\sfdefault}{bx}{n}

\usepackage{xspace}         

\newcommand{\cmark}{\ding{51}}%
\newcommand{\xmark}{\ding{55}}%

\newcommand*\colourcheck[1]{%
  \expandafter\newcommand\csname #1check\endcsname{\textcolor{#1}{\cmark}}%
}
\colourcheck{blue}
\colourcheck{green}
\colourcheck{red}
\colourcheck{black}
\colourcheck{orange}

\definecolor{mygreen}{HTML}{009000}
\definecolor{myorange}{HTML}{FF6700}
\definecolor{mypurple}{HTML}{9900ff}

\renewcommand{\checkmark}{\cmark}
\renewcommand{\checkmark}{\orangecheck}

\newcommand{\ttt}[1]{\texttt{#1}}
\newcommand{\hl}[1]{\textcolor{orange}{#1}}

\newcommand{\model}{\textsc{Espire}\xspace}
\newcommand{\modelfull}{A Diagnostic Benchmark for \hl{E}mbodied \hl{Sp}at\hl{i}al \hl{Re}asoning of Vision-Language Models\xspace}

\usepackage{url}
\usepackage{fontawesome5}

\title{\model: \modelfull}

\author{Yanpeng Zhao$^\dagger\,\textsuperscript{\Letter}$\quad  Wentao Ding$^\dagger$\quad Hongtao Li$^\dagger$\quad Baoxiong Jia \quad Zilong Zheng$\, \textsuperscript{\Letter}$\\[0.2em]
  State Key Laboratory of General Artificial Intelligence, BIGAI
  \\[0.25em]
  \faGlobe~\url{https://spatigen.github.io/espire.io/}
  \hfill
  \faGithub~\url{https://github.com/spatigen/espire}
}

\iclrfinalcopy 
\begin{document}

\maketitle

\blfootnote{
  \begin{tabular}{@{}l@{}}
    Contact: \texttt{yannzhao.ed@gmail.com, \{dingwentao,lihongtao\}@bigai.ai}. \\
    $\dagger$: Core Contributor. \ \Letter: Corresponding Author.
  \end{tabular}
}

%

\begin{abstract}

  A recent trend in vision-language models (VLMs) has been to enhance their spatial cognition for embodied domains. Despite progress, existing evaluations have been limited both in paradigm and in coverage, hindering rapid, iterative model development. To address these limitations, we propose \model, a diagnostic benchmark for embodied spatial reasoning. \model offers a simulated world that physically grounds VLMs and evaluates them on spatial-reasoning-centric robotic tasks, thus narrowing the gap between evaluation and real-world deployment. To adapt VLMs to robotic tasks, we decompose each task into localization and execution, and frame both as generative problems, in stark contrast to predominant discriminative evaluations (e.g., via visual-question answering) that rely on distractors and discard execution. This decomposition further enables a fine-grained analysis beyond passive spatial reasoning toward reasoning to act. We systematically design \model both at the instruction level and at the environment level, ensuring broad coverage of spatial reasoning scenarios. We use \model to diagnose a range of frontier VLMs and provide in-depth analysis of their spatial reasoning behaviors.

\end{abstract}

%
\section{Introduction}

Spatial cognition goes beyond perception; it enables reasoning and interaction with the 3D physical world, forming the foundation for embodied agents. While pivotal, current machine learning models---and in particular, vision-language models (VLMs)---still lag behind humans in this capacity~\citep{VSR,WhatsUp,Blink}, limiting applications in embodied domains such as robotic navigation and manipulation~\citep{VLMaps,VoxPoser,ReKep}. To bridge the gap, extensive efforts have been devoted to enhancing the spatial intelligence of VLMs~\citep{SpatialRGPT,SoFar,Open3DVQA,SpatialVLM, RoboSpatial,RoboRefer,RoboPoint}.

Despite the remarkable progress, the evaluation of spatially intelligent VLMs remains limited. First, most existing benchmarks are static, adopting multiple-choice visual-question answering (VQA), though this facilitates automatic evaluation, the reliance on distractors renders them prone to biases. Moreover, VQA departs from practical scenarios, where VLM agents must \emph{proactively} act upon given instructions in 3D rather than \emph{passively} selecting an answer from a predefined set. Though more reliable real-world evaluations have been explored, the dependence on specific hardware and handcrafted tasks hinders their scalability and reproducibility~\citep{RoboPoint,RoboSpatial}.

Recently, some have eschewed discriminative VQA and proposed \emph{pointing}, a generative evaluation methodology that requires models to locate the target object/space by generating points in 2D pixel space~\citep{RoboPoint,RoboRefer}, but the execution phase that typically follows localization in robotics tasks has been overlooked or overly simplified. Others have attempted to address execution while circumventing the limitations of real-world evaluation using simulated environments~\citep{Libero,Simpler,SoFar,EMBench}.
Yet, both directions lack a systematic design of evaluation tasks that supports detailed analysis of spatial reasoning across different aspects (e.g., relationships and distances) and granularities (e.g., relative vs. precise distance).

\begin{table*}[t!]
\centering
\caption{
Comparisons of spatial-reasoning benchmarks.
`Text Gen.' and `Point Gen.' indicate that models produce answers in natural language and 2D points, respectively. `Fully Gen.' denotes that models generate positions and rotations in 3D. `Tool-Free' means no external tools are used, thus assessing the \emph{intrinsic} spatial reasoning of VLMs.}
\label{tab:bench-comparison}
\resizebox{\linewidth}{!}{
\begin{tabular}{@{}l@{\hspace{0.0cm}}m{0\linewidth}cccccccc
@{\hspace{0.0cm}}m{0\linewidth}
@{\hspace{0.0cm}}m{0\linewidth}
@{\hspace{0.0cm}}m{0\linewidth}@{}}
\toprule 
\multirow{2}{*}{Benchmark} & \multirow{2}{*}{} & \multirow{2}{*}{\makecell[c]{Localization \\ \& Execution}} & \multirow{2}{*}{} & \multicolumn{2}{c}{Evaluation} &\multirow{2}{*}{Systematicity} & \multirow{2}{*}{\makecell[c]{Physically-\\Grounded}} & \multirow{2}{*}{Diagnostic} & \multirow{2}{*}{\makecell[c]{Clutter\\Level}} \\ 
\cmidrule{5-6}
&&&& Paradigm & Tool-Free &&& && \\
\midrule
\multicolumn{12}{c}{\cellcolor[HTML]{EFEFEF}{\textit{image- and video-based}}} \\
\midrule
Blink~\citep{Blink}           &  & \xmark &  & VQA & \checkmark & \xmark & \xmark & \xmark & \emph{high} &  &                           &                                 \\
CV-Bench~\citep{CV-Bench}     &  & \xmark &  & VQA & \checkmark & \xmark & \xmark &  \xmark & \emph{high} &  &                           &                                 \\
VSI-Bench~\citep{VSI-Bench}   &  & \xmark &  & VQA & \checkmark & \xmark & \xmark &  \xmark & \emph{high} &  &                           &                                 \\
\textsc{Where2Place}~\citep{RoboPoint} &  & \xmark &  & {Point} Gen. & \checkmark & \xmark & \xmark & \xmark & \emph{high} &  &                           &                                 \\
SpatialVQA~\citep{SpatialVLM} &  & \xmark &  & VQA, {Text} Gen. & \checkmark & \xmark & \xmark & \xmark & \emph{high} &  &                           &                                 \\
SpatialRGPT-Bench~\citep{SpatialRGPT} &  & \xmark &  & {Text} Gen. & \xmark & \xmark & \xmark & \xmark & \emph{high} &  &                           &                                 \\
RoboSpatial-Home~\citep{RoboSpatial}  &  & \xmark &  & VQA, {Point} Gen. & \checkmark & \xmark & \xmark & \xmark & \emph{high} &  &                           &                                 \\
Point-Bench~\citep{PointArena}  &  & \xmark &  & {Point} Gen. & \checkmark & \xmark & \xmark & \xmark & \emph{high} &  &                           &                                 \\
\midrule
\multicolumn{12}{c}{\cellcolor[HTML]{EFEFEF}{\textit{simulation-based}}} \\
\midrule
Open6DOR~\citep{Open6DOR}     &  & \checkmark &  & VQA & \xmark & \xmark & \checkmark & \xmark & \emph{low} &  &                           &                                 \\
EB-Manipulation~\citep{EMBench}       &  & \checkmark  &  & {Fully} Gen. & \xmark &\xmark & \checkmark & \xmark & \emph{low} &  &                           &                                 \\ 
\cmidrule{1-12}
\model (\emph{ours})                 &  & \checkmark &  & \textbf{\emph{Fully Gen.}} & \checkmark & \checkmark & \checkmark & \checkmark & \textbf{\emph{high}} &  &                           &                                 \\ 
\bottomrule 
\end{tabular}
} 
\vspace{-2em}
\end{table*}

To address these limitations, we propose \model, a simulation-based benchmark for embodied spatial reasoning with physically-grounded VLMs. Since VLMs are inherently not trained to act, to adapt them for robotics tasks, we decompose each task into localization (which identifies manipulable targets) and execution (which performs the corresponding actions), and frame them as goal position and goal pose generation, respectively. This fully generative, unified evaluation paradigm extends passive spatial reasoning toward acting upon understanding, thus reducing the gap between evaluation and real-world deployment.

To serve our diagnostic purpose, we propose a systematic task design that enables assessment and analysis of the native spatial reasoning of VLMs across varying spatial aspects and granularities. We follow a hierarchical design philosophy, ensuring that the evaluation is spatial-centric and has a broad coverage. Specifically, we first identify three primary factors that characterize spatial reasoning: (1) spatial aspects, including attributes, relationships, distances, and orientations, (2) reference objects, including oriented and non-oriented, and (3) reference frames, including relative, intrinsic, and absolute. A particular configuration of these factors defines a context for spatial reasoning. For example, `\emph{place the book behind the picture frame}' requires reasoning about `positional relationship (behind)' relative to an `oriented reference (picture frame)' using the `intrinsic frame (attached to the picture frame)'. Within a given context, we curate tasks to examine reasoning across different granularities, e.g., fine-grained orientations in `\emph{grab a book to the \underline{2 o'clock} of the picture frame}' and precise distances in `\emph{grab a book within \underline{1.2 meters} of you}.' To the best of our knowledge, this systematic design supports the most comprehensive, fine-grained analysis that existing benchmarks lack.\looseness=-1

We build \model on Isaac Sim~\citep{IsaacSim} that provides realistic physics simulation, and incorporate necessary measures to reduce \emph{sim-to-real} gaps. \model offers a total of 148 spatial-reasoning types for localization and covers typical \emph{pick} and \emph{place} actions, enabling a focus on VLM-oriented, embodied native spatial reasoning while maintaining sufficient challenges in tool-free execution. Combined with randomly sampled environments of varied clutter degrees, this provides broad coverage of spatial-centric \emph{reasoning} and \emph{acting}. To support scalable task generation, we represent task instructions in functional programs that can be executed on 3D scene graph representations of environment states and yield ground-truth targets.

We use \model to evaluate a diverse suite of VLMs, spanning proprietary, open-access, unified, and spatially-enhanced models. We find that VLMs perform much better in localization than in execution, indicating good passive spatial understanding but limited capacity for acting-oriented spatial reasoning. Among all spatial aspects, orientation reasoning poses the greatest challenge in both stages, suggesting a critical deficiency in grounding 3D rotational geometry. Overall, these findings highlight promising avenues for advancing the spatial cognition of VLMs.
\emph{We emphasize that \model is not intended to replace real-world evaluation, but to complement it with a scalable, reproducible alternative that facilitates rapid, iterative model improvement.}

In summary, our contributions are the following:
\begin{itemize}[leftmargin=*]
  \item \model, a diagnostic benchmark for embodied spatial reasoning of VLMs in physically-grounded photorealistic environments.
  \item A generative evaluation paradigm that unifies 3D localization and execution, bridging the gap between passive spatial understanding and acting-oriented spatial reasoning. 
  \item A systematic robotic task design that enables fine-grained diagnosis across diverse spatial reasoning contexts and granularities. 
  \item Experiments and analysis that quantify key bottlenecks in 3D rotational geometry and suggest future directions for enhancement.
\end{itemize}

%
\section{Related Work}

\paragraph{Spatial reasoning with vision-language models.}
Extensive research has sought to boost the spatial intelligence of VLMs.
Some rely on enhanced prompting mechanisms for improved 3D spatial reasoning~\citep{SpatialPIN,SpatialVTS}, while many others adopt a data-centric method; in other words, they integrate 3D scene representations (e.g., depth maps and point clouds) into VLMs~\citep{Open3DVQA,SoFar}.
Meanwhile, many benchmarks have been proposed to evaluate their 2D and 3D spatial reasoning ability, including SpatialVQA~\citep{SpatialVLM}, RoboSpatial-Home~\citep{RoboSpatial}, VSI-Bench~\citep{VSI-Bench}, and many others~\citep{VSR,WhatsUp,SpatialBot,Blink,SpatialRGPT,RoboPoint,Robo2VLM,Open3DVQA,CV-Bench,Embodied-R}.
But these benchmarks are limited by their static nature and lack of systematic spatial-centric design. In addition, they predominantly adopt VQA-style evaluations, which are often prone to linguistic biases. In contrast, we propose a systematic task design and a unified generative paradigm, shifting the focus toward active, embodied evaluation.

\paragraph{Simulation-based evaluation through robotic tasks.}
Unlike human-assisted real-world evaluation, simulation-based approaches allow for more scalable and reproducible evaluation of robotics models, and have been widely used to assess robot policies in domains such as navigation and manipulation~\citep{Alfred,Habitat2,Behavior,ManiSkill2,RLBench,MetaWorld,Ravens,Calvin,Open6DOR}.
Due to the inherent limitations of simulators, substantial discrepancies exist between simulated observations and real-world observations. 
To bridge the gap, researchers have been improving physics engines and enhancing synthesis mechanisms to approximate real-world perceptions~\citep{MuJoCo,Gibson,Room2room,IsaacSim}.
Though there have been simulated environments, such as LIBERO~\citep{Libero}, CALVIN~\citep{Calvin}, SIMPLER~\citep{Simpler}, and EmbodiedBench~\citep{EMBench} for \emph{real-to-sim} evaluation, they are limited in overly simplified scenes and tasks or reliance on external tools. In addition, none of them provides a systematic design of spatial-centric reasoning tasks and supports comprehensive diagnoses. 

\paragraph{Foundation models for robotics manipulation.}
Foundation models, including pre-trained LLMs and VLMs, have been applied to robotic manipulation. Early work focuses primarily on task planning while relying on predefined primitives to achieve robot control~\citep{SayCan,PaLM-E,CaP,RoboChatGPT,COME}. Recently, many have attempted to generate trajectories, i.e., sequences of poses, for motion planning~\citep{CoPa,VoxPoser,ReKep,RoboPoint,SoFar} and devise agentic frameworks for reasoning and acting~\citep{GR-ER1.5}. Following the unified design philosophy, more recent efforts have focused on developing integrated vision-language-action models (VLAs) that can directly generate low-level action sequences as control policies~\citep{RT-1,RoboFlamingo,Octo,Pi0,LAPA,UniVLA,UniVLA-BAAI}, but their success hinges on the underlying spatial reasoning of their vision-language components, we focus on diagnosing VLMs to isolate and identify the specialized spatial inductive biases that are required to inform and improve future unified architectures.

\paragraph{6-DoF object rearrangement.}
6-DoF object rearrangement involves predicting a goal state of an object that is described in SE(3) and satisfies the given instruction. With a motion planner, such a formulation enables zero-shot transfer of foundation models from perception to execution~\citep{VoxPoser,Dream2Real}. The approaches to 6-DoF tasks can be roughly divided into generative- and discriminative-based. Generative methods solve for a goal translation and rotation of a directional vector under certain constraints~\citep{CoPa,ReKep}, while discriminative approaches generate random candidates and use a critic to filter and select the best goal pose~\citep{Open6DOR,Dream2Real}. We follow the generative paradigm and prompt VLMs to generate a goal pose and ground it in the simulated physical world. 

%
\section{Spatial-centric Evaluation of Embodied VLMs}

\begin{figure*}[tp]
\centering
\includegraphics[width=\linewidth]{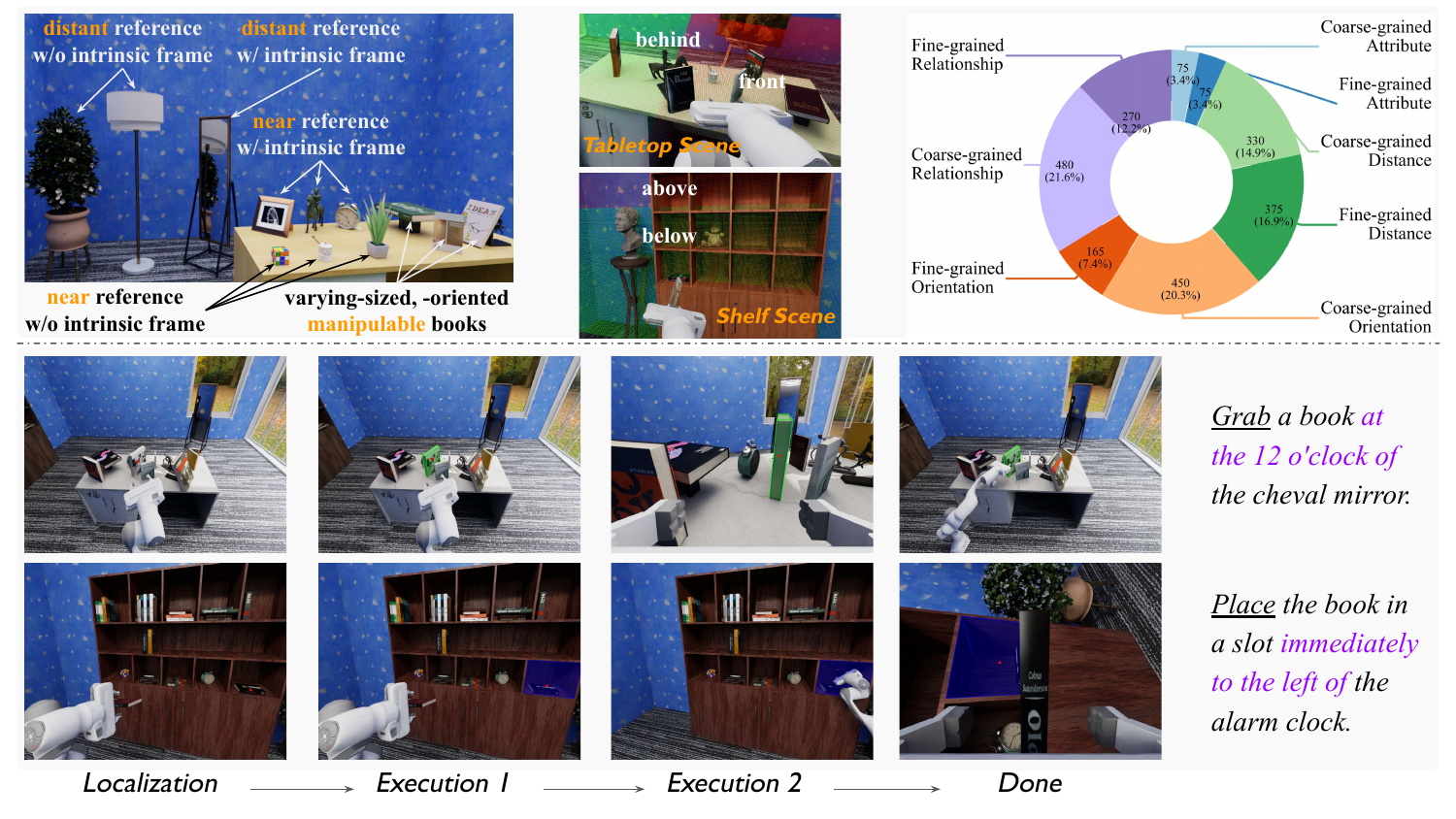} 
\vspace{-2em}
\caption{
\model: a simulated physical world. 
\textbf{Top:} the spatial world of \model covers key factors of spatial reasoning like spatial aspects (e.g., relationship and distance), reference frames, reference objects (\S\ref{sec:espire-task}). It features a tabletop scene for \emph{pick} tasks and a shelf scene for \emph{place} tasks (\S\ref{sec:espire-env}) and supports reasoning at varying granularities (see Table~\ref{tab:spatial-aspect} in Appendix~\ref{sup:env}). 
\textbf{Bottom:} example \model tasks that all inherently rely on spatial reasoning.
}
\label{fig:main}
\vspace{-0em} 
\end{figure*}

We propose evaluating the spatial cognition of VLMs through robotics tasks situated in a simulated physical world, narrowing the gap between evaluation and real-world deployment. To adapt VLMs for robotics tasks, we decompose each task into two sequential subtasks: localization and execution, formulate them as generative tasks, and ensure that spatial reasoning is the key factor.

\begin{itemize}[leftmargin=*]
  \item \textbf{Localization} refers to locating a target that is specified in a given instruction from the paired scene, such as the `book' in `\emph{pick up the farthest book}' and the `empty spot' in `\emph{place the book in an empty spot}'. We follow~\citet{RoboPoint} and~\citet{RoboRefer} and formulate it as a \emph{pointing} task that produces 2D coordinates on scene images.

    \textit{Evaluation Metric.} We measure model performance using accuracy, defined as the fraction of correct localizations. Unlike discriminative VQA-style evaluations that rely on distractors for automatic metrics, our generative formulation allows for directly comparing the predicted point against the target segmentation mask.

  \item \textbf{Execution} follows the localization stage to execute actions (e.g., \emph{pick} or \emph{place}) in the physically grounded environment. Since VLMs cannot directly produce low-level control actions, we simplify execution as a 6-DoF task that predicts the goal pose, including goal position and orientation prediction, in SE(3). We again formulate goal position prediction as a \emph{pointing} task.

    \textit{Evaluation Metric.} We measure model performance using acceptance rate, defined as the fraction of physically achieved poses. The acceptability of a predicted pose is assessed by a motion planner like cuRobo~\citep{cuRobo}, making VLMs physically grounded.
\end{itemize}

In both tasks, native spatial reasoning is inherently needed since VLMs are required to generate positions and orientations in 3D, without relying on external tools. The shared \emph{pointing} formulation between localization and execution further bridges spatial reasoning for understanding and for acting.

%
\section{The \model Benchmark}

We propose \model, a simulated environment that provides a suite of robotics tasks for diagnosing spatial-centric reasoning (see Figure~\ref{fig:main}).
We design \model systematically both in instructions (\S\ref{sec:espire-task}) and environments (\S\ref{sec:espire-env}), ensuring a broad coverage of spatial reasoning scenarios, enabling scalable robotic task generation (\S\ref{sec:espire-sim}), and supporting targeted analysis across contexts and granularities.

\subsection{Spatial Reasoning Tasks}\label{sec:espire-task}

\paragraph{Task specification.}
We group spatial reasoning tasks into four broad classes by the spatial aspects they require to reason about: \emph{relationships}, \emph{distances}, \emph{attributes} (e.g., dimensions and volumes), and \emph{orientations}. A spatial reasoning task typically involves describing an object in relation to another (e.g., `\emph{grab the book to your left}'), thus relying on a frame of reference. Following~\citet{SpaceInLang}, we consider three types of reference frames: \emph{relative}, \emph{intrinsic}, and \emph{absolute} frames. The choice of reference frame depends on the reference object, e.g.. intrinsic-oriented objects like `picture frame' that have a clear front face naturally support intrinsic frames, whereas non-oriented objects like `sphere ball' do not. Moreover, the reference frame may vary with linguistic specifications, e.g., `\emph{pick up a book on the left of the picture frame}' exhibits ambiguity since both a relative frame and an intrinsic frame can be used, but attaching the clause `\emph{relative to the picture frame's front}' makes the intrinsic frame the only valid interpretation.

To disentangle this complexity, we identify three key factors that characterize spatial reasoning: spatial aspect ($S$), reference frame ($F$), and reference object ($O$); we define their combination $C = (S, F, O)$ as the task specification. A particular configuration of these factors specifies a context for spatial reasoning. For example, $c = (\text{relationship}, \text{intrinsic}, \text{table})$ requires using the \emph{intrinsic} frame of the \emph{table} to carry out relationship reasoning; an instance of it can be `\emph{grab a book on the left of the table}.' This disentanglement lets us focus on designing tasks that target reasoning at varying granularities like \emph{left}, \emph{leftmost}, \emph{second leftmost}, and \emph{to your 11 o'clock}.\looseness=-1

\paragraph{Instruction representation.}
We associate each task instruction with a 3-tuple $T = (C, A, P)$, where $C$ denotes the task specification, $A\in\{\text{pick}, \text{place}\}$ represents execution, and $P$ indicates localization. We represent $P$ as a functional program~\citep{CLEVR} that can be evaluated on the 3D scene graph representation $G$ of a given environment state and produces a list of valid answers, i.e., objects to be manipulated or spaces to be filled. Crucially, A functional program is composed of atomic functions and defines a reasoning chain, such as finding a specific object $\texttt{unique}(\texttt{filter}(O, G))$ and querying the objects to its left $\texttt{filterRel}(\text{left}, \texttt{unique}(\texttt{filter}(O, G)))$. This enables flexible control of the task complexity by varying the number of reasoning hops.

\paragraph{Instruction families.}
We define an instruction family on top of a task $T=(C, A, P)$ by associating it with a set of task templates that represent different linguistic expressions of the functional program $P$. Supposing $C=(\text{distance}, \text{viewer}, \text{intrinsic})$, $A=\text{`Pick'}$, and a template `\emph{[A] a book among the books [R] you}', we can create an instruction, which queries the distance between a book and the viewer, by binding the variable $[R]$ with a type of distance reasoning (e.g., Closest or Furthest). Using the same variable $[R]$, the functional program $P$ can be formed as:
$$\ttt{filter}\ttt{Dist}([R], \ttt{filter}(\text{book}, G), \text{viewer})$$

We curate a total of 148 spatial-reasoning task types, distributed across 65 instruction families, including 31 `pick' instruction families and 34 `place' instruction families. For each instruction family, we manually write 3-4 templates to enhance linguistic diversity. Though functional programs enable multi-hop compositional reasoning, we limit reasoning up to 3 hops, as our primary focus is on spatial rather than compositional reasoning.\footnote{Nonetheless, \model can be readily extended by increasing the number of spatial reasoning hops.} In practice, we find that a small number of spatial reasoning hops already poses challenges for existing multimodal foundation models.

\subsection{Simulation Environment}\label{sec:espire-env}

We simulate two task environments in \model: tabletop and shelf scenes.
Both are constructed systematically using a diverse array of photorealistic objects and various spatial layouts and environmental factors like lighting and clutter.
This design ensures that our environments provide a comprehensive instantiation of the task specification $C$, yielding diverse instances that challenge model reasoning across multiple levels of granularity (refer to Appendix~\ref{sup:env} for detailed scene configurations).

\paragraph{Environment representation and generation.}
We initialize each environment from a random state, which is represented by a 3D scene graph that consists of nodes as objects and edges as spatial relationships. All objects are annotated with ground-truth information, including sizes, dimensions, and poses relative to a predefined absolute reference frame. We generate the initial state of an environment by sampling a random 3D scene graph and rendering it in Isaac Sim~\citep{IsaacSim}, ensuring that the environment is physically valid. We adjust the minimum margin of objects and the dimensions of shelf slots; this mitigates the visual ambiguity of spatial aspects and accommodates sufficient, physically feasible tasks in the environment. The Franka robot is initialized in a random pose. We equip it with an on-wrist camera that provides an egocentric view and supplement it with two fixed-position cameras that provide global views of the tabletop and shelf scenes, respectively (referred to as world views). To increase variety and realism, we add external lights. We randomly sample and initialize the positions and orientations of all cameras and external lights.

\paragraph{Reducing the real-to-sim visual gaps.}
Visual gaps mainly arise from distribution shifts in texture, material, lighting, and camera configurations. Instead of performing complex visual-matching mitigation as in SimplerEnv~\citep{Simpler}, we employ a more scalable strategy that focuses on enhancing the diversity of the environment: we use annotated 3D assets with realistic textures and tune their sizes to reflect their real-world counterparts. For essential background assets like the tabletop and shelf, we randomly assign textures derived from real-world materials. Combined with randomization in lighting and camera poses, this produces a diverse and visually realistic set of environments (see details in Appendix~\ref{sup:env} and a discussion on sim-to-real relevance in Appendix~\ref{sup:sim-to-real}).

\subsection{Simulation Tasks}\label{sec:espire-sim}

A simulation task is defined by a pair of an environment state and a task instruction.
We generate `pick' and `place' tasks sequentially. First, we sample and render an environment. The Franka robot is always initialized in a position suitable for performing `pick' tasks, so we start with `pick' task generation, and `place' task generation follows the same procedure.
For each variable in a given instruction family, after sampling a random type, we perform value filtering. This is particularly useful for the reference-object variable, as not all reference objects appear in the task space visible from the world view. Once all variables are bound and instantiated, we obtain the final functional program and execute it on the 3D scene graph representation of the visible portion of the environment state. The yielded answers are further verified using a motion planner. We only retain those that correspond to feasible manipulations.\footnote{We assume the robot can move freely in 3D, with both locomotion across the ground plane and vertical motion along the global up-axis. This relaxation facilitates reliable execution when using VLMs and a large task space that broadens the coverage of spatial reasoning scenarios.}
Finally, we randomly select a task template from the given task family and instantiate it into a natural language instruction.

%
\section{Experiments}

\subsection{Experimental Setups}

\paragraph{Evaluated models.} We consider a diverse range of multimodal foundation models, including proprietary VLMs like Gemini2.5-Pro~\citep{Gemini2.5}, public general-purpose VLMs like instruction-tuned Qwen3-VL~\citep{QWen25vl} and InternVL3~\citep{InternVL3}, and spatial-reasoning enhanced VLMs like RoboBrain2.0~\citep{RoboBrain2.0}. 

\paragraph{Evaluation tasks.} Each task family is paired with at least 15 different scenes, leading to around 15 trials on average. We define the difficulty of a task as the complexity of the accompanying scene. Specifically, we categorize the tabletop and shelf tasks into three difficulty levels: easy, medium, and hard. For tabletop tasks, the three levels correspond to scenes that contain 1-2, 3-5, and 6-8 books on the table, respectively.\footnote{The book number best correlates with task complexity, but overall complexity is driven by instructions and environmental factors like object number and pose, light, and texture.} For shelf tasks, difficulty is defined as the fullness of the associated shelf: easy, medium, and hard correspond to shelves where one-third, two-thirds, and all slots are occupied, respectively (refer to Table~\ref{tab:shelf} in Appendix~\ref{sup:asset} for illustrations).

\paragraph{Evaluation settings.} Our evaluation suite offers a total of 2,220 tasks, consisting of  1,095 \emph{pick} tasks and 1,125 \emph{place} tasks.
We limit the number of attempts to 3 for localization and 5 for execution. If localization fails, we randomly select a gold target for execution; otherwise, we use the target localized by the model for execution. We consider non-reflection and reflection settings.
In the non-reflection setting, the initial observation is provided by the world-view, while all subsequent observations are obtained from the ego-view. In the reflection setting, the model additionally receives as input its reflections from the previous failed attempt (refer to Algorithms~\ref{alg:localization} and~\ref{alg:execution} in Appendix~\ref{sup:eval}). To enhance reflection in execution, both the world-view and the ego-view are provided.
We prompt models to output a 2D point in pixel space while providing them with ground-truth depth. Depending on the settings, we may provide ground-truth rotations of pitch, yaw, and roll or prompt models to generate them directly.

\begin{table*}[t!]
\centering
\caption{The localization accuracy (\%), acceptance rate (\%) in execution, and overall task success rate (\%) across different VLMs.}
\label{tab:main-results}

\resizebox{0.75\linewidth}{!}{%
\begin{tabular}{@{}lcccccc@{}}
\toprule
\multicolumn{1}{c}{\multirow{2}{*}{Models}} & \multicolumn{3}{c}{Pick} & \multicolumn{3}{c}{Place} \\
\cmidrule(l){2-7}
\multicolumn{1}{c}{} & accuracy & acceptance & success & accuracy & \multicolumn{1}{c}{acceptance} & success \\
\midrule
\multicolumn{7}{c}{\cellcolor[HTML]{EFEFEF}{\emph{w/o reflection}}} \\
\midrule
Gemini2.5-Pro & \textbf{57.72} & \textbf{63.93} & \textbf{34.06} & 50.61 & 28.36  & 5.68 \\
InternVL3-78B & 28.31 & 63.01 & 17.26 & 23.66 & 40.94 & 9.67 \\
RoboBrain2.0-7B & \textbf{57.72} & 18.81 & 10.87 & \textbf{50.70} & 15.68 & 8.64 \\
Qwen3-VL-30B-A3B & 54.43 & 62.56 & 32.15 & 45.54 & \textbf{43.47} & \textbf{20.00} \\
Qwen3-VL-8B & 47.03 & 63.20 & 29.32 & 35.71 & 37.31 & 12.41 \\
Qwen3-VL-235B-A22B & 51.96 & 52.79  & 26.76 & 47.42 & 41.22 & 19.34 \\
\midrule
\multicolumn{7}{c}{\cellcolor[HTML]{EFEFEF}{\emph{w/ reflection}}} \\
\midrule
Qwen3-VL-30B-A3B & 54.52 & 27.85 & 17.08 & 51.92 & 23.94 & 13.80 \\
Qwen3-VL-8B & 58.63 & 24.38 & 15.07 & 54.08 & 12.02 & 6.67 \\
Qwen3-VL-235B-A22B & \textbf{64.29} & \textbf{36.71} & \textbf{23.20} & \textbf{59.72} &\textbf{25.45} & \textbf{15.40} \\ 
\bottomrule
\end{tabular}
} 
\vspace{-0em} 
\end{table*}

\subsection{Main Results}

In general, proprietary VLMs like Gemini2.5-Pro show the strongest performance under most metrics, while public VLMs like the Qwen3-VL series is narrowing the gaps and even outperform Gemini2.4-Pro in execution on \emph{place} tasks.
Interestingly, larger models do not necessarily lead to better performance, e.g., Qwen3-VL-30B with 3B activated parameters outperforms Qwen3-VL-8B and Qwen3-VL-235B (w/ 22B activated parameters).
Moreover, all models demonstrate decent localization accuracy except InternVL3-78B, which has an accuracy below 30\% (see Table~\ref{tab:main-results}), possibly because the native multimodal pre-training adopted by InternVL3-78B for aligning vision and language is not as effective as widely-used stage-by-stage alignment learning~\citep{LLaVa}.
Despite that, it is surprising to see that InternVL3-78B performs much better (e.g., >40\%) in execution than in localization.
While RoboBrain2.0-7B achieves impressive localization performance, it is mostly because its post-training involves extensive generic spatial reasoning tasks, but, unfortunately, this does not transfer to improved performance in execution (e.g., <20\%) that requires acting-oriented spatial reasoning about 3D rotational geometry.

\paragraph{`Place' is generally harder than `pick.'} Compared with \emph{pick} tasks, \emph{place} tasks impose much stricter acceptance conditions. Specifically, when predicting a pose for the placement of a book, the model needs to consider additional constraints of the target space, especially when the target space is partially occupied. Moreover, the model usually suffers much more from occlusion when it gets closer to the target space, making it harder to predict a relatively-center point of the space or recover from a non-ideal position, whereas in pick tasks, the model only needs to align with any one of all graspable faces (e.g., the spine or the top edge), and it can optionally move to a better position to facilitate pose prediction (refer to our qualitative studies in Table~\ref{tab:quality_analysis}).

\paragraph{Reflection is helpful for localization but does not necessarily help with execution.} While reflection improves localization performance of all Qwen3-VL models (see Table 1), it does not yield a comparable improvement in execution; conversely, it degrades execution performance. This is likely because strong 3D rotation understanding is the key factor for execution and forms the foundation for reflection; however, as we will see in an analysis of rotation predictions (see Table~\ref{tab:qwenvl-rotation}), current VLMs are weak in this capacity, suggesting that future work is well-suited for curating rotation-reasoning data for fine-tuning.

\begin{figure}[t!]
    \centering
    \includegraphics[width=\linewidth]{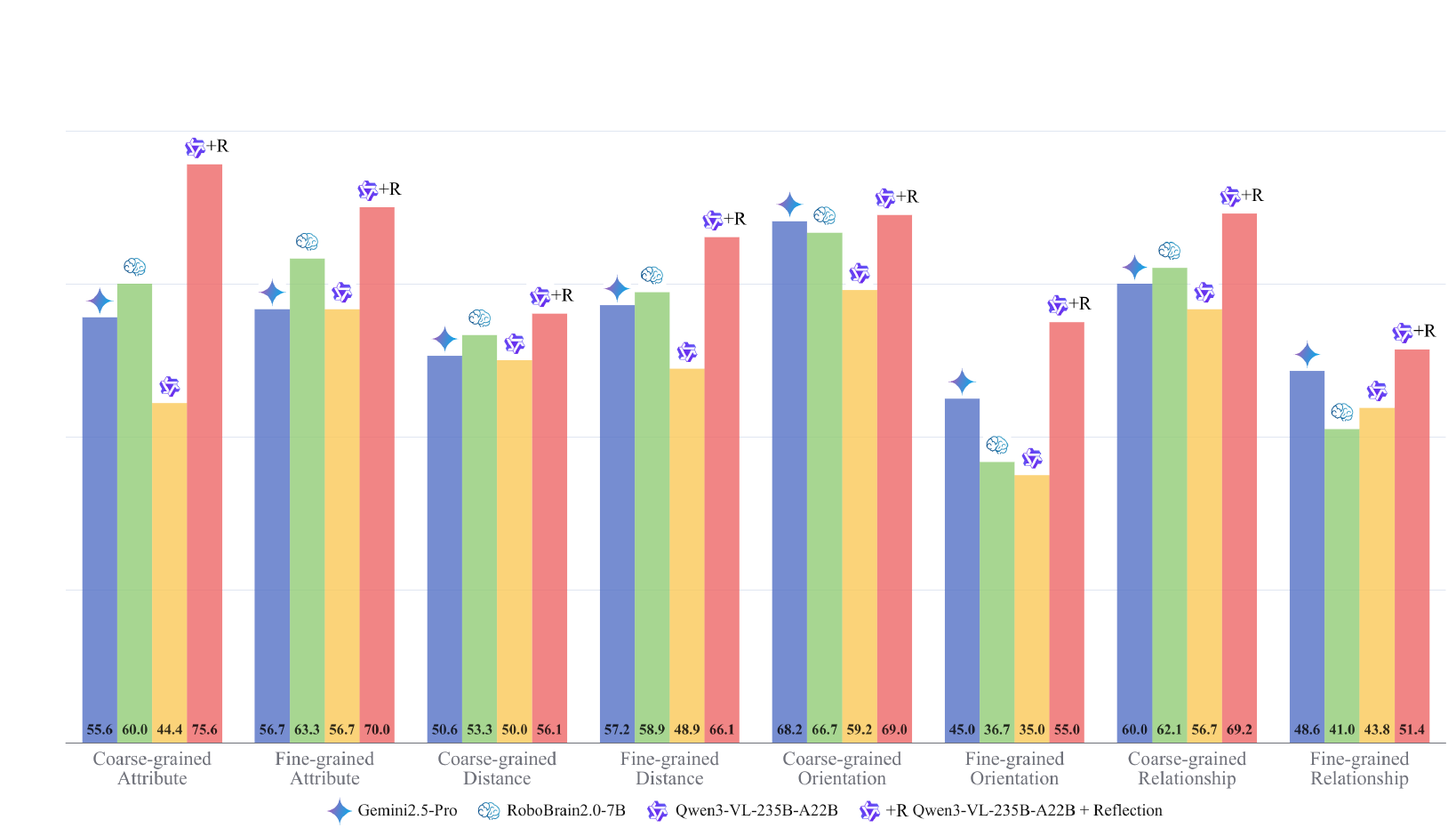}
    \caption{Localization performance across spatial aspects and granularities on \emph{pick} tasks.}
    \label{fig:aspect-pick}
    \vspace{-1em}
\end{figure}

\begin{wraptable}{r}{0.5\linewidth}
\vspace{-1em} 
\centering
\begin{minipage}[t]{\linewidth}
\caption{Localization accuracy (\%) across four primary spatial aspects.}
\label{tab:aspect-pick-place}
\centering
\resizebox{\linewidth}{!}{%
\begin{tabular}{@{}lcccccccc@{}}
\toprule
\multicolumn{1}{c}{\multirow{2}{*}{Models}} 
& \multicolumn{4}{c}{Accuracy} \\ 
\cmidrule(l){2-5}
\multicolumn{1}{c}{}
& attribute & distance & orientation & relationship\\
\midrule
\multicolumn{6}{c}{\cellcolor[HTML]{EFEFEF}{Pick}} \\
\midrule
Gemini2.5-Pro & 56.00 & 53.89 & \textbf{63.81} & \textbf{56.52} \\
InternVL3-78B & 29.33 & 21.67 & 33.65 & 30.14 & \\
RoboBrain2.0-7B & \textbf{61.33} & \textbf{56.11} & 60.95 & 55.65 \\
Qwen3-VL-30B-A3B & 57.33 & 50.83 & 57.14 & 55.07 \\
Qwen3-VL-8B & 42.67 & 40.28 & 53.97 & 48.70 \\
Qwen3-VL-235B-A22B & 49.33 & 49.44 & 54.60 & 52.75 \\
\hdashline
\textbf{\textsc{Average}} & 49.33 & 45.37 & 54.02 & 49.81 \\
\midrule
\multicolumn{6}{c}{\cellcolor[HTML]{EFEFEF}{Place}} \\
\midrule
Gemini2.5-Pro & 53.33 & \textbf{46.09} & \textbf{48.75} & \textbf{55.06} \\
InternVL3-78B & 28.00 & 16.23 & 21.25 & 30.62 \\
RoboBrain2.0-7B & \textbf{69.33} & 45.80 & 47.08 & 53.58 \\
Qwen3-VL-30B-A3B & 48.00 & 38.55 & 39.58 & 54.57  \\
Qwen3-VL-8B & 44.00 & 28.41 & 31.67 & 42.82 \\
Qwen3-VL-235B-A22B & 62.67 & 37.68 & 46.26 & 53.58 \\
\hdashline
\textbf{\textsc{Average}} & 47.20 & 33.33 & 37.17 & 47.03 \\
\bottomrule
\end{tabular}
}
\end{minipage}
\vspace{-2em} 
\end{wraptable}

\subsection{Analysis}

The systematic design of \model enables fine-grained analysis of model behavior. We demonstrate this by examining spatial reasoning performance across spatial aspects and task difficulty levels, and by analyzing behavior during successful task execution.

\paragraph{Localization performance across spatial aspects.}
We group results by spatial aspects (see Table~\ref{tab:aspect-pick-place} and Figure~\ref{fig:aspect-pick}). Overall, all models perform worse on `distance' than on the other spatial aspects, across \emph{pick} and \emph{place} tasks, indicating that current VLMs lack the capacity for precise distance understanding. Among them, Gemini2.5-Pro and RoboBrain2.0-7B exhibit relatively stronger overall performance while showing smaller performance variations across spatial aspects, likely because they have been specifically fine-tuned on related spatial reasoning tasks; this is explicitly the case for RoboBrain2.0-7B.

\paragraph{Model performance across task difficulty levels.}
We further group results by task difficulty levels (see Table~\ref{tab:task_difficulty}). Unsurprisingly, on \emph{pick} tasks, most models demonstrate a decrease in both localization execution with increasing task difficulty, except InternVL3-78B and RoboBrain2.0-7B that, in some cases, perform slightly better on harder tasks. Similarly, on \emph{place} tasks, both the localization performance and the execution performance negatively correlate with the task difficulty. Still, there are exceptions like InternVL3-78B and Qwen3-VL-30B-A3B.

\begin{wraptable}{r}{0.5\linewidth}
\vspace{-1em} 
\centering
\begin{minipage}[t]{\linewidth}
\caption{Performance across difficulty levels.}
\label{tab:task_difficulty}
\centering
\resizebox{\linewidth}{!}{%
\begin{tabular}{@{}lcccccc@{}}
\toprule
\multicolumn{1}{c}{\multirow{2}{*}{Models}} 
& \multicolumn{3}{c}{Accuracy (\%)} 
& \multicolumn{3}{c}{Acceptance (\%)} \\ 
\cmidrule(l){2-7}
\multicolumn{1}{c}{}
& easy & medium & hard
& easy & medium & hard \\
\midrule
\multicolumn{7}{c}{\cellcolor[HTML]{EFEFEF}{Pick}} \\
\midrule
Gemini2.5-Pro & 60.78 & \textbf{60.98} & 52.04 & \textbf{70.96} & 60.98 & 60.71 \\
InternVL3-78B & 24.85 & 29.00 & 30.61 & 60.78 & 65.85 & \textbf{62.24} \\
RoboBrain2.0-7B & \textbf{62.57} & 56.10 & \textbf{55.10} & 21.56 & 15.72  & 19.39 \\
Qwen3-VL-30B-A3B & \textbf{62.57} & 53.39 & 48.87 & 65.87 & \textbf{66.67} & 55.87 \\
Qwen3-VL-8B & 58.08 & 42.82 & 41.58 & 64.97 & 63.96 & 60.97 \\
Qwen3-VL-235B-A22B & 59.58 & 52.57 & 44.90 & 58.68 & 55.56 & 45.15 \\
\midrule
\multicolumn{7}{c}{\cellcolor[HTML]{EFEFEF}{Place}} \\
\midrule
Gemini2.5-Pro & \textbf{57.46} & 48.21 & 46.11 & 36.06 & 28.37 & 20.46 \\
InternVL3-78B & 25.35 & 22.31 & 23.34 & 51.55 & 39.12 & 31.99 \\
RoboBrain2.0-7B & 52.68 & \textbf{52.62} & \textbf{46.69} & 18.31 & 16.80 & 11.82 \\
Qwen3-VL-30B-A3B & 47.61 & 44.08 & 44.96 & \textbf{52.39} & \textbf{43.80} & \textbf{34.01} \\
Qwen3-VL-8B & 37.18 & 36.74 & 33.14 & 42.25 & 40.06 & 29.39 \\
Qwen3-VL-235B-A22B & 51.83 & 48.48 & 41.79 & 48.17 & 43.53 & 31.70 \\
\bottomrule
\end{tabular}
}
\end{minipage}
\vspace{-0em} 
\end{wraptable}

\begin{table}[t!]
\centering
\caption{The average number of attempts to succeed in localization and execution, and average distance (meter) between the target and end-effector upon execution success and before execution success. `Rank' indicates model ranking in execution.}
\label{tab:distance-analysis}
\centering
\resizebox{0.65\linewidth}{!}{%
\begin{tabular}{@{}lcc @{\hspace{1em}} c @{\hspace{1em}} c @{\hspace{1em}} c@{}}
\toprule
\multicolumn{1}{c}{Models} & \#Localization & \#Move & \makecell[c]{Dist. at \\ success} &  \makecell[c]{Dist. before \\ success} & Rank\\
\midrule
\multicolumn{6}{c}{{\cellcolor[HTML]{EFEFEF}{Pick}}} \\
\midrule
Gemini2.5-Pro & 1.20 & 2.54 & 0.07 & 0.47 & 1 \\
InternVL3-78B & 1.05 & 2.56 & 0.05 & 0.48 & 3 \\
RoboBrain2.0-7B & 1.36 & 2.54 & 0.05 & 0.50 & 6 \\
Qwen3-VL-30B-A3B & 1.16 & 2.49 & 0.06 & 0.38 & 4 \\
Qwen3-VL-8B & 1.17 & 2.41 & 0.05 & 0.42 & 2 \\
Qwen3-VL-235B-A22B & 1.18 & 2.53 & 0.05 & 0.40 & 5 \\
\midrule
\multicolumn{6}{c}{{\cellcolor[HTML]{EFEFEF}{Place}}} \\
\midrule
Gemini2.5-Pro & 1.42 & 3.27 & 0.26 & 0.75 & 5 \\
InternVL3-78B & 1.08 & 2.07 & 0.24 & 0.97 & 3 \\
RoboBrain2.0-7B & 1.59 & 2.98 & 0.24 & 0.85 & 6 \\
Qwen3-VL-30B-A3B & 1.33 & 2.12 & 0.24 & 0.95 & 1 \\
Qwen3-VL-8B & 1.30 & 2.10 & 0.24 & 0.97 & 4 \\
Qwen3-VL-235B-A22B & 1.28 & 2.16 & 0.23 & 0.91 & 2 \\
\bottomrule
\end{tabular}
}
\vspace{-0em} 
\end{table}


\paragraph{Prerequisites for successful execution.}
Next, we analyze the prerequisites that are strongly associated with successful execution. To this end, we compute the average number of attempts used to achieve successful localization and execution, and the average distances between the target and the end-effector upon execution success and before execution success (see Table~\ref{tab:distance-analysis}). Though there is no clear correlation between execution success and pre-success distance, we find that InternVL3-78B and Qwen3-VL-30B-A3B, which are relatively better at execution, tend to be far-sighted (e.g., w/ an average distance of 48cm) and near-sighted (e.g., w/ an average distance of 38cm), respectively.
Interestingly, the pre-success distance in \emph{place} tasks is usually twice that in \emph{pick} tasks, presumably because, in \emph{place} tasks, the robot needs to stay reasonably far away from the target space to mitigate occlusion.
Moreover, in \emph{place} tasks, strong models like the Qwen3-VL series often require a moderate number of moves; that is, they tend to try multiple times (i.e., around 2.1) before making the final successful execution. In contrast, models that try much more times (i.e., around 3) are usually weaker in execution, e.g., RoboBrain2.0-7B fails spectacularly because it struggles in acting-oriented spatial reasoning.

Apart from the above quantitative analysis, we present qualitative analysis of both successful runs and failed runs in Table~\ref{tab:quality_analysis}.
\begin{table}[th!]
\centering
\caption{Qualitative analysis. We categorize intermediate executions into following six types: T1 denotes a grasp-favorable viewpoint; T2 denotes a grasp-infeasible viewpoint; T3 denotes manipulator occlusion; T4 denotes object occlusion; T5 denotes unrecognizable target; and T6 denotes physically-achievable execution.}
\label{tab:quality_analysis}

{\setlength{\tabcolsep}{0pt} 
	\makebox[\linewidth]{\resizebox{1.\linewidth}{!}{%
\begin{tabular}{c 
m{0.16\linewidth} @{\hspace{0.1em}}
m{0.16\linewidth} @{\hspace{0.1em}}
m{0.16\linewidth} @{\hspace{0.1em}}
m{0.16\linewidth} @{\hspace{0.1em}}
m{0.16\linewidth} @{\hspace{0.1em}}
m{0.16\linewidth}}
\toprule
\text{Model} & 
\multicolumn{1}{c}{\text{Step 1}} & 
\multicolumn{1}{c}{\text{Step 2}} & 
\multicolumn{1}{c}{\text{Step 3}} & 
\multicolumn{1}{c}{\text{Step 4}} &
\multicolumn{1}{c}{\text{Step 5}} &
\multicolumn{1}{c}{\text{Step 6}}
\tabularnewline
\midrule

& \multicolumn{6}{l}{\textit{\small{Find a book at 12 o'clock of the cheval mirror from the table, and \textbf{grab} it.}}}
\tabularnewline

\multirow{2}{*}{\includegraphics[width=0.7cm,height=0.7cm]{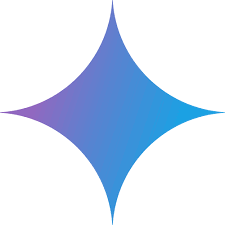}} &
\includegraphics[width=\linewidth,valign=m]{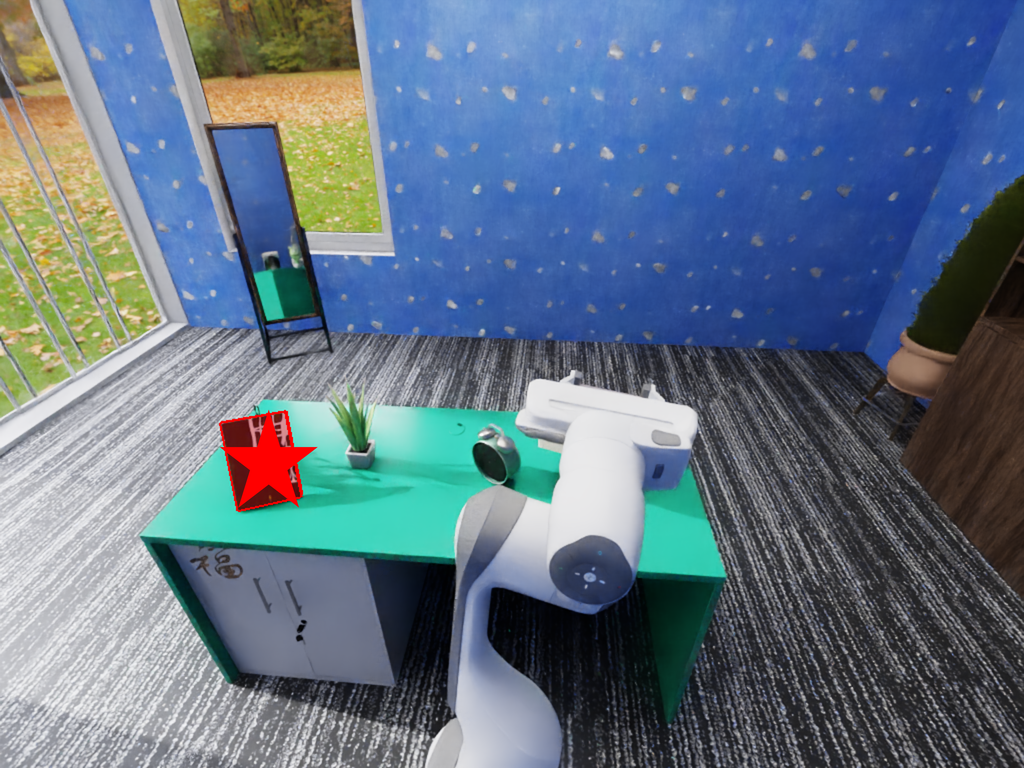} &
\includegraphics[width=\linewidth,valign=m]{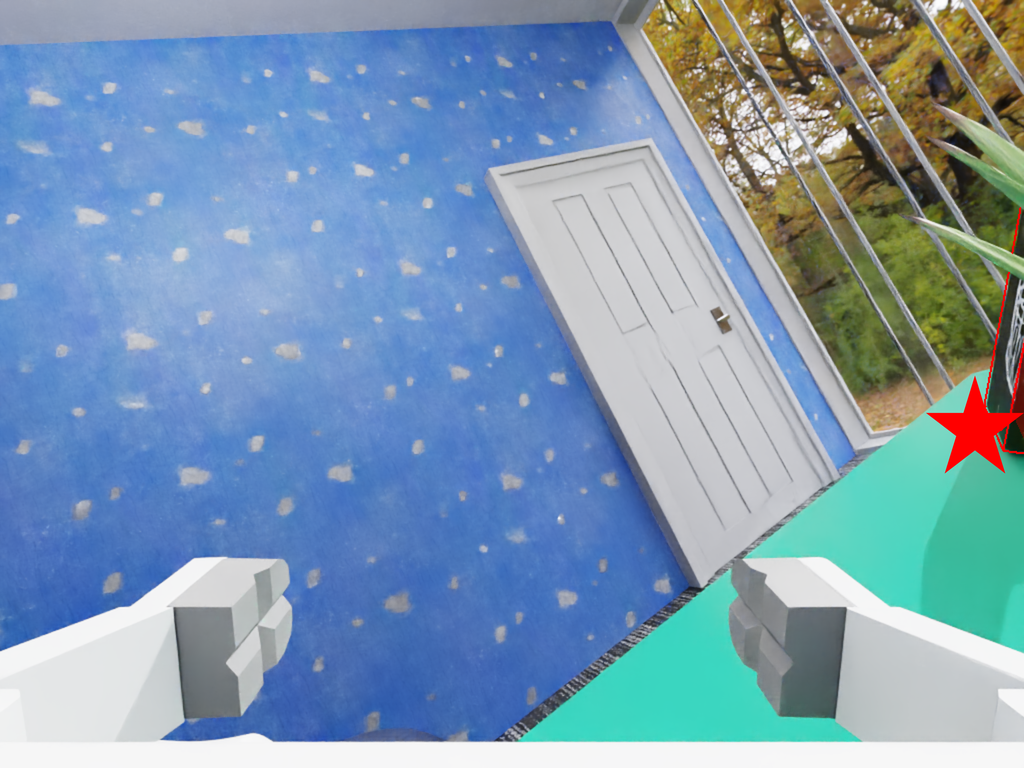} &
\includegraphics[width=\linewidth,valign=m]{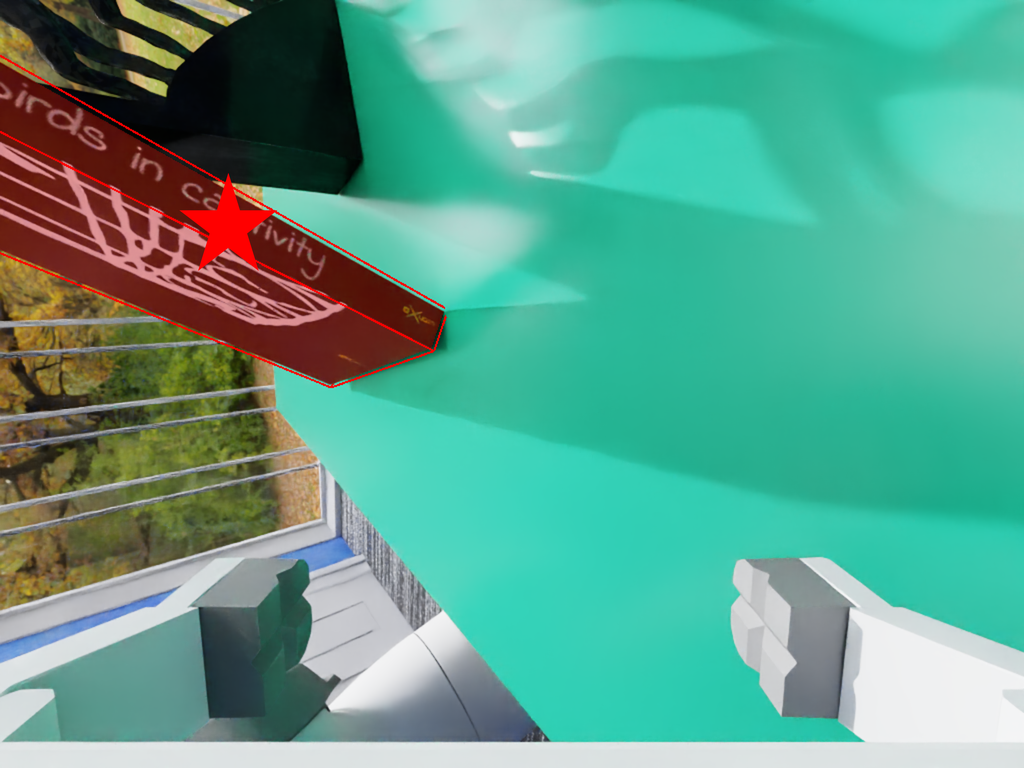} &
\includegraphics[width=\linewidth,valign=m]{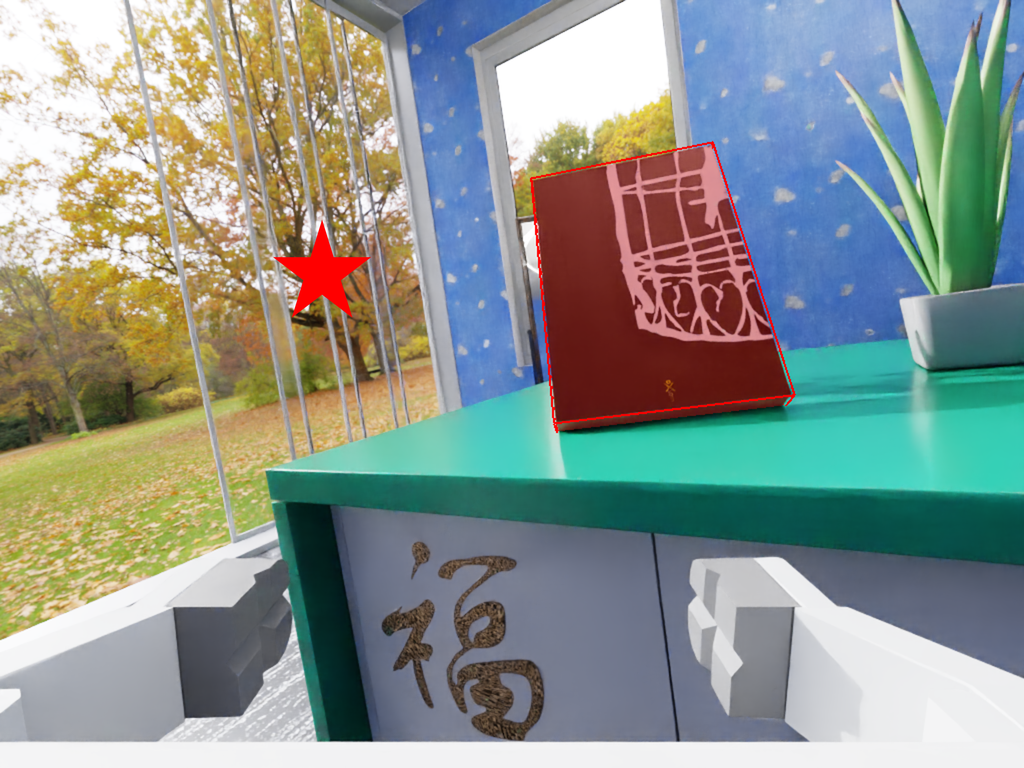} &
\includegraphics[width=\linewidth,valign=m]{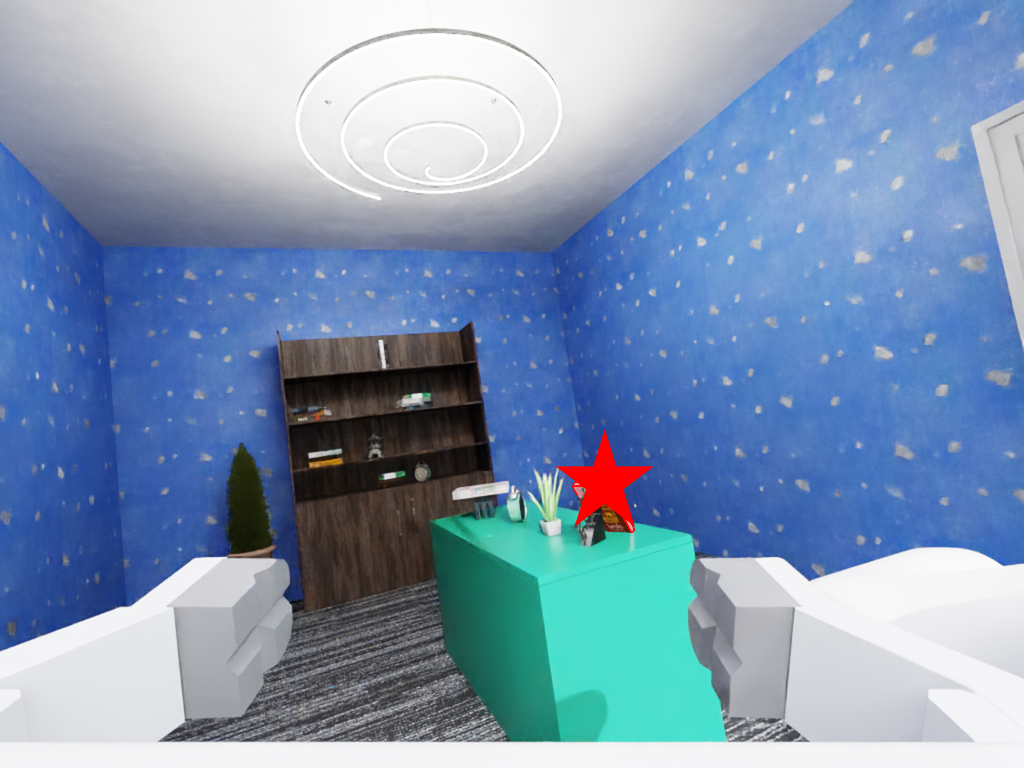} &
\includegraphics[width=\linewidth,valign=m]{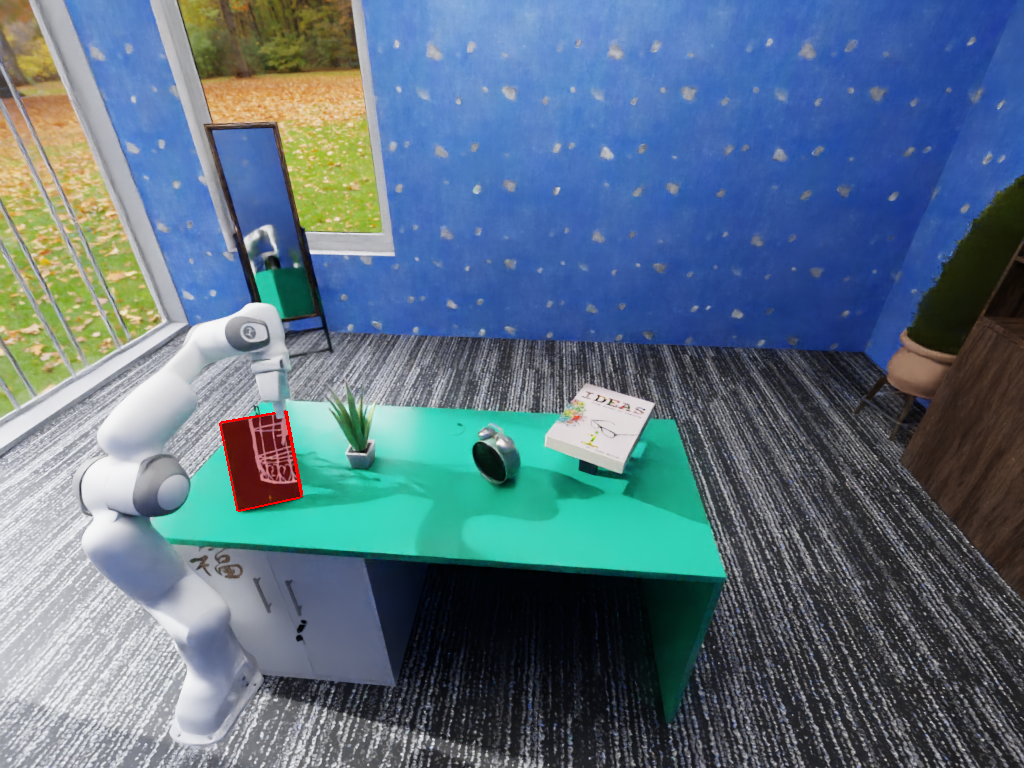} 
\tabularnewline

& \scriptsize \centering T2 & \scriptsize \centering T2 & \scriptsize \centering T2 & \scriptsize \centering T1 & \scriptsize \centering T6 & \scriptsize \centering Execution \tabularnewline


\multirow{2}{*}{\shortstack{
    \includegraphics[width=0.5cm,height=0.5cm]{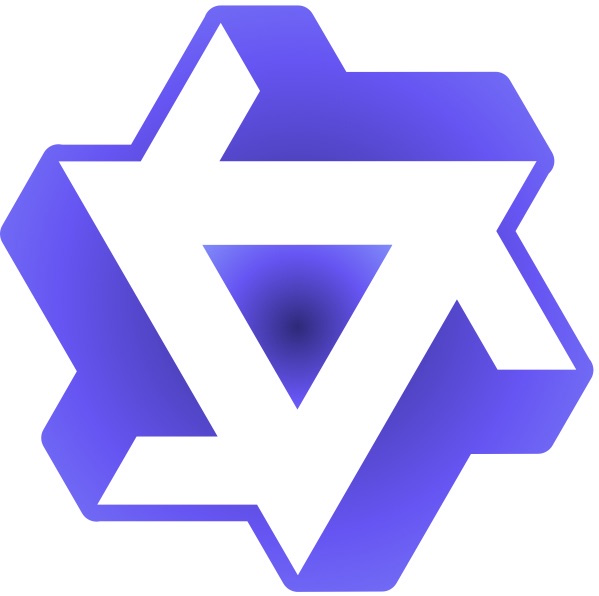}\\
    \scriptsize{235B}
}} &
\includegraphics[width=\linewidth,valign=m]{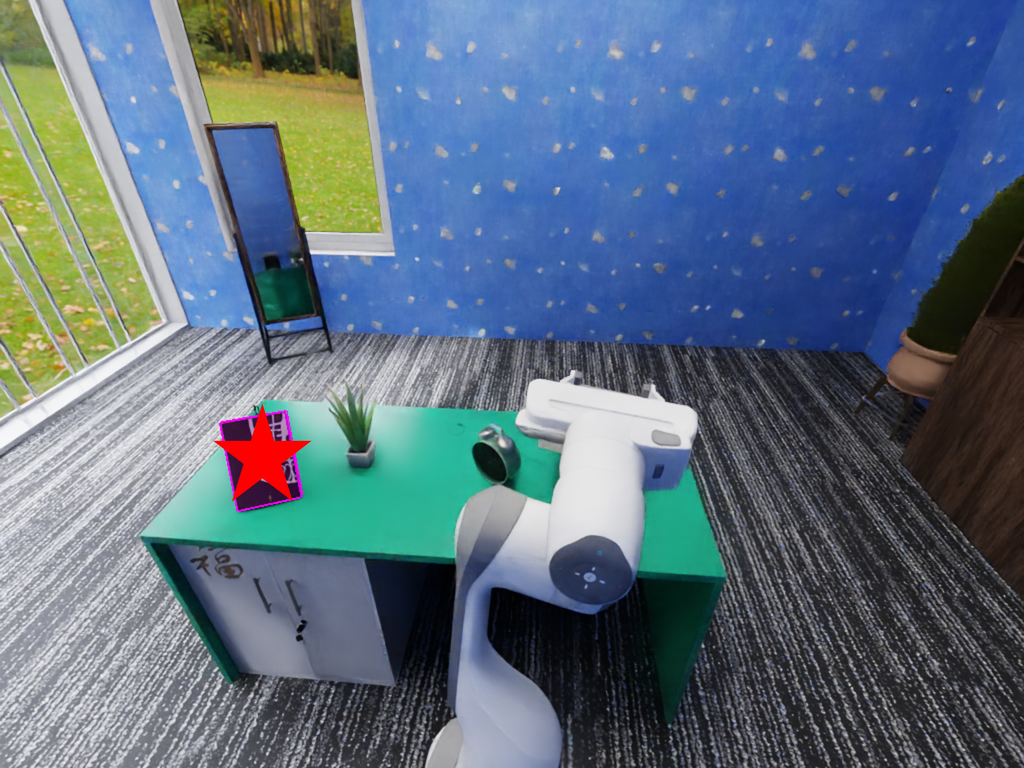} & 
\includegraphics[width=\linewidth,valign=m]{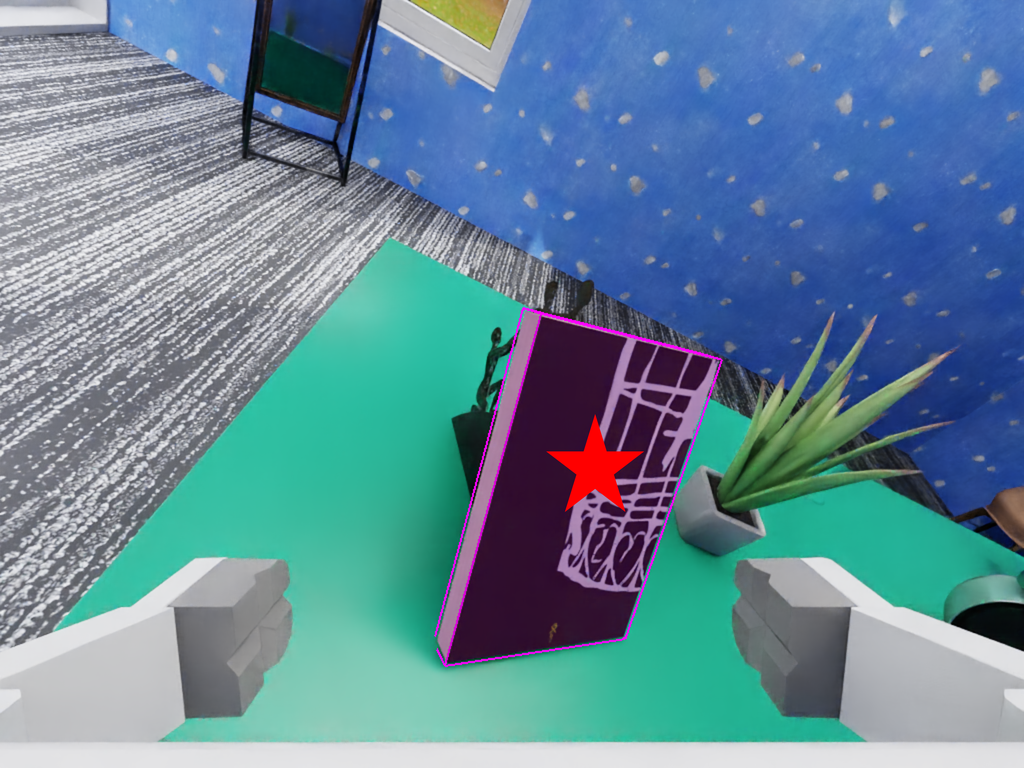} & 
\includegraphics[width=\linewidth,valign=m]{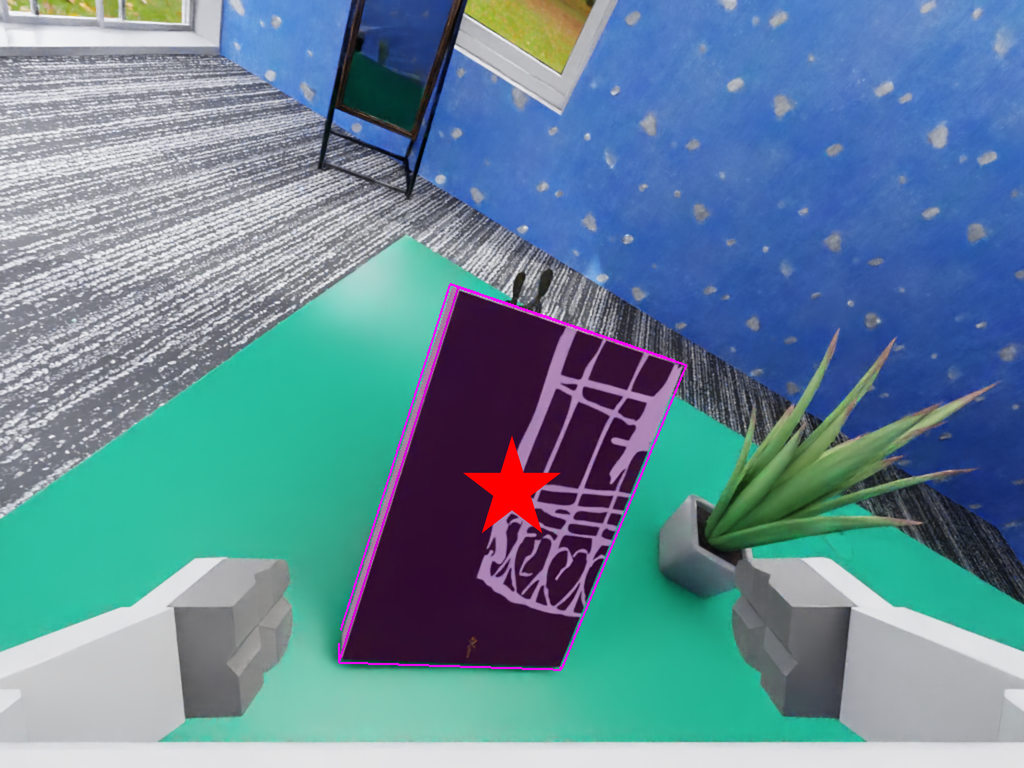} & 
\includegraphics[width=\linewidth,valign=m]{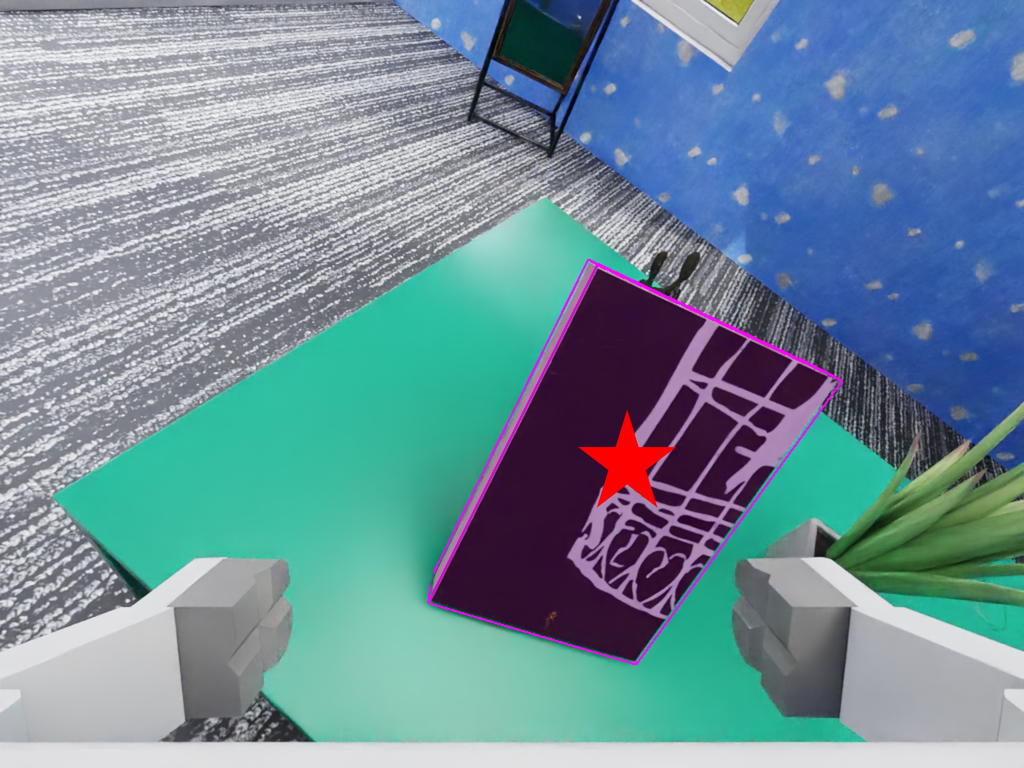} & 
\includegraphics[width=\linewidth,valign=m]{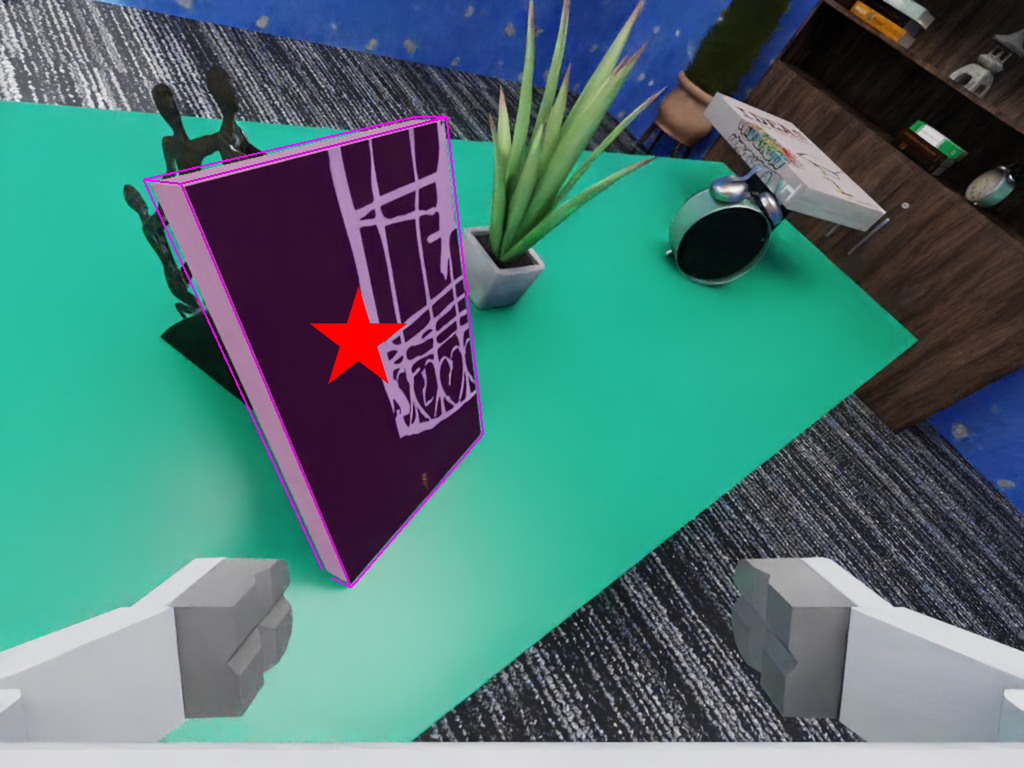} &
\includegraphics[width=\linewidth,valign=m]{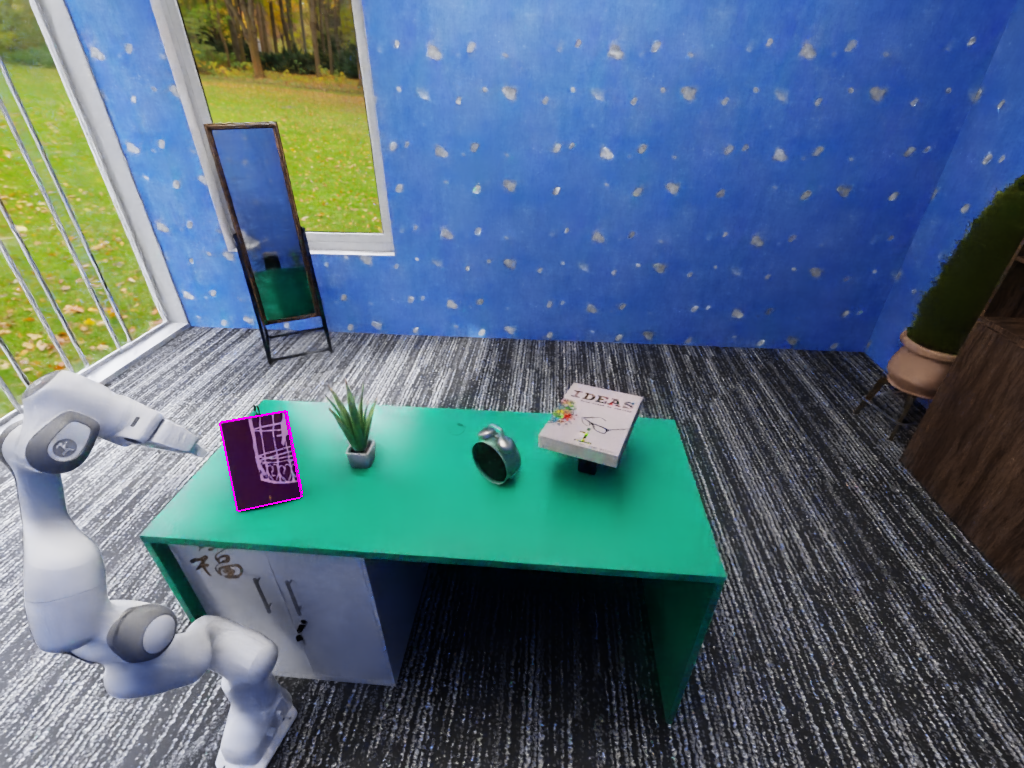} \tabularnewline

& \scriptsize \centering T2 & \scriptsize \centering T2 & \scriptsize \centering T2 & \scriptsize \centering T2 & \scriptsize \centering T2 & \scriptsize \centering Execution \tabularnewline

\midrule

& \multicolumn{6}{l}{\textit{\small{\textbf{Place} the book in the shelf position (row 1, column 5).}}} \tabularnewline

\multirow{2}{*}{\includegraphics[width=0.7cm,height=0.7cm]{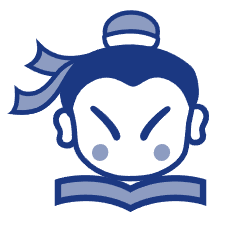}} &
\includegraphics[width=\linewidth,valign=m]{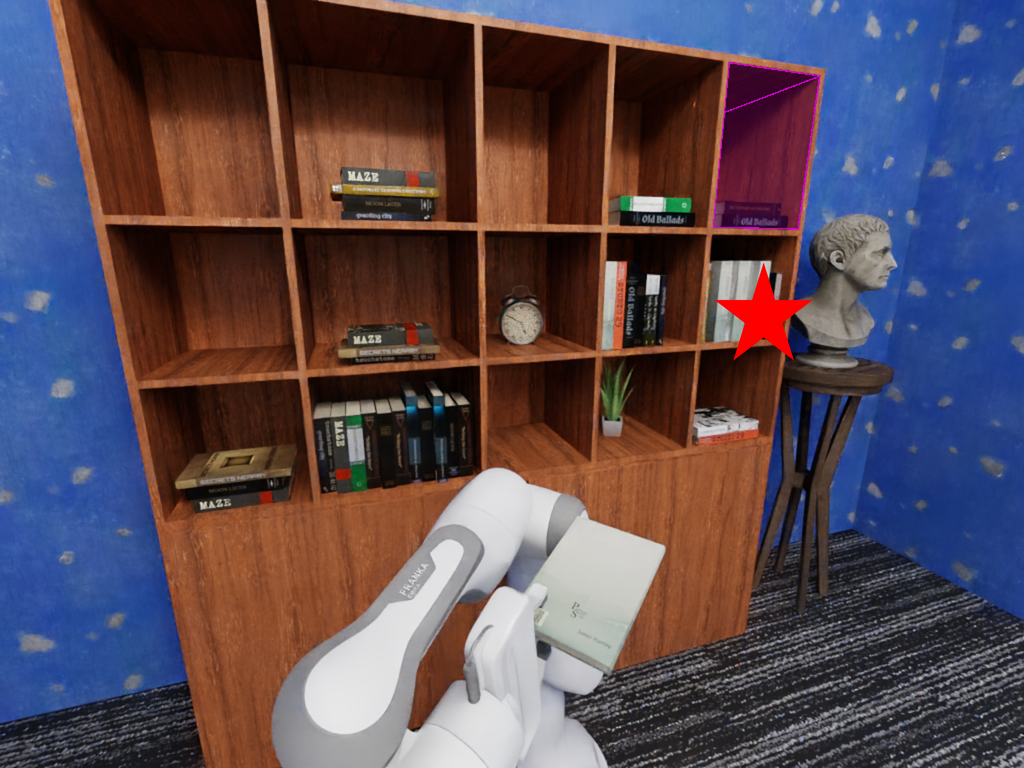} & 
\includegraphics[width=\linewidth,valign=m]{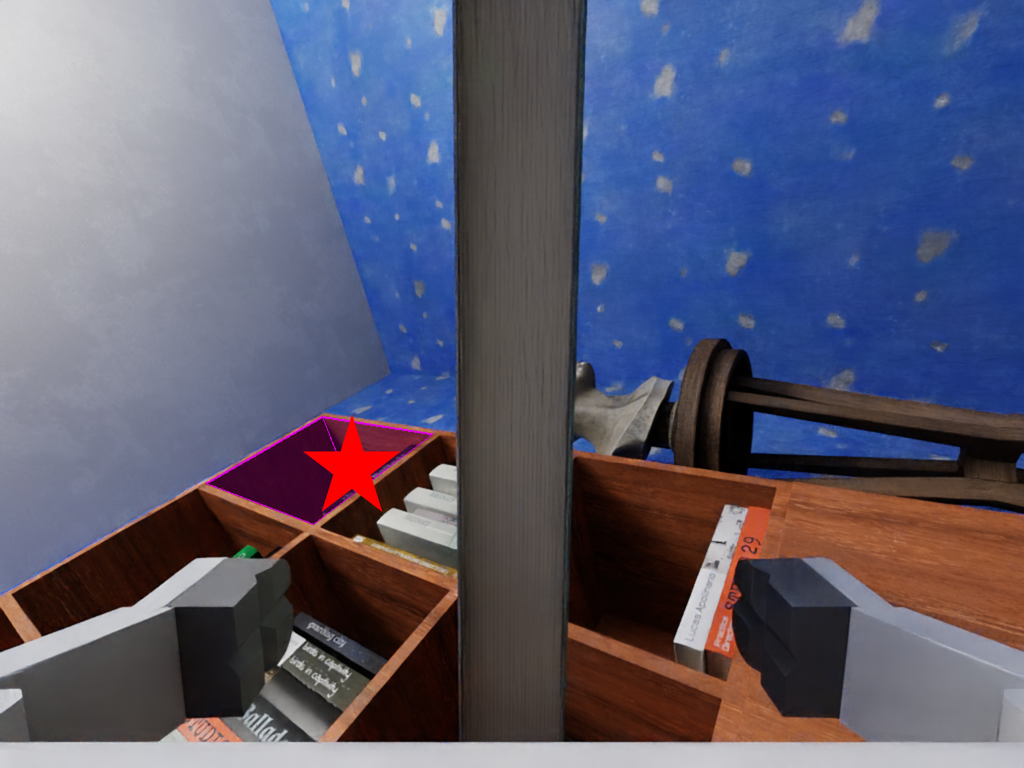} & 
\includegraphics[width=\linewidth,valign=m]{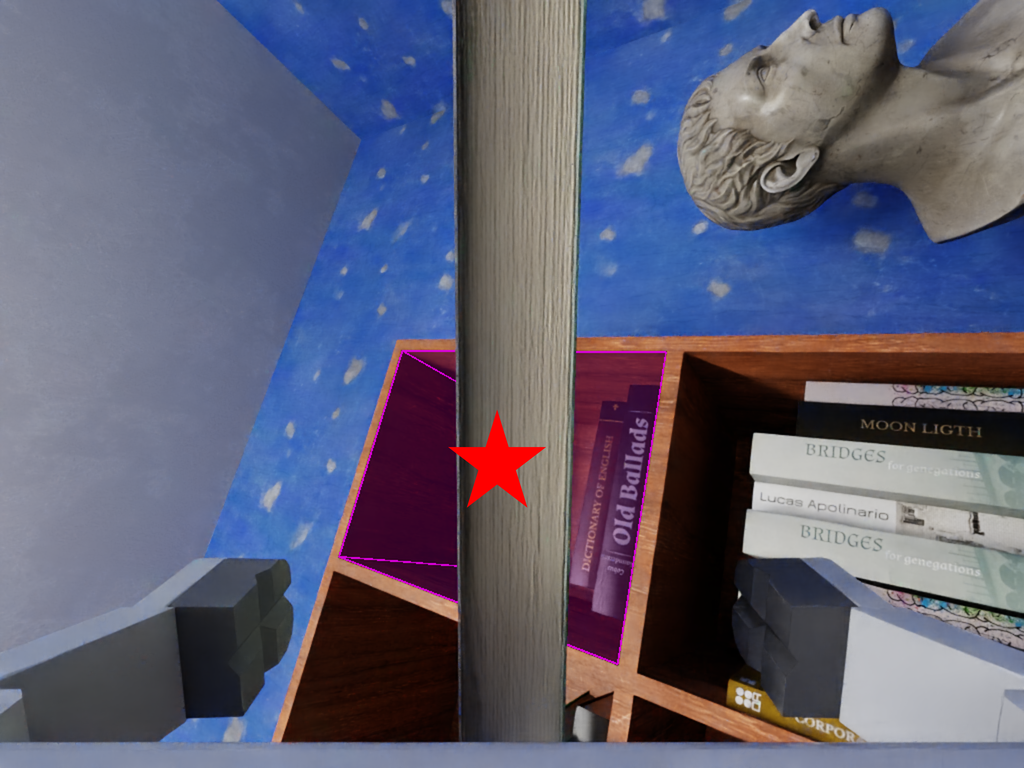} &
\includegraphics[width=\linewidth,valign=m]{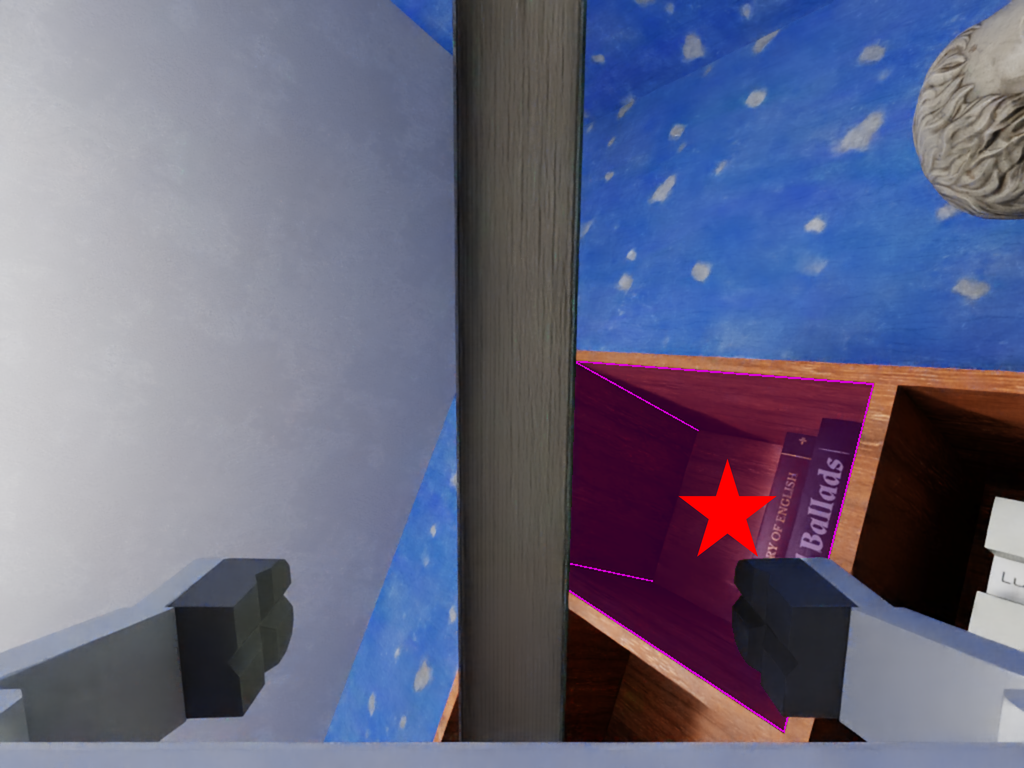} &
\includegraphics[width=\linewidth,valign=m]{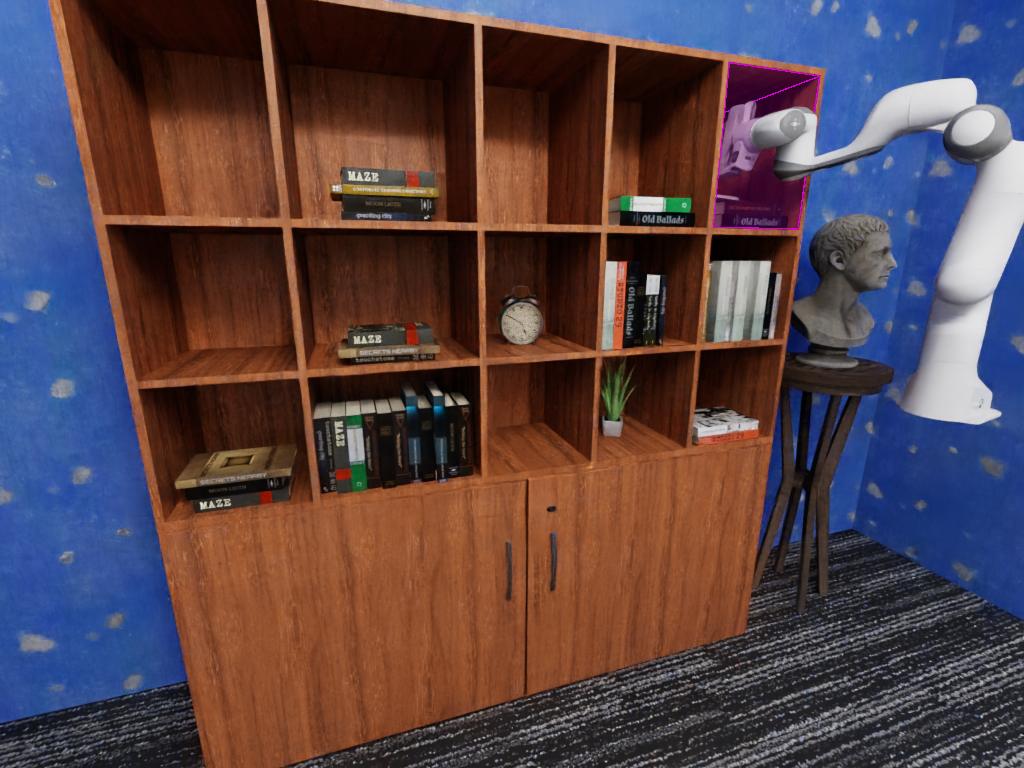} & \tabularnewline

& \scriptsize \centering T5 & \scriptsize \centering T6 & \scriptsize \centering T4 & \scriptsize \centering T6 & \scriptsize \centering Execution & \tabularnewline


\multirow{2}{*}{
    \shortstack{
        \includegraphics[width=0.5cm,height=0.5cm]{figs/quality_examples/qwen_logo.jpeg}\\
        \scriptsize{30B}
    }} &
\includegraphics[width=\linewidth,valign=m]{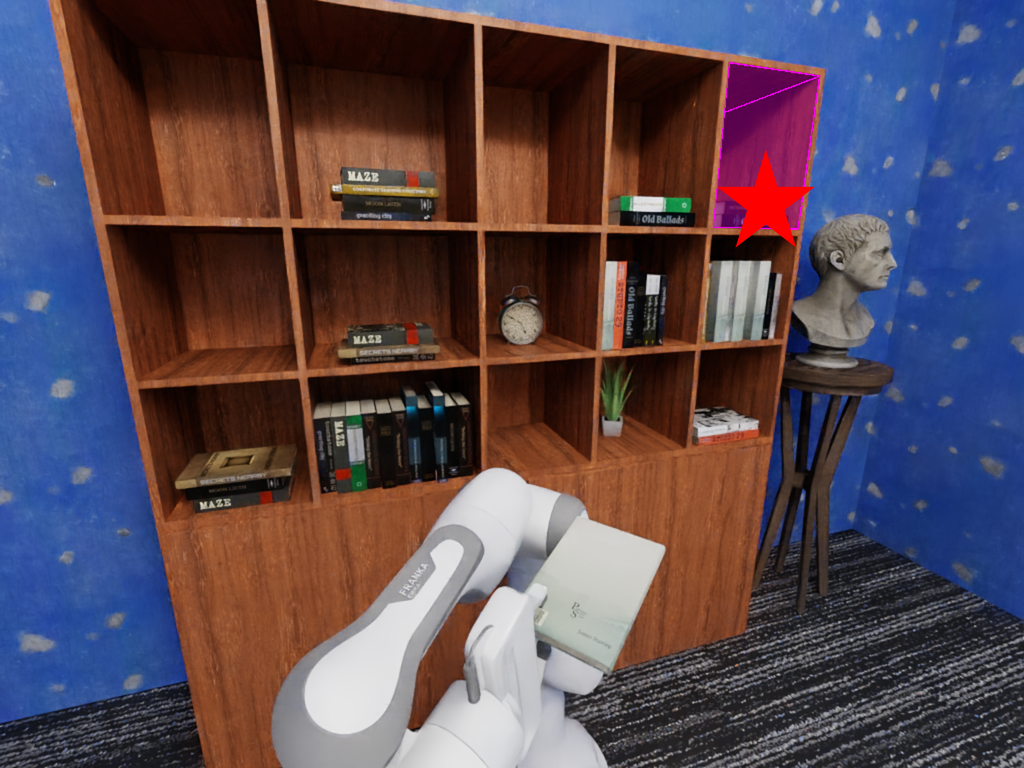} & 
\includegraphics[width=\linewidth,valign=m]{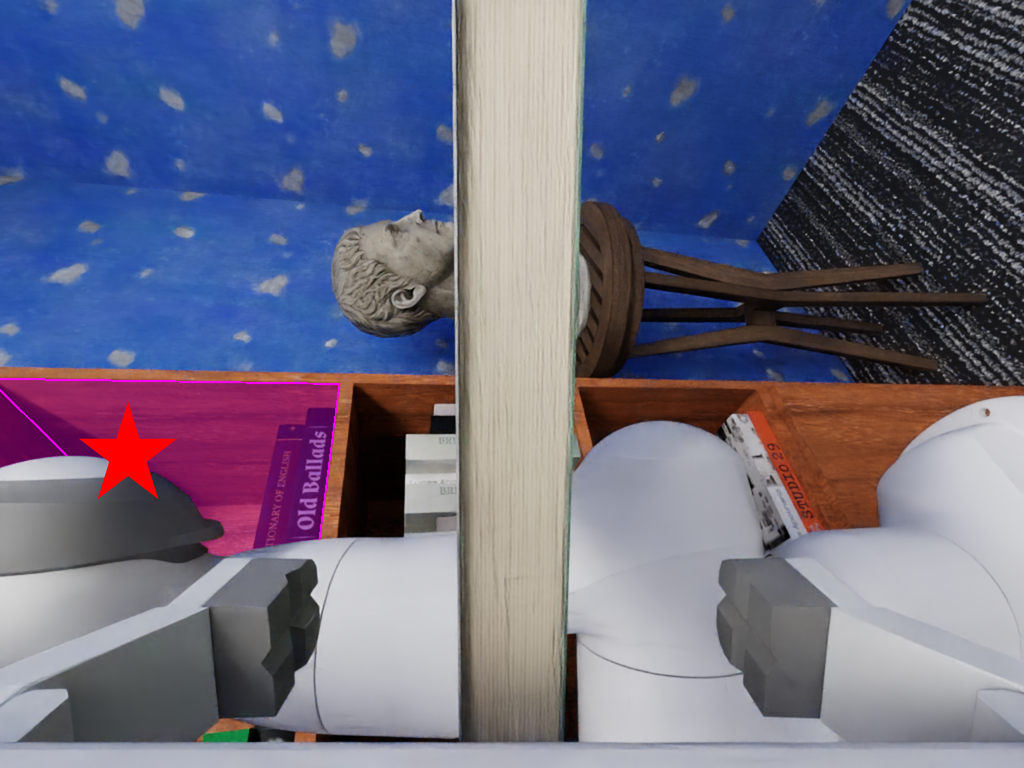} & 
\includegraphics[width=\linewidth,valign=m]{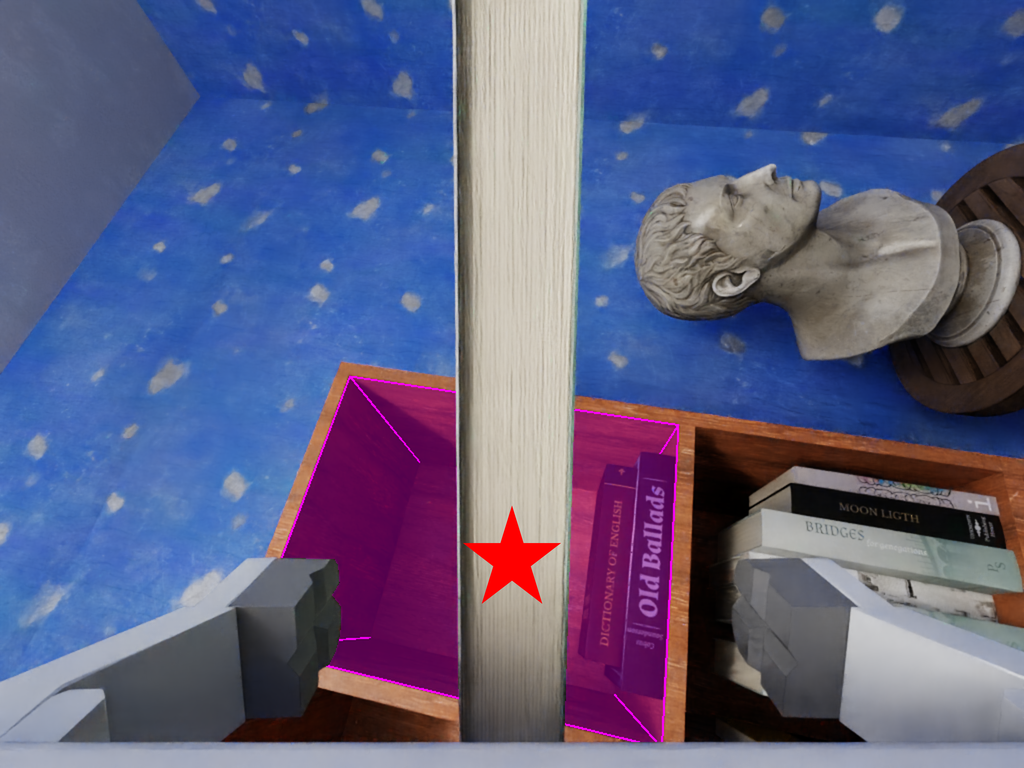} & 
\includegraphics[width=\linewidth,valign=m]{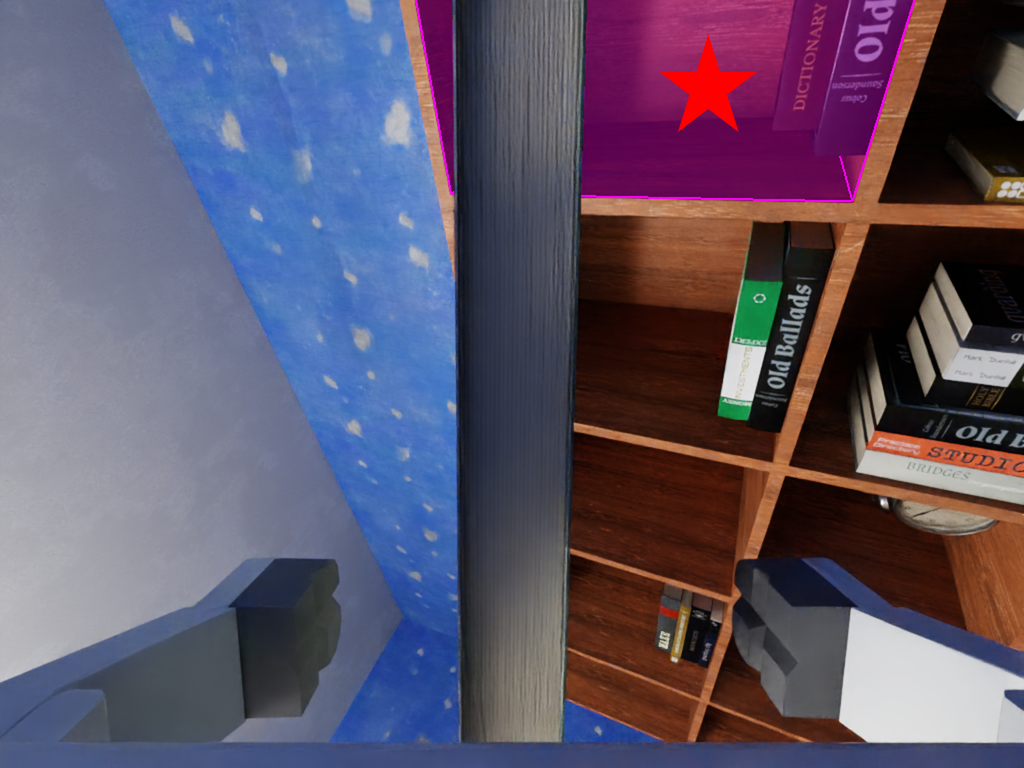} & 
\includegraphics[width=\linewidth,valign=m]{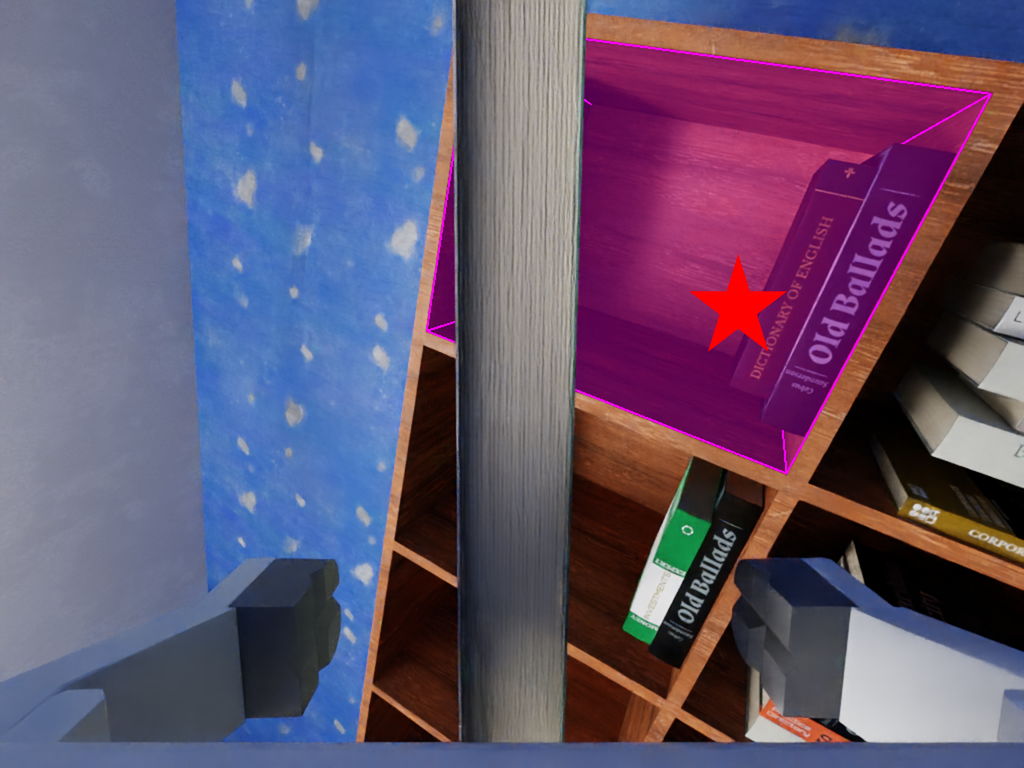} &
\includegraphics[width=\linewidth,valign=m]{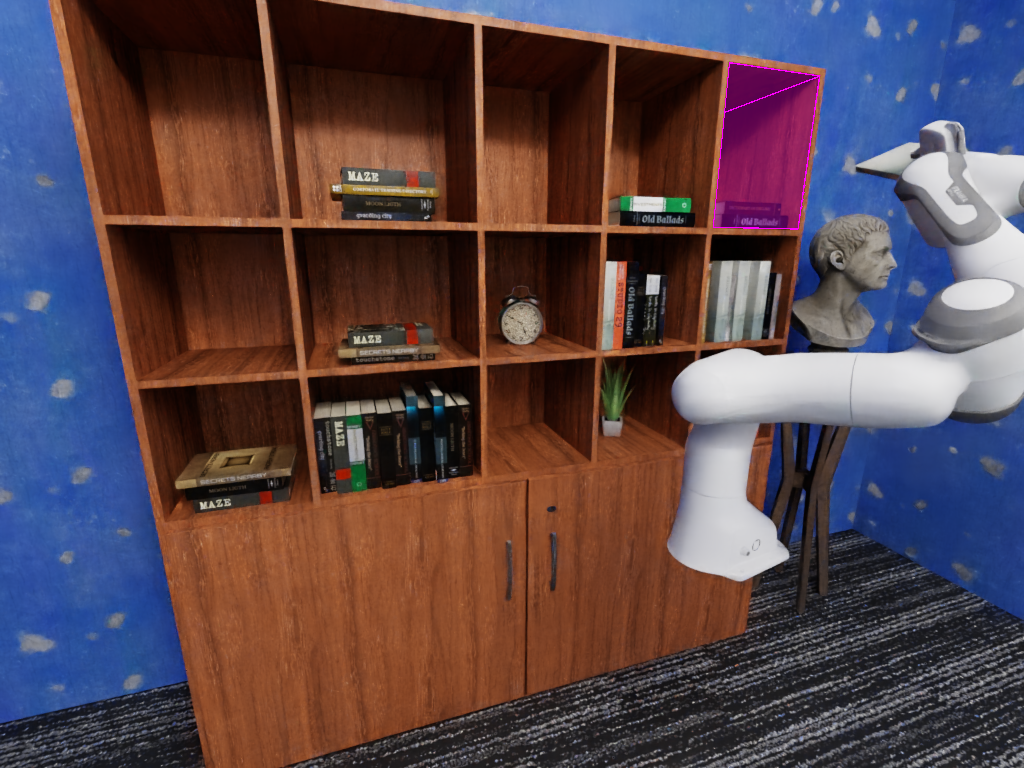} \tabularnewline

& \scriptsize \centering T6 & \scriptsize \centering T3 & \scriptsize \centering T4 & \scriptsize \centering T6 & \scriptsize \centering T6 & \scriptsize \centering Execution \tabularnewline
\bottomrule
\end{tabular}}}}
\end{table}

\subsection{Ablation of Rotation Prediction}

At the core of execution lies the prediction of rotations along the pitch, roll, and yaw axes. This reflects the model’s capability for 3D geometric reasoning and its understanding of object affordances, as the predicted rotations are further composed into a goal pose for execution. To better understand the intrinsic capability of VLMs for rotation prediction, we ablate the set of rotation axes to be predicted using Qwen3-VL-235B-A22B.

\begin{wraptable}{r}{0.4\linewidth}
\vspace{-1em} 
\centering
\begin{minipage}[t]{\linewidth}
\caption{Acceptance rate with rotation axes generated by Qwen3-VL-235B-A22B. Unchecked axes indicate that ground-truth rotations are used. For place tasks, results are reported on tasks with and without explicit pose constraints (C).}
\label{tab:qwenvl-rotation}
\centering
\resizebox{\linewidth}{!}{%
\begin{tabular}{c c c c @{\hspace{1em}} c @{\hspace{1em}} c}
\toprule
\multirow{2}{*}{\footnotesize{Pitch}} & \multirow{2}{*}{\footnotesize{Yaw}} & \multirow{2}{*}{\footnotesize{Roll}} & \multirow{2}{*}{\footnotesize{Pick (\%)}} &
\multicolumn{2}{c}{\footnotesize{Place (\%)}} \\
\cmidrule(l){5-6}
 & & & & \scriptsize{w/o C} & \scriptsize{w/ C} \\
\midrule
 & & & 52.73 & 37.74 & 43.33 \\
\checkmark & & & 20.91 & 26.42  & 26.67 \\
& \checkmark & & 28.18 & 29.25  & 25.00  \\
& & \checkmark & 30.91 & 35.85 & 23.33 \\
\checkmark & \checkmark & & 4.55 & 23.58 & 25.00 \\
\checkmark & & \checkmark & 13.64 & 33.02  & 16.67 \\
& \checkmark & \checkmark & 11.82  & 32.08  & 10.00  \\
\checkmark & \checkmark & \checkmark & 3.64  & 24.53 & 16.00 \\
\bottomrule
\end{tabular}
}
\end{minipage}
\vspace{-1em} 
\end{wraptable}

Specifically, rather than using the ground-truth angles derived from the predicted grasping face of the target book/space, we instruct the VLM to directly generate rotation angles for pitch, yaw, and roll. We randomly sample 110 \emph{pick} and 106 \emph{place} tasks from the test suite for this ablation study. Since most \emph{place} tasks impose no constraints on the final pose, we additionally include 60 \emph{place} tasks with explicit pose constraints (e.g., `\emph{place the book at \underline{a tilt of 60 degrees}.}').

Interestingly, pitch and roll appear to be the key factors for \emph{pick} tasks and constrained \emph{place} tasks, respectively (see Table~\ref{tab:qwenvl-rotation}). We conjecture that \emph{pick} tasks require the model to select a feasible grasping face, which largely depends on the pitch axis, whereas constrained \emph{place} tasks require the model to determine a deviation from the upright direction, primarily governed by the roll axis. As expected, execution becomes harder as more axes need to be predicted. In particular, the pitch-yaw combination adversely affects \emph{pick} the most, while yaw-roll has the largest impact on constrained \emph{place} tasks.

\subsection{Human Study}\label{sec:human-study}

As discussed in Section~\ref{sec:espire-task}, spatial reasoning tasks may exhibit ambiguity when the intended reference frame is not explicitly specified and must be inferred. For example, given the instruction `\emph{grab a book to the left of the picture frame,}' an agent must determine whether to interpret the relation using the intrinsic frame or the relative frame. To investigate to what extent VLMs exhibit frame preferences similar to humans, we use near oriented objects, distant oriented objects, and the table as the reference, construct 91 \emph{pick} tasks involving ambiguous frames, and collect responses from five human participants. Then, we measure human-model agreement by computing the Spearman's rank correlation.

\begin{wraptable}{r}{0.5\linewidth}
\vspace{-1em} 
\centering
\caption{Agreement of humans and models on reference frames. References are categorized into table, near, and distant objects.}
\label{tab:frame-preference}
\resizebox{\linewidth}{!}{
\begin{tabular}{lc  @{\hspace{1em}}  c  @{\hspace{1em}}  c}
\toprule
Model & Near Obj. & Distant Obj. & Table \\
\midrule
Gemini2.5-Pro & -0.573 $\pm$ 0.634 & 0.8 $\pm$ 0.274 & 1.0 $\pm$ 0.0\\
RoboBrain2.0-7B & -0.674 $\pm$ 0.242 &  0.8 $\pm$ 0.274 & 1.0 $\pm$ 0.0 \\
Qwen3-VL-30B-A3B & -0.100 $\pm$ 0.652 &  0.8 $\pm$ 0.274 & 1.0 $\pm$ 0.0\\
Qwen3-VL-8B & -0.674 $\pm$ 0.242 &  0.8 $\pm$ 0.274 & 1.0 $\pm$ 0.0\\
Qwen3-VL-235B-A22B & -0.573 $\pm$ 0.634 &  0.8 $\pm$ 0.274 & 1.0 $\pm$ 0.0\\
\bottomrule
\end{tabular}
}
\vspace{-1em} 
\end{wraptable}

Among the three types of oriented reference objects, humans and models show strong agreement when the table or distant objects are used as references (i.e., w/ high positive correlations above 0.8) but disagree (i.e., w/ negative correlations) when the reference is a near object such as an alarm clock. In such cases, we find that humans tend to prefer the intrinsic frame of the reference object while models favor the relative frame. We hypothesize that when the reference and target objects (i.e., books) are of comparable size, humans perceive the reference as an oriented object with salient geometric cues, making its intrinsic frame more accessible. In contrast, VLMs appear to struggle with object-centric orientation inference and therefore default to the relative frame.

\subsection{Efficiency of \model}
Our analysis regarding the running time of \model reveals two primary sources of latency: API calls and model inference. API response time is largely affected by network stability, whereas the model inference time is determined by the model size and the hardware used for deployment. Taking RoboBrain2.0-7B~\citep{RoboBrain2.0}, when running on an RTX 4090 machine, a single inference takes an average of 9.25 seconds. Another source of latency comes from execution that involves motion planning and environment update. Specifically, the average time for executing a move request is about 18.12 seconds in our experiments on a workstation equipped with an NVIDIA RTX 4090 GPU.

%
\section{Discussion and Future Work}

\model is the first simulated physical environment designed for the diagnostic evaluation of spatial cognition in VLMs, featuring spatial-centric robotic tasks that are explicitly designed to be scalable and diverse.
To evaluate VLMs that cannot directly produce low-level control actions, we have reformulated robotic tasks into localization and execution. While future VLAs are supposed to integrate the two phases, we deliberately prioritize diagnosis over integration. This design choice is further motivated by existing agentic frameworks that decouple reasoning and acting, using VLMs for high-level spatial reasoning and VLAs or controllers for action execution~\citep{GR-ER1.5}. By isolating the reasoning stage, our framework provides a `microscope' to identify where spatial reasoning chains break, offering a concrete roadmap for the specialized spatial inductive biases that future architectures will require.

A limitation of \model is: it is restricted to indoor scenes. Despite our systematic design, it does not cover spatial reasoning scenarios that arise only outdoors, such as reasoning with larger units of measure (e.g., \emph{kilometer}), reasoning with larger-sized reference objects (e.g., \emph{trees}), and reasoning using the global reference frame (e.g., \emph{south} or \emph{east}).
Nonetheless, \model readily supports such extensions, for example, by making outdoor reference objects visible through glass walls.

Beyond that, \model opens several new avenues for the development and analysis of spatially intelligent VLMs.
For example, \model allows for designing long-horizon tasks that require multi-step spatial reasoning, leading to many interesting model analyses, including the modeling of dependencies between reasoning steps and the role of memory in long-horizon spatial reasoning. Moreover, since `pick' and `place' tasks typically occur sequentially in robotics, but are performed in different workspaces in \model, it is well-suited for extending it to evaluate mobile manipulation.

\section{Conclusion}

We have presented \model, a simulated environment that provides an evaluation suite for embodied spatial reasoning with vision-language models. \model evaluates VLMs on robotic tasks in a physically grounded setting, thus mitigating the gap between evaluation and practical deployment. By breaking down each task into localization and execution, \model provides a unified evaluation of passive spatial reasoning and action-oriented spatial reasoning. We systematically design \model to simulate a diverse range of spatial reasoning scenarios, enabling a comprehensive analysis across spatial aspects and at multiple levels of granularity. Our experimental results and analysis reveal future directions for enhancing VLMs in spatial reasoning.

\section{Acknowledgements}
Yanpeng Zhao acknowledges the support of the National Natural Science Foundation of China (12574467).

{\small
  \bibliographystyle{iclr2026_conference}
  \bibliography{11_references}

@inproceedings{Alfred,
  author    = {Shridhar, Mohit and Thomason, Jesse and Gordon, Daniel and Bisk, Yonatan and Han, Winson and Mottaghi, Roozbeh and Zettlemoyer, Luke and Fox, Dieter},
  booktitle = {2020 IEEE/CVF Conference on Computer Vision and Pattern Recognition (CVPR)},
  doi       = {10.1109/CVPR42600.2020.01075},
  number    = {},
  pages     = {10737-10746},
  title     = {ALFRED: A Benchmark for Interpreting Grounded Instructions for Everyday Tasks},
  volume    = {},
  year      = {2020}
}

@inproceedings{Behavior,
  author    = {Srivastava, Sanjana and Li, Chengshu and Lingelbach, Michael and Mart\'in-Mart\'in, Roberto and Xia, Fei and Vainio, Kent Elliott and Lian, Zheng and Gokmen, Cem and Buch, Shyamal and Liu, Karen and Savarese, Silvio and Gweon, Hyowon and Wu, Jiajun and Fei-Fei, Li},
  booktitle = {Proceedings of the 5th Conference on Robot Learning},
  editor    = {Faust, Aleksandra and Hsu, David and Neumann, Gerhard},
  month     = {08--11 Nov},
  pages     = {477--490},
  publisher = {PMLR},
  series    = {Proceedings of Machine Learning Research},
  title     = {BEHAVIOR: Benchmark for Everyday Household Activities in Virtual, Interactive, and Ecological Environments},
  url       = {https://proceedings.mlr.press/v164/srivastava22a.html},
  volume    = {164},
  year      = {2022}
}

@inproceedings{Blink,
  address   = {Berlin, Heidelberg},
  author    = {Fu, Xingyu and Hu, Yushi and Li, Bangzheng and Feng, Yu and Wang, Haoyu and Lin, Xudong and Roth, Dan and Smith, Noah A. and Ma, Wei-Chiu and Krishna, Ranjay},
  booktitle = {Computer Vision – ECCV 2024: 18th European Conference, Milan, Italy, September 29–October 4, 2024, Proceedings, Part XXIII},
  doi       = {10.1007/978-3-031-73337-6_9},
  isbn      = {978-3-031-73336-9},
  location  = {Milan, Italy},
  numpages  = {19},
  pages     = {148–166},
  publisher = {Springer-Verlag},
  title     = {BLINK: Multimodal Large Language Models Can See but Not Perceive},
  url       = {https://doi.org/10.1007/978-3-031-73337-6_9},
  year      = {2024}
}

@article{Calvin,
  author   = {Mees, Oier and Hermann, Lukas and Rosete-Beas, Erick and Burgard, Wolfram},
  doi      = {10.1109/LRA.2022.3180108},
  journal  = {IEEE Robotics and Automation Letters},
  number   = {3},
  pages    = {7327-7334},
  title    = {CALVIN: A Benchmark for Language-Conditioned Policy Learning for Long-Horizon Robot Manipulation Tasks},
  volume   = {7},
  year     = {2022}
}

@inproceedings{CaP,
  author    = {Liang, Jacky and Huang, Wenlong and Xia, Fei and Xu, Peng and Hausman, Karol and Ichter, Brian and Florence, Pete and Zeng, Andy},
  booktitle = {2023 IEEE International Conference on Robotics and Automation (ICRA)},
  doi       = {10.1109/ICRA48891.2023.10160591},
  number    = {},
  pages     = {9493-9500},
  title     = {Code as Policies: Language Model Programs for Embodied Control},
  volume    = {},
  year      = {2023}
}

@inproceedings{CLEVR,
  author    = {Johnson, Justin and Hariharan, Bharath and van der Maaten, Laurens and Fei-Fei, Li and Lawrence Zitnick, C. and Girshick, Ross},
  booktitle = {Proceedings of the IEEE Conference on Computer Vision and Pattern Recognition (CVPR)},
  month     = {July},
  title     = {CLEVR: A Diagnostic Dataset for Compositional Language and Elementary Visual Reasoning},
  year      = {2017}
}

@article{COME,
  author   = {Peiyuan Zhi and Zhiyuan Zhang and Muzhi Han and Zeyu Zhang and Zhitian Li and Ziyuan Jiao and Baoxiong Jia and Siyuan Huang},
  cdate    = {1704067200000},
  journal  = {CoRR},
  publtype = {informal},
  title    = {Closed-Loop Open-Vocabulary Mobile Manipulation with GPT-4V},
  url      = {https://doi.org/10.48550/arXiv.2404.10220},
  volume   = {abs/2404.10220},
  year     = {2024}
}

@inproceedings{CoPa,
  author    = {Huang, Haoxu and Lin, Fanqi and Hu, Yingdong and Wang, Shengjie and Gao, Yang},
  booktitle = {2024 IEEE/RSJ International Conference on Intelligent Robots and Systems (IROS)},
  doi       = {10.1109/IROS58592.2024.10801352},
  number    = {},
  pages     = {9488-9495},
  title     = {CoPa: General Robotic Manipulation through Spatial Constraints of Parts with Foundation Models},
  volume    = {},
  year      = {2024}
}

@inproceedings{cuRobo,
  author    = {Sundaralingam, Balakumar and Hari, Siva Kumar Sastry and Fishman, Adam and Garrett, Caelan and Van Wyk, Karl and Blukis, Valts and Millane, Alexander and Oleynikova, Helen and Handa, Ankur and Ramos, Fabio and Ratliff, Nathan and Fox, Dieter},
  booktitle = {2023 IEEE International Conference on Robotics and Automation (ICRA)},
  doi       = {10.1109/ICRA48891.2023.10160765},
  number    = {},
  pages     = {8112-8119},
  title     = {CuRobo: Parallelized Collision-Free Robot Motion Generation},
  volume    = {},
  year      = {2023}
}

@inproceedings{CV-Bench,
  author    = {Tong, Shengbang and Brown, Ellis and Wu, Penghao and Woo, Sanghyun and Middepogu, Manoj and Akula, Sai Charitha and Yang, Jihan and Yang, Shusheng and Iyer, Adithya and Pan, Xichen and Wang, Austin and Fergus, Rob and LeCun, Yann and Xie, Saining},
  booktitle = {Advances in Neural Information Processing Systems},
  editor    = {A. Globerson and L. Mackey and D. Belgrave and A. Fan and U. Paquet and J. Tomczak and C. Zhang},
  pages     = {87310--87356},
  publisher = {Curran Associates, Inc.},
  title     = {Cambrian-1: A Fully Open, Vision-Centric Exploration of Multimodal LLMs},
  url       = {https://proceedings.neurips.cc/paper_files/paper/2024/file/9ee3a664ccfeabc0da16ac6f1f1cfe59-Paper-Conference.pdf},
  volume    = {37},
  year      = {2024}
}

@inproceedings{Dream2Real,
  author    = {Ivan Kapelyukh and Yifei Ren and Ignacio Alzugaray and Edward Johns},
  booktitle = {First Workshop on Vision-Language Models for Navigation and Manipulation at ICRA 2024},
  title     = {Dream2Real: Zero-Shot 3D Object Rearrangement with Vision-Language Models},
  url       = {https://openreview.net/forum?id=o29sRo5TdE},
  year      = {2024}
}

@inproceedings{EMBench,
  author    = {Yang, Rui and Chen, Hanyang and Zhang, Junyu and Zhao, Mark and Qian, Cheng and Wang, Kangrui and Wang, Qineng and Koripella, Teja Venkat and Movahedi, Marziyeh and Li, Manling and Ji, Heng and Zhang, Huan and Zhang, Tong},
  booktitle = {Proceedings of the 42nd International Conference on Machine Learning},
  editor    = {Singh, Aarti and Fazel, Maryam and Hsu, Daniel and Lacoste-Julien, Simon and Berkenkamp, Felix and Maharaj, Tegan and Wagstaff, Kiri and Zhu, Jerry},
  month     = {13--19 Jul},
  pages     = {70576--70631},
  publisher = {PMLR},
  series    = {Proceedings of Machine Learning Research},
  title     = {{E}mbodied{B}ench: Comprehensive Benchmarking Multi-modal Large Language Models for Vision-Driven Embodied Agents},
  url       = {https://proceedings.mlr.press/v267/yang25f.html},
  volume    = {267},
  year      = {2025}
}

@misc{Embodied-R,
  archiveprefix = {arXiv},
  author        = {Baining Zhao and Ziyou Wang and Jianjie Fang and Chen Gao and Fanhang Man and Jinqiang Cui and Xin Wang and Xinlei Chen and Yong Li and Wenwu Zhu},
  eprint        = {2504.12680},
  primaryclass  = {cs.AI},
  title         = {Embodied-R: Collaborative Framework for Activating Embodied Spatial Reasoning in Foundation Models via Reinforcement Learning},
  url           = {https://arxiv.org/abs/2504.12680},
  year          = {2025}
}

@misc{Gemini2.5,
  archiveprefix = {arXiv},
  author        = {Gemini 2.5 Team
                   and Gheorghe Comanici and Eric Bieber and Mike Schaekermann and Ice Pasupat and Noveen Sachdeva and Inderjit Dhillon and Marcel Blistein and Ori Ram and Dan Zhang and Others      
                   },
  eprint        = {2507.06261},
  primaryclass  = {cs.CL},
  title         = {Gemini 2.5: Pushing the Frontier with Advanced Reasoning, Multimodality, Long Context, and Next Generation Agentic Capabilities},
  url           = {https://arxiv.org/abs/2507.06261},
  year          = {2025}
}

@inproceedings{Gibson,
  author    = {Xia, Fei and Zamir, Amir R. and He, Zhiyang and Sax, Alexander and Malik, Jitendra and Savarese, Silvio},
  booktitle = {2018 IEEE/CVF Conference on Computer Vision and Pattern Recognition},
  doi       = {10.1109/CVPR.2018.00945},
  number    = {},
  pages     = {9068-9079},
  title     = {Gibson Env: Real-World Perception for Embodied Agents},
  volume    = {},
  year      = {2018}
}

@misc{GR-ER1.5,
  author = {Gemini-Robotics-Team      
            },
  title  = {Gemini Robotics 1.5: Pushing the Frontier of Generalist Robots with Advanced Embodied Reasoning, Thinking, and Motion Transfer},
  url    = {https://deepmind.google/discover/blog/gemini-robotics-15-brings-ai-agents-into-the-physical-world/},
  year   = {2025}
}

@inproceedings{Habitat2,
  address   = {Red Hook, NY, USA},
  articleno = {20},
  author    = {Szot, Andrew and Clegg, Alex and Undersander, Eric and Wijmans, Erik and Zhao, Yili and Turner, John and Maestre, Noah and Mukadam, Mustafa and Chaplot, Devendra and Maksymets, Oleksandr and Gokaslan, Aaron and Vondrus, Vladimir and Dharur, Sameer and Meier, Franziska and Galuba, Wojciech and Chang, Angel and Kira, Zsolt and Koltun, Vladlen and Malik, Jitendra and Savva, Manolis and Batra, Dhruv},
  booktitle = {Proceedings of the 35th International Conference on Neural Information Processing Systems},
  isbn      = {9781713845393},
  numpages  = {16},
  publisher = {Curran Associates Inc.},
  series    = {NIPS '21},
  title     = {Habitat 2.0: training home assistants to rearrange their habitat},
  year      = {2021}
}

@misc{InternVL3,
  archiveprefix = {arXiv},
  author        = {Jinguo Zhu and Weiyun Wang and Zhe Chen and Zhaoyang Liu and Shenglong Ye and Lixin Gu and Hao Tian and Yuchen Duan and Weijie Su and Jie Shao and Zhangwei Gao and Erfei Cui and Xuehui Wang and Yue Cao and Yangzhou Liu and Xingguang Wei and Hongjie Zhang and Haomin Wang and Weiye Xu and Hao Li and Jiahao Wang and Nianchen Deng and Songze Li and Yinan He and Tan Jiang and Jiapeng Luo and Yi Wang and Conghui He and Botian Shi and Xingcheng Zhang and Wenqi Shao and Junjun He and Yingtong Xiong and Wenwen Qu and Peng Sun and Penglong Jiao and Han Lv and Lijun Wu and Kaipeng Zhang and Huipeng Deng and Jiaye Ge and Kai Chen and Limin Wang and Min Dou and Lewei Lu and Xizhou Zhu and Tong Lu and Dahua Lin and Yu Qiao and Jifeng Dai and Wenhai Wang},
  eprint        = {2504.10479},
  primaryclass  = {cs.CV},
  title         = {InternVL3: Exploring Advanced Training and Test-Time Recipes for Open-Source Multimodal Models},
  url           = {https://arxiv.org/abs/2504.10479},
  year          = {2025}
}

@misc{IsaacSim,
  author  = {{NVIDIA}},
  license = {Apache-2.0},
  title   = {{Isaac Sim}},
  url     = {https://github.com/isaac-sim/IsaacSim},
  version = {4.5.0},
  year    = {2025}
}

@inproceedings{LAPA,
  author    = {Seonghyeon Ye and Joel Jang and Byeongguk Jeon and Se June Joo and Jianwei Yang and Baolin Peng and Ajay Mandlekar and Reuben Tan and Yu-Wei Chao and Bill Yuchen Lin and Lars Liden and Kimin Lee and Jianfeng Gao and Luke Zettlemoyer and Dieter Fox and Minjoon Seo},
  booktitle = {The Thirteenth International Conference on Learning Representations},
  title     = {Latent Action Pretraining from Videos},
  url       = {https://openreview.net/forum?id=VYOe2eBQeh},
  year      = {2025}
}

@inproceedings{Libero,
  author    = {Bo Liu and Yifeng Zhu and Chongkai Gao and Yihao Feng and qiang liu and Yuke Zhu and Peter Stone},
  booktitle = {Thirty-seventh Conference on Neural Information Processing Systems Datasets and Benchmarks Track},
  title     = {{LIBERO}: Benchmarking Knowledge Transfer for Lifelong Robot Learning},
  url       = {https://openreview.net/forum?id=xzEtNSuDJk},
  year      = {2023}
}

@inproceedings{LLaVa,
  author    = {Haotian Liu and Chunyuan Li and Qingyang Wu and Yong Jae Lee},
  booktitle = {Thirty-seventh Conference on Neural Information Processing Systems},
  title     = {Visual Instruction Tuning},
  url       = {https://openreview.net/forum?id=w0H2xGHlkw},
  year      = {2023}
}

@inproceedings{ManiSkill2,
  author    = {Jiayuan Gu and Fanbo Xiang and Xuanlin Li and Zhan Ling and Xiqiang Liu and Tongzhou Mu and Yihe Tang and Stone Tao and Xinyue Wei and Yunchao Yao and Xiaodi Yuan and Pengwei Xie and Zhiao Huang and Rui Chen and Hao Su},
  booktitle = {The Eleventh International Conference on Learning Representations },
  title     = {ManiSkill2: A Unified Benchmark for Generalizable Manipulation Skills},
  url       = {https://openreview.net/forum?id=b_CQDy9vrD1},
  year      = {2023}
}

@inproceedings{MetaWorld,
  author    = {Yu, Tianhe and Quillen, Deirdre and He, Zhanpeng and Julian, Ryan and Hausman, Karol and Finn, Chelsea and Levine, Sergey},
  booktitle = {Proceedings of the Conference on Robot Learning},
  editor    = {Kaelbling, Leslie Pack and Kragic, Danica and Sugiura, Komei},
  month     = {30 Oct--01 Nov},
  pages     = {1094--1100},
  publisher = {PMLR},
  series    = {Proceedings of Machine Learning Research},
  title     = {Meta-World: A Benchmark and Evaluation for Multi-Task and Meta Reinforcement Learning},
  url       = {https://proceedings.mlr.press/v100/yu20a.html},
  volume    = {100},
  year      = {2020}
}

@inproceedings{MuJoCo,
  author    = {Todorov, Emanuel and Erez, Tom and Tassa, Yuval},
  booktitle = {2012 IEEE/RSJ International Conference on Intelligent Robots and Systems},
  doi       = {10.1109/IROS.2012.6386109},
  number    = {},
  pages     = {5026-5033},
  title     = {MuJoCo: A physics engine for model-based control},
  volume    = {},
  year      = {2012}
}

@inproceedings{Octo,
  author    = {Oier Mees and Dibya Ghosh and Karl Pertsch and Kevin Black and Homer Rich Walke and Sudeep Dasari and Joey Hejna and Tobias Kreiman and Charles Xu and Jianlan Luo and You Liang Tan and Dorsa Sadigh and Chelsea Finn and Sergey Levine},
  booktitle = {First Workshop on Vision-Language Models for Navigation and Manipulation at ICRA 2024},
  title     = {Octo: An Open-Source Generalist Robot Policy},
  url       = {https://openreview.net/forum?id=jGrtIvJBpS},
  year      = {2024}
}

@misc{Open3DVQA,
  archiveprefix = {arXiv},
  author        = {Weichen Zhang and Zile Zhou and Zhiheng Zheng and Chen Gao and Jinqiang Cui and Yong Li and Xinlei Chen and Xiao-Ping Zhang},
  eprint        = {2503.11094},
  primaryclass  = {cs.CV},
  title         = {Open3DVQA: A Benchmark for Comprehensive Spatial Reasoning with Multimodal Large Language Model in Open Space},
  url           = {https://arxiv.org/abs/2503.11094},
  year          = {2025}
}

@inproceedings{Open6DOR,
  author    = {Ding, Yufei and Geng, Haoran and Xu, Chaoyi and Fang, Xiaomeng and Zhang, Jiazhao and Wei, Songlin and Dai, Qiyu and Zhang, Zhizheng and Wang, He},
  booktitle = {2024 IEEE/RSJ International Conference on Intelligent Robots and Systems (IROS)},
  doi       = {10.1109/IROS58592.2024.10802733},
  number    = {},
  pages     = {7359-7366},
  title     = {Open6DOR: Benchmarking Open-instruction 6-DoF Object Rearrangement and A VLM-based Approach},
  volume    = {},
  year      = {2024}
}

@inproceedings{PaLM-E,
  author    = {Driess, Danny and Xia, Fei and Sajjadi, Mehdi S. M. and Lynch, Corey and Chowdhery, Aakanksha and Ichter, Brian and Wahid, Ayzaan and Tompson, Jonathan and Vuong, Quan and Yu, Tianhe and Huang, Wenlong and Chebotar, Yevgen and Sermanet, Pierre and Duckworth, Daniel and Levine, Sergey and Vanhoucke, Vincent and Hausman, Karol and Toussaint, Marc and Greff, Klaus and Zeng, Andy and Mordatch, Igor and Florence, Pete},
  booktitle = {Proceedings of the 40th International Conference on Machine Learning},
  editor    = {Krause, Andreas and Brunskill, Emma and Cho, Kyunghyun and Engelhardt, Barbara and Sabato, Sivan and Scarlett, Jonathan},
  month     = {23--29 Jul},
  pages     = {8469--8488},
  publisher = {PMLR},
  series    = {Proceedings of Machine Learning Research},
  title     = {{P}a{LM}-E: An Embodied Multimodal Language Model},
  url       = {https://proceedings.mlr.press/v202/driess23a.html},
  volume    = {202},
  year      = {2023}
}

@misc{Pi0,
  archiveprefix = {arXiv},
  author        = {Kevin Black and Noah Brown and Danny Driess and Adnan Esmail and Michael Equi and Chelsea Finn and Niccolo Fusai and Lachy Groom and Karol Hausman and Brian Ichter and Szymon Jakubczak and Tim Jones and Liyiming Ke and Sergey Levine and Adrian Li-Bell and Mohith Mothukuri and Suraj Nair and Karl Pertsch and Lucy Xiaoyang Shi and James Tanner and Quan Vuong and Anna Walling and Haohuan Wang and Ury Zhilinsky},
  eprint        = {2410.24164},
  primaryclass  = {cs.LG},
  title         = {$\pi_0$: A Vision-Language-Action Flow Model for General Robot Control},
  url           = {https://arxiv.org/abs/2410.24164},
  year          = {2024}
}

@misc{PointArena,
  archiveprefix = {arXiv},
  author        = {Long Cheng and Jiafei Duan and Yi Ru Wang and Haoquan Fang and Boyang Li and Yushan Huang and Elvis Wang and Ainaz Eftekhar and Jason Lee and Wentao Yuan and Rose Hendrix and Noah A. Smith and Fei Xia and Dieter Fox and Ranjay Krishna},
  eprint        = {2505.09990},
  primaryclass  = {cs.CV},
  title         = {PointArena: Probing Multimodal Grounding Through Language-Guided Pointing},
  url           = {https://arxiv.org/abs/2505.09990},
  year          = {2025}
}

@misc{QWen25vl,
  archiveprefix = {arXiv},
  author        = {Shuai Bai and Keqin Chen and Xuejing Liu and Jialin Wang and Wenbin Ge and Sibo Song and Kai Dang and Peng Wang and Shijie Wang and Jun Tang and Humen Zhong and Yuanzhi Zhu and Mingkun Yang and Zhaohai Li and Jianqiang Wan and Pengfei Wang and Wei Ding and Zheren Fu and Yiheng Xu and Jiabo Ye and Xi Zhang and Tianbao Xie and Zesen Cheng and Hang Zhang and Zhibo Yang and Haiyang Xu and Junyang Lin},
  eprint        = {2502.13923},
  primaryclass  = {cs.CV},
  title         = {Qwen2.5-VL Technical Report},
  url           = {https://arxiv.org/abs/2502.13923},
  year          = {2025}
}

@inproceedings{Ravens,
  author    = {Zeng, Andy and Florence, Pete and Tompson, Jonathan and Welker, Stefan and Chien, Jonathan and Attarian, Maria and Armstrong, Travis and Krasin, Ivan and Duong, Dan and Sindhwani, Vikas and Lee, Johnny},
  booktitle = {Proceedings of the 2020 Conference on Robot Learning},
  editor    = {Kober, Jens and Ramos, Fabio and Tomlin, Claire},
  month     = {16--18 Nov},
  pages     = {726--747},
  publisher = {PMLR},
  series    = {Proceedings of Machine Learning Research},
  title     = {Transporter Networks: Rearranging the Visual World for Robotic Manipulation},
  url       = {https://proceedings.mlr.press/v155/zeng21a.html},
  volume    = {155},
  year      = {2021}
}

@inproceedings{ReKep,
  author    = {Wenlong Huang and Chen Wang and Yunzhu Li and Ruohan Zhang and Li Fei-Fei},
  booktitle = {8th Annual Conference on Robot Learning},
  title     = {ReKep: Spatio-Temporal Reasoning of Relational Keypoint Constraints for Robotic Manipulation},
  url       = {https://openreview.net/forum?id=9iG3SEbMnL},
  year      = {2024}
}

@article{RLBench,
  author   = {James, Stephen and Ma, Zicong and Arrojo, David Rovick and Davison, Andrew J.},
  doi      = {10.1109/LRA.2020.2974707},
  journal  = {IEEE Robotics and Automation Letters},
  number   = {2},
  pages    = {3019-3026},
  title    = {RLBench: The Robot Learning Benchmark \& Learning Environment},
  volume   = {5},
  year     = {2020}
}

@misc{Robo2VLM,
  archiveprefix = {arXiv},
  author        = {Kaiyuan Chen and Shuangyu Xie and Zehan Ma and Pannag R Sanketi and Ken Goldberg},
  eprint        = {2505.15517},
  primaryclass  = {cs.RO},
  title         = {Robo2VLM: Visual Question Answering from Large-Scale In-the-Wild Robot Manipulation Datasets},
  url           = {https://arxiv.org/abs/2505.15517},
  year          = {2025}
}

@misc{RoboBrain2.0,
  archiveprefix = {arXiv},
  author        = {BAAI RoboBrain-Team and Mingyu Cao and Huajie Tan and Yuheng Ji and Xiansheng Chen and Minglan Lin and Zhiyu Li and Zhou Cao and Pengwei Wang and Enshen Zhou and Yi Han and Yingbo Tang and Xiangqi Xu and Wei Guo and Yaoxu Lyu and Yijie Xu and Jiayu Shi and Mengfei Du and Cheng Chi and Mengdi Zhao and Xiaoshuai Hao and Junkai Zhao and Xiaojie Zhang and Shanyu Rong and Huaihai Lyu and Zhengliang Cai and Yankai Fu and Ning Chen and Bolun Zhang and Lingfeng Zhang and Shuyi Zhang and Dong Liu and Xi Feng and Songjing Wang and Xiaodan Liu and Yance Jiao and Mengsi Lyu and Zhuo Chen and Chenrui He and Yulong Ao and Xue Sun and Zheqi He and Jingshu Zheng and Xi Yang and Donghai Shi and Kunchang Xie and Bochao Zhang and Shaokai Nie and Chunlei Men and Yonghua Lin and Zhongyuan Wang and Tiejun Huang and Shanghang Zhang},
  eprint        = {2507.02029},
  primaryclass  = {cs.RO},
  title         = {RoboBrain 2.0 Technical Report},
  url           = {https://arxiv.org/abs/2507.02029},
  year          = {2025}
}

@inproceedings{RoboChatGPT,
  author    = {Bing Xie and Xiangming Xi and Xinan Zhao and Yuhan Wang and Wei Song and Jianjun Gu and Shiqiang Zhu},
  booktitle = {ICIRA (5)},
  cdate     = {1672531200000},
  pages     = {365-376},
  title     = {ChatGPT for Robotics: A New Approach to Human-Robot Interaction and Task Planning},
  url       = {https://doi.org/10.1007/978-981-99-6495-6_31},
  year      = {2023}
}

@inproceedings{RoboFlamingo,
  author    = {Xinghang Li and Minghuan Liu and Hanbo Zhang and Cunjun Yu and Jie Xu and Hongtao Wu and Chilam Cheang and Ya Jing and Weinan Zhang and Huaping Liu and Hang Li and Tao Kong},
  booktitle = {The Twelfth International Conference on Learning Representations},
  title     = {Vision-Language Foundation Models as Effective Robot Imitators},
  url       = {https://openreview.net/forum?id=lFYj0oibGR},
  year      = {2024}
}

@inproceedings{RoboPoint,
  author    = {Wentao Yuan and Jiafei Duan and Valts Blukis and Wilbert Pumacay and Ranjay Krishna and Adithyavairavan Murali and Arsalan Mousavian and Dieter Fox},
  booktitle = {8th Annual Conference on Robot Learning},
  title     = {RoboPoint: A Vision-Language Model for Spatial Affordance Prediction in Robotics},
  url       = {https://openreview.net/forum?id=GVX6jpZOhU},
  year      = {2024}
}

@misc{RoboRefer,
  archiveprefix = {arXiv},
  author        = {Enshen Zhou and Jingkun An and Cheng Chi and Yi Han and Shanyu Rong and Chi Zhang and Pengwei Wang and Zhongyuan Wang and Tiejun Huang and Lu Sheng and Shanghang Zhang},
  eprint        = {2506.04308},
  primaryclass  = {cs.RO},
  title         = {RoboRefer: Towards Spatial Referring with Reasoning in Vision-Language Models for Robotics},
  url           = {https://arxiv.org/abs/2506.04308},
  year          = {2025}
}

@inproceedings{RoboSpatial,
  author    = {Song, Chan Hee and Blukis, Valts and Tremblay, Jonathan and Tyree, Stephen and Su, Yu and Birchfield, Stan},
  booktitle = {Proceedings of the Computer Vision and Pattern Recognition Conference (CVPR)},
  month     = {June},
  pages     = {15768-15780},
  title     = {RoboSpatial: Teaching Spatial Understanding to 2D and 3D Vision-Language Models for Robotics},
  year      = {2025}
}

@inproceedings{Room2room,
  author    = {Anderson, Peter and Wu, Qi and Teney, Damien and Bruce, Jake and Johnson, Mark and Sünderhauf, Niko and Reid, Ian and Gould, Stephen and van den Hengel, Anton},
  booktitle = {2018 IEEE/CVF Conference on Computer Vision and Pattern Recognition},
  doi       = {10.1109/CVPR.2018.00387},
  number    = {},
  pages     = {3674-3683},
  title     = {Vision-and-Language Navigation: Interpreting Visually-Grounded Navigation Instructions in Real Environments},
  volume    = {},
  year      = {2018}
}

@inproceedings{RT-1,
  author    = {Anthony Brohan and Noah Brown and Justice Carbajal and Yevgen Chebotar and Joseph Dabis and Chelsea Finn and Keerthana Gopalakrishnan and Karol Hausman and Alexander Herzog and Jasmine Hsu and Julian Ibarz and Brian Ichter and Alex Irpan and Tomas Jackson and Sally Jesmonth and Nikhil J. Joshi and Ryan Julian and Dmitry Kalashnikov and Yuheng Kuang and Isabel Leal and Kuang-Huei Lee and Sergey Levine and Yao Lu and Utsav Malla and Deeksha Manjunath and Igor Mordatch and Ofir Nachum and Carolina Parada and Jodilyn Peralta and Emily Perez and Karl Pertsch and Jornell Quiambao and Kanishka Rao and Michael S. Ryoo and Grecia Salazar and Pannag R. Sanketi and Kevin Sayed and Jaspiar Singh and Sumedh Sontakke and Austin Stone and Clayton Tan and Huong Tran and Vincent Vanhoucke and Steve Vega and Quan Vuong and Fei Xia and Ted Xiao and Peng Xu and Sichun Xu and Tianhe Yu and Brianna Zitkovich},
  booktitle = {Robotics: Science and Systems},
  cdate     = {1672531200000},
  title     = {RT-1: Robotics Transformer for Real-World Control at Scale},
  url       = {https://doi.org/10.15607/RSS.2023.XIX.025},
  year      = {2023}
}

@inproceedings{SayCan,
  author    = {Brian Ichter and Anthony Brohan and Yevgen Chebotar and Chelsea Finn and Karol Hausman and Alexander Herzog and Daniel Ho and Julian Ibarz and Alex Irpan and Eric Jang and Ryan Julian and Dmitry Kalashnikov and Sergey Levine and Yao Lu and Carolina Parada and Kanishka Rao and Pierre Sermanet and Alexander T Toshev and Vincent Vanhoucke and Fei Xia and Ted Xiao and Peng Xu and Mengyuan Yan and Noah Brown and Michael Ahn and Omar Cortes and Nicolas Sievers and Clayton Tan and Sichun Xu and Diego Reyes and Jarek Rettinghouse and Jornell Quiambao and Peter Pastor and Linda Luu and Kuang-Huei Lee and Yuheng Kuang and Sally Jesmonth and Kyle Jeffrey and Rosario Jauregui Ruano and Jasmine Hsu and Keerthana Gopalakrishnan and Byron David and Andy Zeng and Chuyuan Kelly Fu},
  booktitle = {6th Annual Conference on Robot Learning},
  title     = {Do As I Can, Not As I Say: Grounding Language in Robotic Affordances},
  url       = {https://openreview.net/forum?id=bdHkMjBJG_w},
  year      = {2022}
}

@inproceedings{Simpler,
  author    = {Xuanlin Li and Kyle Hsu and Jiayuan Gu and Oier Mees and Karl Pertsch and Homer Rich Walke and Chuyuan Fu and Ishikaa Lunawat and Isabel Sieh and Sean Kirmani and Sergey Levine and Jiajun Wu and Chelsea Finn and Hao Su and Quan Vuong and Ted Xiao},
  booktitle = {8th Annual Conference on Robot Learning},
  title     = {Evaluating Real-World Robot Manipulation Policies in Simulation},
  url       = {https://openreview.net/forum?id=LZh48DTg71},
  year      = {2024}
}

@misc{SoFar,
  archiveprefix = {arXiv},
  author        = {Zekun Qi and Wenyao Zhang and Yufei Ding and Runpei Dong and Xinqiang Yu and Jingwen Li and Lingyun Xu and Baoyu Li and Xialin He and Guofan Fan and Jiazhao Zhang and Jiawei He and Jiayuan Gu and Xin Jin and Kaisheng Ma and Zhizheng Zhang and He Wang and Li Yi},
  eprint        = {2502.13143},
  primaryclass  = {cs.RO},
  title         = {SoFar: Language-Grounded Orientation Bridges Spatial Reasoning and Object Manipulation},
  url           = {https://arxiv.org/abs/2502.13143},
  year          = {2025}
}

@book{SpaceInLang,
  author     = {Levinson, Stephen C.},
  collection = {Language Culture and Cognition},
  place      = {Cambridge},
  publisher  = {Cambridge University Press},
  series     = {Language Culture and Cognition},
  title      = {Space in Language and Cognition: Explorations in Cognitive Diversity},
  year       = {2003}
}

@article{SpatialBot,
  author   = {Wenxiao Cai and Yaroslav Ponomarenko and Jianhao Yuan and Xiaoqi Li and Wankou Yang and Hao Dong and Bo Zhao},
  cdate    = {1704067200000},
  journal  = {CoRR},
  publtype = {informal},
  title    = {SpatialBot: Precise Spatial Understanding with Vision Language Models},
  url      = {https://doi.org/10.48550/arXiv.2406.13642},
  volume   = {abs/2406.13642},
  year     = {2024}
}

@inproceedings{SpatialPIN,
  author    = {Ma, Chenyang and Lu, Kai and Cheng, Ta-Ying and Trigoni, Niki and Markham, Andrew},
  booktitle = {Advances in Neural Information Processing Systems},
  editor    = {A. Globerson and L. Mackey and D. Belgrave and A. Fan and U. Paquet and J. Tomczak and C. Zhang},
  pages     = {68803--68832},
  publisher = {Curran Associates, Inc.},
  title     = {SpatialPIN: Enhancing Spatial Reasoning Capabilities of Vision-Language Models through Prompting and Interacting 3D Priors},
  url       = {https://proceedings.neurips.cc/paper_files/paper/2024/file/7f2257d2b291b8d7e712c70b67e09412-Paper-Conference.pdf},
  volume    = {37},
  year      = {2024}
}

@inproceedings{SpatialRGPT,
  author    = {An-Chieh Cheng and Hongxu Yin and Yang Fu and Qiushan Guo and Ruihan Yang and Jan Kautz and Xiaolong Wang and Sifei Liu},
  booktitle = {The Thirty-eighth Annual Conference on Neural Information Processing Systems},
  title     = {Spatial{RGPT}: Grounded Spatial Reasoning in Vision-Language Models},
  url       = {https://openreview.net/forum?id=JKEIYQUSUc},
  year      = {2024}
}

@inproceedings{SpatialVLM,
  author    = {Chen, Boyuan and Xu, Zhuo and Kirmani, Sean and Ichter, Brain and Sadigh, Dorsa and Guibas, Leonidas and Xia, Fei},
  booktitle = {Proceedings of the IEEE/CVF Conference on Computer Vision and Pattern Recognition (CVPR)},
  month     = {June},
  pages     = {14455-14465},
  title     = {SpatialVLM: Endowing Vision-Language Models with Spatial Reasoning Capabilities},
  year      = {2024}
}

@misc{SpatialVTS,
  archiveprefix = {arXiv},
  author        = {Xun Liang and Xin Guo and Zhongming Jin and Weihang Pan and Penghui Shang and Deng Cai and Binbin Lin and Jieping Ye},
  eprint        = {2507.20529},
  primaryclass  = {cs.CV},
  title         = {Enhancing Spatial Reasoning through Visual and Textual Thinking},
  url           = {https://arxiv.org/abs/2507.20529},
  year          = {2025}
}

@misc{UniVLA,
  archiveprefix = {arXiv},
  author        = {Qingwen Bu and Yanting Yang and Jisong Cai and Shenyuan Gao and Guanghui Ren and Maoqing Yao and Ping Luo and Hongyang Li},
  eprint        = {2505.06111},
  primaryclass  = {cs.RO},
  title         = {UniVLA: Learning to Act Anywhere with Task-centric Latent Actions},
  url           = {https://arxiv.org/abs/2505.06111},
  year          = {2025}
}

@misc{UniVLA-BAAI,
  archiveprefix = {arXiv},
  author        = {Yuqi Wang and Xinghang Li and Wenxuan Wang and Junbo Zhang and Yingyan Li and Yuntao Chen and Xinlong Wang and Zhaoxiang Zhang},
  eprint        = {2506.19850},
  primaryclass  = {cs.CV},
  title         = {Unified Vision-Language-Action Model},
  url           = {https://arxiv.org/abs/2506.19850},
  year          = {2025}
}

@inproceedings{VLMaps,
  author    = {Huang, Chenguang and Mees, Oier and Zeng, Andy and Burgard, Wolfram},
  booktitle = {2023 IEEE International Conference on Robotics and Automation (ICRA)},
  doi       = {10.1109/ICRA48891.2023.10160969},
  number    = {},
  pages     = {10608-10615},
  title     = {Visual Language Maps for Robot Navigation},
  volume    = {},
  year      = {2023}
}

@inproceedings{VoxPoser,
  author    = {Huang, Wenlong and Wang, Chen and Zhang, Ruohan and Li, Yunzhu and Wu, Jiajun and Fei-Fei, Li},
  booktitle = {Proceedings of The 7th Conference on Robot Learning},
  editor    = {Tan, Jie and Toussaint, Marc and Darvish, Kourosh},
  month     = {06--09 Nov},
  pages     = {540--562},
  publisher = {PMLR},
  series    = {Proceedings of Machine Learning Research},
  title     = {VoxPoser: Composable 3D Value Maps for Robotic Manipulation with Language Models},
  url       = {https://proceedings.mlr.press/v229/huang23b.html},
  volume    = {229},
  year      = {2023}
}

@misc{VSI-Bench,
  archiveprefix = {arXiv},
  author        = {Jihan Yang and Shusheng Yang and Anjali W. Gupta and Rilyn Han and Li Fei-Fei and Saining Xie},
  eprint        = {2412.14171},
  primaryclass  = {cs.CV},
  title         = {Thinking in Space: How Multimodal Large Language Models See, Remember, and Recall Spaces},
  url           = {https://arxiv.org/abs/2412.14171},
  year          = {2024}
}

@article{VSR,
  address   = {Cambridge, MA},
  author    = {Liu, Fangyu  and
               Emerson, Guy  and
               Collier, Nigel},
  doi       = {10.1162/tacl_a_00566},
  journal   = {Transactions of the Association for Computational Linguistics},
  pages     = {635--651},
  publisher = {MIT Press},
  title     = {Visual Spatial Reasoning},
  url       = {https://aclanthology.org/2023.tacl-1.37/},
  volume    = {11},
  year      = {2023}
}

@inproceedings{WhatsUp,
  author    = {Amita Kamath and Jack Hessel and Kai-Wei Chang},
  booktitle = {The 2023 Conference on Empirical Methods in Natural Language Processing},
  title     = {What's ''up'' with vision-language models? Investigating their struggle with spatial reasoning},
  url       = {https://openreview.net/forum?id=RN5KLywTll},
  year      = {2023}
}
}


\appendix

%
\clearpage
\setcounter{page}{1}

\begin{abstract}
  This supplementary material includes
  (1) details of task definitions (\S\ref{sup:task}), including a taxonomy of spatial aspects in Table~\ref{tab:spatial-aspect} and curated functional programs in Table~\ref{tab:task-family-pick} and ~\ref{tab:task-family-place},
  (2) setups of the tabletop scene and the shelf scene (\S\ref{sup:env}),
  (3) a discussion on the sim-to-real relevance (\S\ref{sup:sim-to-real}),
  (4) evaluation details, such as prompting procedures, essential prompts, and evaluation efficiency (\S\ref{sup:eval}),
  and (5) details of \model assets, including their visualizations and dimensions (\S\ref{sup:asset}).

\end{abstract}

\section{Appendix}

\subsection{Participants of a robotics task}\label{sup:task}

A robotics task, in its simplest form, can be defined by an action and a manipulable object (e.g., `\emph{pick up the book}').
We employ two primitive actions, `pick' and `place', to initiate robotics tasks. Keeping the action space simple helps isolate spatial reasoning behaviors, allowing for a focus on their analysis.

To highlight the facets of spatial reasoning and support systematic task design and experimental analysis, we categorize key spatial aspects ($S$), reference frames ($F$), and reference objects ($O$) that characterize spatial reasoning and combine them to define task specifications (see Section 4.1).

\begin{itemize}[leftmargin=*]

  \item \textbf{Reference frames.} Reference frames refer to coordinate systems essential for describing one object in relation to another. They can be made explicit via linguistic specifications, but are usually implicitly conveyed within the context. Following~\citet{SpaceInLang}, we consider three types of reference frames.
    \begin{itemize}[leftmargin=*]
      \item \textbf{Relative frames} are viewer-centered; for example, `\emph{behind the mirror}' may refer to the space further from the viewer, from the viewer's perspective toward the mirror.
      \item \textbf{Intrinsic frames} are object-centered; for example, `\emph{behind the mirror}' may indicate the space opposite to the mirror's facing direction, independent of the viewer.
      \item \textbf{Absolute frames} are defined with respect to fixed global coordinates, such as elevation and altitude (useful for describing \emph{below} and \emph{above}) and cardinal directions (e.g., \emph{north} and \emph{east}), but are only used in a few indoor scenarios.
    \end{itemize}

  \item \textbf{Objects.} \model contains two primary object types: manipulable and reference objects.
    \begin{itemize}[leftmargin=*]
      \item \textbf{Manipulable objects} are instantiated as cuboid-shaped books (see Table~\ref{tab:book}). The regular geometries make it easier to verify their final states, facilitating automated evaluation. Moreover, they yield a relatively high likelihood of generating valid grasping/placement poses, without relying on external tools for pose proposal, yet remain sufficiently challenging for 6-DoF tasks.
      \item \textbf{Reference objects} participate in describing an object in relation to another. In cases where a reference frame is not explicitly specified, the intrinsic frame of the referenced object may be used. Thus, we divide reference objects into \emph{intrinsic-oriented} objects that have a clear front face (e.g., a chair or mirror) and \emph{non-oriented} objects that do not (e.g., a jar or ball).
        To support fine-grained analysis of spatial reasoning, such as distinguishing units of measure in distance estimation (meter vs. centimeter), we further divide reference objects into \emph{near} and \emph{distant} categories (see Table~\ref{tab:reference-near} and~\ref{tab:reference-distant}). Near objects appear on the shelf or tabletop, whereas distant objects are located outside these areas.
    \end{itemize}

  \item \textbf{Spatial aspects.} We group spatial aspects into four broad classes: attributes, distances, relationships, and orientations (see an overview in Table~\ref{tab:spatial-aspect}). Whenever applicable, we consider both coarse- and fine-grained expressions, such as relative distance and precise distance:
    \begin{itemize}[leftmargin=*]
      \item \textbf{Attributes} primarily refer to intrinsic size attributes (i.e., dimensions) of an object, such as \emph{height}, \emph{length}, \emph{width}, \emph{volume}, and \emph{diameter/radius}. They may be implicitly used to check space fitness and describe object volume, e.g., \emph{a large/small book}.

      \item \textbf{Distances} describe the proximity between objects. Apart from relative distance descriptions like \emph{nearest}, \emph{farthest}, and \emph{second farthest}, we include precise distance descriptions using different units of measure, e.g., \emph{within 1 meter of the mirror} and \emph{20 centimeters away from the jar.}'

      \item \textbf{Relationships} primarily describes positional relations, i.e., how one object is positioned relative to another. They can be expressed in diverse ways in natural language, but we consider only the most commonly used basic forms like \emph{left}, \emph{right}, \emph{in front of}, \emph{behind}, \emph{below}, and \emph{above}, and their comparative and superlative forms like \emph{leftmost}, \emph{rightmost}, and \emph{second leftmost}.

      \item \textbf{Orientations} cover directional expressions, including coarse-grained state descriptions (e.g., \emph{upright} and \emph{at a tilt}) and fine-grained clock positions (e.g., \emph{to your 6 o'clock}) and degrees of a tilt (e.g., \emph{at a 45-degree tilt}).
    \end{itemize}

\end{itemize}

We note that our definition of the task specification ($C=(S, F, O)$) primarily disentangles the complexity of spatial reasoning over `Relationships' and `Orientations', as they rely on a frame of reference, but `Attributes' and `Distances' do not. Nonetheless, we use this definition across all four spatial aspects to keep consistent.

\begin{figure*}[t!]
\centering
\includegraphics[width=\linewidth]{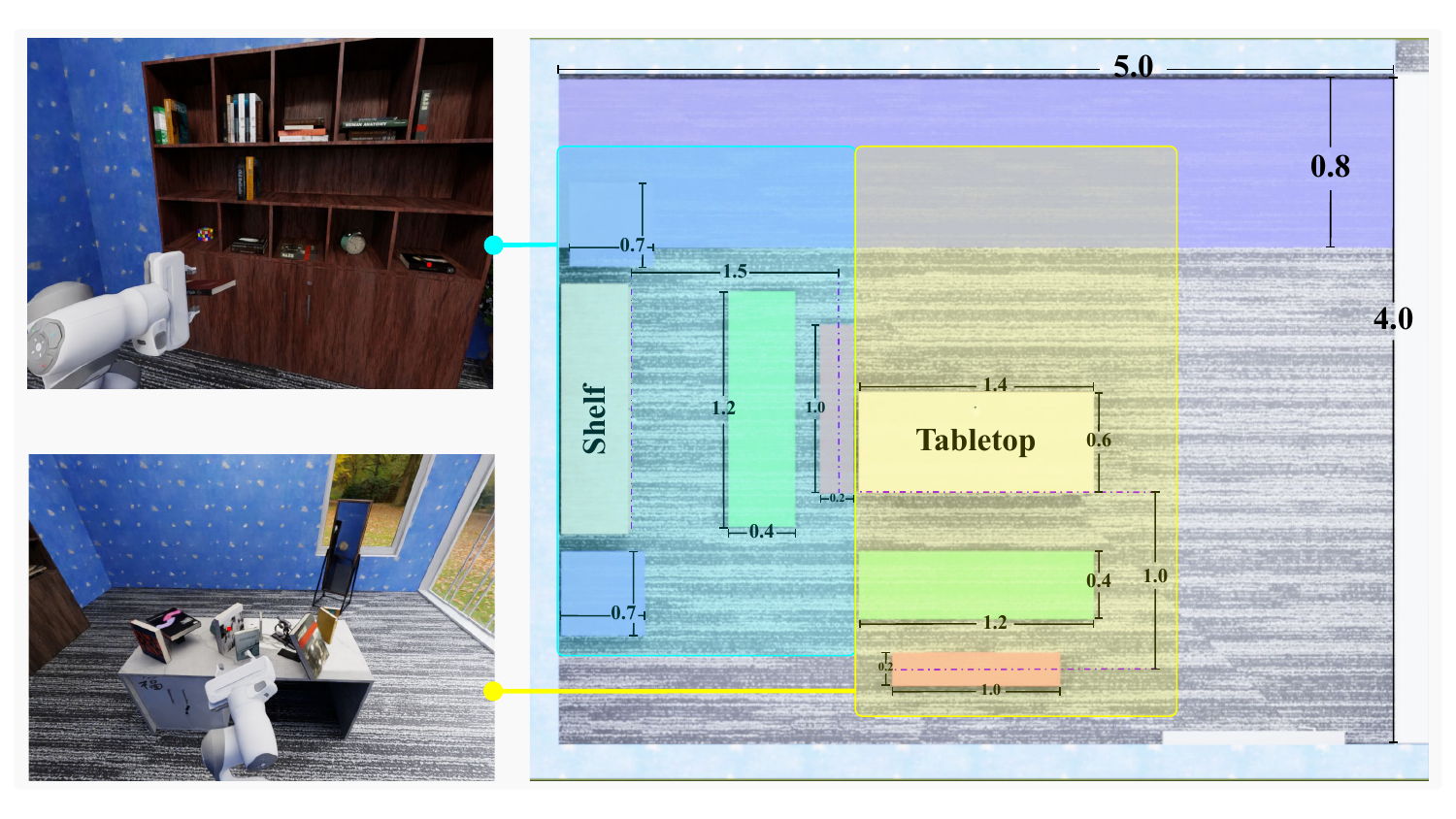}
\caption{Layouts of the tabletop and shelf scenes within \model (best viewed in color). 
The light red region denotes the camera viewpoint sampling area, the light green region indicates where the robot end effector may appear, and the light blue region denotes where distant reference objects are placed. All labeled dimensions are in meters.}
\label{fig:layout}
\end{figure*}

\subsection{Simulated Environment}\label{sup:env}

We focus on tasks that involve picking up an object from the table and placing it on the shelf (see Figure~\ref{fig:layout}). Though the reverse direction---picking up an object from the shelf and placing it on the table---is also feasible and would increase task diversity, we consider only the former because it alone suffices to cover a diverse range of spatial reasoning scenarios.

\paragraph{Tabletop tasks.}
The table scene is initialized with manipulable books, support ornaments, and reference ornaments. For books, we consider three different sizes (i.e., \emph{small}, \emph{medium}, and \emph{large}), and three different initial poses (i.e., \emph{standing upright}, \emph{lying flat}, and \emph{at a tilt}). We use a small set of support ornaments to create \emph{flat} and \emph{tilting} poses while ensuring that the books are pickable via 6-DoF pose prediction. Near references are small-sized, appearing on the table (e.g., a picture frame or ceramic jar), whereas distant references are large-sized, located on the floor behind the table (e.g., a floor lamp or cheval mirror). Note that the reference ornaments are carefully selected to cover intrinsic-oriented and non-oriented categories, and the robot itself is an intrinsic-oriented reference, always facing the front of the table. We initialize the tabletop with two random near objects and one random distant object. The degree of clutter in the scene is controlled by varying the number of books on the table, while the overall complexity is driven by instructions and environmental factors like varying poses, lights, and textures.

The global camera looks at a random point on the front edge of the table. We randomly sample its elevation 0.5--1m above the table surface.
The elevation of the end-effector is randomly sampled from 0.3--0.5m below the global camera, with pitch, yaw, roll randomly sampled from $[-22.5^\circ, 22.5^\circ]$, $[-22.5^\circ, 22.5^\circ]$, and $[-45^\circ, 45^\circ]$, respectively.

\paragraph{Shelf tasks.}
The shelf scene contains the same types of objects as in the tabletop scene. Analogously to the tabletop scenes, where books are initialized in random grasping poses, we initialize the shelf with random support ornaments to let books lean against, creating various placement poses. Compared with the tabletop scene, the shelf scene supports reasoning of two additional spatial relationships: \emph{above} and \emph{below}. We locate distant references on the floor, either to the left or to the right of the shelf. The robot always faces the front of the shelf. We control the complexity of shelf tasks by varying shelf layouts, including horizontal panels, grids of slots, and their combinations.

The global camera always faces the shelf center. We randomly sample its elevation 1.2--1.5m above the ground.
The elevation of the end-effector is randomly sampled from 0.3--0.5m below the global camera, with pitch, yaw, roll randomly sampled from $[-22.5^\circ, 22.5^\circ]$, $[-22.5^\circ, 22.5^\circ]$, and $[-120^\circ, -60^\circ]\cup[60^\circ, 120^\circ]$, respectively.

\paragraph{Definitions of spatial relationships and orientations.}
In natural language, spatial relationships can exhibit ambiguity due to the reliance on reference frames and contexts. 
To address this issue, we use a unified definition to assign them unambiguous geometric interpretations. Specifically, under a given reference frame, we use its forward axis to represent the front-facing direction, then \emph{left} and \emph{right} are defined relative to it. The definition of \emph{behind} is, however, more involved, as it depends on the reference frame. Suppose the description `$O_1$ is \emph{behind} $O_2$', when the reference frame is independent of $O_2$, it is interpreted as: $O_1$ is further than $O_2$ along the front-facing direction; when the reference frame is attached to $O_2$, the meaning changes to: $O1$ is further along the opposite of the front-facing direction.

We account for two fine-grained types of orientation: direction and tilt. Directions are represented using clock positions that provide a granular description relative to a specific reference frame. In this setup, the forward axis is assigned to 12 o'clock, with all other positions mapped relative to this heading. To describe precise tilts, we measure the tilt angle in degrees and define it as the deviation between the global up-axis and the upright axis of the object.

\paragraph{Definitions of `above' and `below'.} The global up-axis corresponds to the up-direction of an absolute reference frame, defined as the surface normal of the floor in our simulation environment.
We also rely on this global up-axis to define spatial relationships like \emph{above} and \emph{below}.
Specifically, `$O_1$ is \emph{above} $O_2$' indicates that $O_1$ lies further along the global up-axis; equivalently, `$O_2$ is \emph{below} $O_1$'.
Following the standard convention, we define the tilt angle as the angle between the up axis and the surface normal of an object. 

\paragraph{Mitigation of ambiguity.}
For objects involved in tabletop tasks, we randomly initialize their locations while ensuring that they are spaced at least 5cm apart. Note that due to physical rendering constraints, the final spacing may be smaller than 5cm. We require that at least 20\% of the pixels of each object are visible in the global view. For shelf tasks, we require that at least 50\% of the pixels of the book in hand are visible. When a target satisfies multiple constraints (e.g., a book can be behind the picture while also being to its left), we select the most salient one for task generation.

\begin{algorithm*}[t!]
\caption{Balanced Task Sampling}
\label{alg:balanced_task_sampling}
\begin{algorithmic}[1]\footnotesize
    \State \textbf{Input:} Task families $\mathcal{T}$, scenes $\mathcal{S}$, difficulty levels $\mathcal{L}=\{\text{easy},\text{medium},\text{hard}\}$, the number of tasks per family $N_{\mathrm{t}}$, the number of tasks in total $N_{\mathrm{all}}$.
    \State Initialize task counts $\{C_{t,l}\}_{t\in\mathcal{T},\,l\in\mathcal{L}} \gets 0$
    \State Initialize scene attempts $\{A_s\}_{s \in \mathcal{S}} \gets 0$
    \State Initialize the task set $\mathcal{Q} \gets \emptyset$
    \State $N \gets 0$
    \While{\textsc{EligibleFamily}$(\mathcal{T}, \mathcal{S}) \neq \emptyset$}
        \If{$N_{\mathrm{all}}$ is defined \textbf{and} $N \geq N_{\mathrm{all}}$}
            \State \textbf{break}
        \EndIf
        \State $\mathcal{T}_{\mathrm{sub}} \gets \{t \in \mathcal{T} \mid \sum_l C_{t,l} < N_{\mathrm{t}},\ \textsc{GetCompatibleScenes}(t,\mathcal{S}) \neq \emptyset\}$
        \If{$\mathcal{T}_{\mathrm{sub}} = \emptyset$}
            \State \textbf{break}
        \EndIf
        \ForAll{$t \in \mathcal{T}_{\mathrm{sub}}$}
            \State $w_t \gets \dfrac{1}{\sum_l C_{t,l} + 1}$ \Comment{Under-sampled task families}
        \EndFor
        \State $t^\star \gets \textsc{WeightedSampling}(\mathcal{T}_{\mathrm{sub}}, \{w_t\})$
        \State $\mathcal{S}_{t^\star} \gets \textsc{GetCompatibleScenes}(t^\star, \mathcal{S})$
        \ForAll{$s \in \mathcal{S}_{t^\star}$}
            \State $l \gets \textsc{GetDifficultyLevel}(t^\star, s)$
            \State $w_s \gets \dfrac{1}{\sum_{t'} C_{t',l} + 1} \cdot \dfrac{1}{(A_s + 1)^2}$ \Comment{Under-sampled scenes and difficulty levels}
        \EndFor
        \State $s^\star \gets \textsc{WeightedSampling}(\mathcal{S}_{t^\star}, \{w_s\})$
        \State $q \gets \textsc{GenerateAnswerSet}(t^\star, s^\star)$
        \State $A_{s^\star} \gets A_{s^\star} + 1$
        \If{$|q| > 1$} \Comment{Retain only non-trivial tasks}
            \State $l^\star \gets \textsc{GetDifficultyLevel}(s^\star, t^\star)$
            \State $\mathcal{Q} \gets \mathcal{Q} \cup \{(t^\star, s^\star)\}$
            \State $C_{t^\star,l^\star} \gets C_{t^\star,l^\star} + 1$
            \State $N \gets N + 1$
        \EndIf
    \EndWhile
    \State \Return $\mathcal{Q}$
\end{algorithmic}
\end{algorithm*}

\paragraph{Balanced task sampling.}
To ensure that \model tasks are approximately uniformly distributed across task families $\mathcal{T}$ and difficulty levels $\mathcal{L}=\{\text{easy},\text{medium},\text{hard}\}$, we propose a balanced task sampling strategy (see Algorithm~\ref{alg:balanced_task_sampling}). Specifically, we maintain a counter $C_{t,l}$ for each combination of task family $t\in\mathcal{T}$ and difficulty level $l\in\mathcal{L}$. We also record the number of times each scene $s\in\mathcal{S}$ has been attempted so far, denoted by $A_s$. These counters are used to dynamically adjust the sampling weights (lines~15 and~21).

During task generation, we first select a task family with preference for underrepresented families (lines~12–15). Given the selected task family, we collect all scenes that yield a non-empty answer set and randomly sample one, favoring underrepresented difficulty levels while penalizing scenes that have been repeatedly attempted (lines~16–19).
We repeat this process until the desired number of tasks has been generated.

\subsection{Sim-to-real relevance}\label{sup:sim-to-real}
To confirm \model serves as a reliable proxy for embodied spatial reasoning, we establish the benchmark's validity through the following two lens:
\paragraph{Performance alignment.}
We evaluated Qwen3-VL (8B/30B/235B), RoboBrain2.0-7B and Gemini2.5-Pro on \model and the pointing tasks of the natural-image benchmark RefSpatial~\citep{RoboRefer}. We then compute Spearman’s rank correlation between the model performance rankings on the two benchmarks.
The resulting correlation is 96.4\% (with $p=0.00498$), indicating strong alignment between the two evaluations and suggesting that \model serves as a high-fidelity proxy for real-world embodied spatial reasoning.

\paragraph{Human study.}
We conducted a study with five humans to assess environment realism and model alignment (see Section~\ref{sec:human-study}).
First, we observe an average $94.9\pm3.4\%$ human success rate across all tasks, suggesting that the simulated scenarios are readily interpretable and solvable by humans.

We further analyze agreement on the ground-truth reference frame across three reference categories: near oriented objects, distant oriented objects, and the table.
Specifically, we measure the proportion of examples with unanimous agreement among the five annotators.
For examples involving distant oriented references and the table, humans agree on the reference frame in more than 97\% of cases, suggesting that the intended frame is clearly interpretable in these settings.
However, only 31.03\% of the examples involving near oriented references yield unanimous agreement. This lower agreement likely reflects the inherent ambiguity of reasoning with nearby oriented objects, as discussed in Section~\ref{sec:human-study}.

\begin{algorithm*}[t!]
\caption{Localization Procedures}
\label{alg:localization}
\begin{subalgorithm}[t]{.48\linewidth}
\caption{Localization w/o Reflection}
\label{alg:eval_grounding}
\begin{algorithmic}[1]\footnotesize
  \State \textbf{Input:} Task instruction $T$, Scene observation $O$, Maximum trials $N$
  \For{$i = 1$ to $N$}
      \State $P \gets \textsc{Predict}(T, O)$
      \State $R \gets \textsc{Evaluate}(P)$
      \If{$R.\text{success}$}
          \State \textbf{break} \Comment{Stop if successful}
      \EndIf
  \EndFor
\end{algorithmic}
\end{subalgorithm}
\hfill
\begin{subalgorithm}[t]{.48\linewidth}
\caption{Localization w/ Reflection}
\label{alg:eval_reflective_grounding}
\begin{algorithmic}[1]\footnotesize
  \State \textbf{Input:} Task instruction $T$, Scene observation $O$, Maximum trials $N$
  \State $F, R_0 \gets \text{None}$
  \For{$i = 1$ to $N$}
      \State \textcolor{myorange}{$P \gets \textsc{Predict}(T, O, F, R_0)$}
      \State $R_1 \gets \textsc{Evaluate}(P)$
      \If{$R.\text{success}$}
          \State \textbf{break}
      \EndIf
      \State \textcolor{myorange}{$F \gets \textsc{Reflect}(T, O, R)$}
      \State $R_0 \gets R_1$
  \EndFor
\end{algorithmic}
\end{subalgorithm}

\end{algorithm*}
\begin{algorithm*}[t!]
\caption{Execution Procedures}
\label{alg:execution}
\begin{subalgorithm}[t]{.48\linewidth}
\caption{Execution w/o Reflection}
\label{alg:eval_moving}
\begin{algorithmic}[1]\footnotesize
  \State \textbf{Input:} Task configuration $T$, Scene observation $O$, Maximum trials $N$
  \For{$i = 1$ to $N$}
      \State $P \gets \textsc{Predict}(T, O)$
      \State $R \gets \textsc{Evaluate}(P)$
      \If{$R.\text{success}$}
          \State \textbf{break} \Comment{Stop if task is done}
      \EndIf
      \State $O \gets \textsc{GetObservation}()$
  \EndFor
\end{algorithmic}
\end{subalgorithm}
\hfill
\begin{subalgorithm}[t]{.48\linewidth}
\caption{Execution w/ Reflection}
\label{alg:eval_reflective_moving}
\begin{algorithmic}[1]\footnotesize
  \State \textbf{Input:} Task configuration $T$, Initial observation $O$, Maximum trials $N$
  \State $F, O_0, R_0 \gets \text{None}$
  \For{$i = 1$ to $N$}
      \State $P \gets \textsc{Predict}(T, O, F, O_0, R_0)$
      \State $R_1 \gets \textsc{Evaluate}(P)$
      \If{$R_1.\text{success}$}
          \State \textbf{break}
      \EndIf
      \State \textcolor{myorange}{$O_0 \gets O$}
      \State \textcolor{myorange}{$R_0 \gets R_1$}
      \State \textcolor{myorange}{$O \gets \textsc{GetObservation}()$}
      \State \textcolor{myorange}{$F \gets \textsc{Reflect}(T, O, O_0, R_0)$}
  \EndFor
\end{algorithmic}
\end{subalgorithm}

\end{algorithm*}

\subsection{Evaluation}\label{sup:eval}

\paragraph{Task status checking.}
After execution, we obtain task status (e.g., failure or success) by checking if the final environment state satisfies the constraints specified in the instruction.
(1) \emph{Distance} is measured as the minimum distance between the boundaries of the target and the reference object (which can be a 3D point).
The final distance within $\pm3$ cm of the expected distance is considered correct in the evaluation.
(2) \emph{Orientation and Relationship} are determined by checking if the center of the target lies in the target area defined by a reference frame.
The final tilt angle within $\pm 10^\circ$ of the expected angle is considered correct in the evaluation.

\paragraph{Algorithms.}
We illustrate the evaluation procedures used for localization in Algorithm~\ref{alg:eval_grounding} (w/o reflection) and Algorithm~\ref{alg:eval_reflective_grounding} (w/ reflection).
Reflection is performed following a localization failure (line 9 of Algorithm~\ref{alg:eval_reflective_grounding}). The generated reflection tokens are added to the inputs to the next iteration (line 4 of Algorithm~\ref{alg:eval_reflective_grounding}).
The evaluation procedures for execution are illustrated in Algorithm~\ref{alg:eval_moving} (w/o reflection) and Algorithm~\ref{alg:eval_reflective_moving} (w/ reflection). They are similar to those used for localization except that the reflection for execution relies on an additional view of the failure state (line 9--12 of Algorithm~\ref{alg:eval_reflective_moving}).

\paragraph{Prompts.}
We provide our customized prompts for \emph{pick} tasks with Qwen3-VL~\citep{QWen25vl}. Figures~\ref{prompt:grounding_prompt}, \ref{prompt:moving_prompt}, \ref{prompt:reflect_grounding_prompt}, and \ref{prompt:query_rotation_prompt} show the prompts used for localization, execution, localization w/ reflection, and rotation, respectively. They are different across VLMs primarily in the output format, e.g., Gemini2.5-Pro~\citep{Gemini2.5} outputs point coordinates in $[y, x]$ while others in $[x, y]$. The differences in the prompts for \emph{pick} and \emph{place} tasks arise primarily in the task descriptions. For example, a localization instruction for \emph{place} tasks could be: `\emph{Given a scene image and a textual description specifying the placement conditions for a book currently held by a robot gripper, you are required to determine the exact placement location in the image.}'

\paragraph{Evaluation Time.}
We break down the evaluation time along the evaluation stage (see Table~\ref{tab:time-analysis}). The results are averaged across all tasks and attempts. We record the time taken until a successful attempt is achieved. During execution, the environment updates after each move and in each observation query, so we also report the average number of environment updates over successful tasks.
Compared to the model inference time in localization and execution, the environment update is quite quick, i.e., it takes an average of 11.65 seconds per update. Models with reflection enabled generally take longer because they require additional API calls. A higher number of environment updates indicates more execution attempts. For example, RoboBrain2.0-7B~\citep{RoboBrain2.0} not only requires the largest number of environment updates but also achieves the lowest success rate, suggesting its weaker capability in execution.

\paragraph{Running Examples.}
We provide running examples with Qwen3-VL-235B-A22B~\citep{QWen25vl}. Figure~\ref{fig:pick_localization_example_part1} shows an excerpt of the localization (w/ reflection) logs of a \emph{pick} task, and Figure~\ref{fig:place_localization_example} shows an example run on a \emph{place} task without reflection.

Figure~\ref{fig:pick_execution_example} shows an excerpt of the execution (w/o reflection) logs of a \emph{pick} task. Note that, in this example, the model also needs to predict rotations for the \emph{pitch} axis. Figure~\ref{fig:place_execution_example} presents an excerpt of the execution (w/ reflection) logs of a \emph{place} task.

In this task, the model must predict a goal position for the end-effector. Interestingly, the model fails in the first attempt but succeeds by leveraging reflection in the second attempt.

\begin{table*}[t!]
\centering
\caption{Breakdowns of evaluation time in seconds (s).}
\label{tab:time-analysis}
\resizebox{\linewidth}{!}{%
\begin{tabular}{@{}lcccccccc@{}}
\toprule
\multicolumn{1}{c}{\multirow{2}{*}{Models}} & \multicolumn{4}{c}{Pick} & \multicolumn{4}{c}{Place} \\
\cmidrule(l){2-9}
\multicolumn{1}{c}{} &
  localization (s) & execution (s) & \# update & success (\%) &
  localization (s) & execution (s) & \# update & success (\%) \\
\midrule
\multicolumn{9}{c}{\cellcolor[HTML]{EFEFEF}{w/o reflection}} \\
\midrule
Gemini2.5-Pro           & 22.37 & 40.53 & 6.86 & 34.06 & 33.00 & 41.15 & 9.01 & 15.70 \\
InternVL3-78B           & 14.44 & 32.21 & 6.92 & 17.26 & 19.06 & 34.04 & 7.59 & 9.67 \\
RoboBrain2.0-7B         & 14.17 & 27.56 & 9.07 & 10.87 & 19.08 & 30.83 & 9.36 & 8.64 \\
Qwen3-VL-8B & 15.44 & 34.87 & 6.73 & 29.32 & 19.48 & 35.86 & 7.83 & 12.41 \\
Qwen3-VL-30B-A3B & 12.50 & 30.61 & 6.86 & 32.15 & 17.38 & 32.59 & 7.39 & 20.00 \\
Qwen3-VL-235B-A22B & 16.08 & 34.49 & 7.38 & 26.76 & 20.40 & 37.55 & 7.64 & 19.34 \\
\midrule
\multicolumn{9}{c}{\cellcolor[HTML]{EFEFEF}{w/ reflection}} \\
\midrule
Qwen3-VL-8B & 23.94 & 44.88 & 8.58 & 15.07 & 31.22 & 47.58 & 9.31 & 6.67 \\
Qwen3-VL-30B-A3B & 21.51 & 44.70 & 8.37 & 17.08 & 27.99 & 49.29 & 8.13 & 13.80 \\
Qwen3-VL-235B-A22B & 33.23 & 66.83 & 8.12 & 23.20 & 44.03 & 68.58 & 8.53 & 15.40 \\
\bottomrule
\end{tabular}
} 
\end{table*}

\begin{figure*}[t!]
\centering
\begin{subfigure}[b]{\textwidth}
\begin{tcolorbox}[colback=lightgray!5, colframe=darkgray, arc=1mm, boxrule=.5pt]
You are an expert Visual-Language Assistant specialized in embodied pointing tasks. Note that any occurrence of ``You'' denotes the first-person camera viewpoint of the RGB image (i.e., the perceptual perspective). It does NOT refer to any object, robot, or robotic end-effector visible in the image.

\textbf{Task:}
Identify the object in an RGB image according to a detailed textual description and provide precise coordinates for the object that match the input conditions.

\textbf{Inputs:}
1. A text description specifying the book that needs to be located, including all required attributes and conditions.
2. An RGB image representing the scene, captured from your perceptual viewpoint.

\textbf{Output:}
Locate the target object with a point, report its point coordinate in JSON format like this:
{``point\_2d'': [x, y]}
\end{tcolorbox}
\caption{Localization in \emph{pick} tasks.}
\label{prompt:grounding_prompt}
\end{subfigure}

\vspace{1em}

\begin{subfigure}[b]{\textwidth}
\begin{tcolorbox}[colback=lightgray!5, colframe=darkgray, arc=1mm, boxrule=.5pt]
You are an expert Visual-Language Assistant specialized for embodied robot-arm manipulation.

\textbf{Task:}
Determine the target position of the robotic arm's end-effector to grasp a specific book, where the book is marked with a colored bounding box. Choose a grasp point that is easily reachable and suitable for the robot's end-effector, avoiding edges or positions that are obstructed or unstable. The grasp point should allow the robot to securely pick up the book without collisions or slippage.

\textbf{Note:}
If the predicted target point lies on the target book, the robot will attempt to directly grasp the book at that location.
If the predicted point lies on another object or a free-space region, the robot will move its end-effector to the indicated position.

\textbf{Inputs:}
1. An RGB image representing the scene, where the target book is marked with a colored bounding box.
2. A textual instruction that specifies how the robot should grasp the book.

\textbf{Output:}
Locate the target position with a point, report its point coordinate in JSON format like this:
{``point\_2d'': [x, y]}
\end{tcolorbox}
\caption{Execution in \emph{pick} tasks.}
\label{prompt:moving_prompt}
\end{subfigure}
\caption{Example prompts with Qwen3-VL (continued).}
\end{figure*}

\begin{figure*}[t!]
\centering
\begin{subfigure}[b]{\textwidth}
\begin{tcolorbox}[colback=lightgray!5, colframe=darkgray, arc=1mm, boxrule=.5pt]
You are an expert Visual-Language Assistant specialized in reflective analysis of embodied pointing tasks.

\textbf{Task:}
Analyze the previous prediction error in the image grounding task. The incorrect predicted point has been marked with a red circle on the RGB image. Reflect on why the model's prediction failed and explain what visual or textual cues were misunderstood or ignored.

\textbf{Inputs:}
1. The original textual description specifying the target object and its required attributes or conditions.
2. The same RGB image used in the grounding task, now containing the red circle marking the incorrect prediction.

\textbf{Output:}
Provide a detailed textual reflection explaining:
- The likely cause of the failure (e.g., visual confusion, incorrect attribute matching, spatial misinterpretation, etc.).
- The correct reasoning process that should have been applied.
- Suggestions for how to improve the next prediction.
\end{tcolorbox}
\caption{Localization w/ reflection in \emph{pick} tasks.}
\label{prompt:reflect_grounding_prompt}
\end{subfigure}

\vspace{1em}

\begin{subfigure}[b]{\textwidth}
\begin{tcolorbox}[colback=lightgray!5, colframe=darkgray, arc=1mm, boxrule=.5pt]
You are an expert Visual-Language Assistant specialized for embodied robot-arm manipulation and precise grasp orientation control.

\textbf{Task:}
Given an RGB image with a colored bounding box marking the target book and a red dot indicating the moving target position, together with a textual instruction specifying the book attributes and required rotation axes (pitch, yaw, roll), predict the end-effector's **rotation offsets** relative to its current orientation. All rotations are defined in the **target book's local coordinate frame**.

\textbf{Inputs:}
1. RGB image:
   - Target book marked by a colored bounding box.
   - Moving target position marked by a red dot.
2. Text instruction describing the target book and specifying which rotation axes to output.

\textbf{Reference Frame:}
Rotations follow the book's intrinsic axes (forward–backward, left–right, up–down):
- **Pitch:** tilt up/down.
- **Yaw:** swing left/right.
- **Roll:** roll clockwise/counterclockwise.
Positive/negative follow standard right-hand conventions.

\textbf{Output:}
Return the predicted rotation offsets (degrees) in JSON:
```json
{``rotation'': [pitch, yaw, roll]}
\end{tcolorbox}
\caption{Rotation prediction in \emph{pick} tasks, summarized by GPT-5.1 for the demonstration purpose.}
\label{prompt:query_rotation_prompt}
\end{subfigure}
\caption{Example prompts with Qwen3-VL.}
\end{figure*}

\newcommand{\examplekey}[1]{\textcolor{mygreen}{#1}}
\newcommand{\examplehl}[1]{\textcolor{myorange}{#1}}
\newcommand{\exampleoutput}[1]{\textcolor{mypurple}{#1}}

\begin{figure*}[t]
\centering
\begin{tcolorbox}[
    colback=lightgray!5,
    colframe=darkgray,
    arc=1mm,
    boxrule=.5pt,
    width=\textwidth
]
\begin{tabular}{
p{0.08\textwidth} 
p{0.65\textwidth} 
p{0.2\textwidth}}
\vspace{0pt} \textbf{ENV:} &
\vspace{0pt} \{\examplekey{task\_instruction}: from the items on the table, find a book more than 120cm away from you, and fetch it. \examplekey{category}: pick\} &
\vspace{0pt} \includegraphics[width=\linewidth]{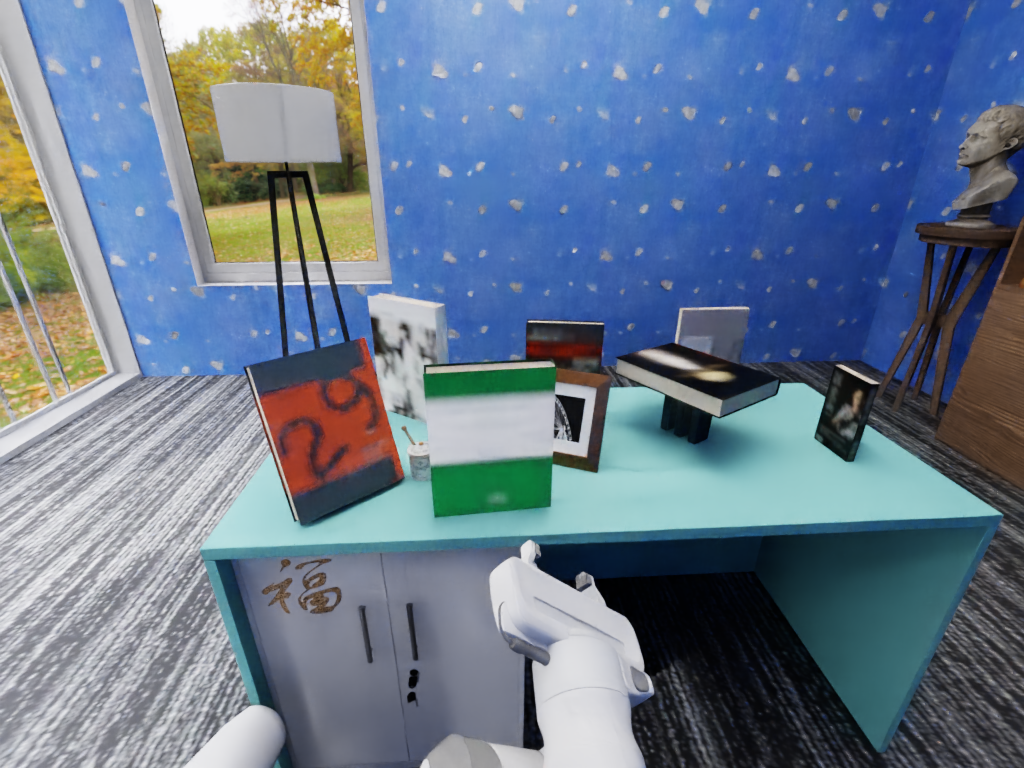} \\

\vspace{0pt} \textbf{VLM:} &
\vspace{0pt} \{\examplekey{system\_prompt}: [see Figure~\ref{prompt:grounding_prompt}], \examplekey{user}: from the items on the table, find a book more than 120cm away from you, and fetch it. \examplekey{assistant}: \{point\_2d: \exampleoutput{[757, 448]}, label: book more than 120cm away from you\}\} &
\vspace{0pt} \includegraphics[width=\linewidth]{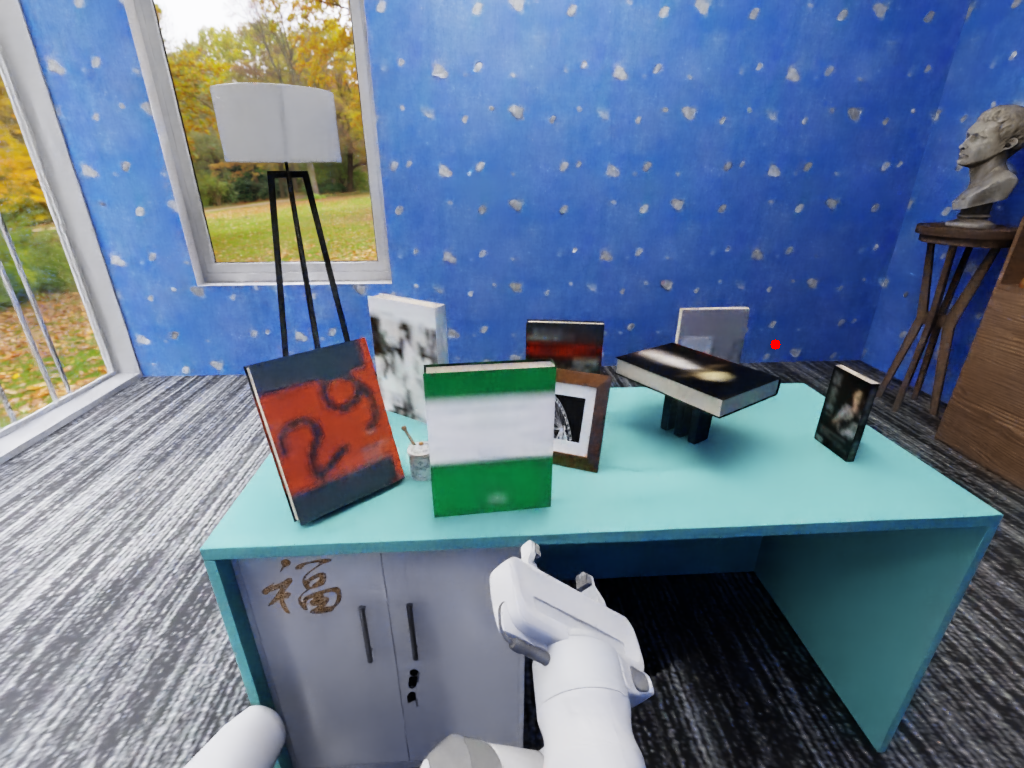} \\

\vspace{0pt} \textbf{ENV:} &
\vspace{0pt} \{\examplekey{moved}: false, \examplekey{success}: false\} &
\vspace{0pt} \includegraphics[width=\linewidth]{figs/running_examples/localize/pick/0.png} \\

\vspace{0pt} \textbf{VLM:} &
\vspace{0pt} \{\examplekey{system\_prompt}: [see Figure~\ref{prompt:reflect_grounding_prompt}], \examplekey{user}: The previous visual grounding attempt for the [task\_instruction] was incorrect. The red circles in the image mark the wrong grounding regions. Please carefully analyze the image and the instruction again, understand why the previous prediction was wrong, and describe what should be corrected in the next attempt. \examplekey{assistant}: The previous prediction was incorrect because it failed to properly interpret the spatial constraint in the instruction... \examplehl{Why the Previous Prediction Was Wrong:} 1. Misinterpretation of Spatial Distance 2. Focus on Object Type Only... \examplehl{What Should Be Corrected in the Next Attempt:} 1. Incorporate Spatial Reasoning 2. Re-evaluate All Books on the Table...\} &
\vspace{0pt} \includegraphics[width=\linewidth]{figs/running_examples/localize/pick/0.png} \includegraphics[width=\linewidth]{figs/running_examples/localize/pick/1.png} \\

\vspace{0pt} \textbf{VLM:} &
\vspace{0pt} \{\examplekey{system\_prompt}: [see Figure~\ref{prompt:moving_prompt}], \examplekey{user}: Here is a reflection on the previous failed attempt and some suggestions: [reflection]. Now, complete this task: from the items..., \examplekey{assistant}: \{point\_2d: \exampleoutput{[745, 400]}\}\} &
\vspace{0pt} \includegraphics[width=\linewidth]{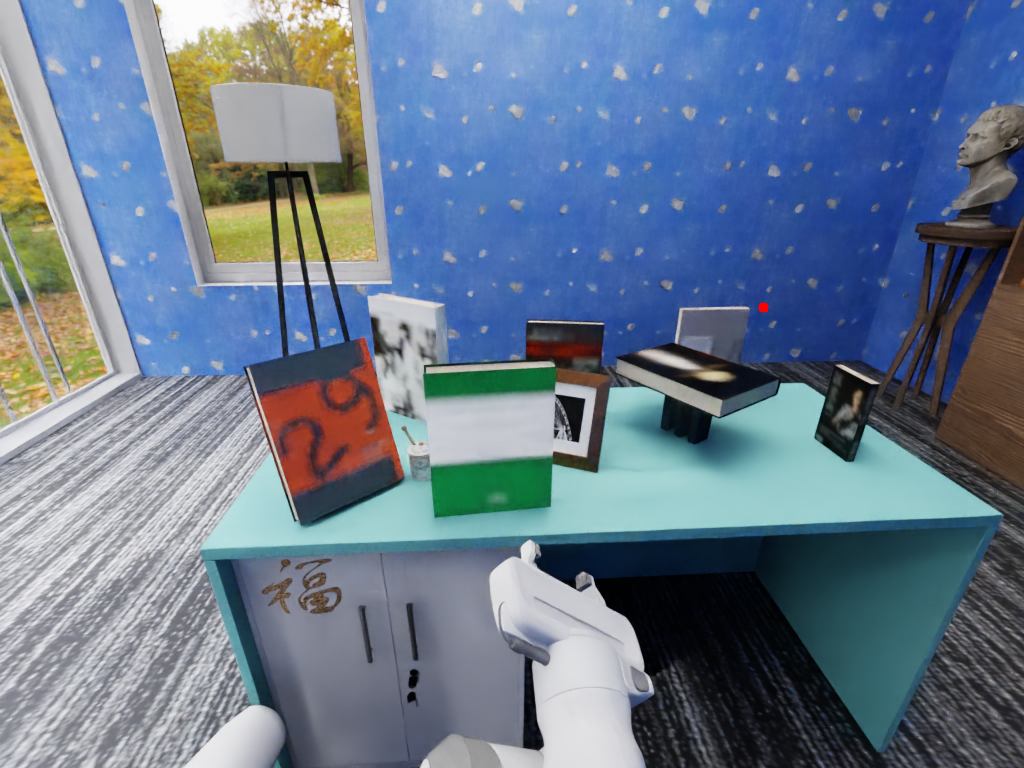} \\

\vspace{0pt} \textbf{ENV:} &
\vspace{0pt} \{\examplekey{moved}: false, \examplekey{success}: false\} &
\vspace{0pt} \includegraphics[width=\linewidth]{figs/running_examples/localize/pick/0.png} \\
\end{tabular}
\end{tcolorbox}

\caption{Pick localization example with reflection, where the prompt simplified for better demonstration.}
\label{fig:pick_localization_example_part1}
\end{figure*}


\begin{figure*}[t]
\centering
\begin{tcolorbox}[
    colback=lightgray!5,
    colframe=darkgray,
    arc=1mm,
    boxrule=.5pt,
    width=\textwidth
]
\begin{tabular}{
p{0.08\textwidth} 
p{0.65\textwidth} 
p{0.2\textwidth}}
\vspace{0pt} \textbf{VLM:} &
\vspace{0pt} \{\examplekey{system\_prompt}: [see Figure~\ref{prompt:reflect_grounding_prompt}], \examplekey{user}: The previous visual grounding attempt for the [task\_instruction] was incorrect. The red circles in the image mark the wrong grounding regions. Please carefully analyze the image and the instruction again, understand why the previous prediction was wrong, and describe what should be corrected in the next attempt. \examplekey{assistant}: The previous prediction was incorrect because it focused on a region on the wall... \examplehl{Why the Error Occurred:} - Misinterpretation of ``items on the table'' - Failure to filter by object class... \examplehl{What Should Be Corrected in the Next Attempt:} 1. Identify all books on the table 2. Estimate distance from the agent (robot arm)...\} &
\vspace{0pt} \includegraphics[width=\linewidth]{figs/running_examples/localize/pick/0.png} \includegraphics[width=\linewidth]{figs/running_examples/localize/pick/2.png} \\

\vspace{0pt} \textbf{VLM:} &
\vspace{0pt} \{\examplekey{system\_prompt}: [see Figure~\ref{prompt:moving_prompt}], \examplekey{user}: Here is a reflection on the previous failed attempt and some suggestions: [reflection]. Now, complete this task: from the items..., \examplekey{assistant}: \{point\_2d: \exampleoutput{[835, 555]}\}\} &
\vspace{0pt} \includegraphics[width=\linewidth]{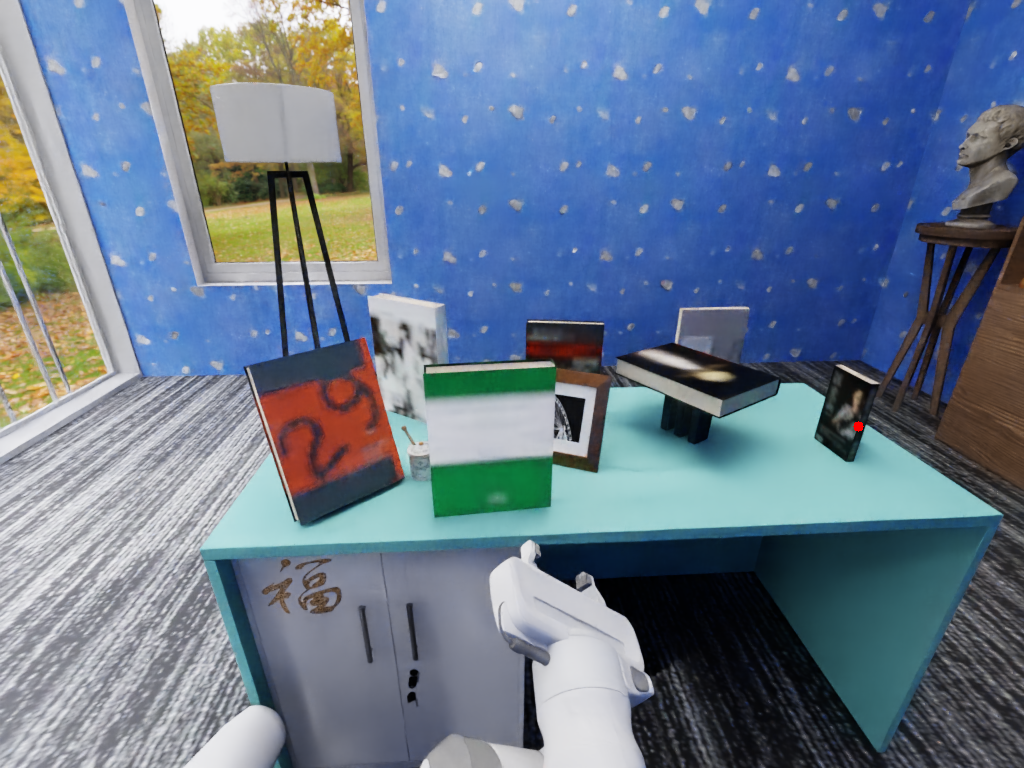} \\

\vspace{0pt} \textbf{ENV:} &
\vspace{0pt} \{\examplekey{moved}: false, \examplekey{success}: true\} &
\vspace{0pt} \includegraphics[width=\linewidth]{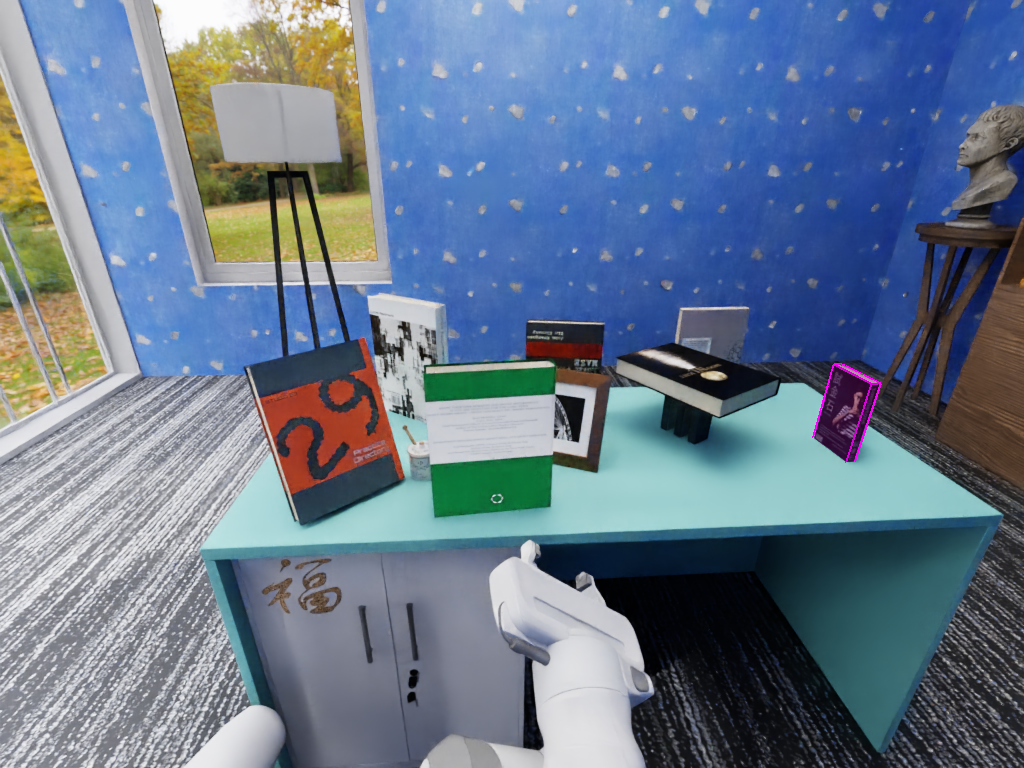} \\
\end{tabular}
\end{tcolorbox}

\caption{Pick localization example (subsequent attempts and successful execution).}
\label{fig:pick_localization_example_part2}
\end{figure*}


\begin{figure*}[t]
\centering
\begin{tcolorbox}[
    colback=lightgray!5,
    colframe=darkgray,
    arc=1mm,
    boxrule=.5pt,
    width=\textwidth
]
\begin{tabular}{p{0.08\textwidth} p{0.65\textwidth} p{0.2\textwidth}}
\vspace{0pt} \textbf{ENV:} &
\vspace{0pt} \{\examplekey{task\_instruction}: position the book in row 1, column 3 of the shelf. \examplekey{category}: place\} &
\vspace{0pt} \includegraphics[width=\linewidth]{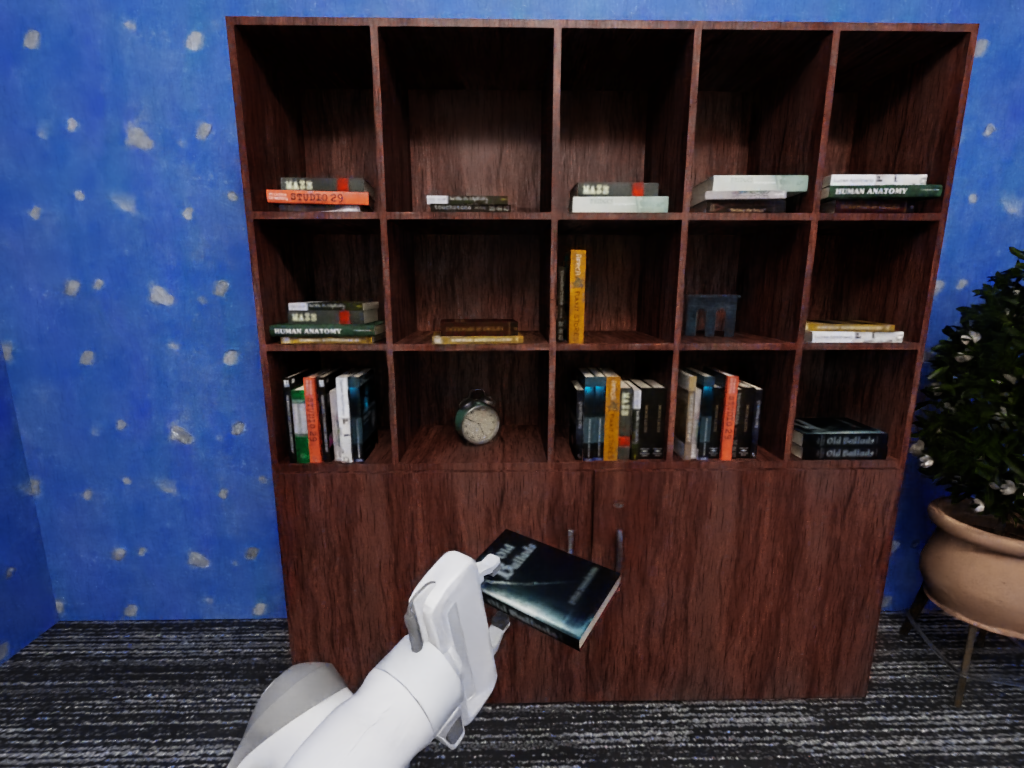} \\

\vspace{0pt} \textbf{VLM:} &
\vspace{0pt} \{\examplekey{system\_prompt}: [see Figure~\ref{prompt:grounding_prompt}], \examplekey{user}: position the book in row 1, column 3 of the shelf. \examplekey{assistant}: \{point\_2d: \exampleoutput{[640, 224]}\}\} &
\vspace{0pt} \includegraphics[width=\linewidth]{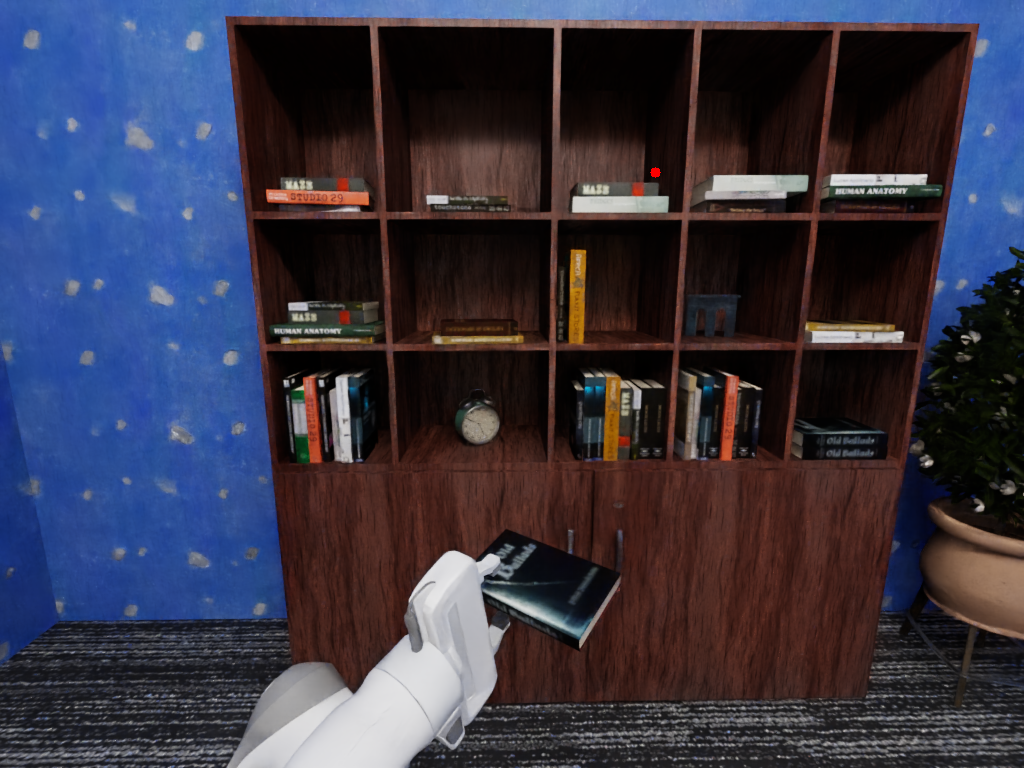} \\

\vspace{0pt} \textbf{ENV:} &
\vspace{0pt} \{\examplekey{moved}: false, \examplekey{success}: true\} &
\vspace{0pt} \includegraphics[width=\linewidth]{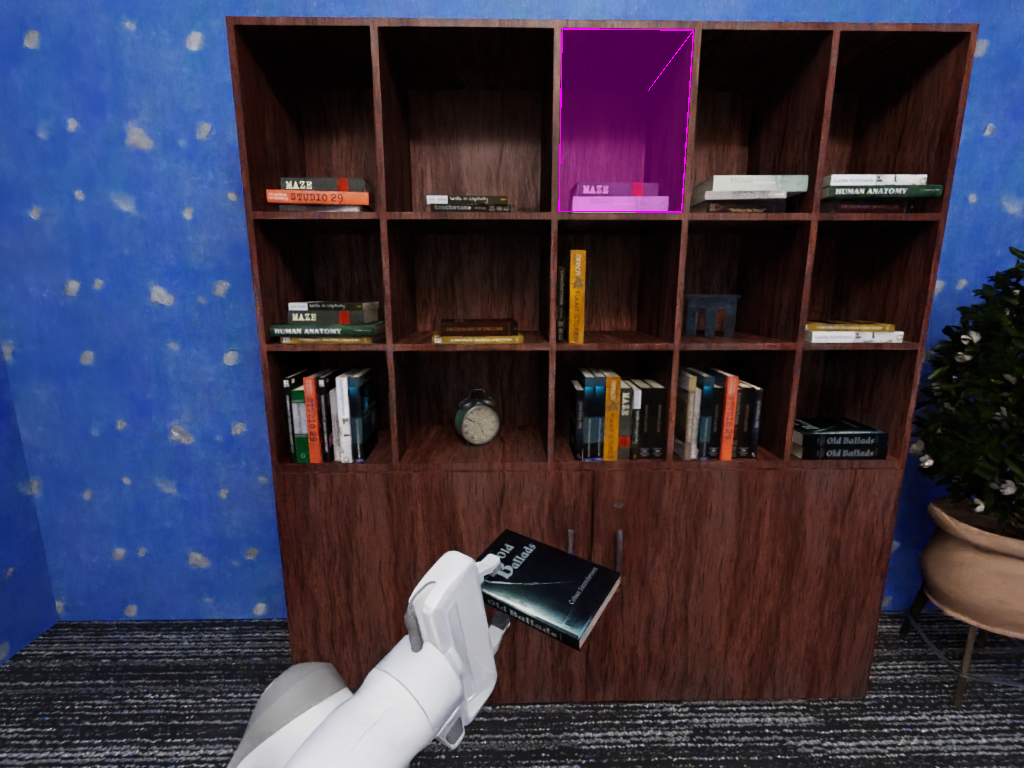} \\
\end{tabular}
\end{tcolorbox}

\caption{Place localization example without reflection, where the prompt simplified for better demonstration.}
\label{fig:place_localization_example}
\end{figure*}


\begin{figure*}[t]
\centering
\begin{tcolorbox}[
    colback=lightgray!5,
    colframe=darkgray,
    arc=1mm,
    boxrule=.5pt,
    width=\textwidth
]
\begin{tabular}{p{0.08\textwidth} p{0.65\textwidth} p{0.2\textwidth}}
\vspace{0pt} \textbf{ENV:} &
\vspace{0pt} \{\examplekey{task\_instruction}: from the items on the table, find a book more than 120cm away from you, and fetch it. \examplekey{category}: pick\} &
\vspace{0pt} \includegraphics[width=\linewidth]{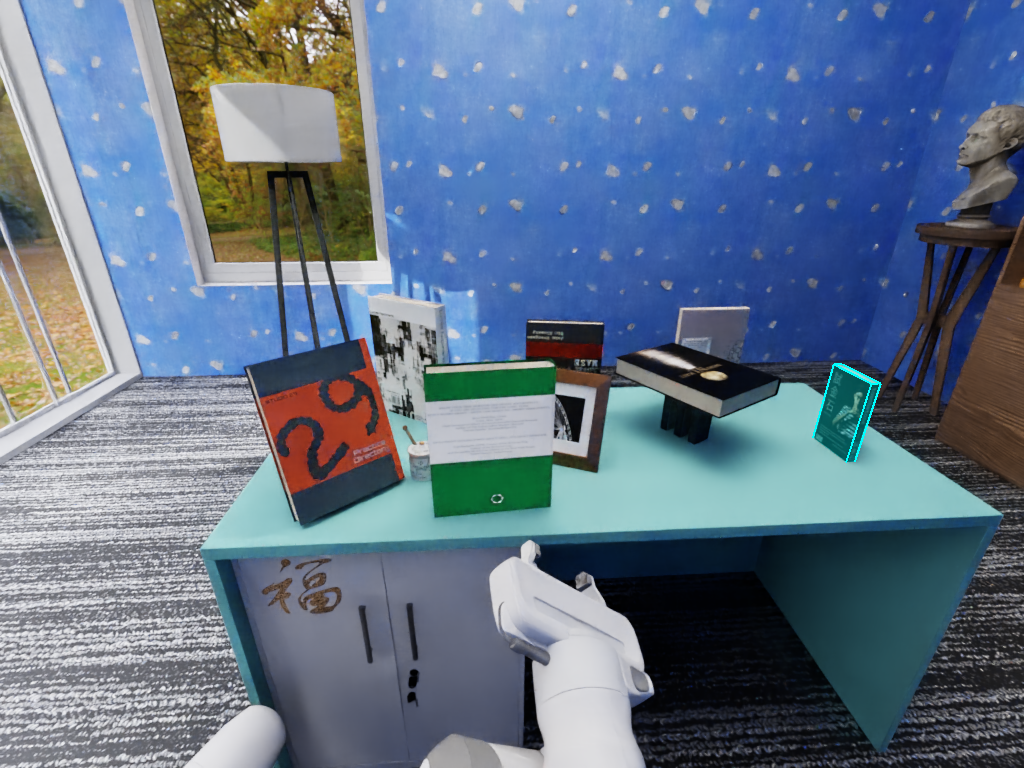} \\

\vspace{0pt} \textbf{VLM:} &
\vspace{0pt} \{\examplekey{system\_prompt}: [see Figure~\ref{prompt:moving_prompt}], \examplekey{user}: reach for the book that needs to be grabbed; this book is marked by a bounding box of the red color. \examplekey{assistant}: \{point\_2d: \exampleoutput{[825, 541]}, label: target book\}\} \{\examplekey{system\_prompt}: [see Figure~\ref{prompt:query_rotation_prompt}], \examplekey{user}: [task\_instruction] + The rotation delta(s) you should predict: pitch, \examplekey{assistant}: \{rotation: \exampleoutput{[5.0, null, null]}\}\} &
\vspace{0pt} \includegraphics[width=\linewidth]{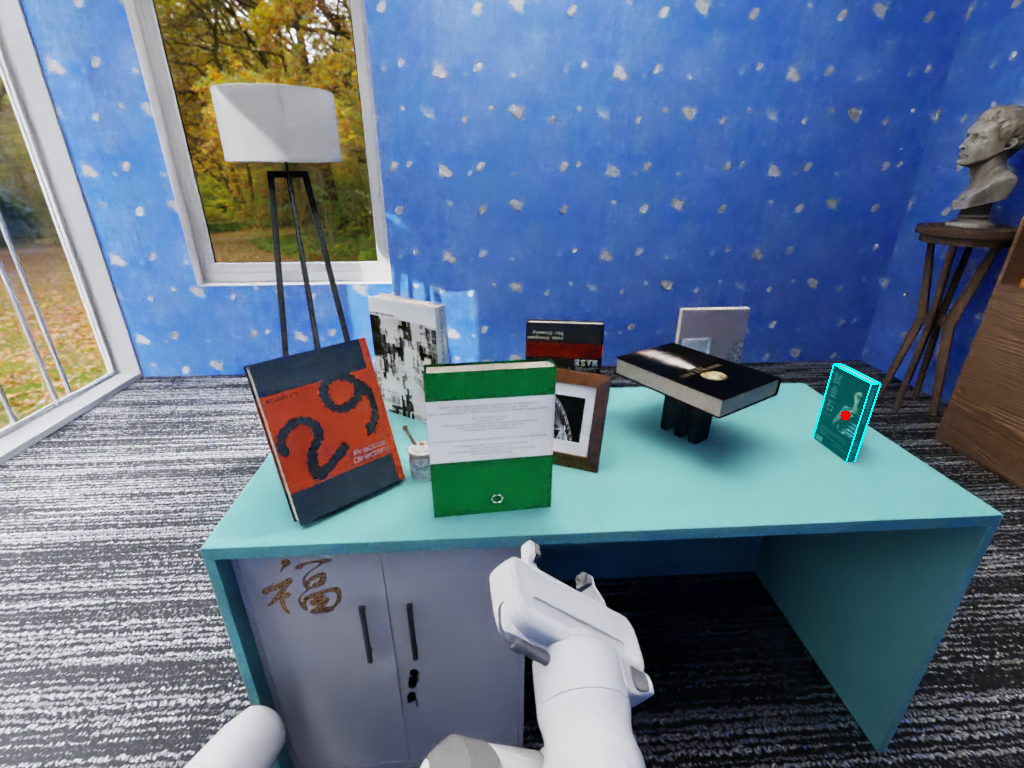} \\

\vspace{0pt} \textbf{ENV:} &
\vspace{0pt} \{\examplekey{moved}: true, \examplekey{success}: true\} &
\vspace{0pt} \includegraphics[width=\linewidth]{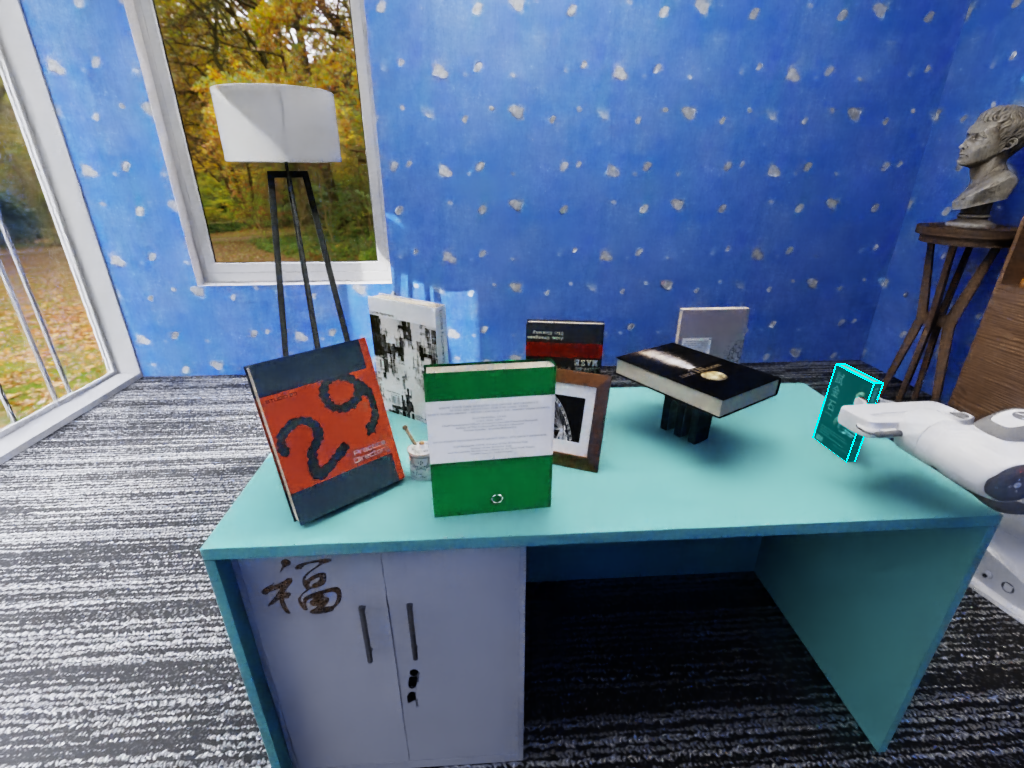} \\
\end{tabular}
\end{tcolorbox}

\caption{Pick execution example without reflection, where the prompt simplified for better demonstration.}
\label{fig:pick_execution_example}
\end{figure*}


\begin{figure*}[t]
\centering
\begin{tcolorbox}[
    colback=lightgray!5,
    colframe=darkgray,
    arc=1mm,
    boxrule=.5pt,
    width=\textwidth
]
\begin{tabular}{
p{0.08\textwidth} 
p{0.65\textwidth} 
p{0.2\textwidth}}
\vspace{0pt} \textbf{ENV:} &
\vspace{0pt} \{\examplekey{task\_instruction}: position the book in row 1, column 3 of the shelf. \examplekey{category}: place\} &
\vspace{0pt} \includegraphics[width=\linewidth]{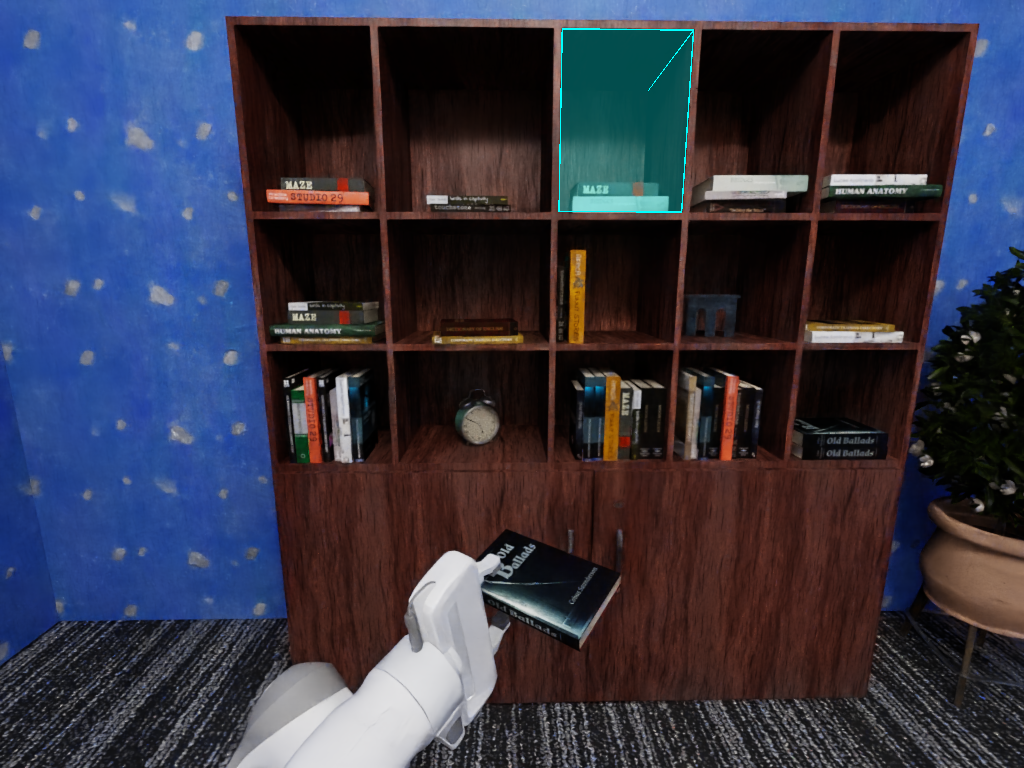} \\

\vspace{0pt} \textbf{VLM:} &
\vspace{0pt} \{\examplekey{system\_prompt}: [see Figure~\ref{prompt:moving_prompt}], \examplekey{user}: Reach for the location where the book in your hand needs to be placed. The book in your hand should be placed in the vertical manner, and the placement area is marked by a bounding box in the cyan color. \examplekey{assistant}: \{point\_2d: \exampleoutput{[587, 194]}\}\} &
\vspace{0pt} \includegraphics[width=\linewidth]{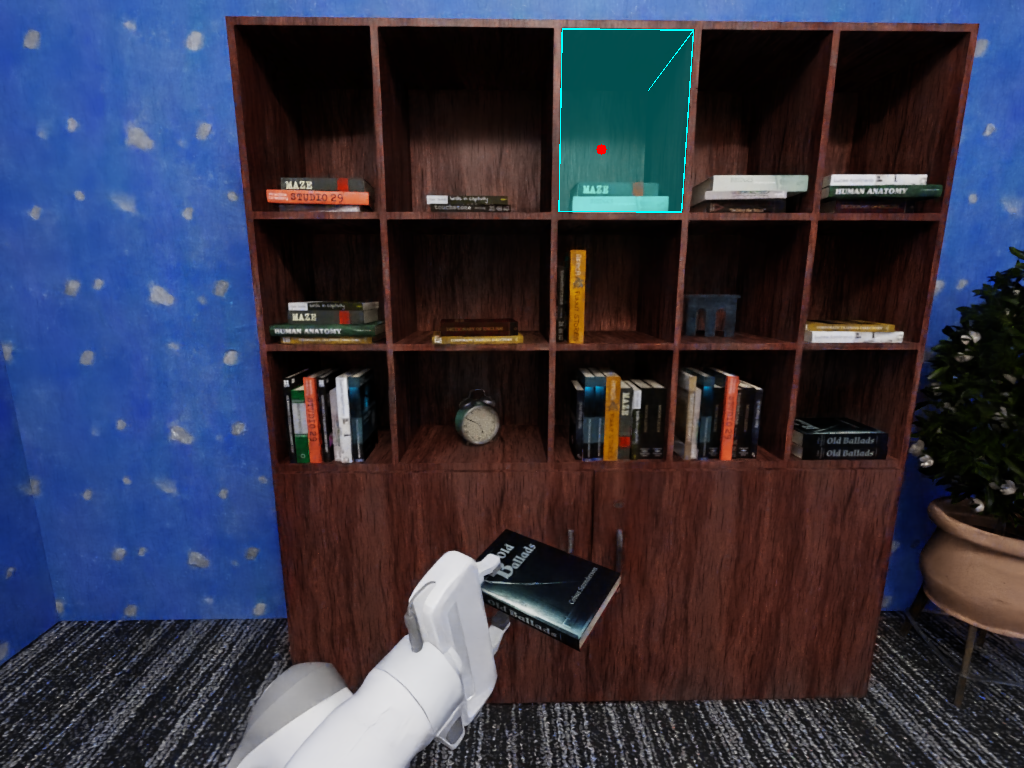} \\

\vspace{0pt} \textbf{ENV:} &
\vspace{0pt} \{\examplekey{moved}: true, \examplekey{success}: false\} &
\vspace{0pt} \includegraphics[width=\linewidth]{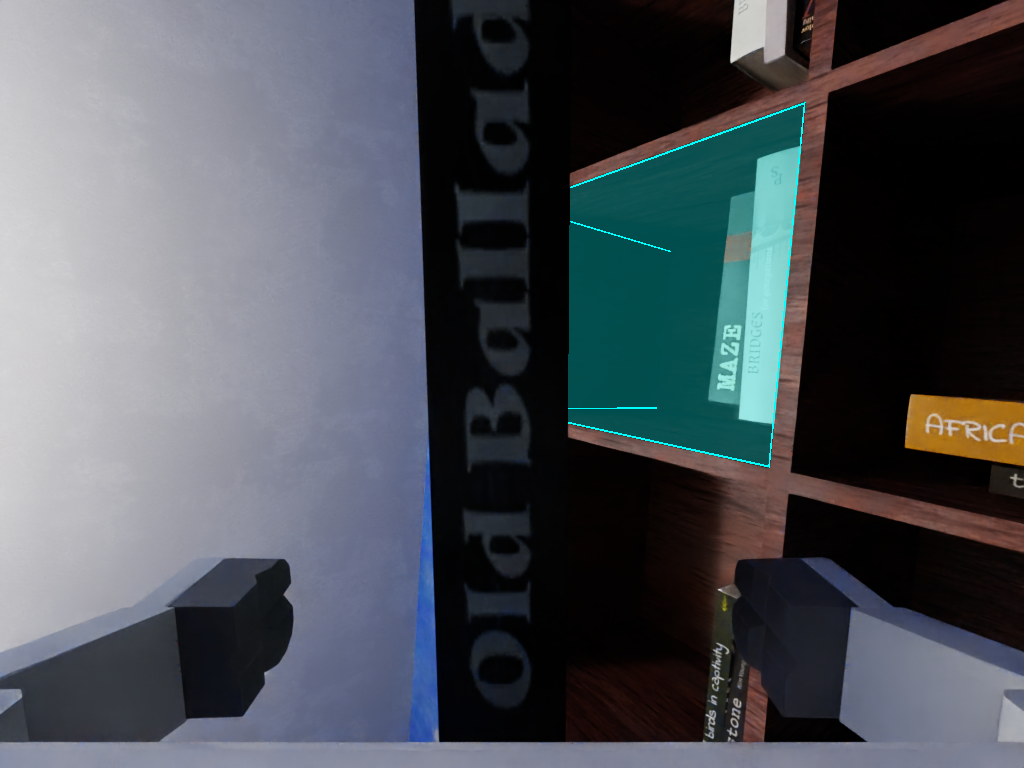} \\

\vspace{0pt} \textbf{VLM:} &
\vspace{0pt} \{\examplekey{system\_prompt}: [see Figure~\ref{prompt:reflect_grounding_prompt}], \examplekey{user}: The previous visual moving attempt for the [task\_instruction] was incorrect. The red circles in the image mark the wrong moving regions. Please carefully analyze the image and the instruction again, understand why the previous prediction was wrong, and describe what should be corrected in the next attempt. \examplekey{assistant}: The failure in the previous attempt stems from a misinterpretation of the spatial relationship between the cyan bounding box and the robot’s reachable workspace... \examplehl{Likely Cause of Failure:} Target Point Misplacement... Kinematic limits... \examplehl{Suggestions for Improving the Next Prediction:} Focus on the shelf surface, not the back wall... Prioritize reachable points...\} &
\vspace{0pt} \includegraphics[width=\linewidth]{figs/running_examples/execution/place/1.png} \includegraphics[width=\linewidth]{figs/running_examples/execution/place/2.png} \\

\vspace{0pt} \textbf{VLM:} &
\vspace{0pt} \{\examplekey{system\_prompt}: [see Figure~\ref{prompt:moving_prompt}], \examplekey{user}: Here is a reflection on the previous failed attempt and some suggestions: [reflection]. Now, complete this task: Reach for the location ..., \examplekey{assistant}: \{point\_2d: \exampleoutput{[684, 418]}\}\} &
\vspace{0pt} \includegraphics[width=\linewidth]{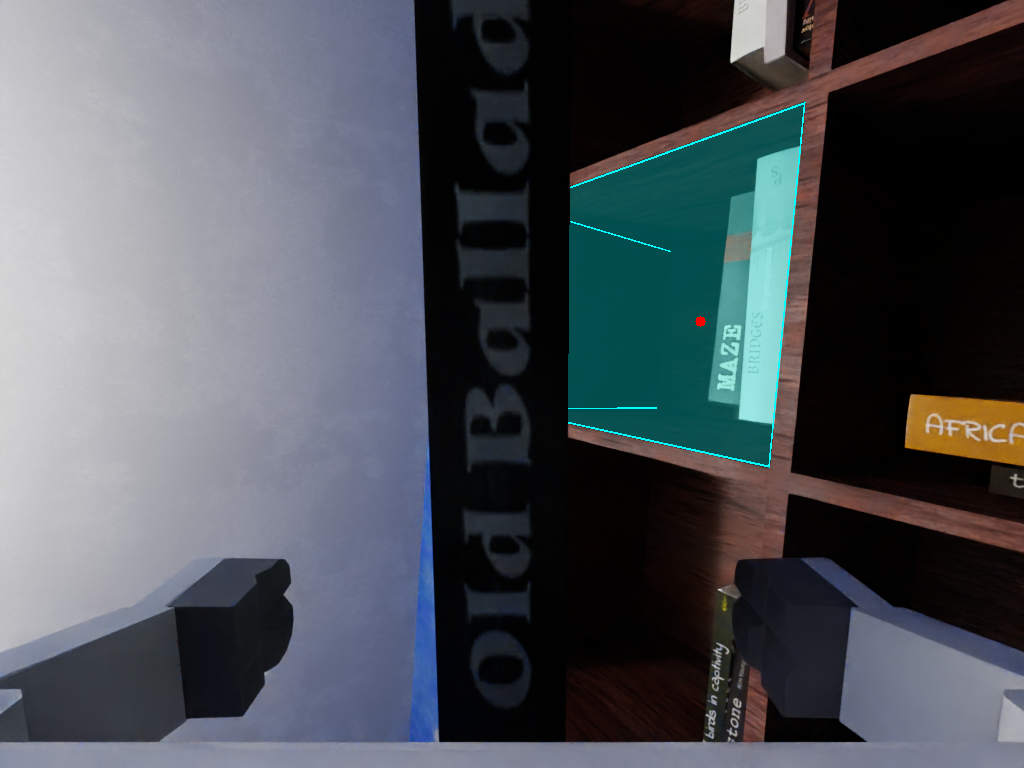} \\

\vspace{0pt} \textbf{ENV:} &
\vspace{0pt} \{\examplekey{moved}: true, \examplekey{success}: true\} &
\vspace{0pt} \includegraphics[width=\linewidth]{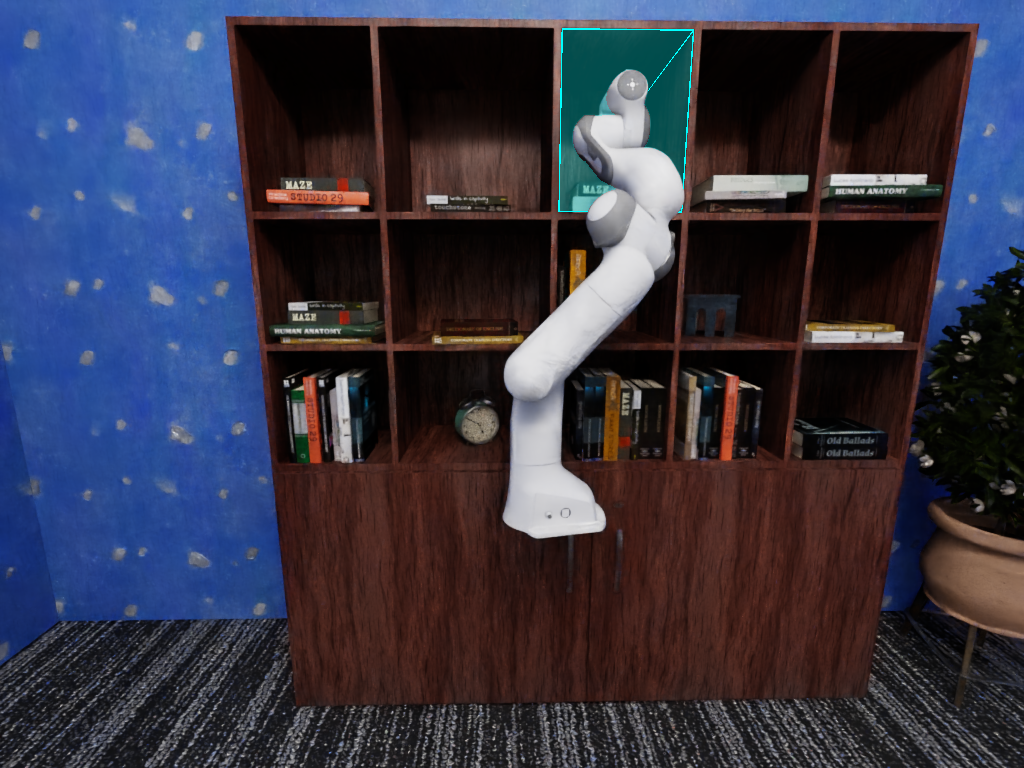} \\
\end{tabular}
\end{tcolorbox}

\caption{Place execution example with reflection, where the prompt simplified for better demonstration.}
\label{fig:place_execution_example}
\end{figure*}

\begin{table*}[t!]\small
\caption{Spatial aspects of varying granularities. \ttt{nil} indicates no input parameters are required. Note that \emph{left}, \emph{right}, \emph{front}, \emph{behind}, \emph{above}, and \emph{below} are reloaded as \emph{directional} relations in `Orientation' (\emph{cf.} \emph{positional} relations in `Relationship').}
\label{tab:spatial-aspect}
\resizebox{\linewidth}{!}{%
\begin{tabular}{
@{}
c @{\hspace{0.2cm}} c @{\hspace{0.2cm}} l @{\hspace{0.2cm}} c @{\hspace{0.2cm}} l
@{}}
\toprule
\makecell{Spatial \\ Aspect}      & \makecell{Gran-\\ularity}          & \makecell{Type}          & Input & Example Instruction            \\ \midrule
\multirow{10}{*}{\rotatebox[origin=c]{90}{{\footnotesize Attribute}}}            & \multirow{6}{*}{\rotatebox[origin=c]{90}{{\footnotesize Coarse}}}               & Small         &   \ttt{nil}    & \textit{take a small book from the table}                          \\
                     &                      & Medium        &   \ttt{nil}    & \textit{take a medium-sized book}                   \\
                     &                      & Large         &   \ttt{nil}    & \textit{take a large book}                          \\
\multicolumn{1}{l}{} &                      & Empty         &   \ttt{nil}    & \textit{place the book in an empty slot}                               \\
\multicolumn{1}{l}{} &                      & NonEmpty      &   \ttt{nil}    & \textit{place the book in a partly occupied slot}                              \\
\multicolumn{1}{l}{} & \multicolumn{1}{l}{} & Emptiest  &   \ttt{nil}    & \textit{place the book in the emptiest slot}                               \\ \cdashlinelr{2-5}
                     & \multirow{4}{*}{\rotatebox[origin=c]{90}{{\footnotesize Fine}}}                     & Height        &   (\ttt{\#},)    & \textit{take a book around \hl{20} centimeters high}            \\
                     &                      & Width         &   (\ttt{\#},)    & \textit{place the book in a slot around \hl{45} centimeters wide}           \\
\multicolumn{1}{l}{} & \multicolumn{1}{l}{} & Index1D       &   (\ttt{\#},)    & \textit{place the book at row \hl{2} of the shelf}                            \\
\multicolumn{1}{l}{} & \multicolumn{1}{l}{} & Index2D       &   (\ttt{\#},\ttt{\#})    & \textit{place the book at row \hl{2}, column \hl{3} of the shelf}                               \\ \midrule
\multirow{8}{*}{\rotatebox[origin=c]{90}{{\footnotesize Distance}}}            & \multirow{4}{*}{\rotatebox[origin=c]{90}{{\footnotesize Coarse}}}               & Closest       &   \ttt{nil}    & \textit{take a book among the books closest to you}                     \\
                     &                      & Farthest      &   \ttt{nil}    & \textit{take a book among the books farthest from you}                  \\ 
                  & & LessThan      &   (\ttt{\#},)    & \textit{place the book in a slot within \hl{1.5} meters of you}               \\
\multicolumn{1}{l}{} &                      & MoreThan      &   (\ttt{\#},)    & \textit{take a book more than \hl{1.5} meters away from you} \\ \cdashlinelr{2-5}
                     &  \multirow{4}{*}{\rotatebox[origin=c]{90}{{\footnotesize Fine}}}                    & RankClosest   &   (\ttt{\#},)    & \textit{take a book among the books \hl{second} closest to you}                 \\
\multicolumn{1}{l}{} &                      & RankFarthest  &   (\ttt{\#},)    & \textit{take a book among the books \hl{second} farthest from you}               \\ 
&                      & EqualTo      &   (\ttt{\#},)    & \textit{take a book about \hl{1.5} meters away from you} \\
                     &                      & Range         &   (\ttt{\#},\ttt{\#})    & \textit{take a book \hl{1.5} to \hl{2} meters away from you}    \\ \midrule
\multirow{11}{*}{\rotatebox[origin=c]{90}{{\footnotesize Relationship}}}         & \multirow{10}{*}{\rotatebox[origin=c]{90}{{\footnotesize Coarse}}}               & Left          &   \ttt{nil}    & \textit{take a book on the left of the table}                               \\
                     &                      & Right         &   \ttt{nil}    & \textit{place the book in a slot on the left of the shelf}                               \\
                     &                      & Front         &   \ttt{nil}    &  \textit{take a book at the front of the table}                              \\
                     &                      & Behind        &   \ttt{nil}    & \textit{take a book at the back of the table}                               \\
\multicolumn{1}{l}{} & \multicolumn{1}{l}{} & Upper         &   \ttt{nil}    & \textit{place the book in a slot in the upper part of the shelf}                               \\
\multicolumn{1}{l}{} & \multicolumn{1}{l}{} & Lower         &   \ttt{nil}    & \textit{place the book in a slot in the lower part of the shelf}                               \\
                     &                      & LeftMost      &   \ttt{nil}    &  \textit{take the leftmost book from the table}                              \\
                     &                      & RightMost     &   \ttt{nil}    & \textit{place the the book in a leftmost slot on the shelf}                                \\ \cdashlinelr{2-5}
                     &   \multirow{3}{*}{\rotatebox[origin=c]{90}{{\footnotesize Fine}}}                 & RankLeftMost  &   (\ttt{\#},)    & \textit{take the \hl{second} leftmost book on the table}                               \\
                     &                      & RankRightMost &   (\ttt{\#},)    & \textit{place the book in a \hl{second} rightmost slot on the shelf}                               \\ 
                     &                      & Between &   (\ttt{\#},\ttt{\#})    & \textit{place the book between the \hl{alarm clock} and the \hl{succulents}}                               \\\midrule
\multirow{15}{*}{\rotatebox[origin=c]{90}{{\footnotesize Orientation}}}& \multirow{9}{*}{\rotatebox[origin=c]{90}{{\footnotesize Coarse}}}               & Flat          &   \ttt{nil}    & \textit{take a flat-lying book from the table}                               \\
                     &                      & Vertical      &   \ttt{nil}    & \textit{place the book upright on the shelf}                                \\
                     &                      & Tilted        &   \ttt{nil}    & \textit{place the book at a tilt on the shelf}                                \\ 
                     &                      & Left      &   \ttt{nil}    & \textit{take a book to your left}                                \\ 
                     &                      & Right      &   \ttt{nil}    & \textit{place the book in a slot to your right}                                \\
                     &                      & Front         &   \ttt{nil}    &  \textit{place the book in front of the teddy bear}                                       \\
                     &                      & Behind        &   \ttt{nil}    & \textit{take a book behind the picture frame}                               \\
\multicolumn{1}{l}{} & \multicolumn{1}{l}{} & Above         &   \ttt{nil}    & \textit{place the book in a slot above the picture frame}                               \\
\multicolumn{1}{l}{} & \multicolumn{1}{l}{} & Below         &   \ttt{nil}    & \textit{place the book in a slot below the picture frame}                               \\                     
                     \cdashlinelr{2-5}
\multicolumn{1}{l}{} & \multirow{6}{*}{\rotatebox[origin=c]{90}{{\footnotesize Fine}}}            & DirectLeft         &   \ttt{nil}    & \textit{place the book immediately to the left of the alarm clock}                               \\  
&& DirectRight         &   \ttt{nil}    & \textit{place the book immediately to the right of the succulents}  \\
&& DirectAbove         &   \ttt{nil}    & \textit{place the book in a slot directly above the alarm clock}                               \\
&& DirectBelow         &   \ttt{nil}    & \textit{place the book in a slot directly below the picture frame} \\ 
&& ClockPosition &   (\ttt{\#},)    & \textit{place the book in a slot to your \hl{6} o'clock}                                \\
\multicolumn{1}{l}{} &                      & TiltDegree    &   (\ttt{\#},)    & \textit{place the book at a tilt angle of about \hl{30} degrees}                               \\ \bottomrule
\end{tabular}
} 
\end{table*}

\begin{table*}\small
\caption{%
Example instruction families of \emph{pick} tasks. The outermost \ttt{pick}($\cdot$) is discarded for simplicity. \ttt{unique}($\cdot$) ensures a unique item from the input set. \ttt{TABLE} returns all items in the tabletop scene.
}\label{tab:task-family-pick}
\resizebox{\linewidth}{!}{%
\begin{tabular}{
@{}
c  @{\hspace{0.2cm}} c @{\hspace{0.2cm}} c @{\hspace{0.2cm}} l @{\hspace{0.2cm}} l @{\hspace{0.2cm}} m{0.6\linewidth} 
@{}}
\toprule
 $S$ & $F$ & $O$ & $R$ & $I$ & Example Program \\
\midrule
\multirow{5}{*}{\rotatebox[origin=c]{90}{{\footnotesize Attribute}}} 
&  & & Small & &\multirow{3}{*}{\rotatebox[origin=c]{0}{{\footnotesize {filterAttr\hl{\$R}}({filterBook}(TABLE))}}} \\
&  & & Large & &\\
&  & & Medium & &\\ \cdashlinelr{4-6}
&  & & Height & \multirow{2}{*}{\footnotesize\emph{float}} &
\multirow{2}{*}{\rotatebox[origin=c]{0}{{\footnotesize {filterAttr\hl{\$R}}($I$, {filterBook}(TABLE))}}} \\
&  & & Width & &\\
\midrule
\multirow{6}{*}{\rotatebox[origin=c]{90}{{\footnotesize Distance}}} 
&  & \multirow{6}{*}{\makecell[c]{\footnotesize
\emph{viewer}\\\emph{distant obj.}\\\emph{near obj.}
}}
& RankClosest & \multirow{2}{*}{\footnotesize\emph{int}} &\multirow{2}{*}{\rotatebox[origin=c]{0}{{\footnotesize {filterDist\hl{\$R}}($I$, {filterBook}(TABLE), $O$)}}} \\
&  & & RankFarthest & &\\  \cdashlinelr{4-6}
&  & &LessThan & \multirow{4}{*}{\footnotesize\emph{list}} & \multirow{4}{*}{\rotatebox[origin=c]{0}{{\footnotesize {filterDist\hl{\$R}}($I$, {filterBook}(TABLE), $O$)}}} \\
&  & &MoreThan & &\\
&  & &EqualTo & &\\
&  & &Range & &\\
\midrule
\multirow{11}{*}{\rotatebox[origin=c]{90}{{\footnotesize Relationship}}} 
& \multirow{7}{*}{\rotatebox[origin=c]{90}{{\footnotesize intrinsic}}} & \multirow{7}{*}{\makecell[c]{\footnotesize\emph{table}}}
& Left & \multirow{2}{*}{\footnotesize\emph{}} &\multirow{4}{*}{\rotatebox[origin=c]{0}{{\footnotesize {filterRel\hl{\$R}}({filterBook}(TABLE), $O$)}}} \\
&  & &Right & &\\
&  & &Front & &\\
&  & &Behind & &\\ \cdashlinelr{4-6}
&  & &RankLeftMost & \multirow{2}{*}{\footnotesize\emph{int}} & \multirow{2}{*}{\rotatebox[origin=c]{0}{{\footnotesize {filterRel\hl{\$R}}($I$, {filterBook}(TABLE), $O$)}}} \\
&  & &RankRightMost & &\\ \cdashlinelr{4-6}
&  & &Between & \multirow{1}{*}{\footnotesize\emph{list}} & \footnotesize\makecell[l]{{filterRel\hl{\$R}}({filterBook}(TABLE), \\ \quad filter($I_1$, TABLE), filter($I_2$, TABLE))} \\
\cmidrule{2-6} 
& \multirow{4}{*}{\rotatebox[origin=c]{90}{{\footnotesize relative}}} & \multirow{4}{*}{\makecell[c]{\footnotesize\emph{viewer}}}
& Left & \multirow{2}{*}{\footnotesize\emph{}} &\multirow{2}{*}{\rotatebox[origin=c]{0}{{\footnotesize {filterRel\hl{\$R}}({filterBook}(TABLE), $O$)}}} \\
&  & &Right & &\\ \cdashlinelr{4-6}
&  & &RankLeftMost & \multirow{2}{*}{\footnotesize\emph{int}} & \multirow{2}{*}{\rotatebox[origin=c]{0}{{\footnotesize {filterRel\hl{\$R}}($I$, {filterBook}(TABLE), $O$)}}} \\
&  & &RankRightMost & &\\ 
\midrule
\multirow{19}{*}{\rotatebox[origin=c]{90}{{\footnotesize Orientation}}} 
& \multirow{11}{*}{\rotatebox[origin=c]{90}{{\footnotesize intrinsic}}} & \multirow{11}{*}{\makecell[c]{\footnotesize\emph{viewer}\\\emph{oriented}}}
& Left & \multirow{2}{*}{\footnotesize\emph{}} &\multirow{4}{*}{\rotatebox[origin=c]{0}{{\footnotesize {filterOri\hl{\$R}}({filterBook}(TABLE), $O$)}}} \\
&  & &Right & &\\
&  & &Front & &\\
&  & &Behind & &\\ \cdashlinelr{4-6}
&  & &RankLeftMost & \multirow{2}{*}{\footnotesize\emph{int}} & \multirow{2}{*}{\rotatebox[origin=c]{0}{{\footnotesize {filterOri\hl{\$R}}($I$, {filterBook}(TABLE), $O$)}}} \\
&  & &RankRightMost & &\\ \cdashlinelr{4-6}
&  & &Flat & \multirow{3}{*}{\footnotesize\emph{}} & \multirow{3}{*}{\footnotesize\makecell[l]{{filterOri\hl{\$R}}({filterBook}(TABLE))}} \\
&  & &Vertical & &\\
&  & &Tilted & &\\ \cdashlinelr{4-6}
&  & &ClockPosition & \multirow{1}{*}{\footnotesize\emph{int}} & \multirow{1}{*}{\footnotesize\makecell[l]{{filterOri\hl{\$R}}($I$, {filterBook}(TABLE), $O$)}} \\ \cdashlinelr{4-6}
&  & &TiltDegree & \multirow{1}{*}{\footnotesize\emph{float}} & \multirow{1}{*}{\footnotesize\makecell[l]{{filterOri\hl{\$R}}($I$, {filterBook}(TABLE), $O$)}} \\
\cmidrule{2-6} 
& \multirow{8}{*}{\rotatebox[origin=c]{90}{{\footnotesize relative}}} & \multirow{8}{*}{\makecell[c]{\footnotesize\emph{viewer}\\\emph{non-oriented}}}
& Left & \multirow{2}{*}{\footnotesize\emph{}} &\multirow{4}{*}{\rotatebox[origin=c]{0}{{\footnotesize {filterOri\hl{\$R}}({filterBook}(TABLE), $O$)}}} \\
&  & &Right & &\\
&  & &Front & &\\
&  & &Behind & &\\ \cdashlinelr{4-6}
&  & &RankLeftMost & \multirow{2}{*}{\footnotesize\emph{int}} & \multirow{2}{*}{\rotatebox[origin=c]{0}{{\footnotesize {filterOri\hl{\$R}}($I$, {filterBook}(TABLE), $O$)}}} \\
&  & &RankRightMost & &\\ \cdashlinelr{4-6}
&  & &ClockPosition & \multirow{1}{*}{\footnotesize\emph{int}} & \multirow{1}{*}{\footnotesize\makecell[l]{{filterOri\hl{\$R}}($I$, {filterBook}(TABLE), $O$)}} \\ \cdashlinelr{4-6}
&  & &TiltDegree & \multirow{1}{*}{\footnotesize\emph{float}} & \multirow{1}{*}{\footnotesize\makecell[l]{{filterOri\hl{\$R}}($I$, {filterBook}(TABLE), $O$)}} \\
\bottomrule
\end{tabular}
} 
\end{table*}

\begin{table*}\small
\caption{%
Example instruction families of \emph{place} tasks. The outermost \ttt{place}($\cdot$) is discarded for simplicity. \ttt{unique}($\cdot$) ensures a unique item from the input set. \ttt{SHELF} returns all shelf-scene items.
}\label{tab:task-family-place}
\resizebox{\linewidth}{!}{%
\begin{tabular}{
@{}
c  @{\hspace{0.2cm}} c @{\hspace{0.2cm}} c @{\hspace{0.2cm}} l @{\hspace{0.2cm}} l @{\hspace{0.2cm}} m{0.6\linewidth} 
@{}}
\toprule
 $S$ & $F$ & $O$ & $R$ & $I$ & Example Program \\
\midrule
\multirow{5}{*}{\rotatebox[origin=c]{90}{{\footnotesize Attribute}}} 
&  & & Small & &\multirow{3}{*}{\rotatebox[origin=c]{0}{{\footnotesize {filterAttr\hl{\$R}}({filterSlot}(SHELF))}}} \\
&  & & Large & &\\
&  & & Medium & &\\ \cdashlinelr{4-6}
&  & & Height & \multirow{2}{*}{\footnotesize\emph{float}} &
\multirow{2}{*}{\rotatebox[origin=c]{0}{{\footnotesize {filterAttr\hl{\$R}}($I$, {filterSlot}(SHELF))}}} \\
&  & & Width & &\\
\midrule
\multirow{6}{*}{\rotatebox[origin=c]{90}{{\footnotesize Distance}}} 
&  & \multirow{6}{*}{\makecell[c]{\footnotesize
\emph{viewer}\\\emph{distant obj.}\\\emph{near obj.}
}}
& RankClosest & \multirow{2}{*}{\footnotesize\emph{int}} &\multirow{2}{*}{\rotatebox[origin=c]{0}{{\footnotesize {filterDist\hl{\$R}}($I$, {filterSlot}(SHELF), $O$)}}} \\
&  & & RankFarthest & &\\  \cdashlinelr{4-6}
&  & & LessThan & \multirow{4}{*}{\footnotesize\emph{list}} & \multirow{4}{*}{\makecell[l]{\footnotesize {filterDist\hl{\$R}}($I$, {filterSlot}(SHELF), $O$) \\ \footnotesize {filterDist\hl{\$R}}($I$, {filterSpace}(SHELF), $O$)}} \\
&  & &MoreThan & &\\
&  & &EqualTo & &\\
&  & &Range & &\\
\midrule
\multirow{13}{*}{\rotatebox[origin=c]{90}{{\footnotesize Relationship}}} 
& \multirow{7}{*}{\rotatebox[origin=c]{90}{{\footnotesize intrinsic}}} & \multirow{7}{*}{\makecell[c]{\footnotesize\emph{shelf}}}
& Left & \multirow{2}{*}{\footnotesize\emph{}} &\multirow{4}{*}{\makecell[l]{\footnotesize {filterRel\hl{\$R}}({filterSlot}(SHELF), $O$) \\ \footnotesize {filterRel\hl{\$R}}({filterSpace}(SHELF), $O$)}} \\
&  & &Right & &\\
&  & &Upper & &\\
&  & &Lower & &\\ \cdashlinelr{4-6}
&  & &RankLeftMost & \multirow{2}{*}{\footnotesize\emph{int}} & \multirow{2}{*}{\makecell[l]{\footnotesize {filterRel\hl{\$R}}($I$, {filterSlot}(SHELF), $O$)}} \\
&  & &RankRightMost & &\\ \cdashlinelr{4-6}
&  & &Between & \multirow{1}{*}{\footnotesize\emph{list}} & \footnotesize\makecell[l]{{filterRel\hl{\$R}}({filterSpace}(SHELF), \\ \quad filter($I_1$, SHELF), filter($I_2$, SHELF))} \\
\cmidrule{2-6} 
& \multirow{4}{*}{\rotatebox[origin=c]{90}{{\footnotesize relative}}} & \multirow{4}{*}{\makecell[c]{\footnotesize\emph{viewer}}}
& Left & \multirow{2}{*}{\footnotesize\emph{}} &\multirow{2}{*}{\makecell[l]{\footnotesize {filterRel\hl{\$R}}({filterSlot}(SHELF), $O$) \\ \footnotesize {filterRel\hl{\$R}}({filterSpace}(SHELF), $O$)}} \\
&  & &Right & &\\ \cdashlinelr{4-6}
&  & &RankLeftMost & \multirow{2}{*}{\footnotesize\emph{int}} & \multirow{2}{*}{\rotatebox[origin=c]{0}{{\footnotesize {filterRel\hl{\$R}}($I$, {filterSlot}(SHELF), $O$)}}} \\
&  & &RankRightMost & &\\ 
\midrule
\multirow{20}{*}{\rotatebox[origin=c]{90}{{\footnotesize Orientation}}} 
& \multirow{7}{*}{\rotatebox[origin=c]{90}{{\footnotesize intrinsic}}} & \multirow{7}{*}{\makecell[c]{\footnotesize\emph{viewer}\\\emph{oriented}}}
& Left & \multirow{2}{*}{\footnotesize\emph{}} & \multirow{4}{*}{\makecell[l]{\footnotesize {filterOri\hl{\$R}}({filterSlot}(SHELF), $O$) \\ \footnotesize {filterOri\hl{\$R}}({filterSpace}(SHELF), $O$)}} \\
&  & &Right & &\\
&  & &Front & &\\
&  & &Behind & &\\ \cdashlinelr{4-6}
&  & &RankLeftMost & \multirow{2}{*}{\footnotesize\emph{int}} & \multirow{2}{*}{\rotatebox[origin=c]{0}{{\footnotesize {filterOri\hl{\$R}}($I$, {filterSlot}(SHELF), $O$)}}} \\
&  & &RankRightMost & &\\ \cdashlinelr{4-6}
&  & &ClockPosition & \multirow{1}{*}{\footnotesize\emph{int}} & \multirow{1}{*}{\footnotesize\makecell[l]{{filterOri\hl{\$R}}($I$, {filterSpace}(SHELF), $O$)}} \\
\cmidrule{2-6} 
& \multirow{7}{*}{\rotatebox[origin=c]{90}{{\footnotesize relative}}} & \multirow{7}{*}{\makecell[c]{\footnotesize\emph{viewer}\\\emph{non-oriented}}}
& Left & \multirow{2}{*}{\footnotesize\emph{}} & \multirow{4}{*}{\makecell[l]{\footnotesize {filterOri\hl{\$R}}({filterSlot}(SHELF), $O$) \\ \footnotesize {filterOri\hl{\$R}}({filterSpace}(SHELF), $O$)}} \\
&  & &Right & &\\
&  & &Front & &\\
&  & &Behind & &\\ \cdashlinelr{4-6}
&  & &RankLeftMost & \multirow{2}{*}{\footnotesize\emph{int}} & \multirow{2}{*}{\rotatebox[origin=c]{0}{{\footnotesize {filterOri\hl{\$R}}($I$, {filterSlot}(SHELF), $O$)}}} \\
&  & &RankRightMost & &\\ \cdashlinelr{4-6}
&  & &ClockPosition & \multirow{1}{*}{\footnotesize\emph{int}} & \multirow{1}{*}{\footnotesize\makecell[l]{{filterOri\hl{\$R}}($I$, {filterSpace}(SHELF), $O$)}} \\
\cmidrule{2-6}
& \multirow{6}{*}{\rotatebox[origin=c]{90}{{\footnotesize absolute}}} & \multirow{7}{*}{\makecell[c]{\footnotesize\emph{distant obj.}\\\emph{near obj.}}} & Flat & \multirow{4}{*}{\footnotesize\emph{float}} & \multirow{4}{*}{\makecell[l]{\footnotesize {placeOri\hl{\$R}}($I$, unique({filterSlot}(SHELF))) \\ \footnotesize {placeOri\hl{\$R}}($I$, unique({filterSpace}(SHELF)))}} \\
&  & &Vertical & &\\
&  & &Tilted & & \\
&  & &TiltDegree & & \\ \cdashlinelr{4-6}
&  & &Above & \multirow{2}{*}{\footnotesize\emph{}} & \multirow{2}{*}{\makecell[l]{\footnotesize {filterOri\hl{\$R}}({filterSlot}(SHELF), $O$) \\ \footnotesize {filterOri\hl{\$R}}({filterSpace}(SHELF), $O$)}} \\
&  & &Below & &\\
\bottomrule
\end{tabular}
} 
\end{table*}

\subsection{Asset Visualization}\label{sup:asset}

We visualize primary assets of \model, including near reference objects in Table~\ref{tab:reference-near}, distant reference objects in Table~\ref{tab:reference-distant}, tables in Table~\ref{tab:table}, shelf layouts in Table~\ref{tab:shelf}, shelf textures in Table~\ref{tab:shelf-texture}, manipulable books in Table~\ref{tab:book}, and support ornaments in Table~\ref{tab:bookend}.

\begin{table*}[t!]\small
	\centering
	{\setlength{\tabcolsep}{0pt} 
	\makebox[\linewidth]{\resizebox{1.\linewidth}{!}{%
    \begin{tabular}{@{} 
            m{0.24\linewidth} @{\hspace{0.01\linewidth}}  
            m{0.24\linewidth} @{\hspace{0.01\linewidth}}  
            m{0.24\linewidth} @{\hspace{0.01\linewidth}}  
            m{0.24\linewidth} 
            @{}}
	\raisebox{-\totalheight}{\includegraphics[width=\linewidth]{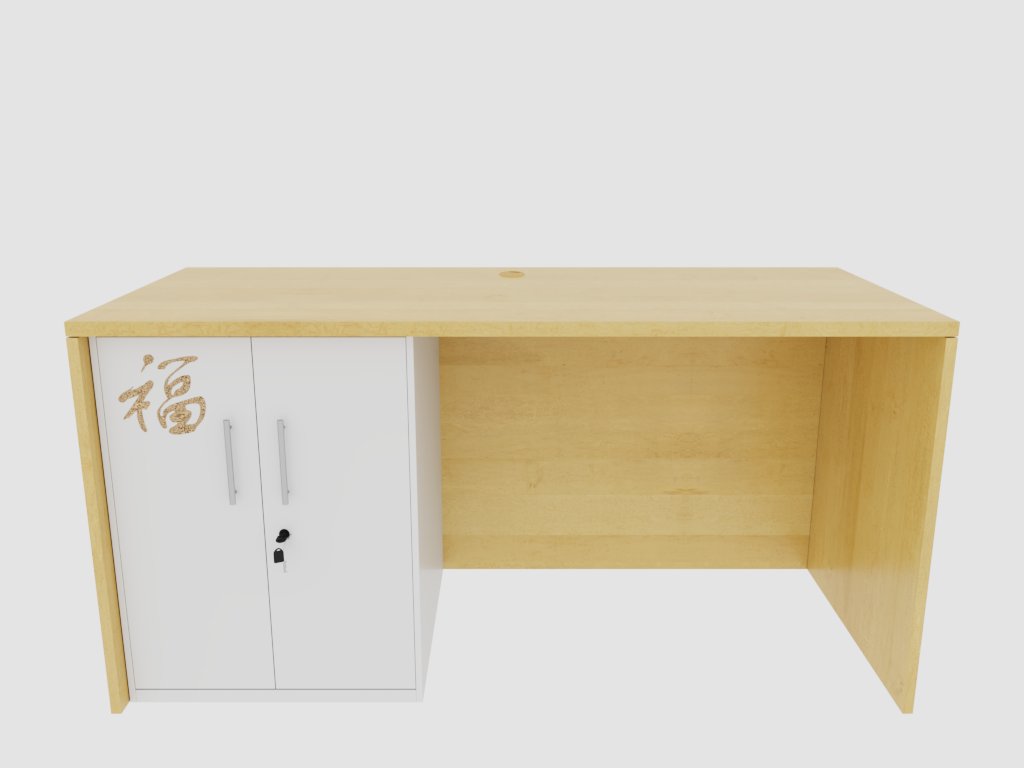}} &
	\raisebox{-\totalheight}{\includegraphics[width=\linewidth]{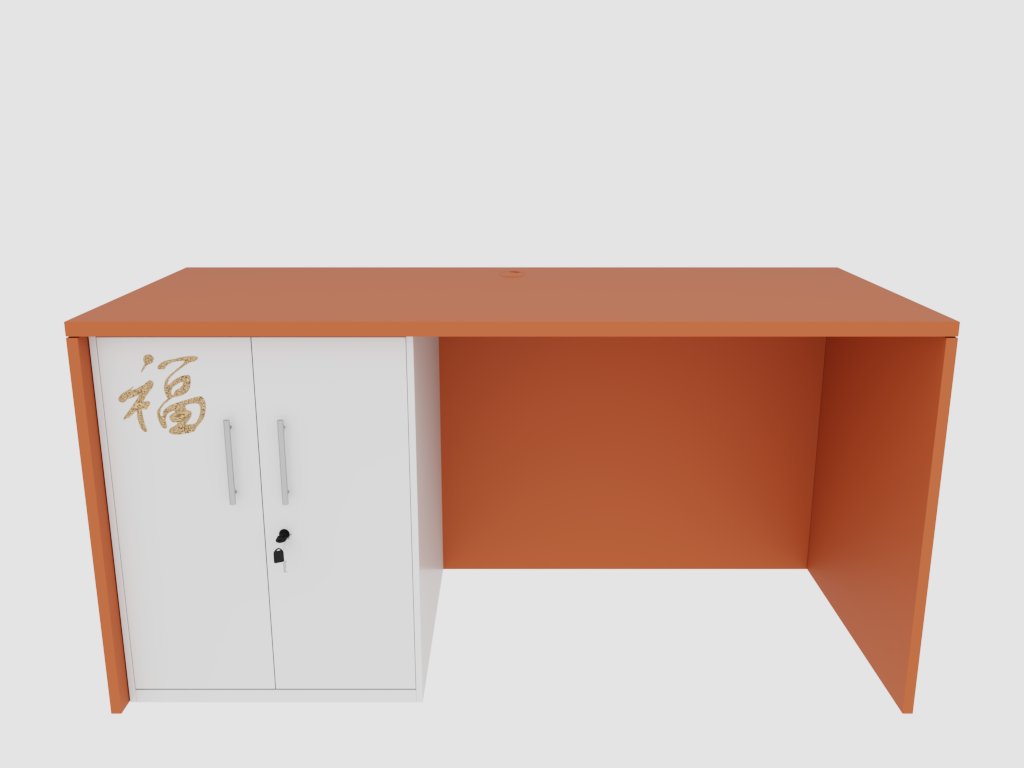}} &
	\raisebox{-\totalheight}{\includegraphics[width=\linewidth]{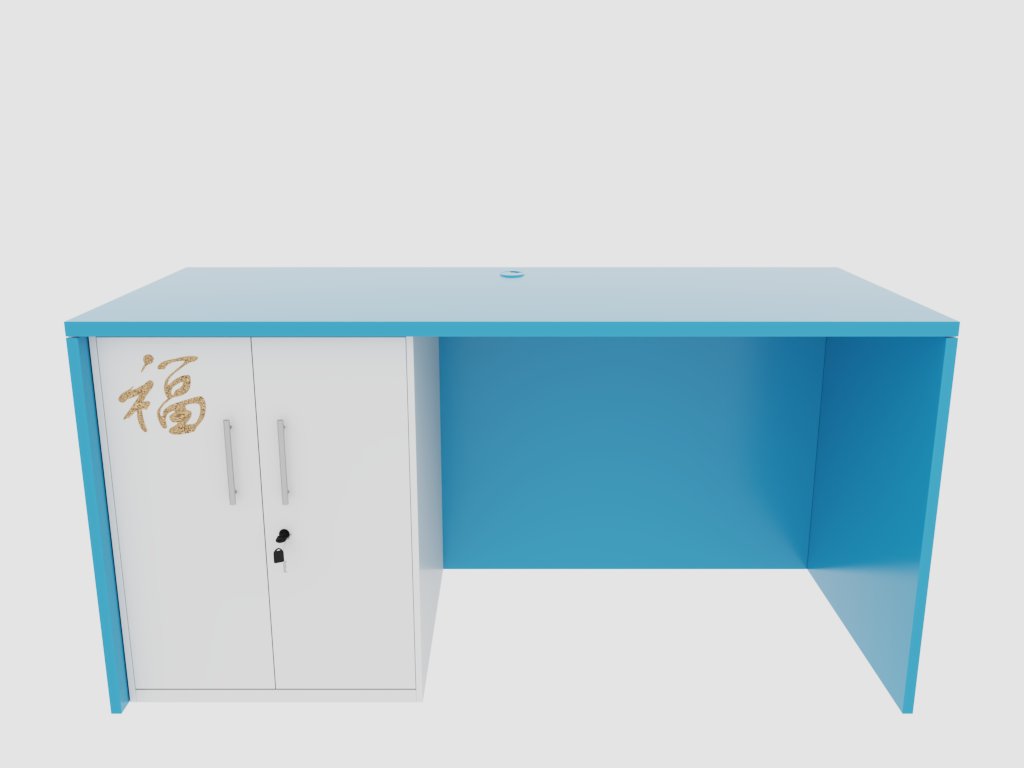}} &
	\raisebox{-\totalheight}{\includegraphics[width=\linewidth]{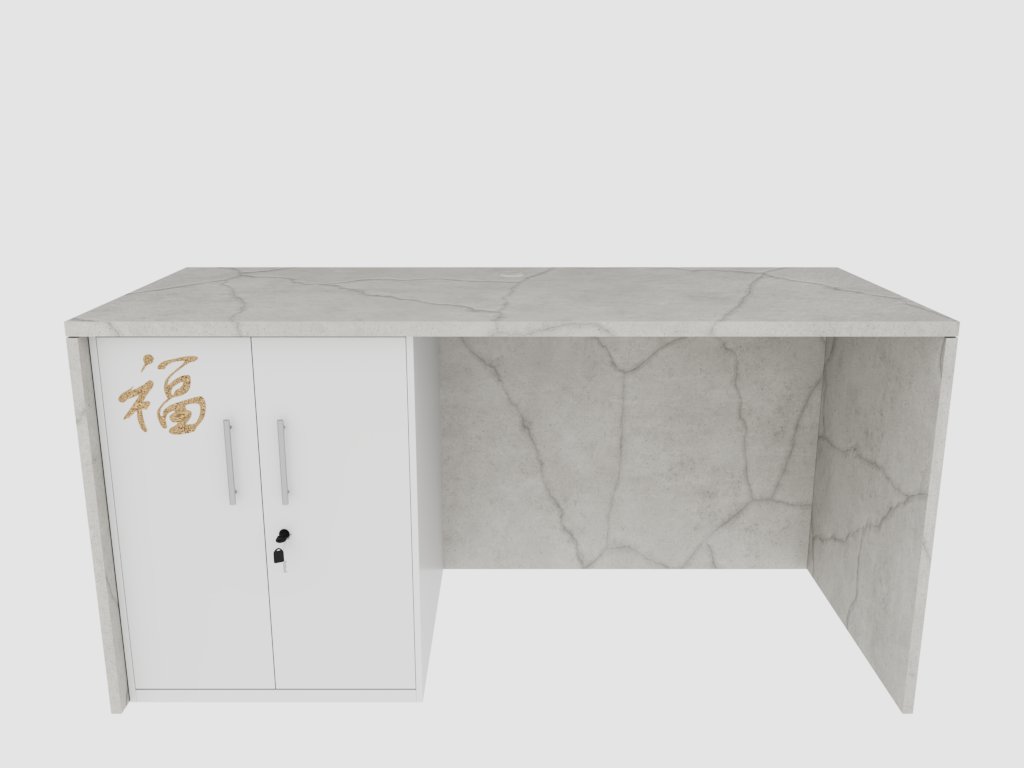}} \\
    Table 1 & Table 2 & Table 3 & Table 4 \\
    \raisebox{-\totalheight}
    {\includegraphics[width=\linewidth]{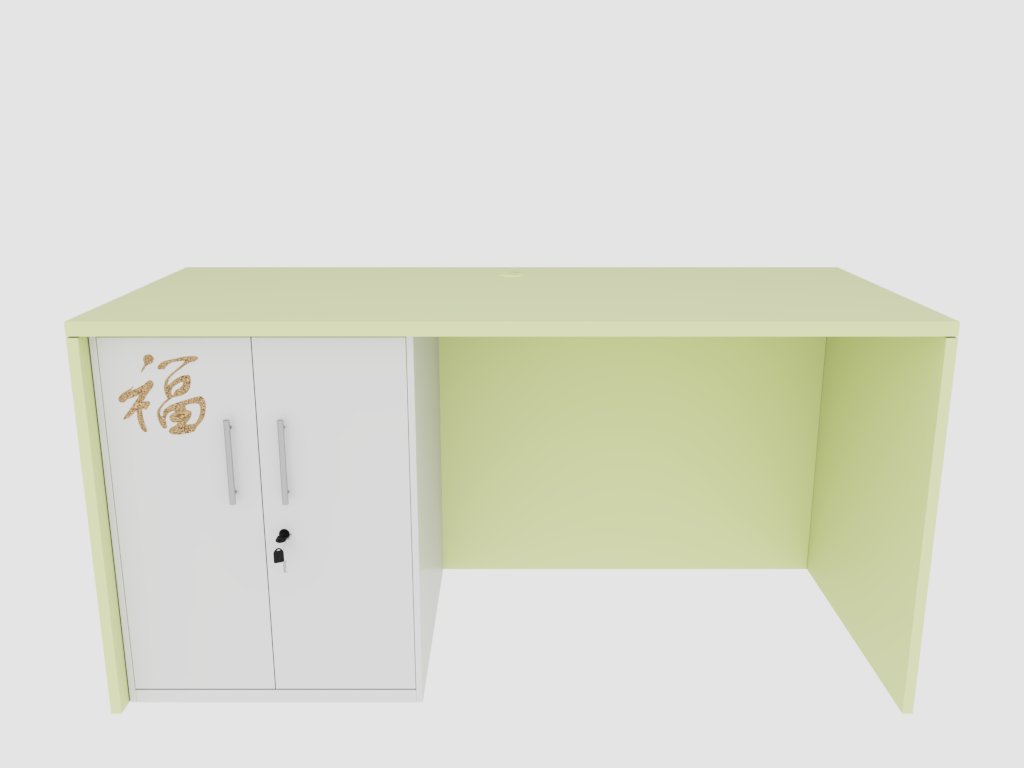}} &
	\raisebox{-\totalheight}{\includegraphics[width=\linewidth]{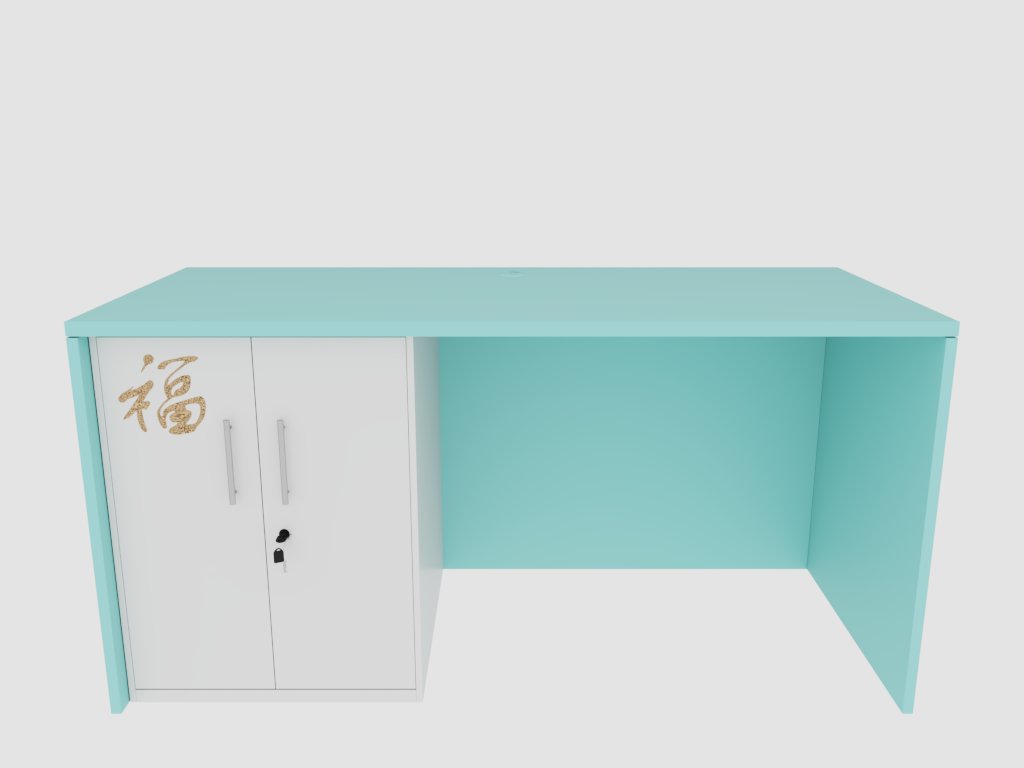}} &
	\raisebox{-\totalheight}{\includegraphics[width=\linewidth]{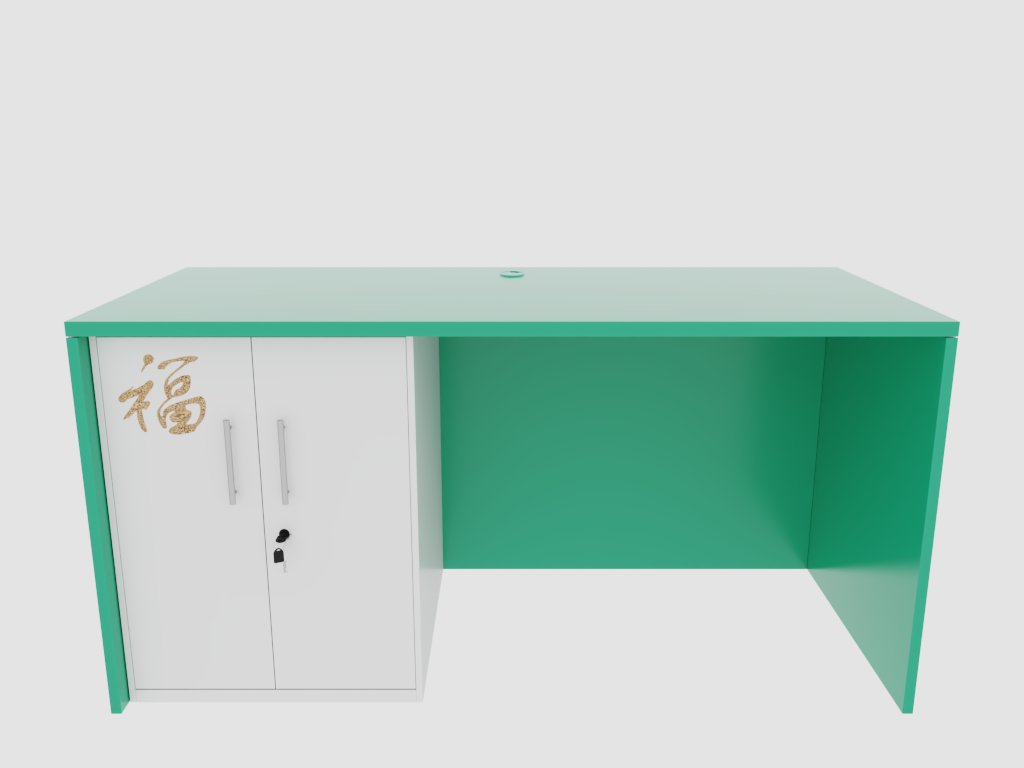}} &
	\raisebox{-\totalheight}{\includegraphics[width=\linewidth]{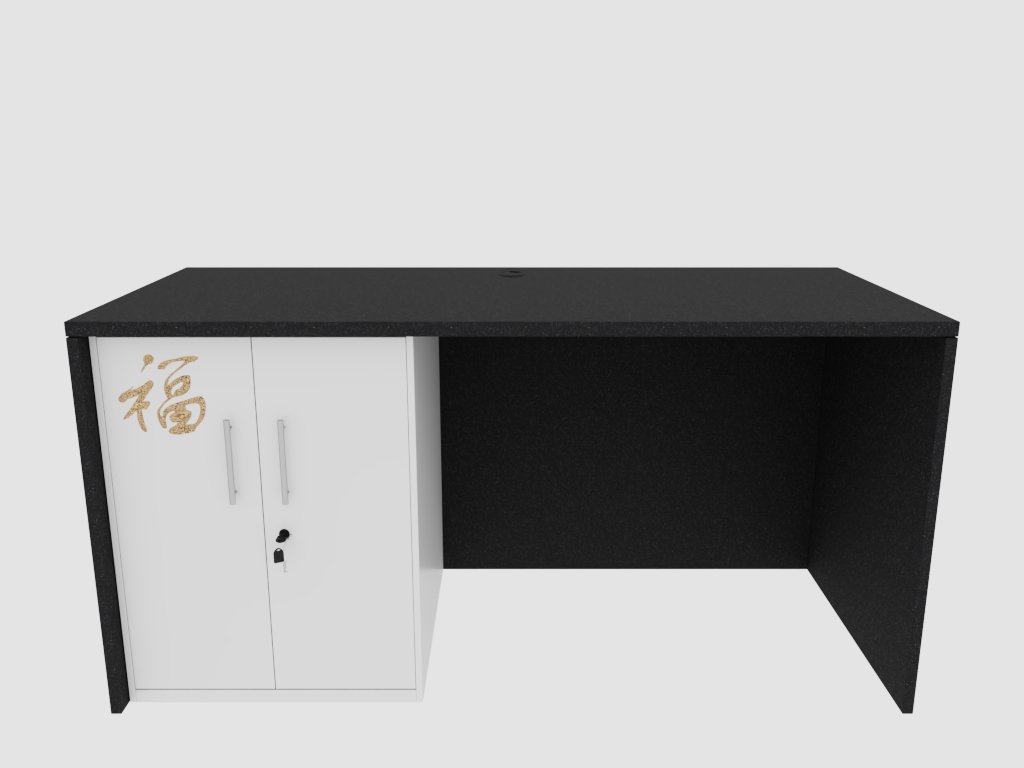}} \\
    Table 5 & Table 6 & Table 7 & Table 8 \\
	\end{tabular}}}}	
	\caption{
	Tables with different colors and textures.
	}
    \label{tab:table}
	\vspace{-1em}
\end{table*}
\begin{table*}[t!]\small
	\centering
	{\setlength{\tabcolsep}{0pt} 
	\makebox[\linewidth]{\resizebox{1.\linewidth}{!}{%
    \begin{tabular}{@{} 
            m{0.32\linewidth} @{\hspace{0.01\linewidth}}  
            m{0.32\linewidth} @{\hspace{0.01\linewidth}}  
            m{0.32\linewidth} 
            @{}}
	\raisebox{-\totalheight}{\includegraphics[width=\linewidth]{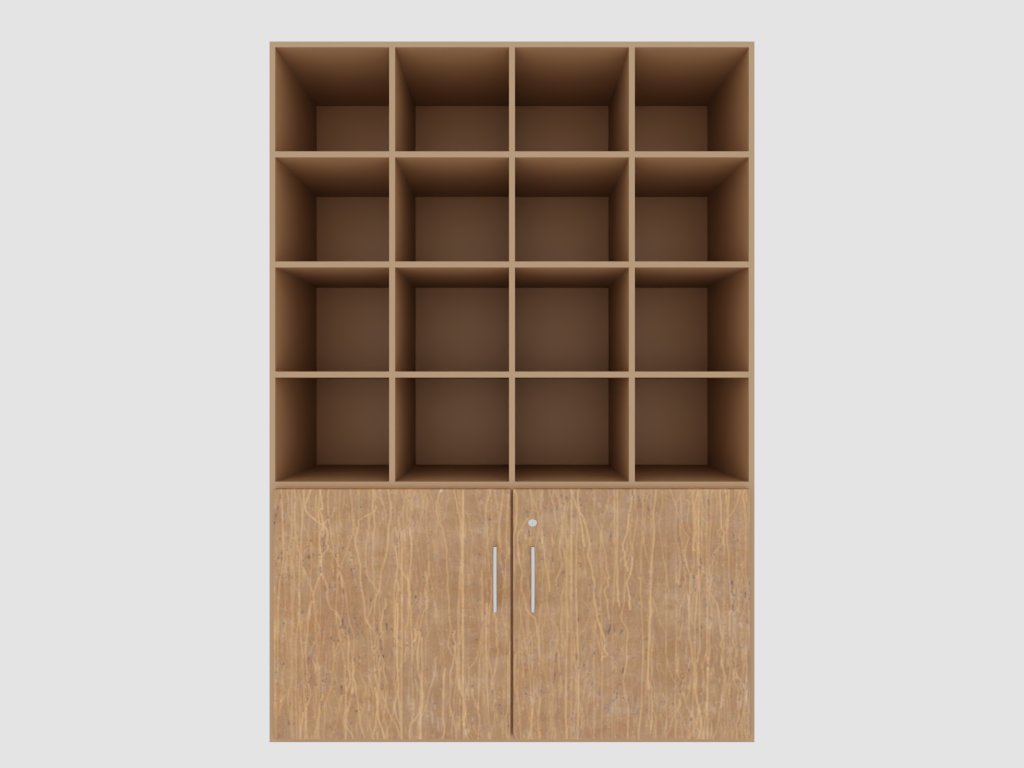}} &
	\raisebox{-\totalheight}{\includegraphics[width=\linewidth]{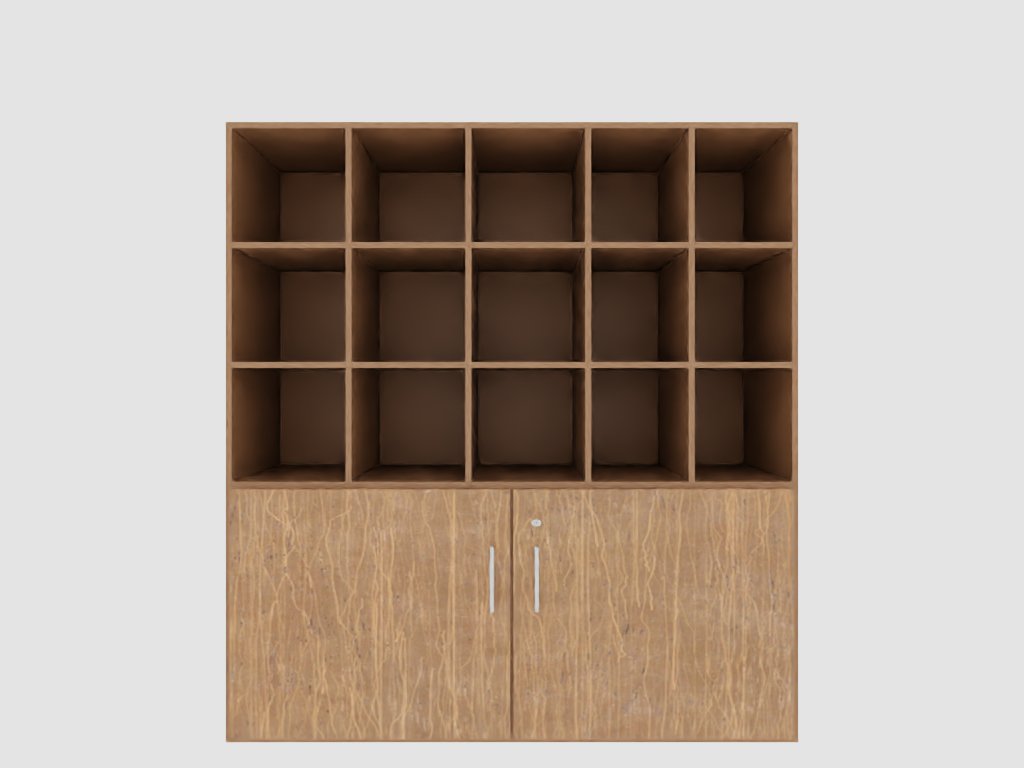}} &
	\raisebox{-\totalheight}{\includegraphics[width=\linewidth]{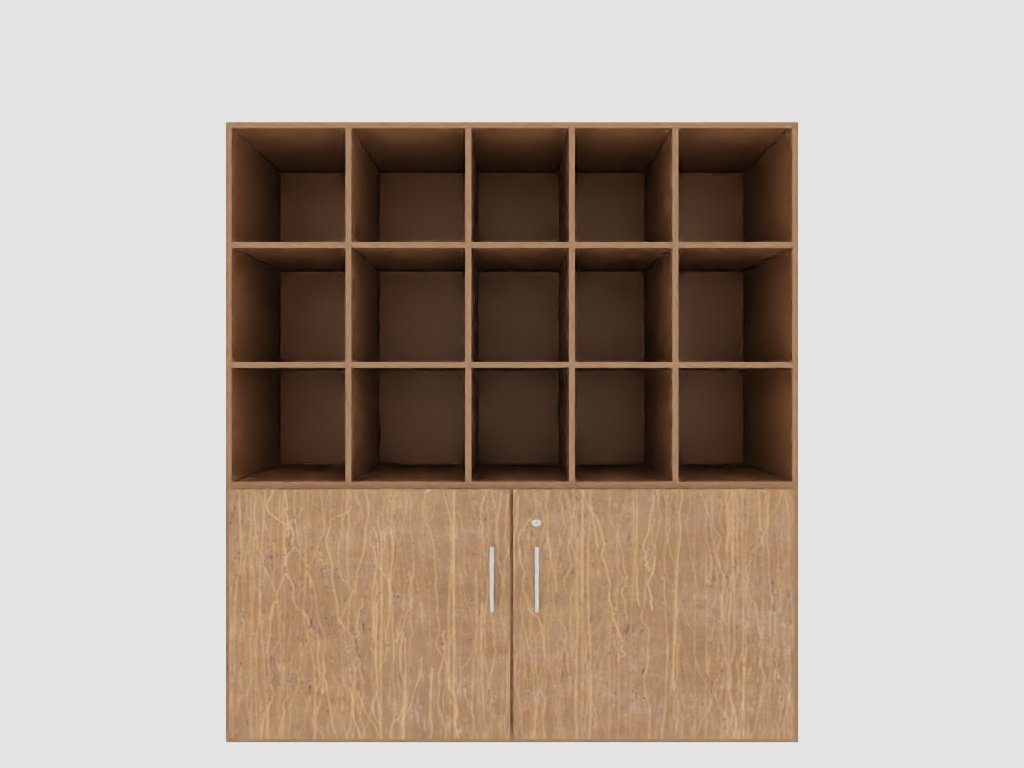}} \\
    Shelf 1 & Shelf 2 & Shelf 3 \\
    \raisebox{-\totalheight}{\includegraphics[width=\linewidth]{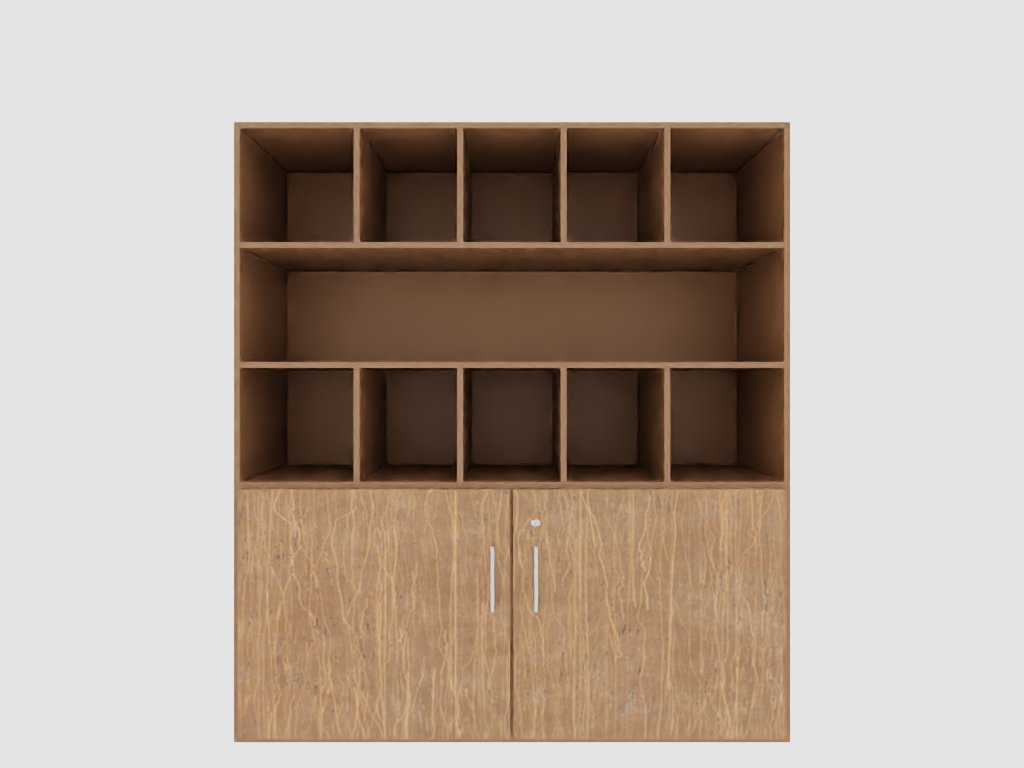}} &
	\raisebox{-\totalheight}{\includegraphics[width=\linewidth]{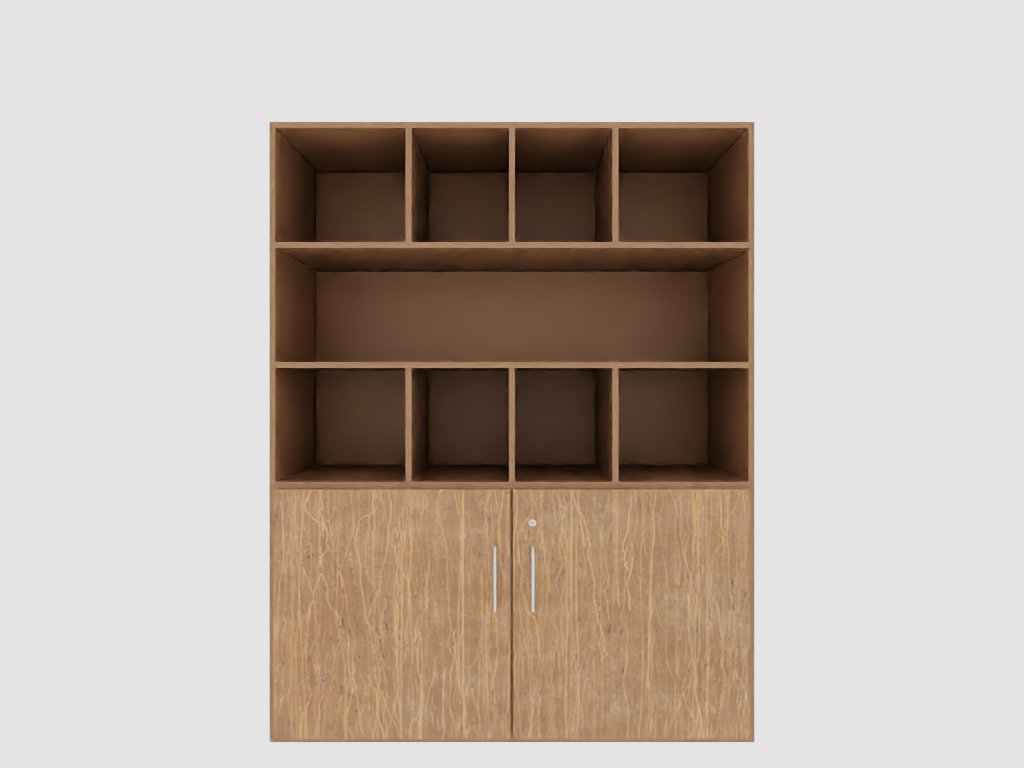}} &
	\raisebox{-\totalheight}{\includegraphics[width=\linewidth]{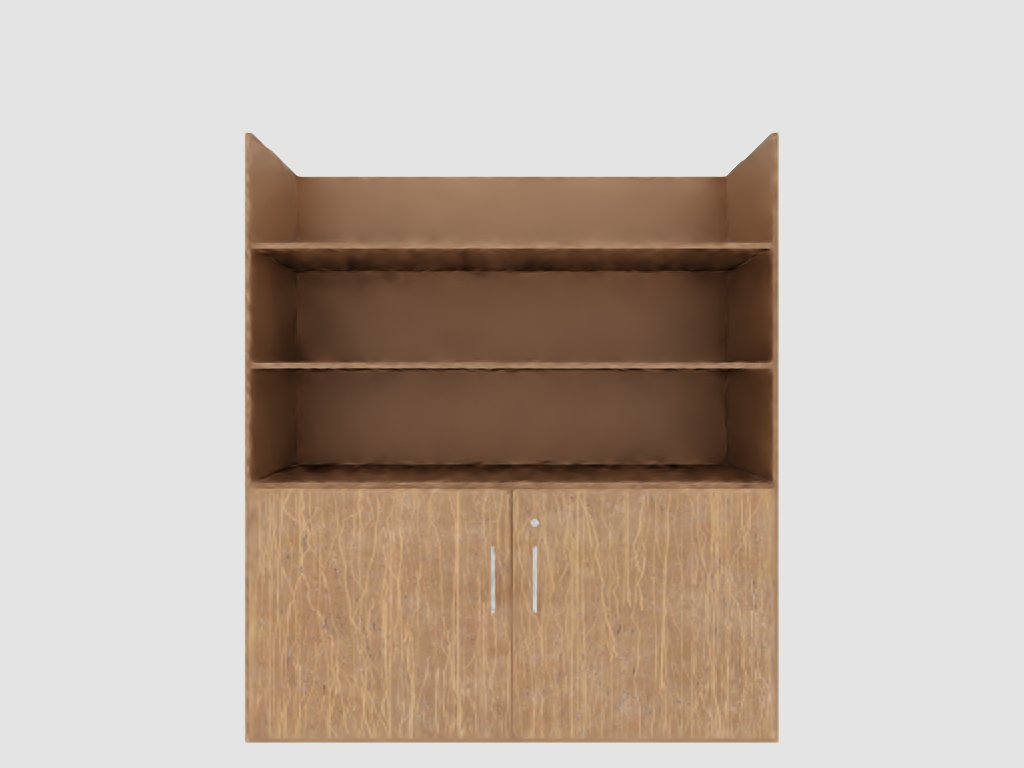}} \\
    Shelf 4 & Shelf 5 & Shelf 6
	\end{tabular}}}}
	\caption{
	Shelf with different layouts.
	}
    \label{tab:shelf}
	\vspace{-1em}
\end{table*}

\begin{table*}[t!]\small
	\centering
	{\setlength{\tabcolsep}{0pt} 
	\makebox[\linewidth]{\resizebox{1.\linewidth}{!}{%
    \begin{tabular}{@{} 
            m{0.19\linewidth} @{\hspace{0.01\linewidth}}  
            m{0.19\linewidth} @{\hspace{0.01\linewidth}}  
            m{0.19\linewidth} @{\hspace{0.01\linewidth}}  
            m{0.19\linewidth} @{\hspace{0.01\linewidth}} 
            m{0.19\linewidth} 
            @{}}
	\raisebox{-\totalheight}{\includegraphics[width=\linewidth]{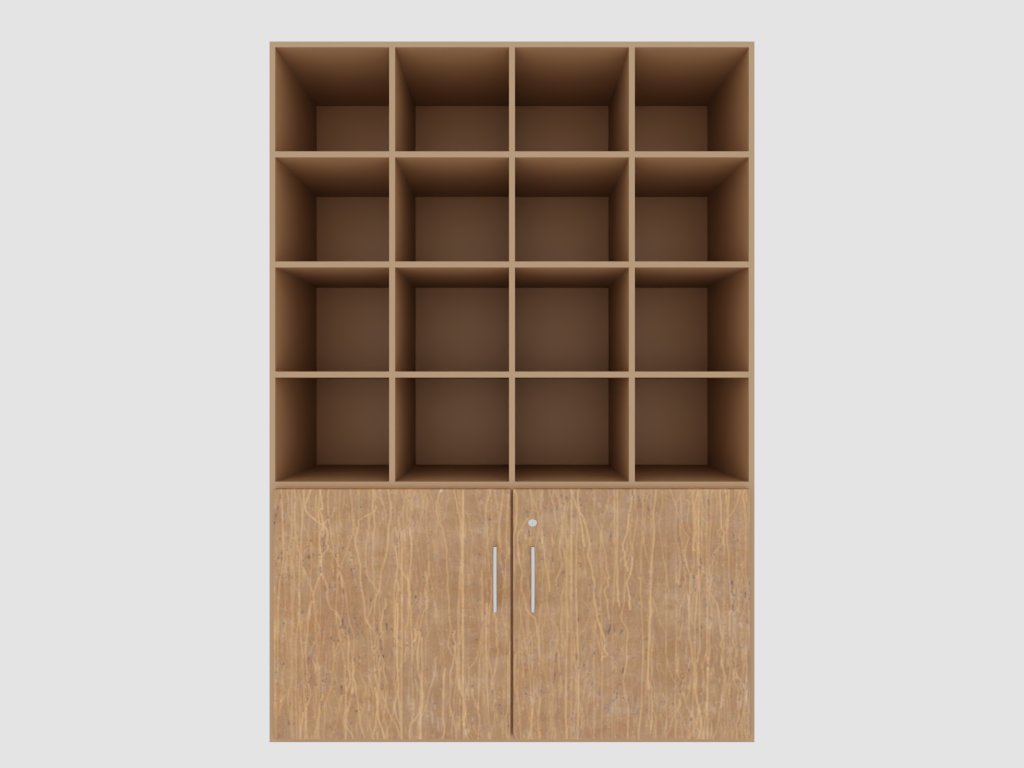}} &
	\raisebox{-\totalheight}{\includegraphics[width=\linewidth]{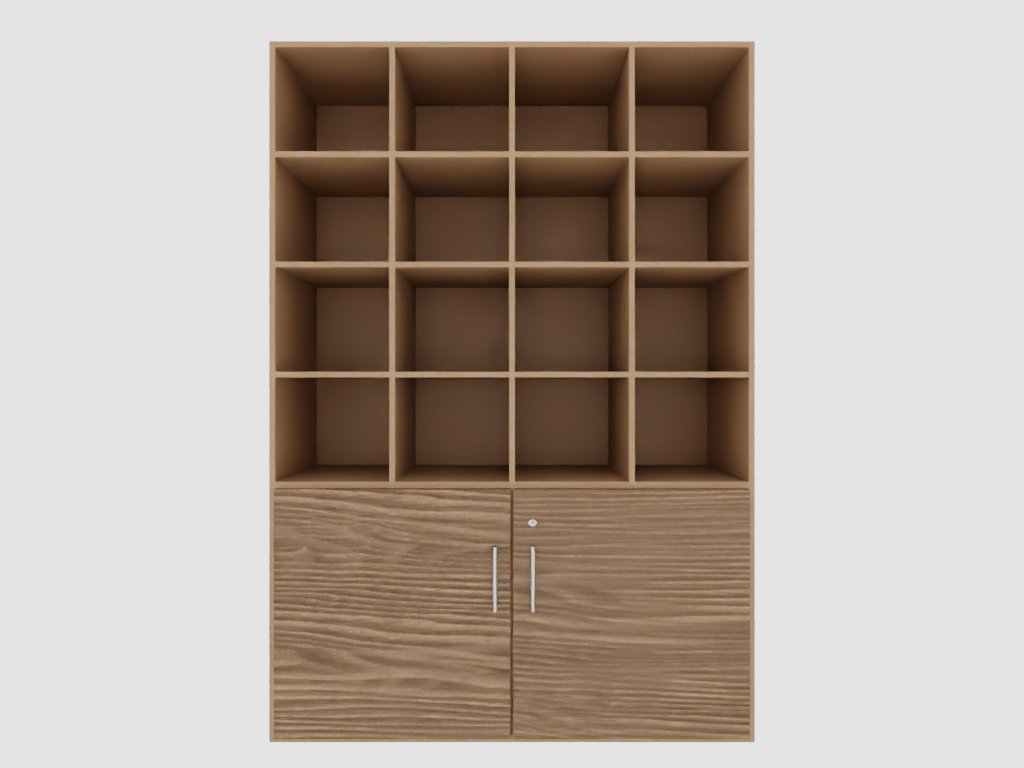}} &
	\raisebox{-\totalheight}{\includegraphics[width=\linewidth]{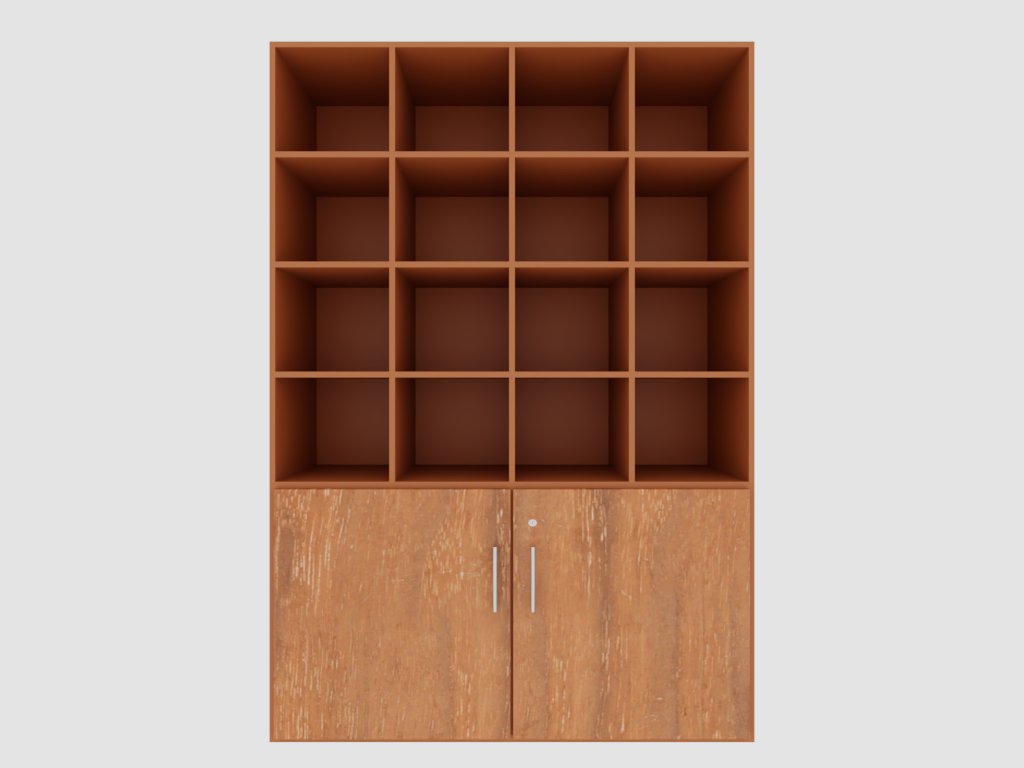}} &
    \raisebox{-\totalheight}{\includegraphics[width=\linewidth]{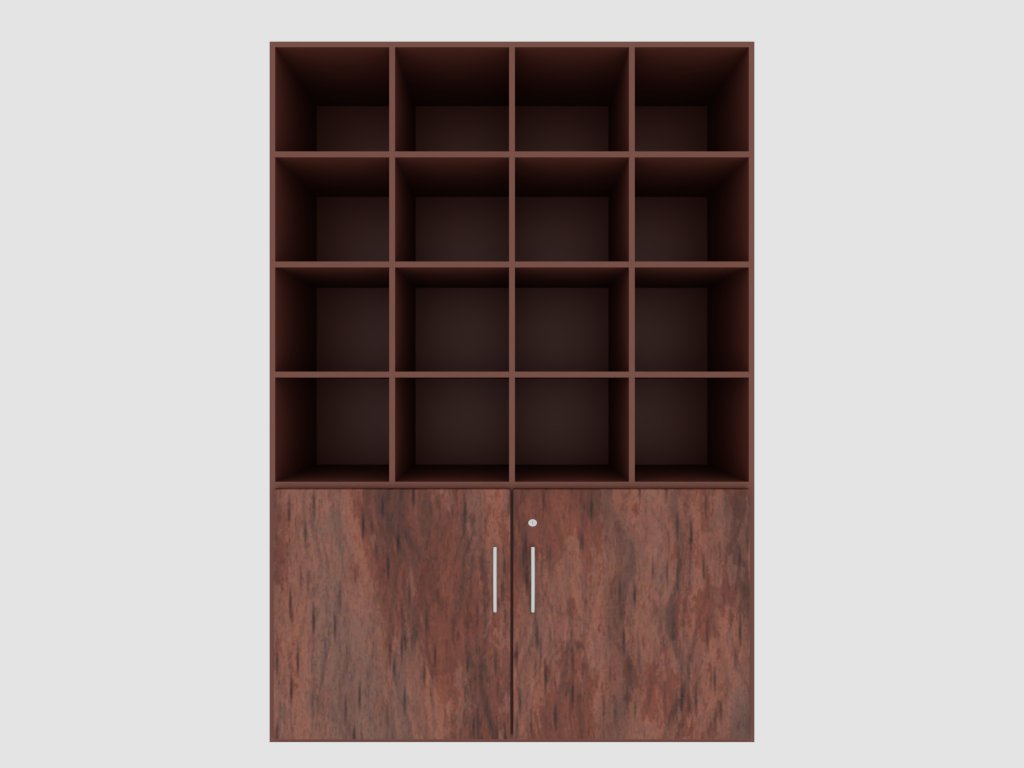}} &
	\raisebox{-\totalheight}{\includegraphics[width=\linewidth]{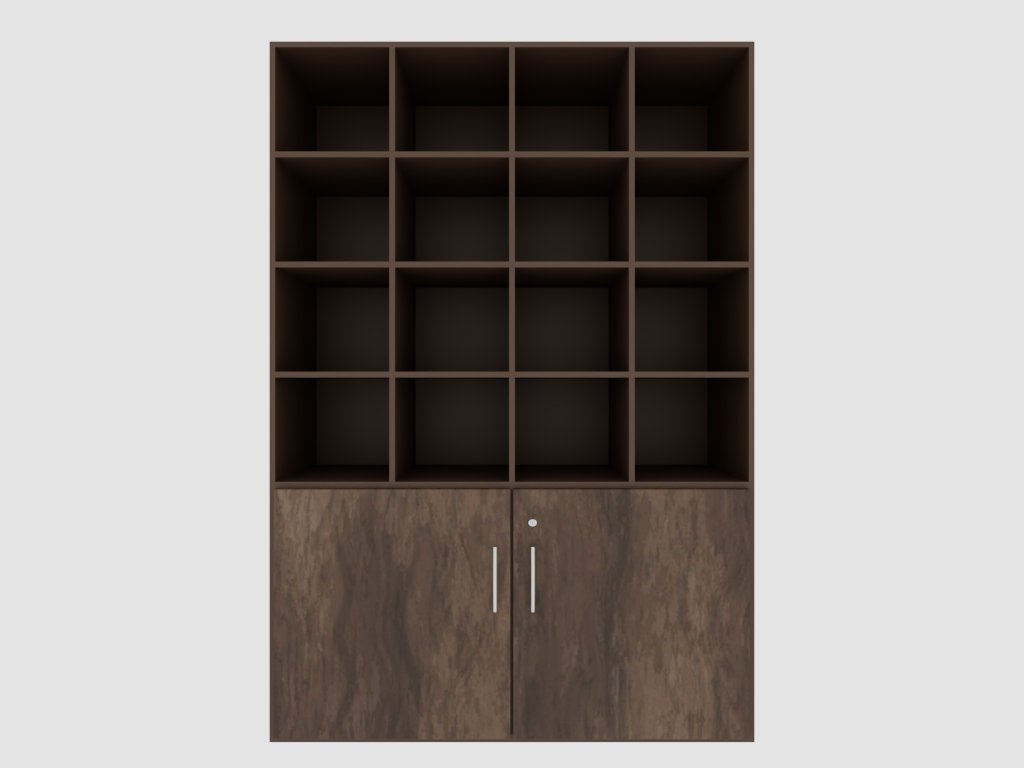}}
	\end{tabular}}}}
	\caption{
	Shelf with different textures.
	}
    \label{tab:shelf-texture}
	\vspace{-.0em}
\end{table*}

\begin{table*}[t!]\small
	\centering
	{\setlength{\tabcolsep}{0pt} 
	\makebox[\linewidth]{\resizebox{1.\linewidth}{!}{%
    \begin{tabular}{@{} 
            m{0.24\linewidth} @{\hspace{0.01\linewidth}}  
            m{0.24\linewidth} @{\hspace{0.01\linewidth}}  
            m{0.24\linewidth} @{\hspace{0.01\linewidth}}  
            m{0.24\linewidth} 
            @{}}
	\raisebox{-\totalheight}{\includegraphics[width=\linewidth]{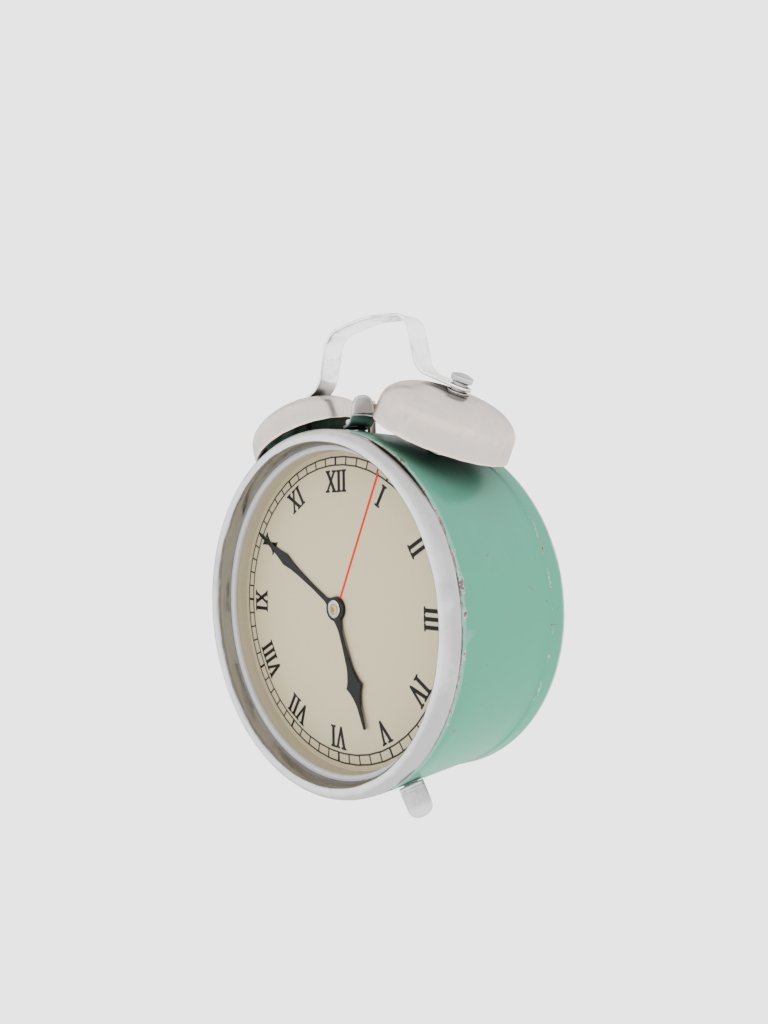}} &
	\raisebox{-\totalheight}{\includegraphics[width=\linewidth]{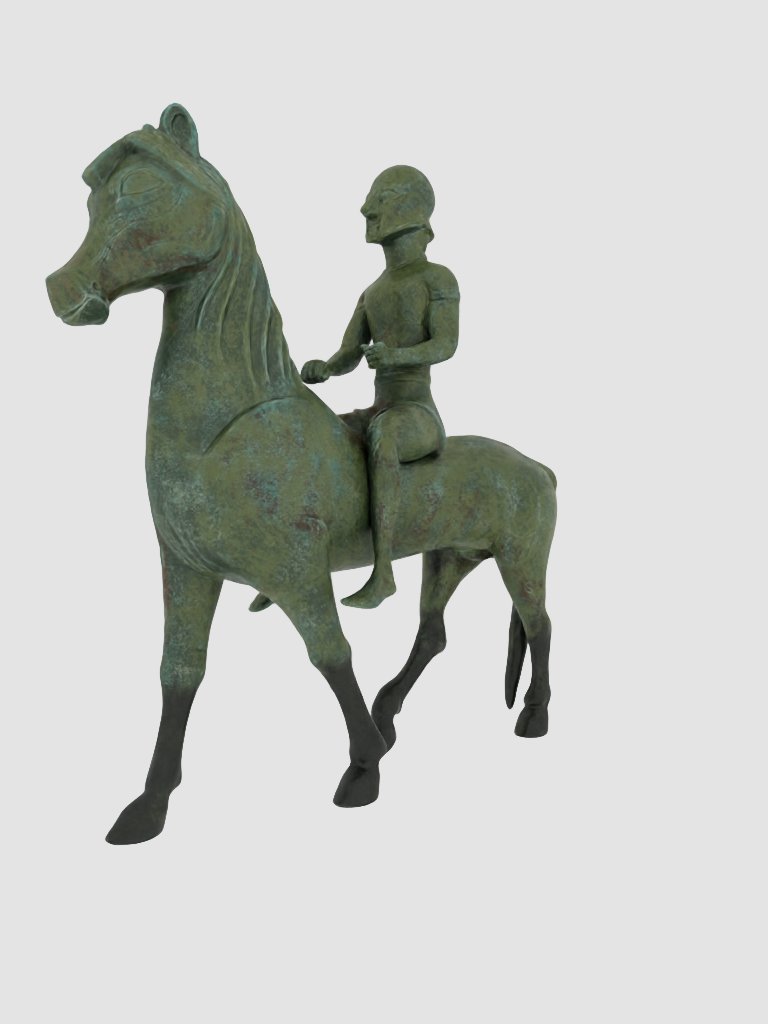}} &
	\raisebox{-\totalheight}{\includegraphics[width=\linewidth]{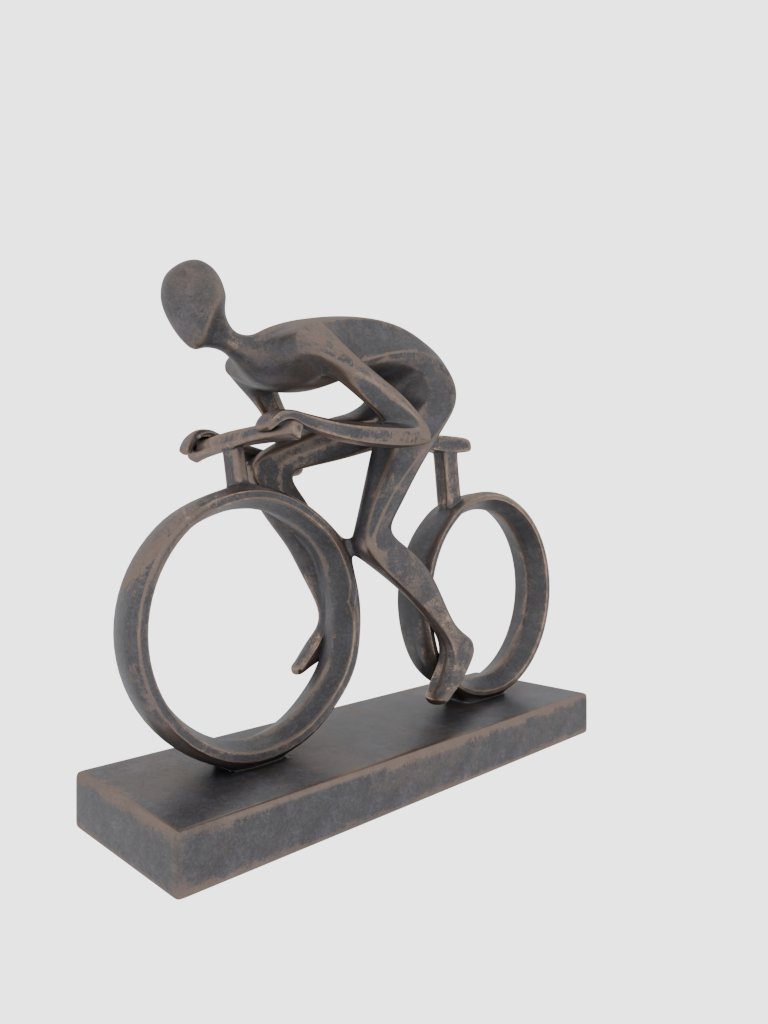}} &
	\raisebox{-\totalheight}{\includegraphics[width=\linewidth]{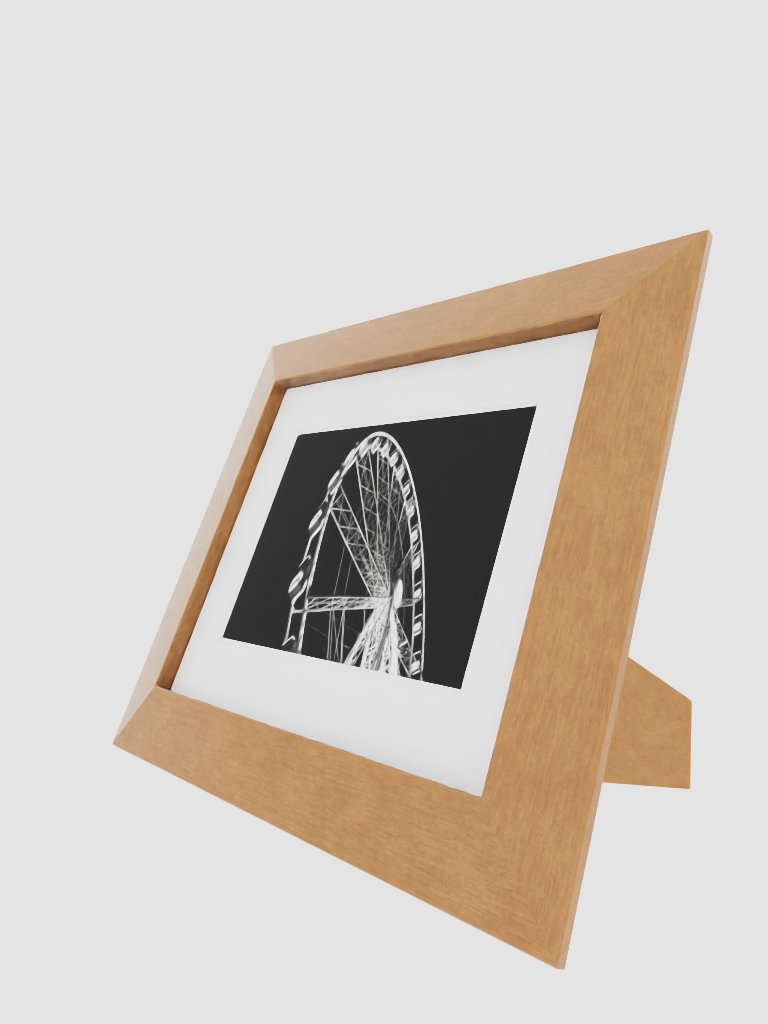}} \\
    Alarm Clock & Armento Rider & Bicycle Sculpture & Picture Frame \\
	\raisebox{-\totalheight}{\includegraphics[width=\linewidth]{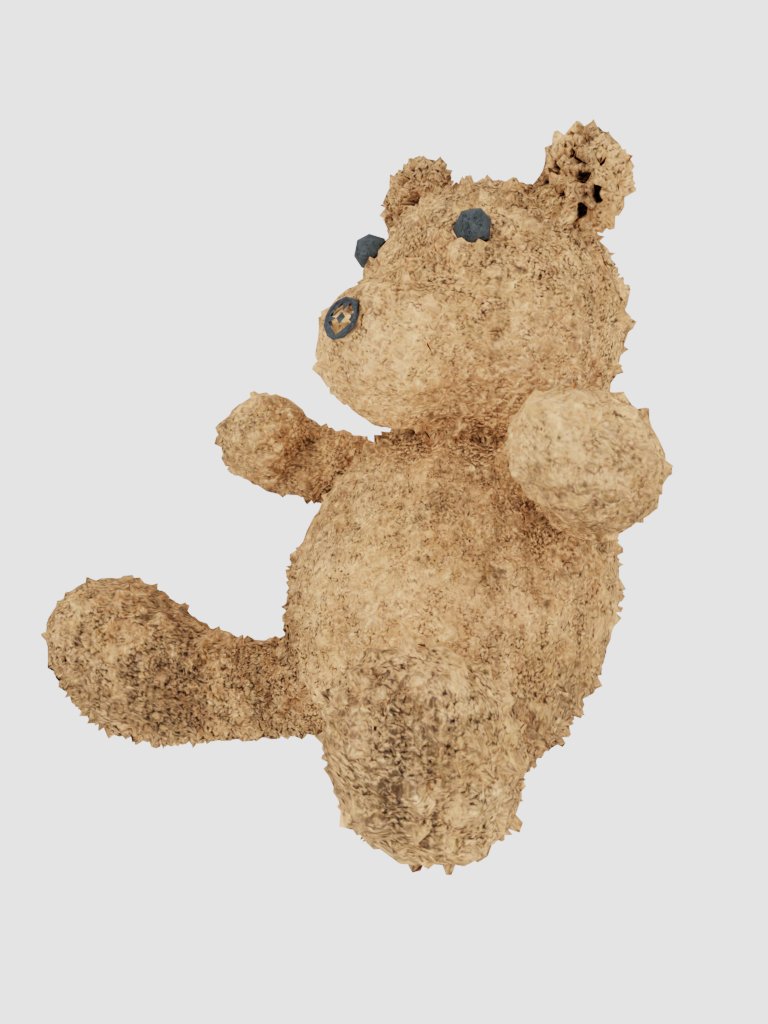}} &
	\raisebox{-\totalheight}{\includegraphics[width=\linewidth]{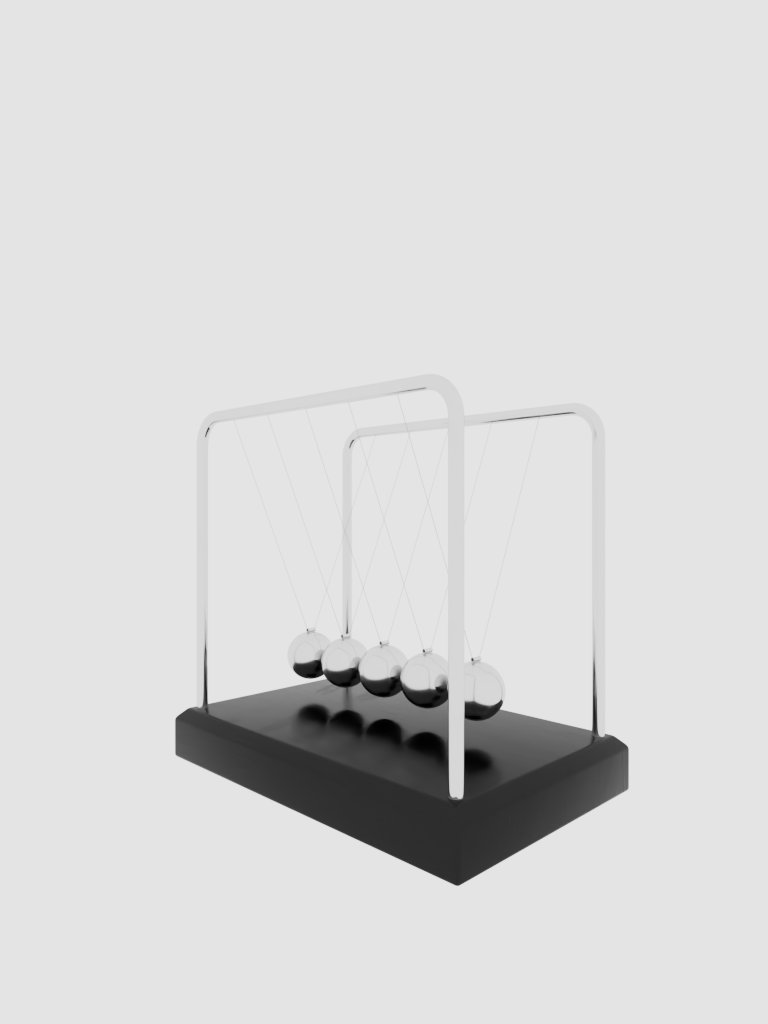}} &
	\raisebox{-\totalheight}{\includegraphics[width=\linewidth]{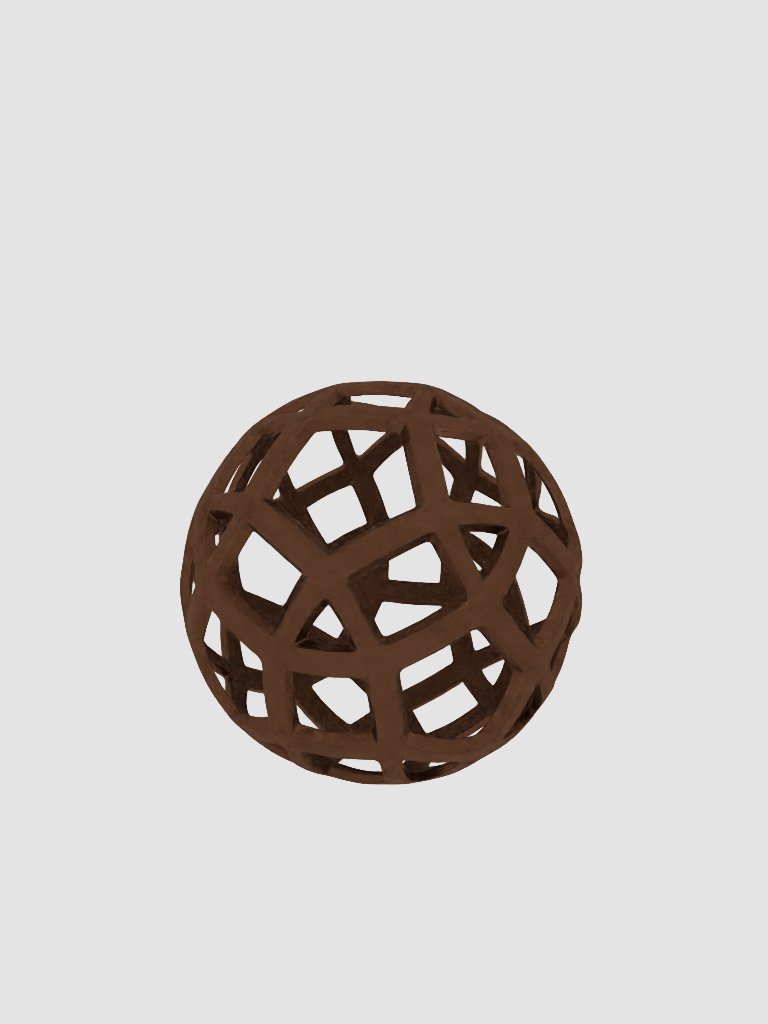}} &
	\raisebox{-\totalheight}{\includegraphics[width=\linewidth]{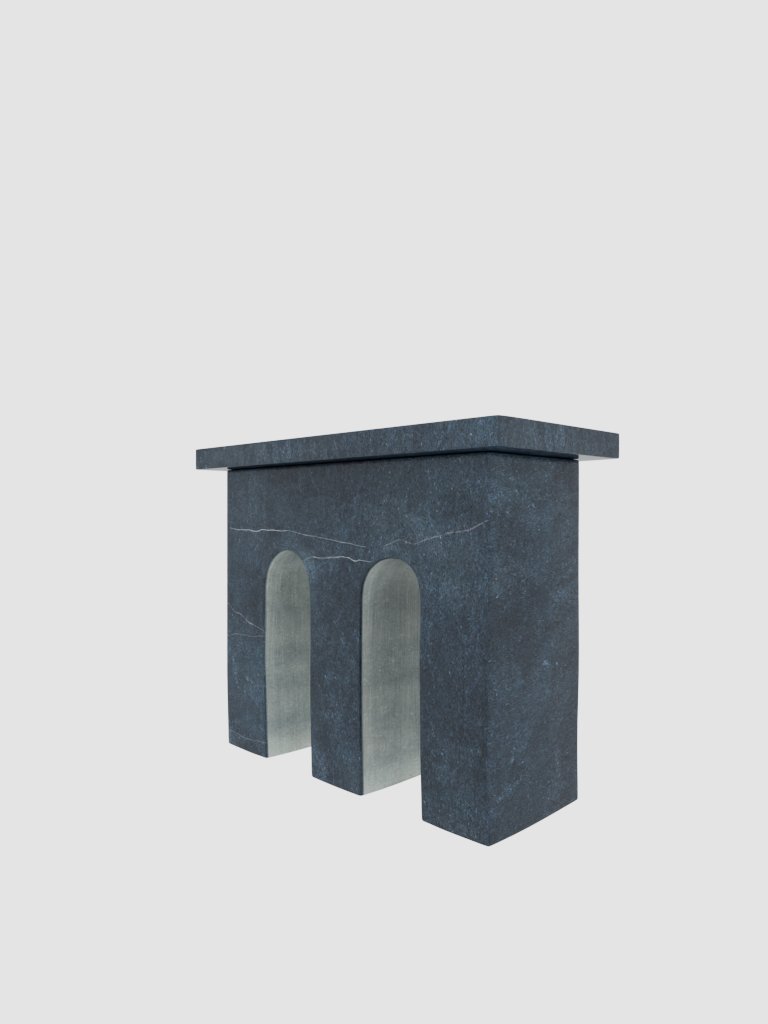}} \\
    Teddy Bear & Newton's Cradle & Geosphere Sculpture & Pillar Bookend \\
	\raisebox{-\totalheight}{\includegraphics[width=\linewidth]{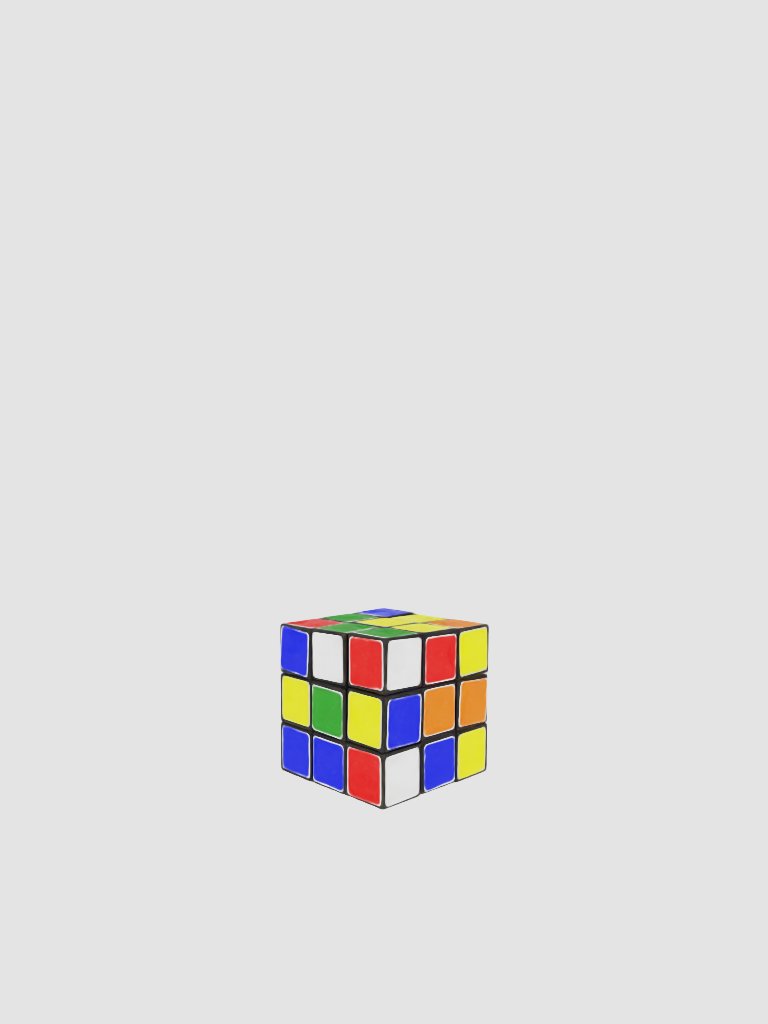}} &
	\raisebox{-\totalheight}{\includegraphics[width=\linewidth]{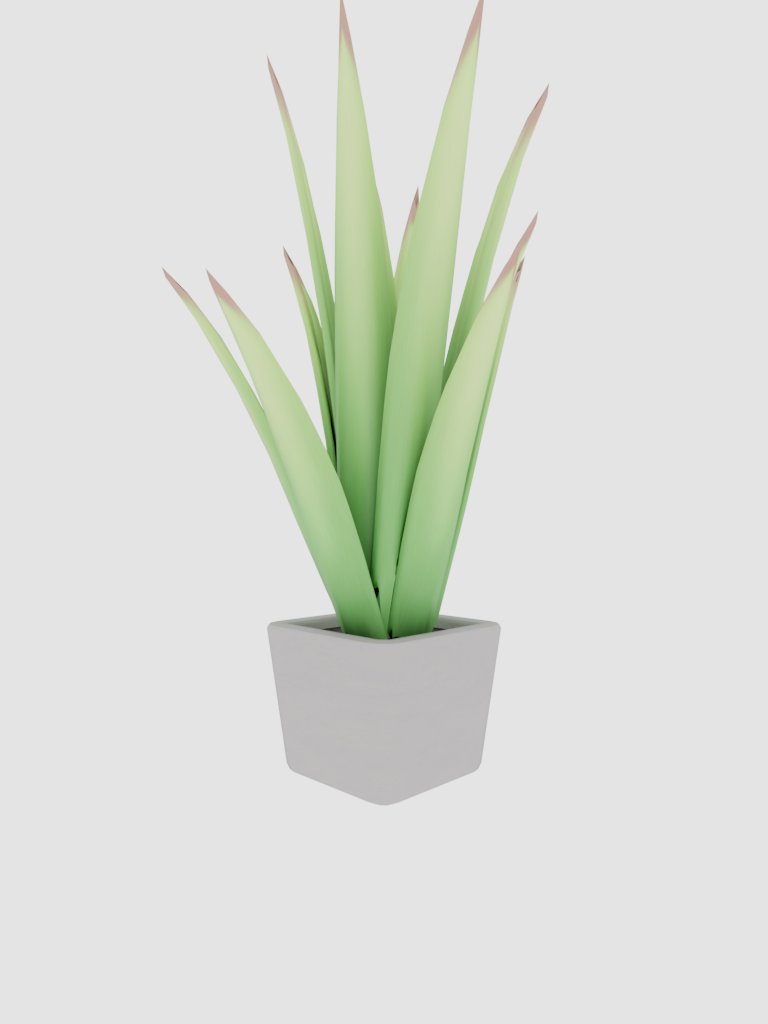}} &
    \raisebox{-\totalheight}{\includegraphics[width=\linewidth]{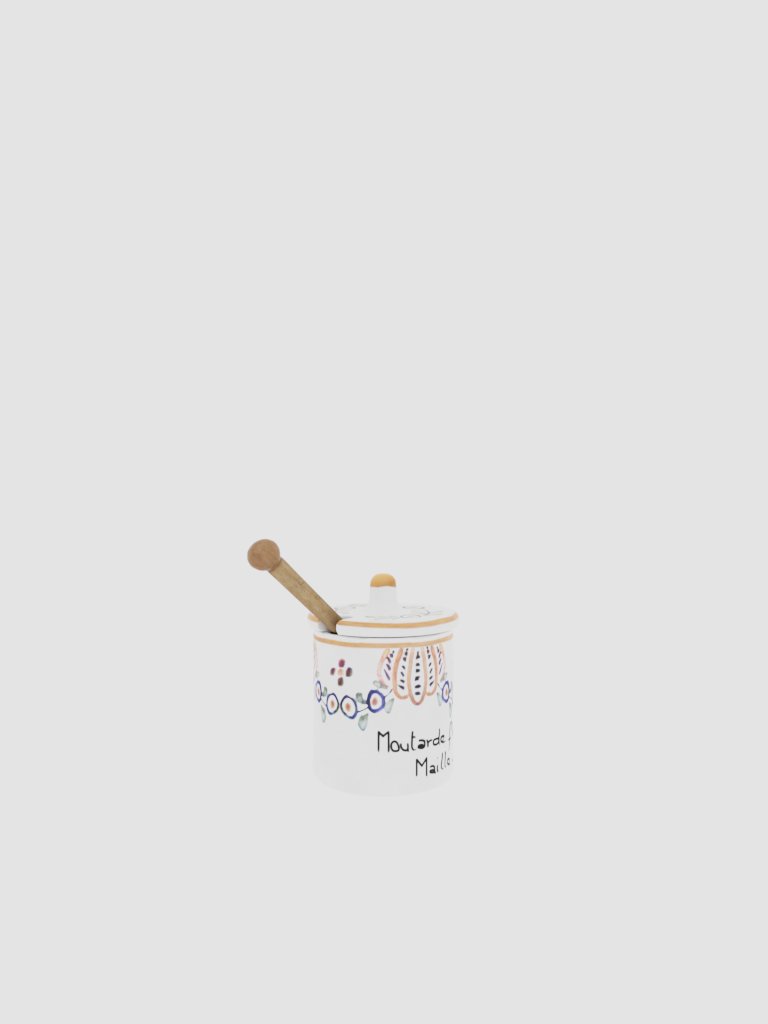}} &
	\raisebox{-\totalheight}{\includegraphics[width=\linewidth]{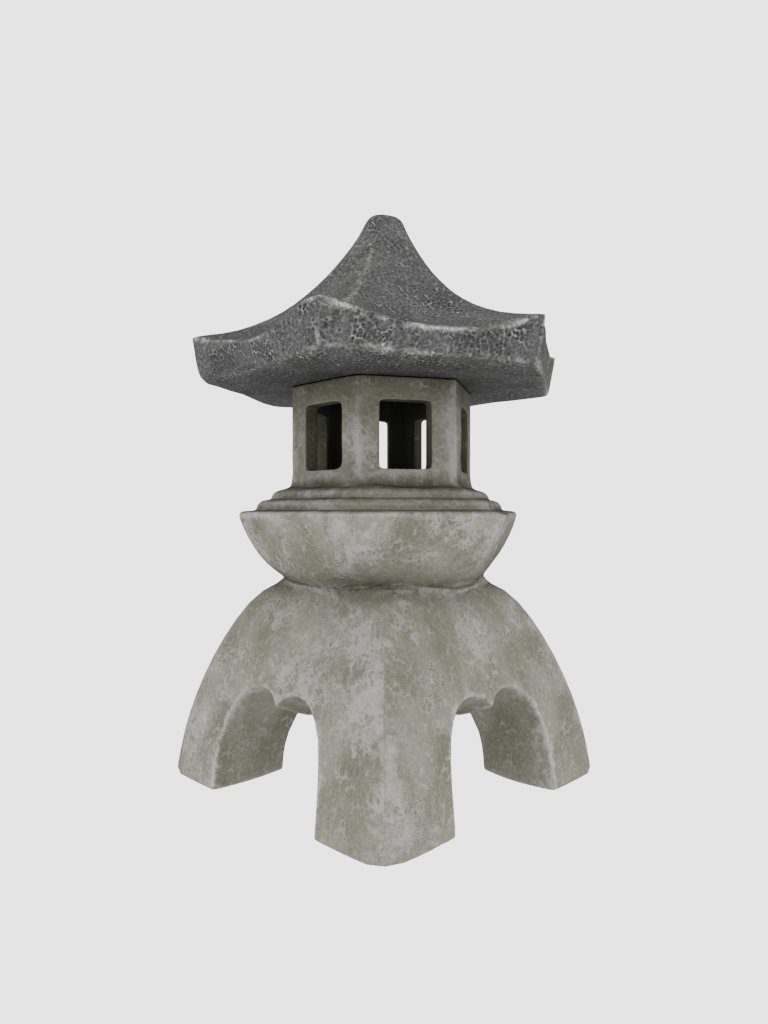}} \\
    Rubik's Cube & Succulents & Ceramic Jar & Pagoda Statue
	\end{tabular}}}}	
	\caption{
	Near reference objects (w/ and w/o an intrinsic frame of reference) appear on the table or shelf.
	}
    \label{tab:reference-near}
	\vspace{-.0em}
\end{table*}
\begin{table*}[t!]\small
	\centering
	{\setlength{\tabcolsep}{0pt} 
	\makebox[\linewidth]{\resizebox{1.\linewidth}{!}{%
    \begin{tabular}{@{} 
            m{0.24\linewidth} @{\hspace{0.01\linewidth}}  
            m{0.24\linewidth} @{\hspace{0.01\linewidth}}  
            m{0.24\linewidth} @{\hspace{0.01\linewidth}}  
            m{0.24\linewidth} 
            @{}}
	\raisebox{-\totalheight}{\includegraphics[width=\linewidth]{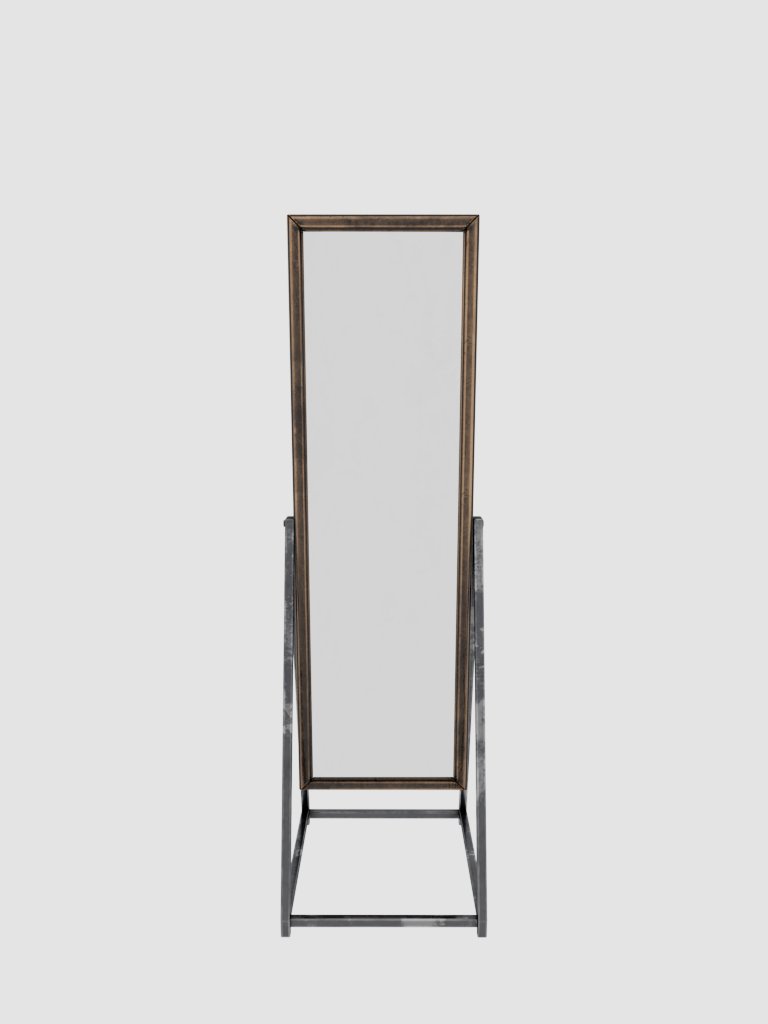}} &
    \raisebox{-\totalheight}{\includegraphics[width=\linewidth]{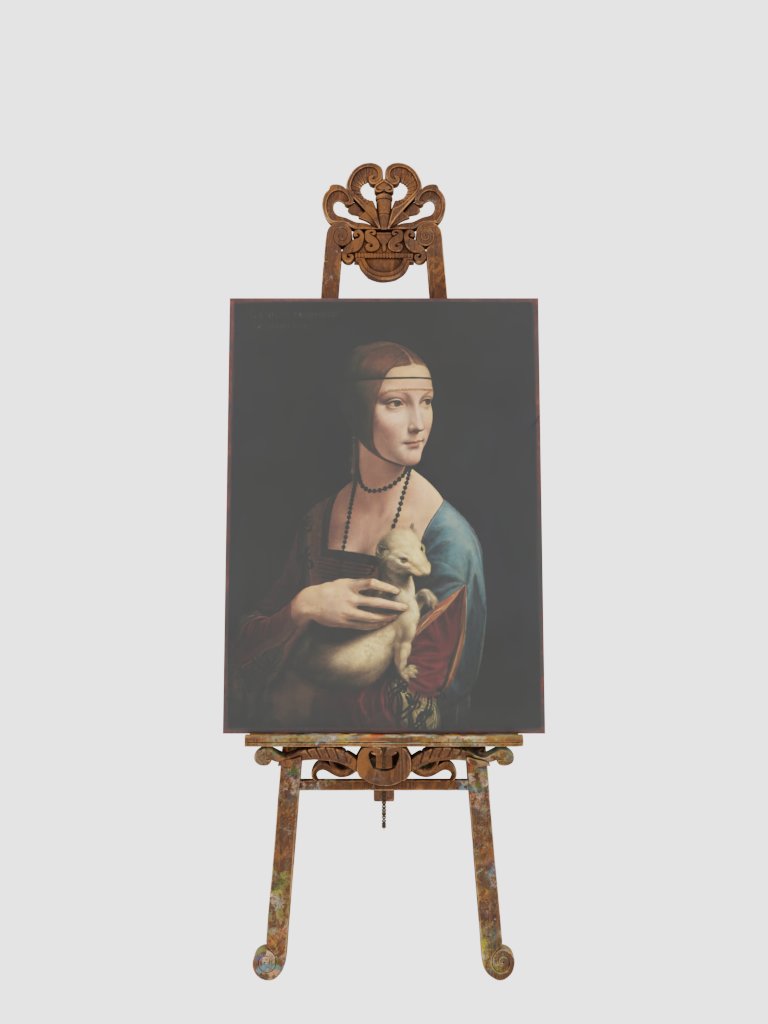}} &
	\raisebox{-\totalheight}{\includegraphics[width=\linewidth]{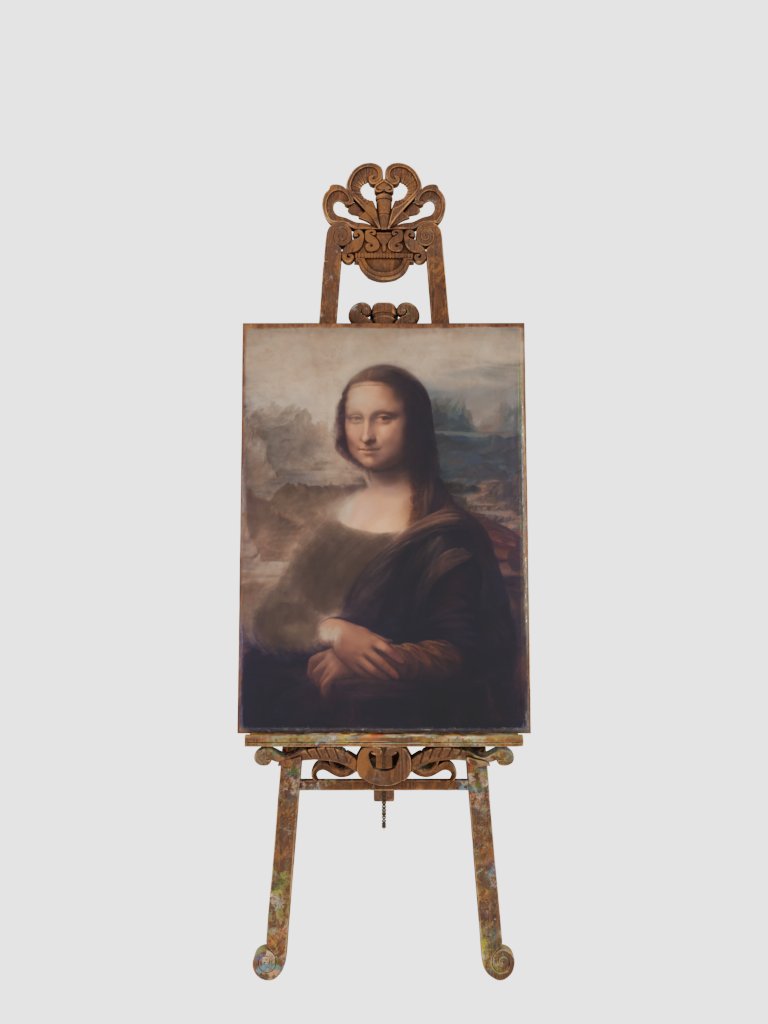}} &
	\raisebox{-\totalheight}{\includegraphics[width=\linewidth]{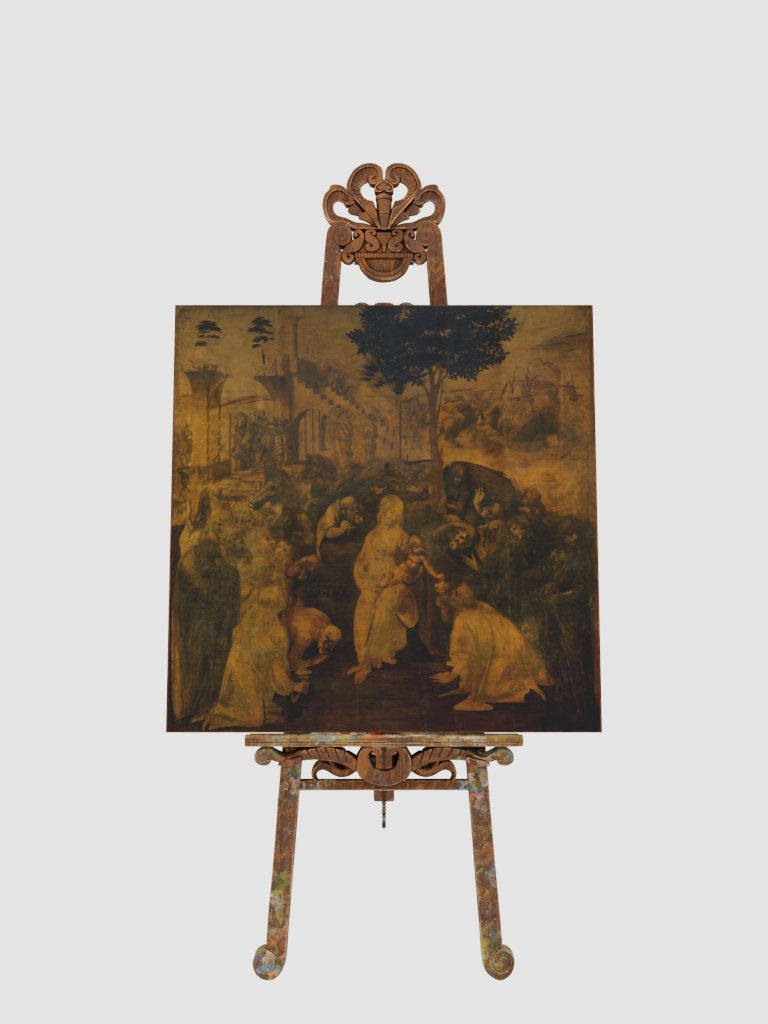}} \\
    Cheval Mirror & Lady with an Ermine & Mona Lisa & Adoration of the Magi \\
	\raisebox{-\totalheight}{\includegraphics[width=\linewidth]{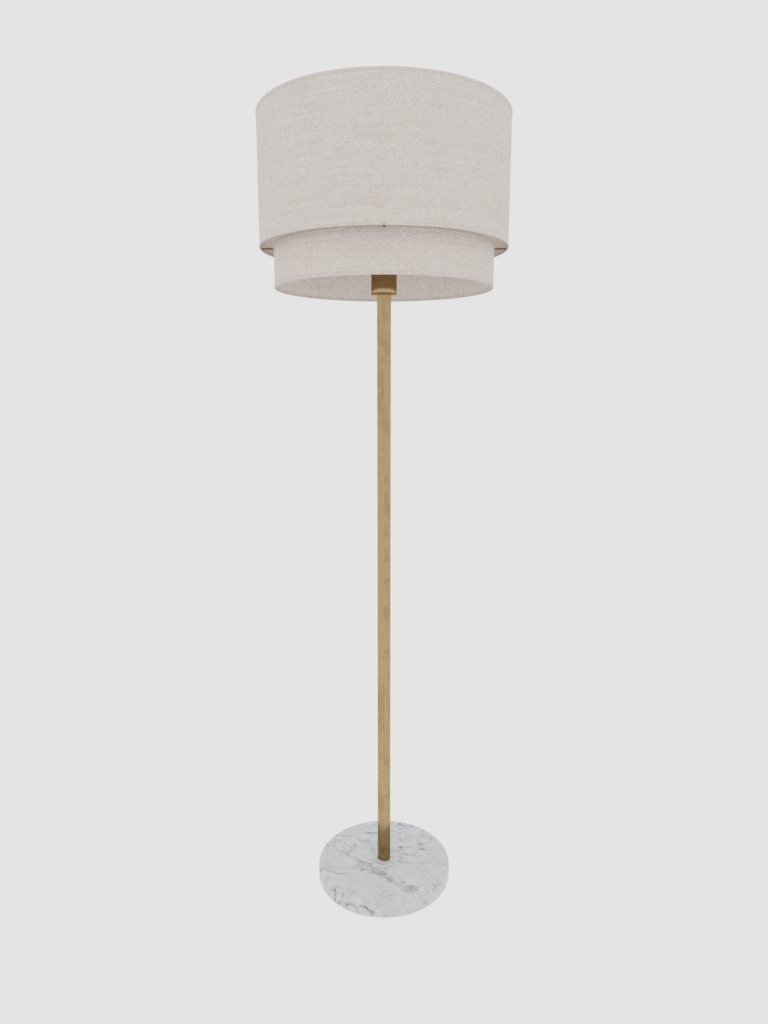}} &
	\raisebox{-\totalheight}{\includegraphics[width=\linewidth]{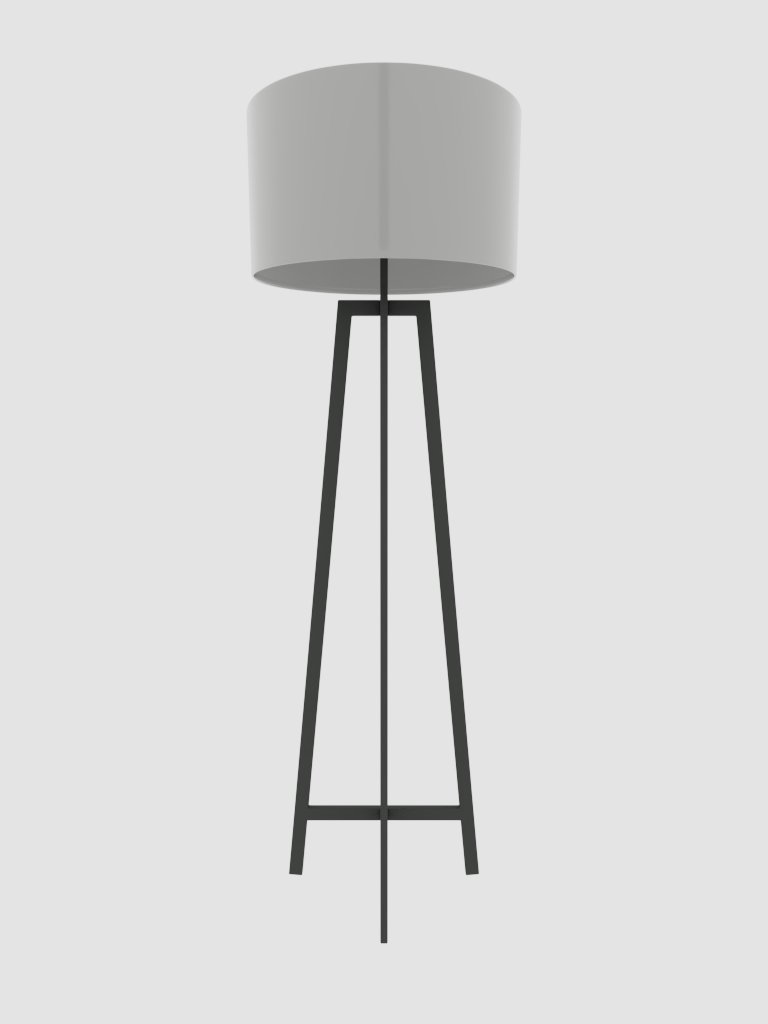}} &
	\raisebox{-\totalheight}{\includegraphics[width=\linewidth]{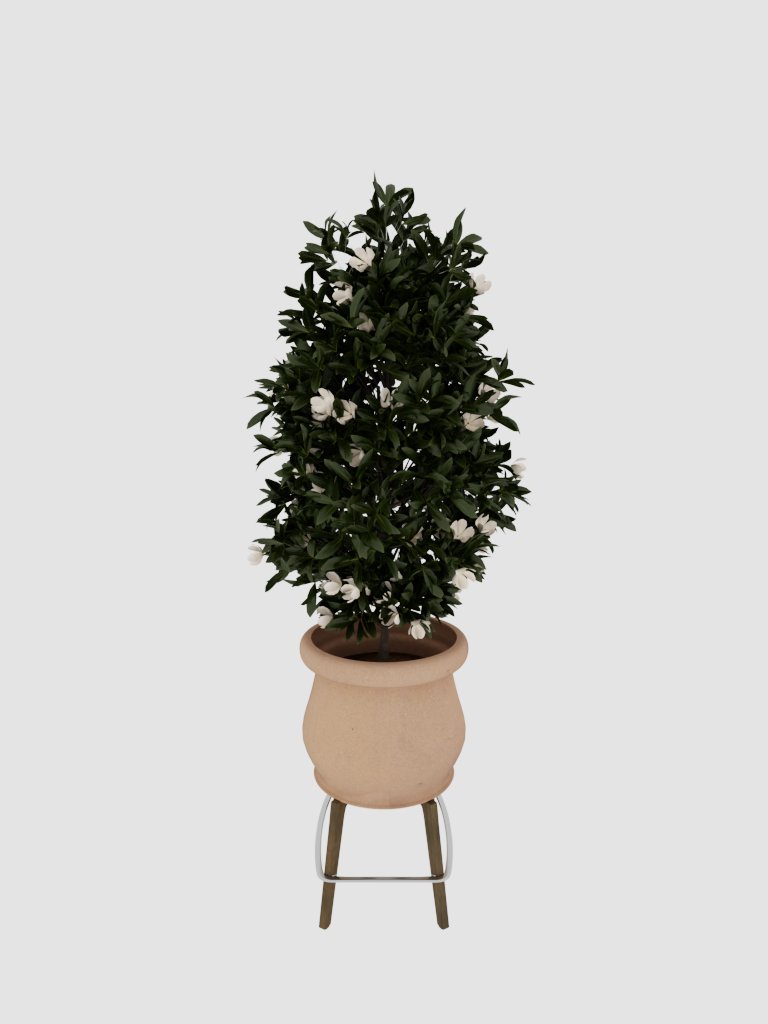}} &
	\raisebox{-\totalheight}{\includegraphics[width=\linewidth]{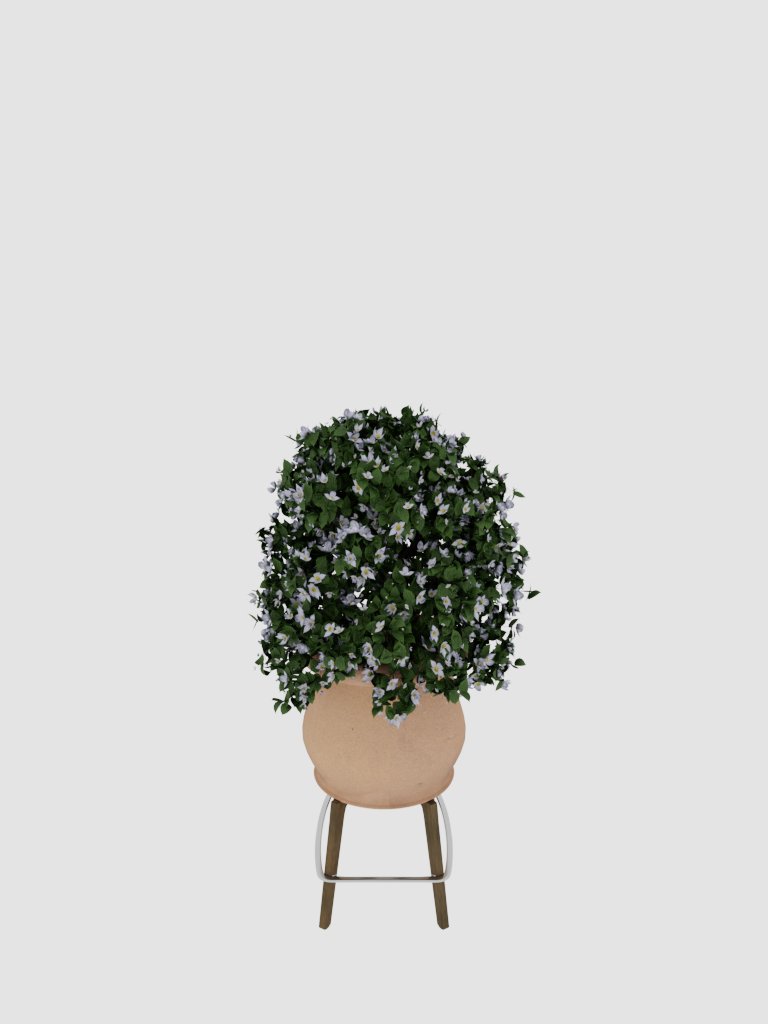}} \\
    Floor lamp 1 & Floor lamp 2 & Magnolia sieboldii & Philadelphus shrub \\
	\raisebox{-\totalheight}{\includegraphics[width=\linewidth]{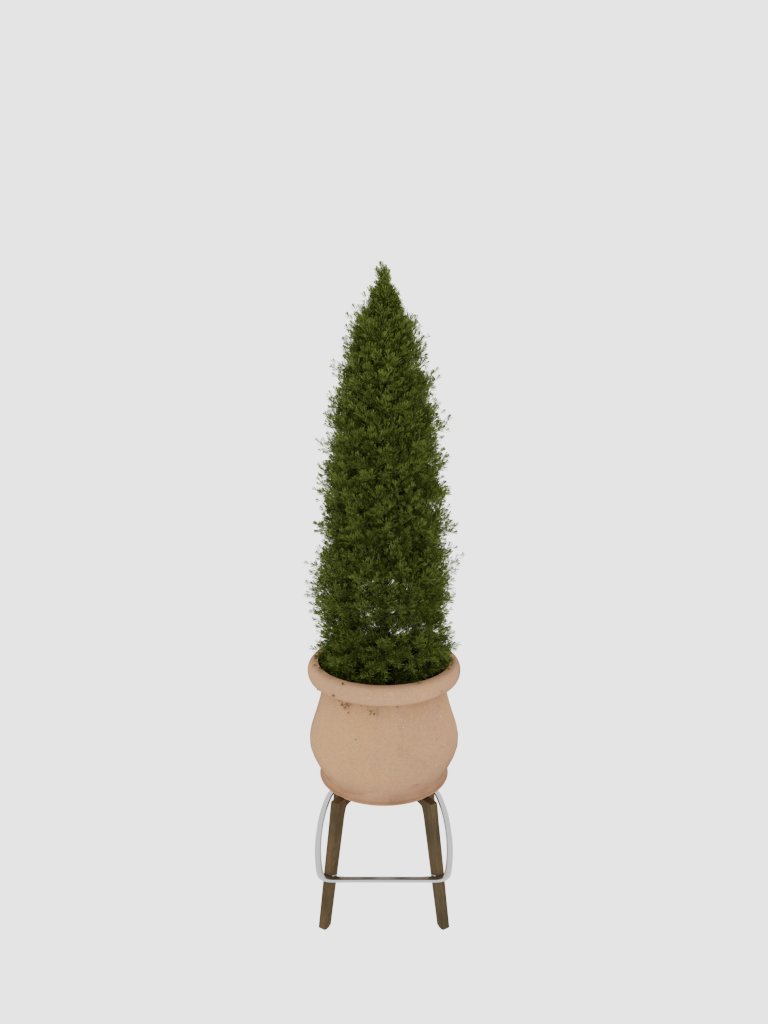}} &
	\raisebox{-\totalheight}{\includegraphics[width=\linewidth]{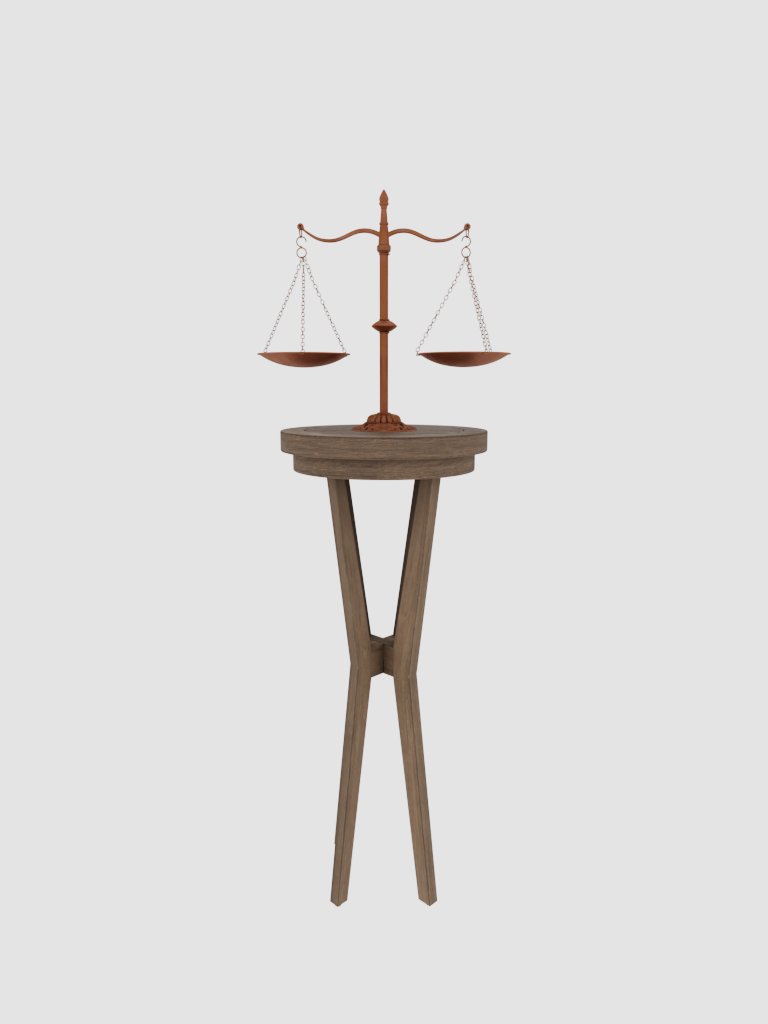}} &
	\raisebox{-\totalheight}{\includegraphics[width=\linewidth]{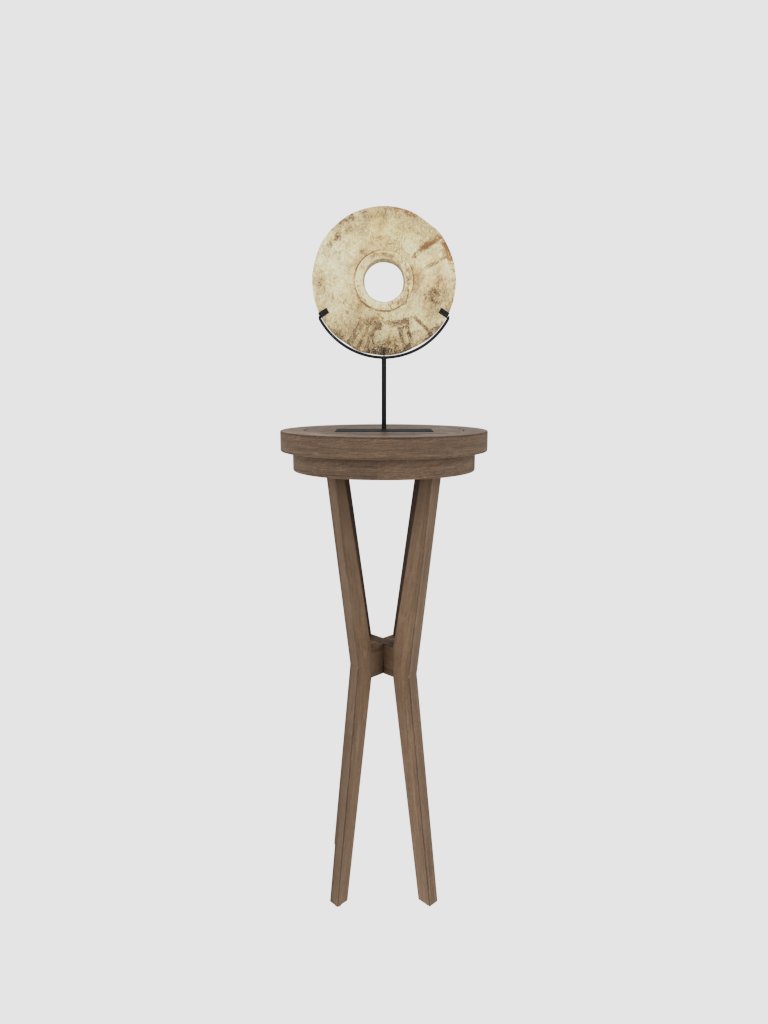}} &
	\raisebox{-\totalheight}{\includegraphics[width=\linewidth]{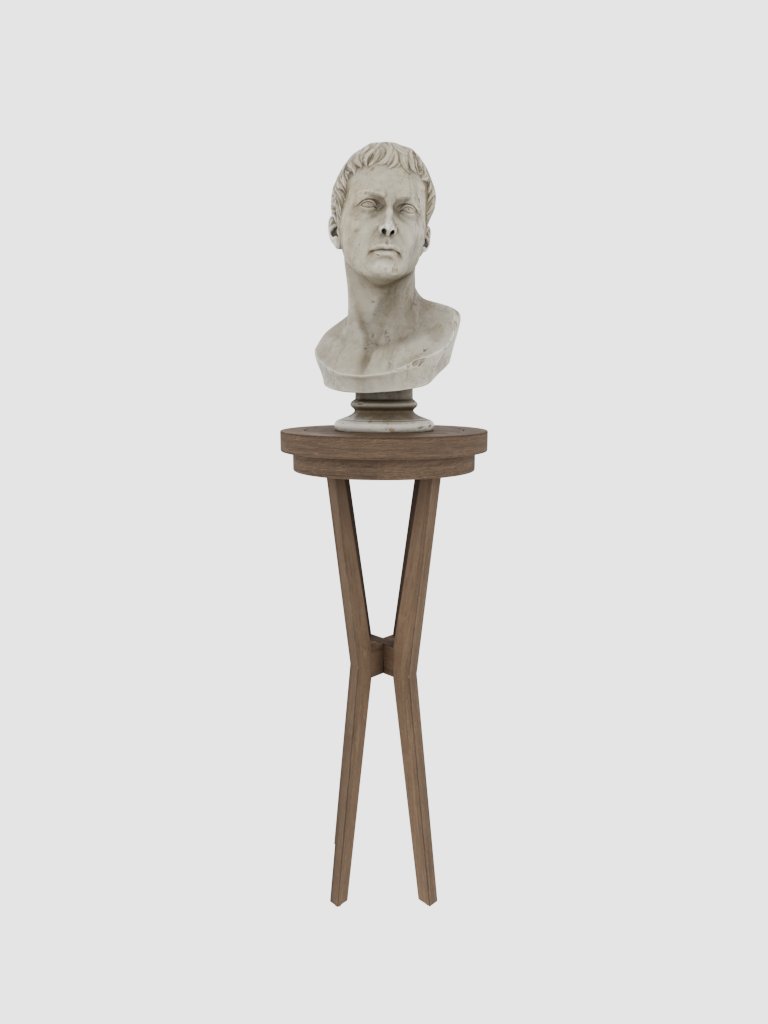}} \\
    Juniperus communis & Copper scale & Decorative disk & Marble bust
	\end{tabular}}}}	
	\caption{
	Distant reference objects (w/ and w/o an intrinsic frame of reference) appear behind the table or besides the shelf.
	}
    \label{tab:reference-distant}
	\vspace{-.0em}
\end{table*}

\begin{table*}[t!]\small
	\centering
	{\setlength{\tabcolsep}{0pt} 
	\makebox[\linewidth]{\resizebox{1.\linewidth}{!}{%
    \begin{tabular}{@{} 
            m{0.24\linewidth} @{\hspace{0.01\linewidth}}  
            m{0.24\linewidth} @{\hspace{0.01\linewidth}}  
            m{0.24\linewidth} @{\hspace{0.01\linewidth}}  
            m{0.24\linewidth} 
            @{}}
	\raisebox{-\totalheight}{\includegraphics[width=\linewidth]{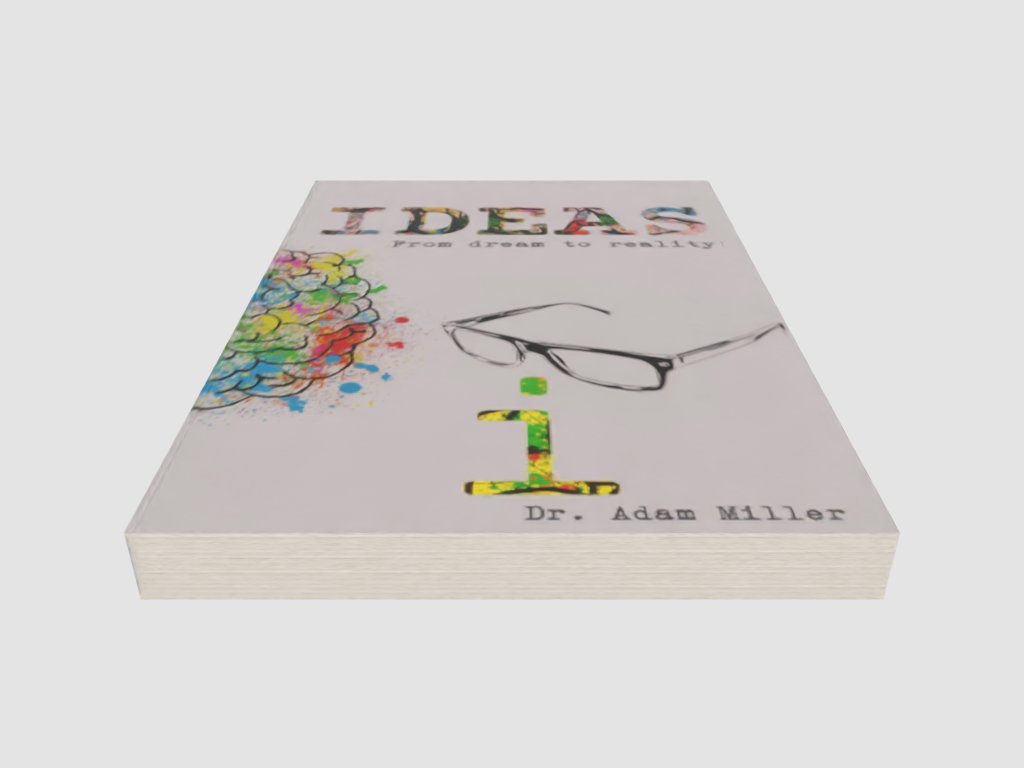}} &
	\raisebox{-\totalheight}{\includegraphics[width=\linewidth]{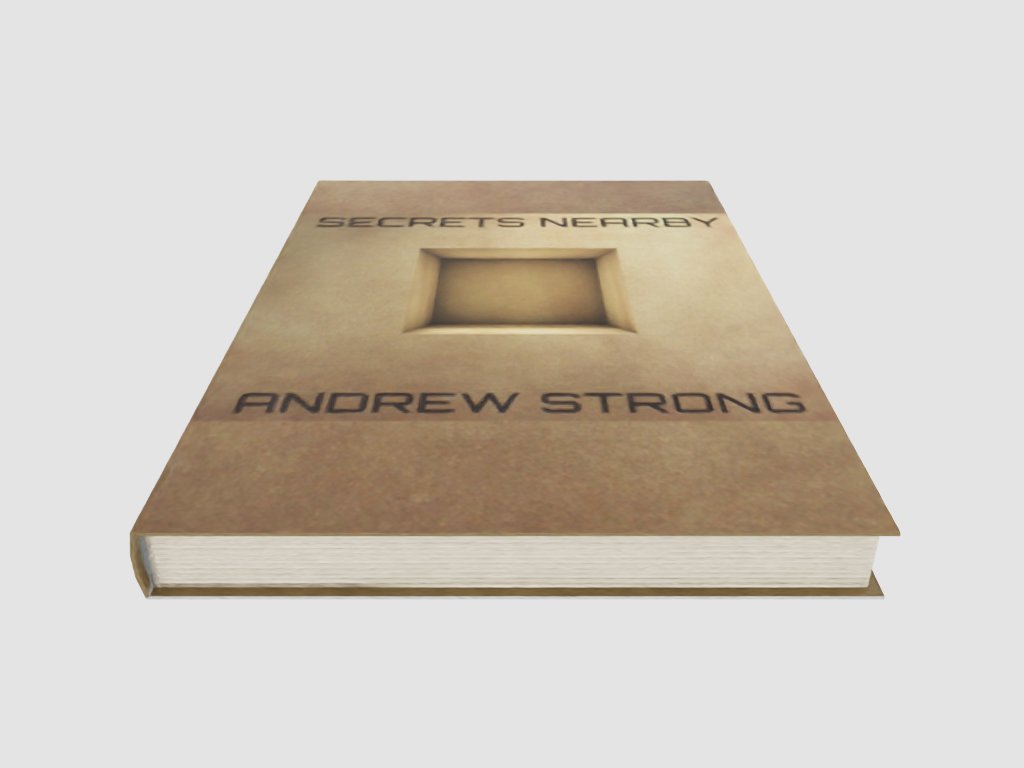}} &
	\raisebox{-\totalheight}{\includegraphics[width=\linewidth]{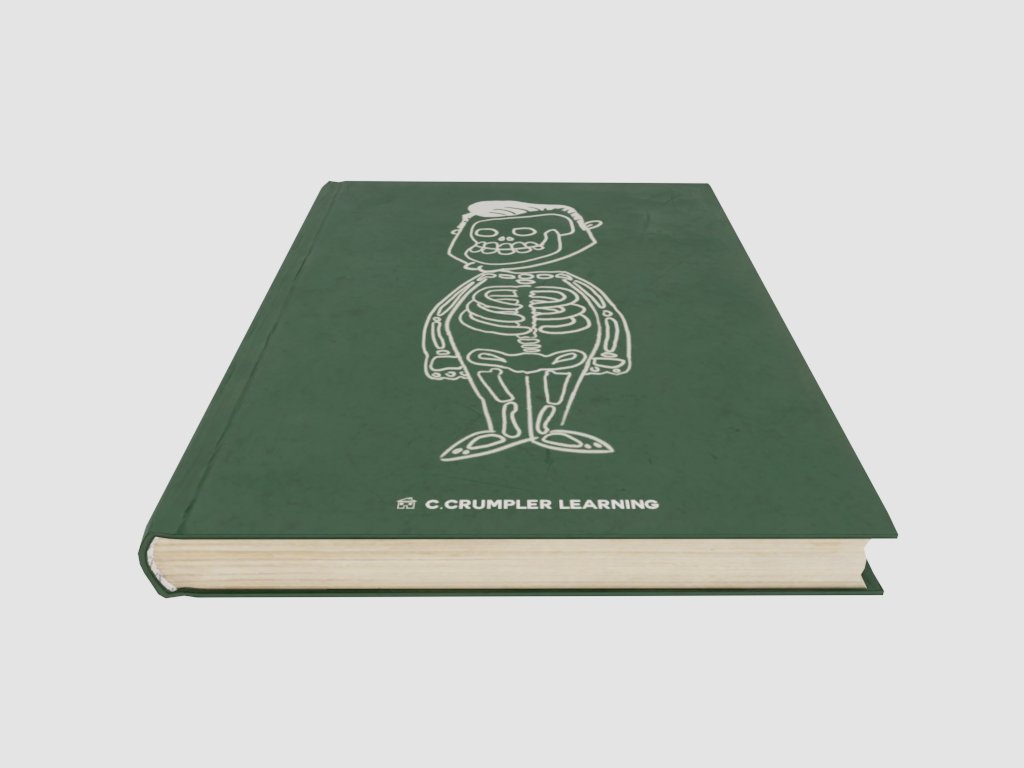}} &
	\raisebox{-\totalheight}{\includegraphics[width=\linewidth]{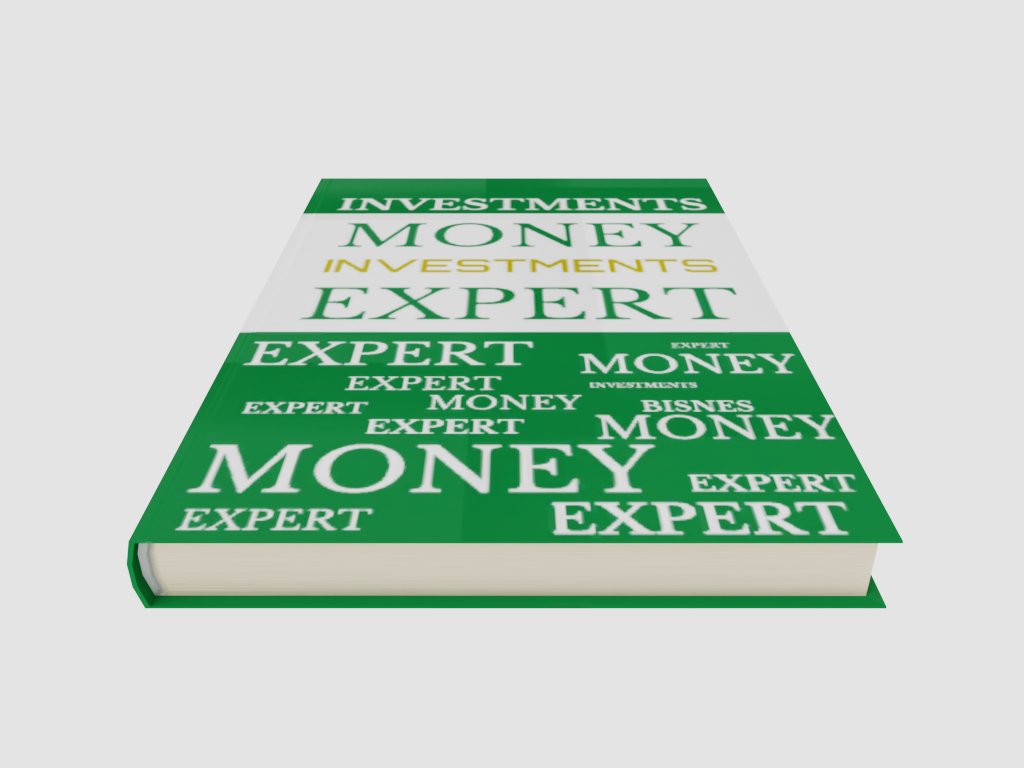}} \\
	\raisebox{-\totalheight}{\includegraphics[width=\linewidth]{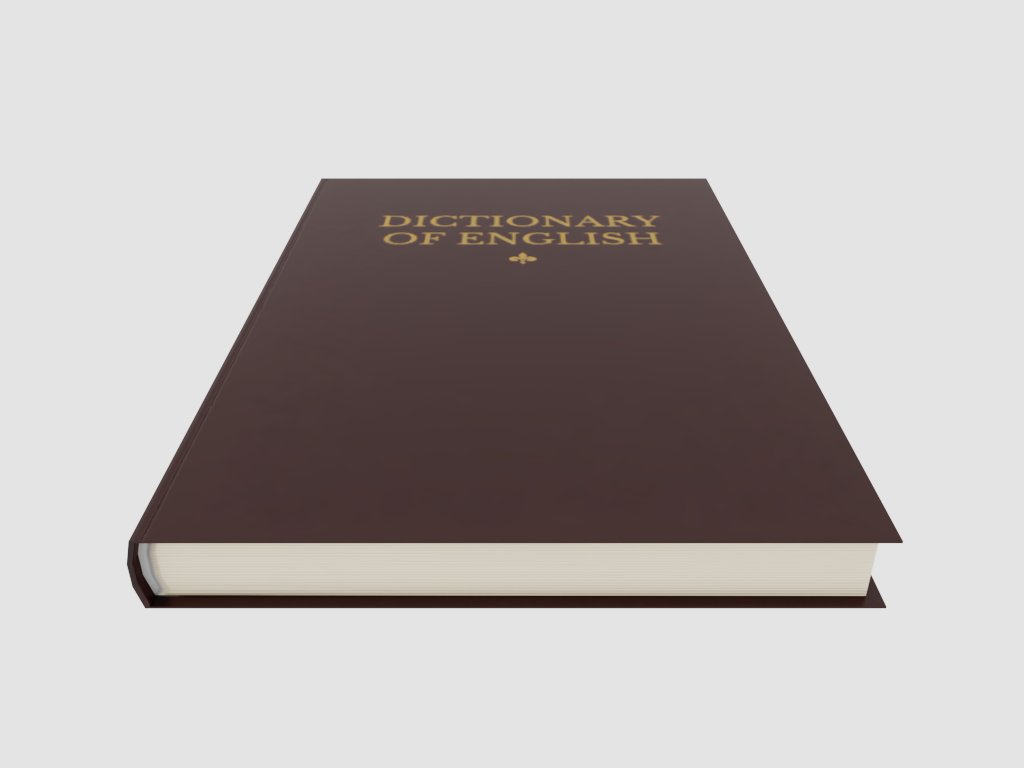}} &
	\raisebox{-\totalheight}{\includegraphics[width=\linewidth]{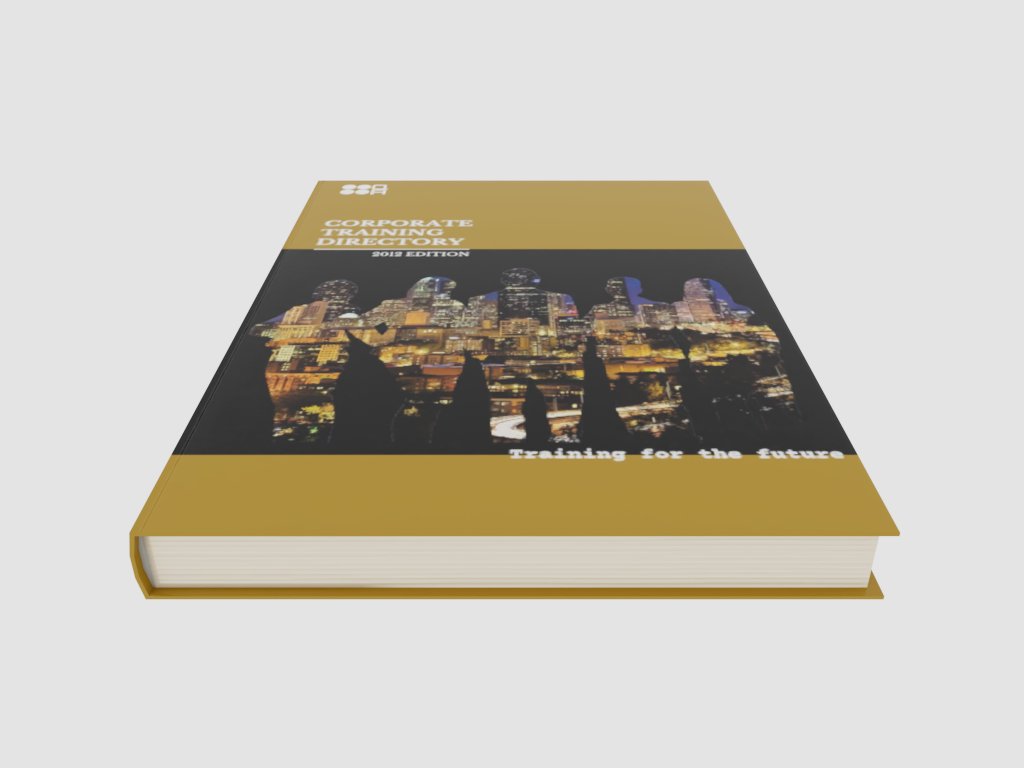}} &
	\raisebox{-\totalheight}{\includegraphics[width=\linewidth]{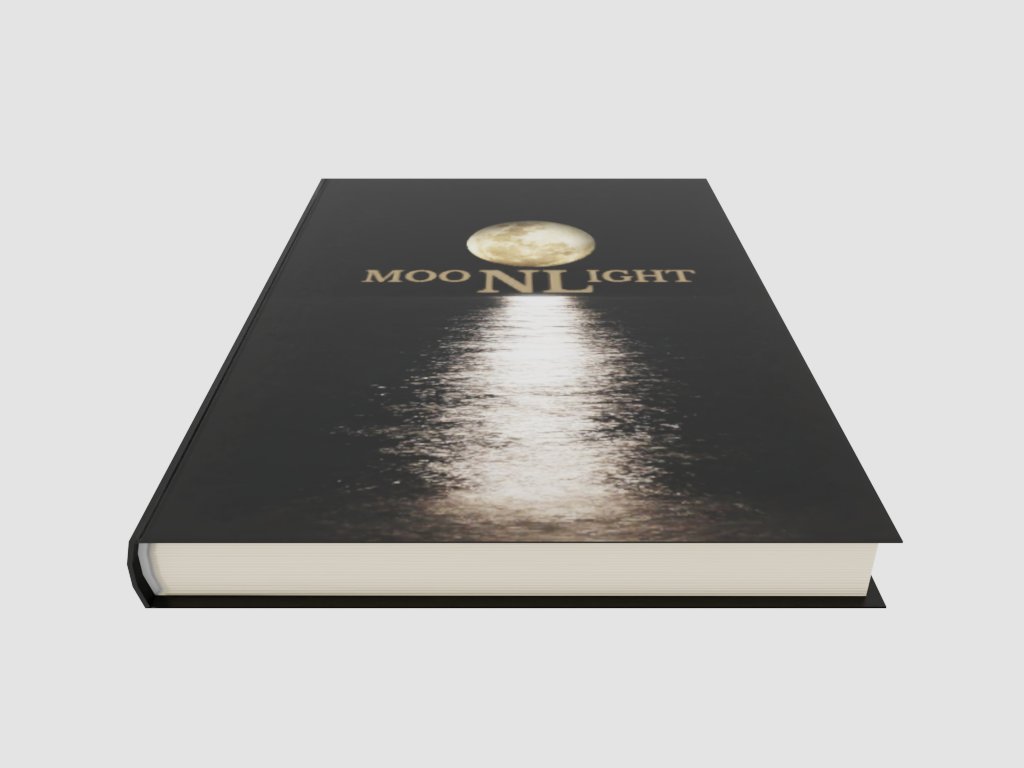}} &
	\raisebox{-\totalheight}{\includegraphics[width=\linewidth]{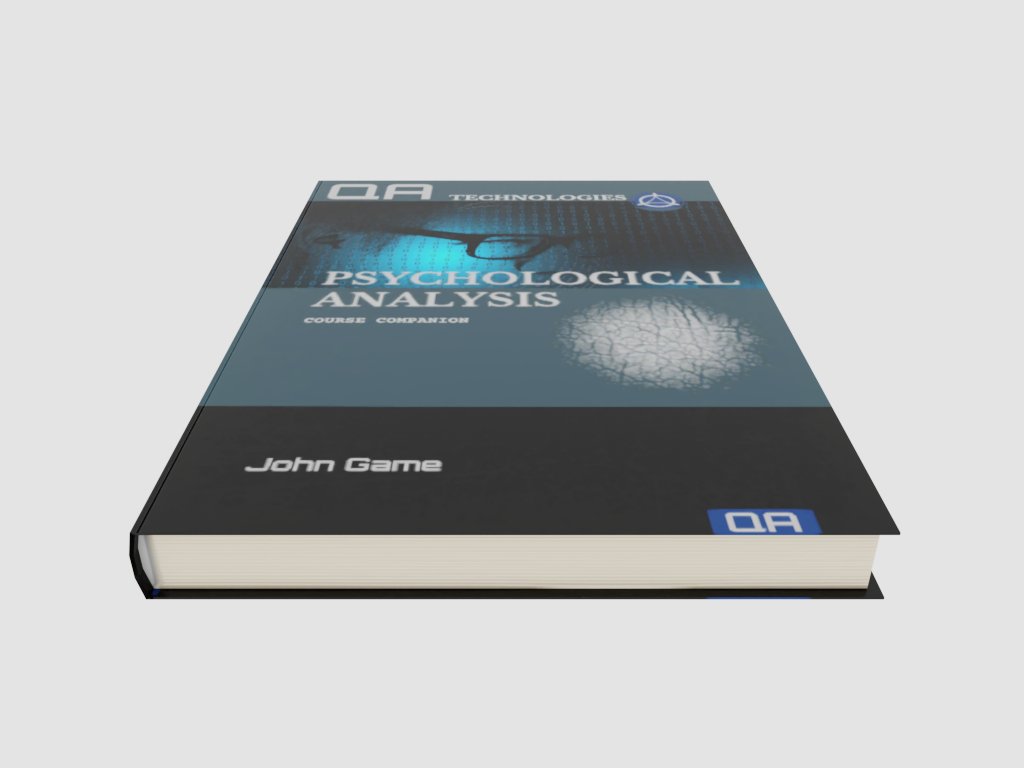}} \\
	\raisebox{-\totalheight}{\includegraphics[width=\linewidth]{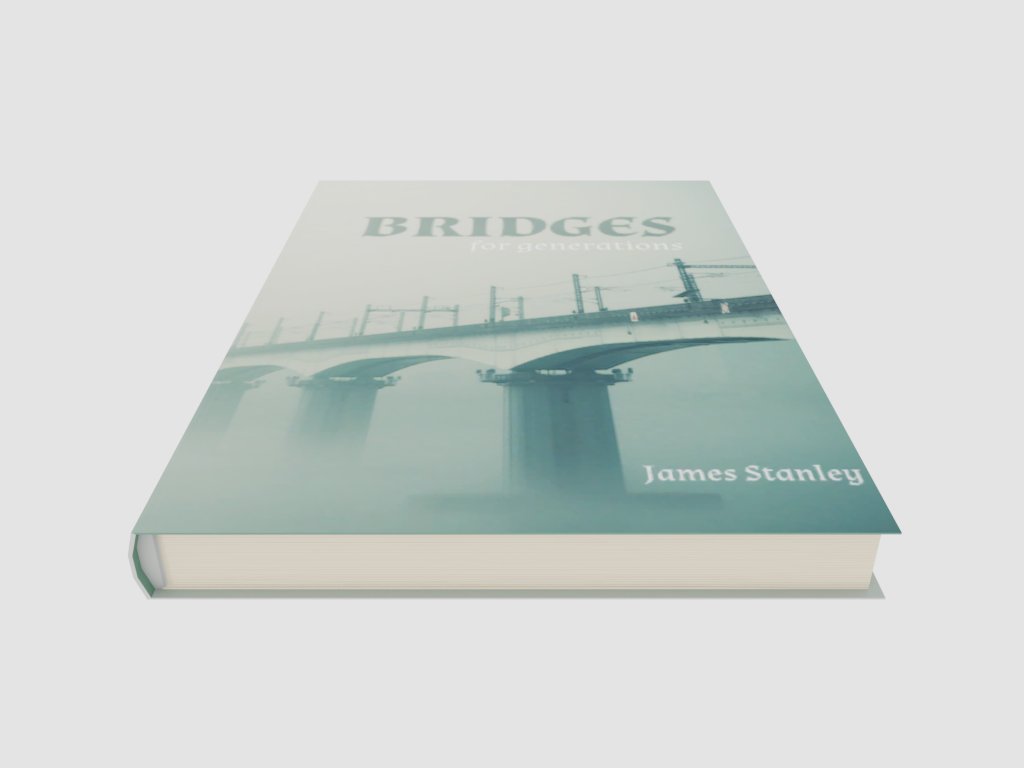}} &
	\raisebox{-\totalheight}{\includegraphics[width=\linewidth]{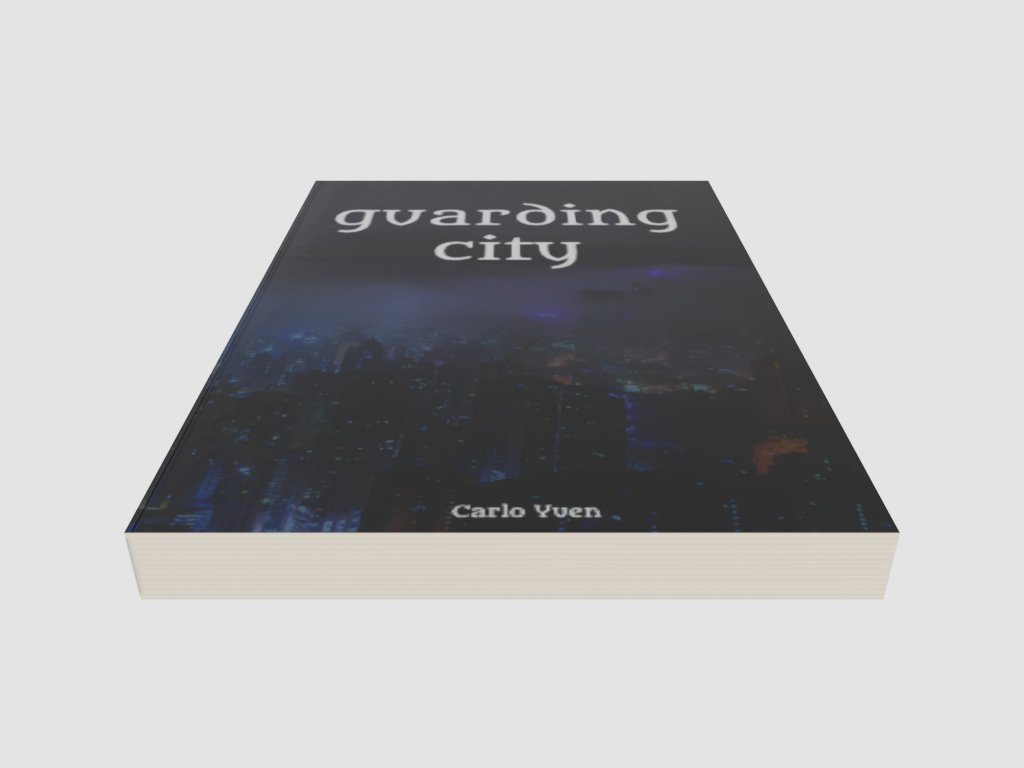}} &
	\raisebox{-\totalheight}{\includegraphics[width=\linewidth]{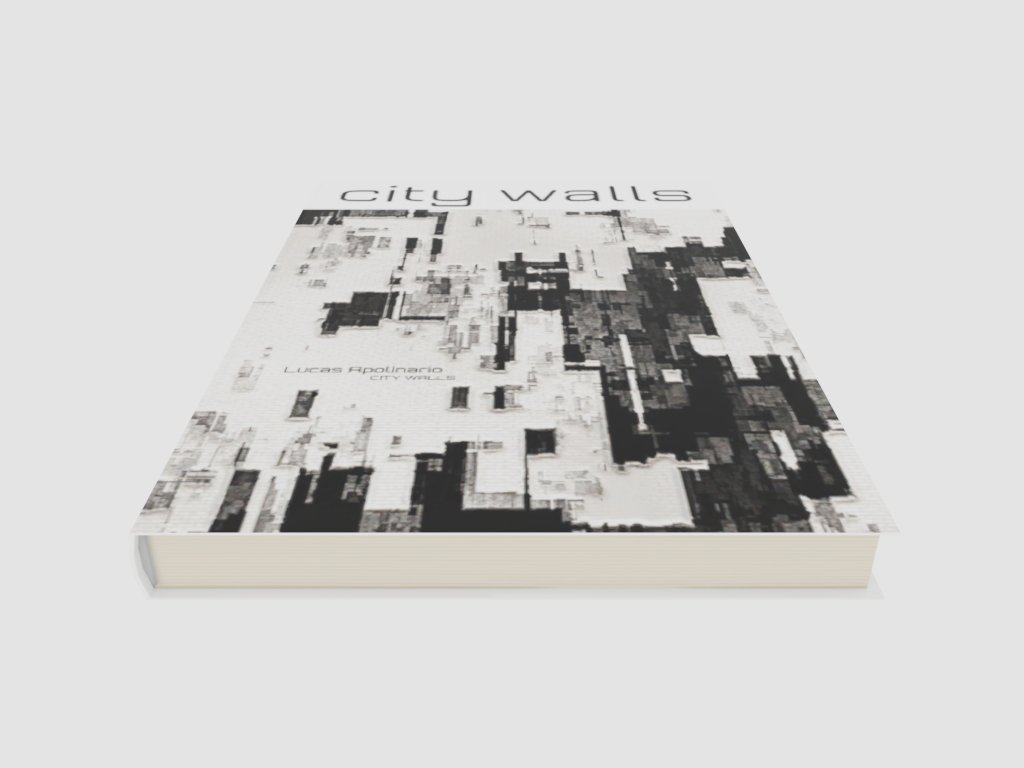}} &
	\raisebox{-\totalheight}{\includegraphics[width=\linewidth]{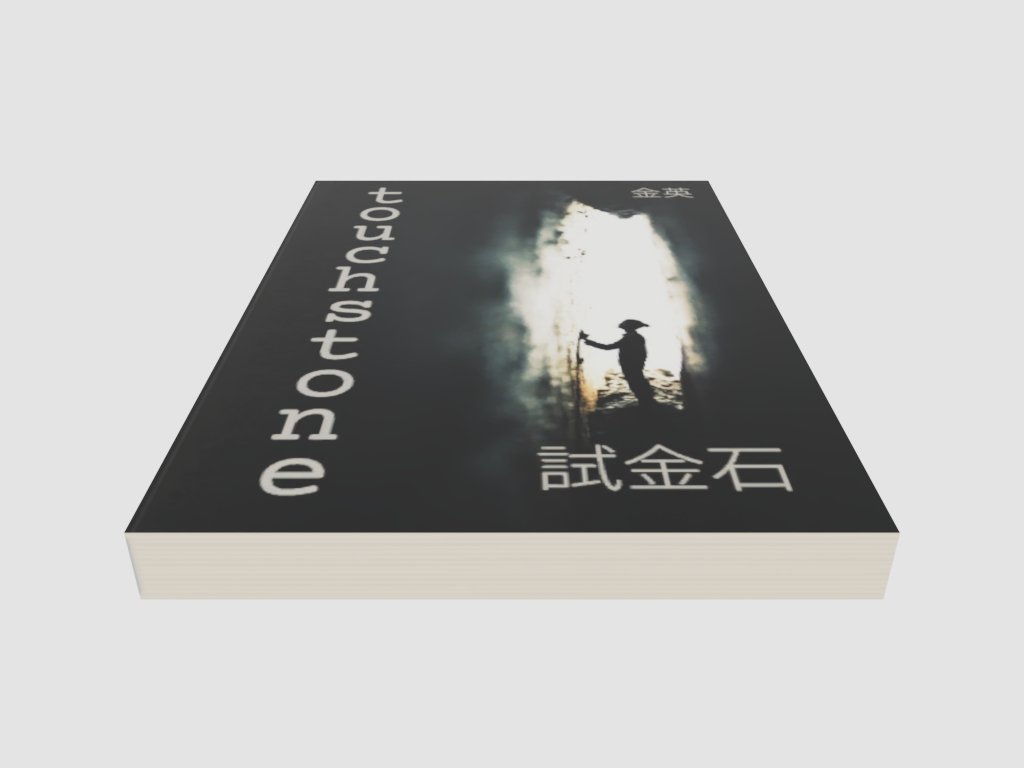}} \\
	\raisebox{-\totalheight}{\includegraphics[width=\linewidth]{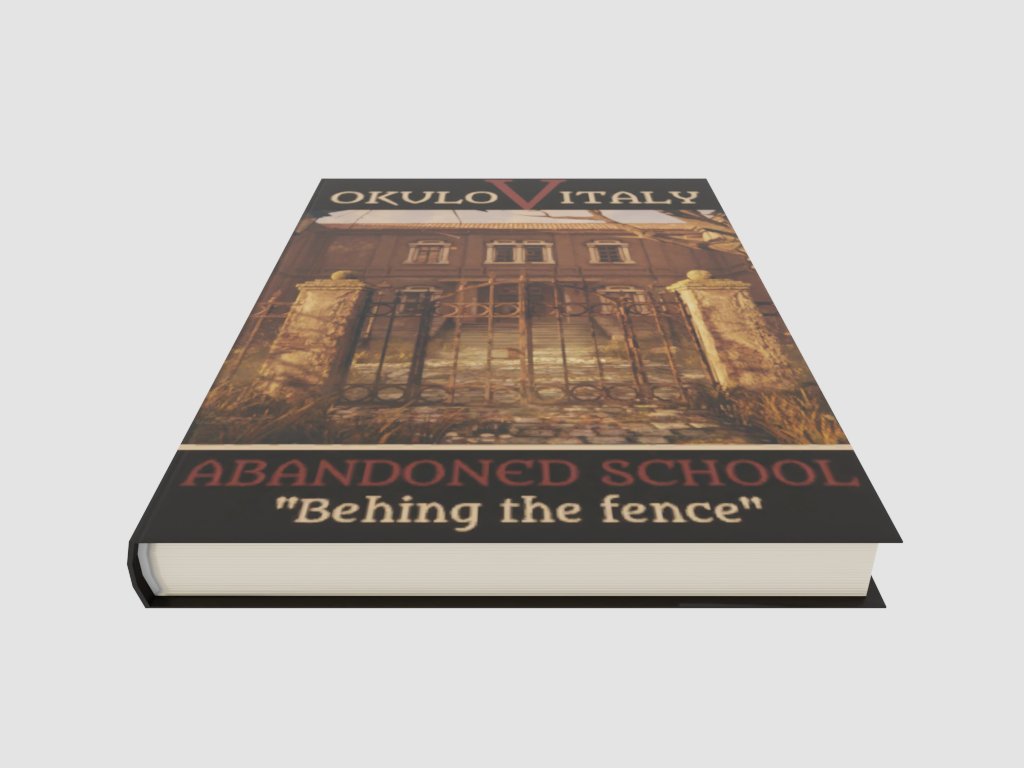}} &
	\raisebox{-\totalheight}{\includegraphics[width=\linewidth]{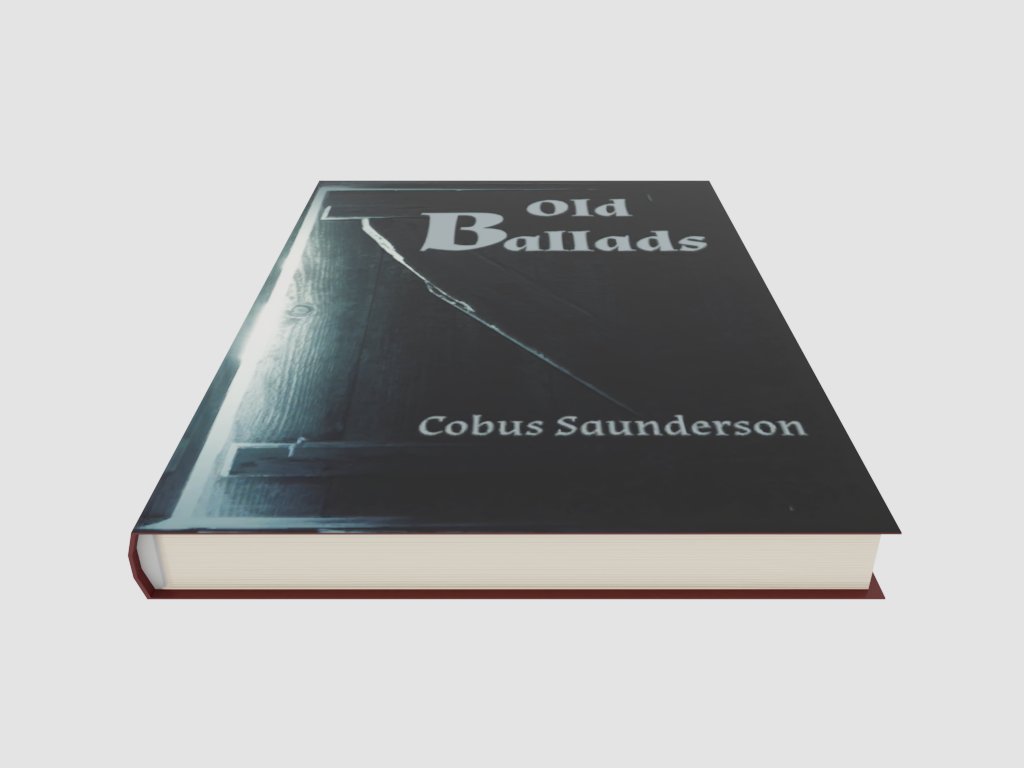}} &
	\raisebox{-\totalheight}{\includegraphics[width=\linewidth]{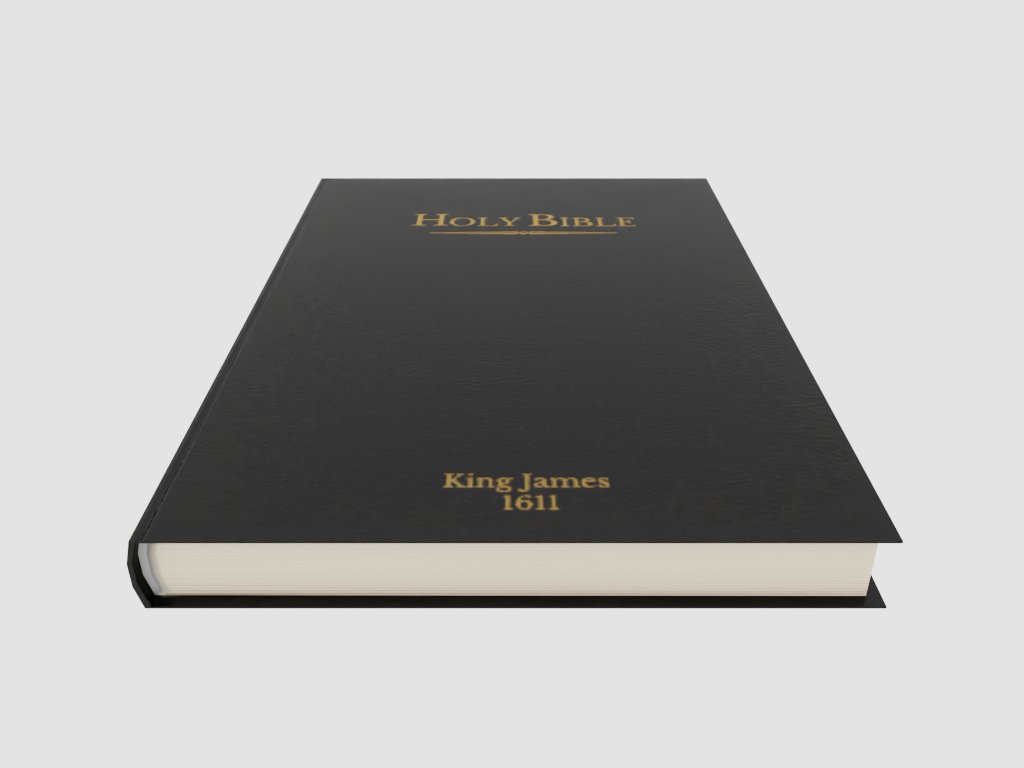}} &
	\raisebox{-\totalheight}{\includegraphics[width=\linewidth]{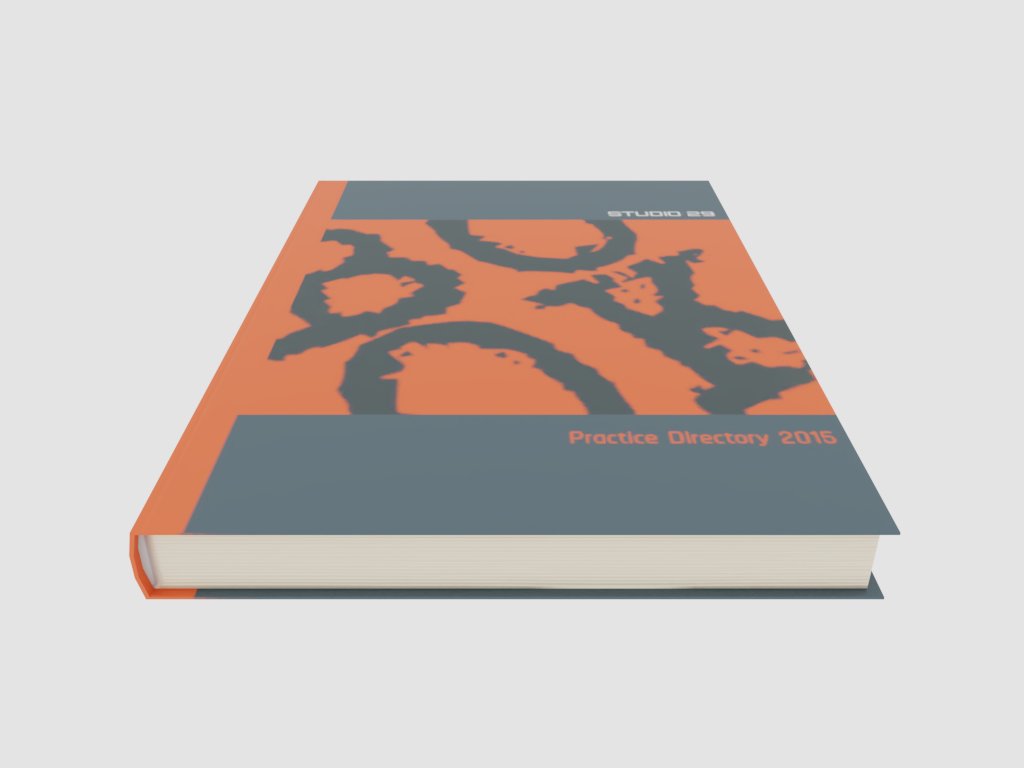}} \\
    \raisebox{-\totalheight}{\includegraphics[width=\linewidth]{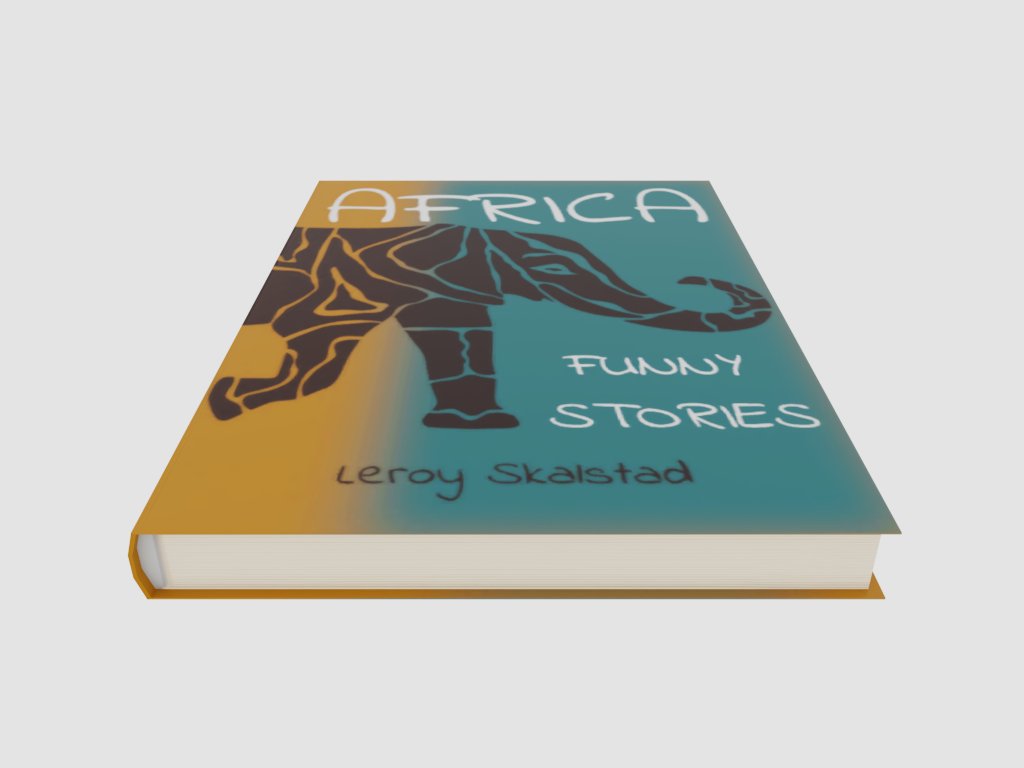}} &
	\raisebox{-\totalheight}{\includegraphics[width=\linewidth]{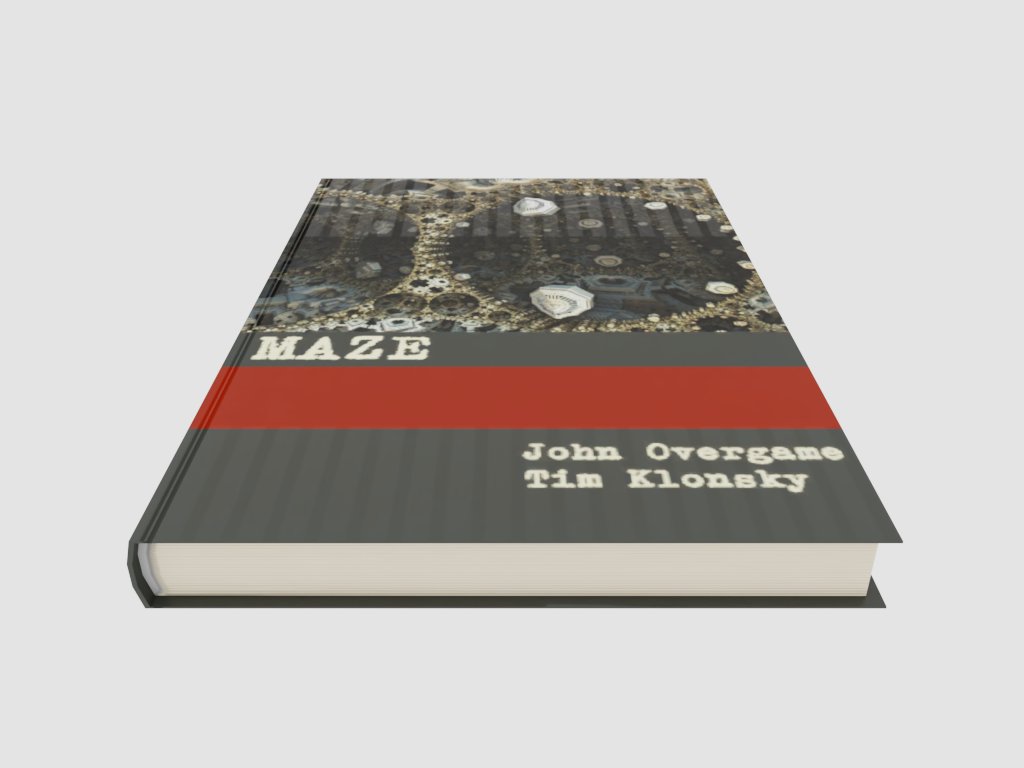}} &
	\raisebox{-\totalheight}{\includegraphics[width=\linewidth]{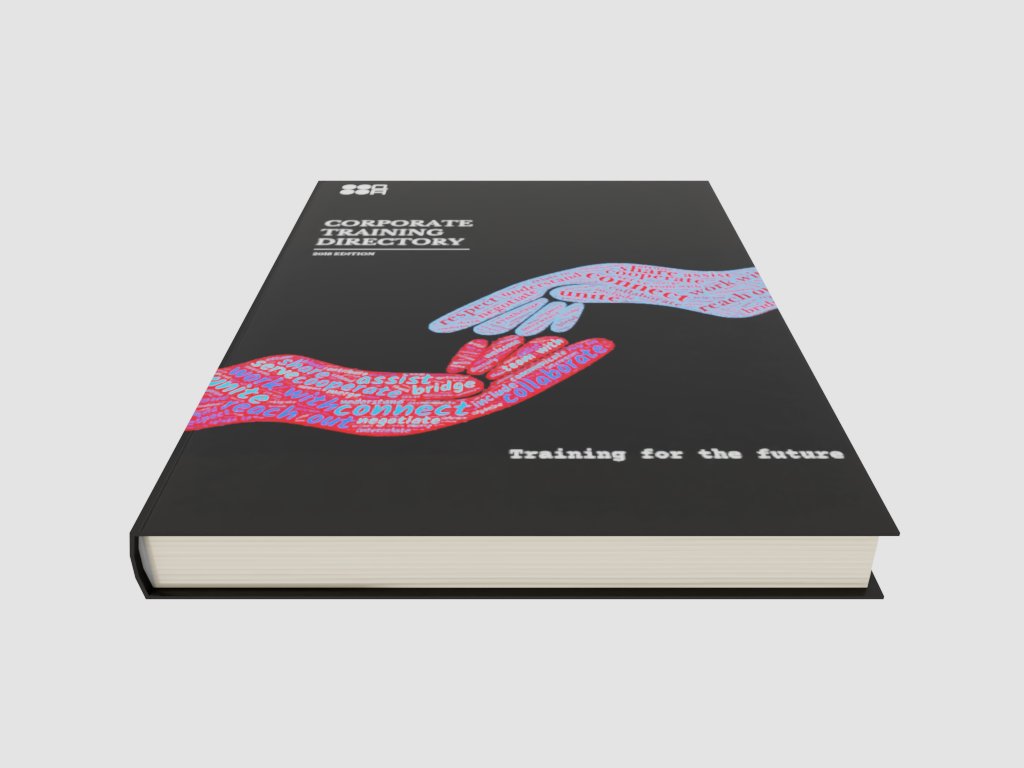}} &
	\raisebox{-\totalheight}{\includegraphics[width=\linewidth]{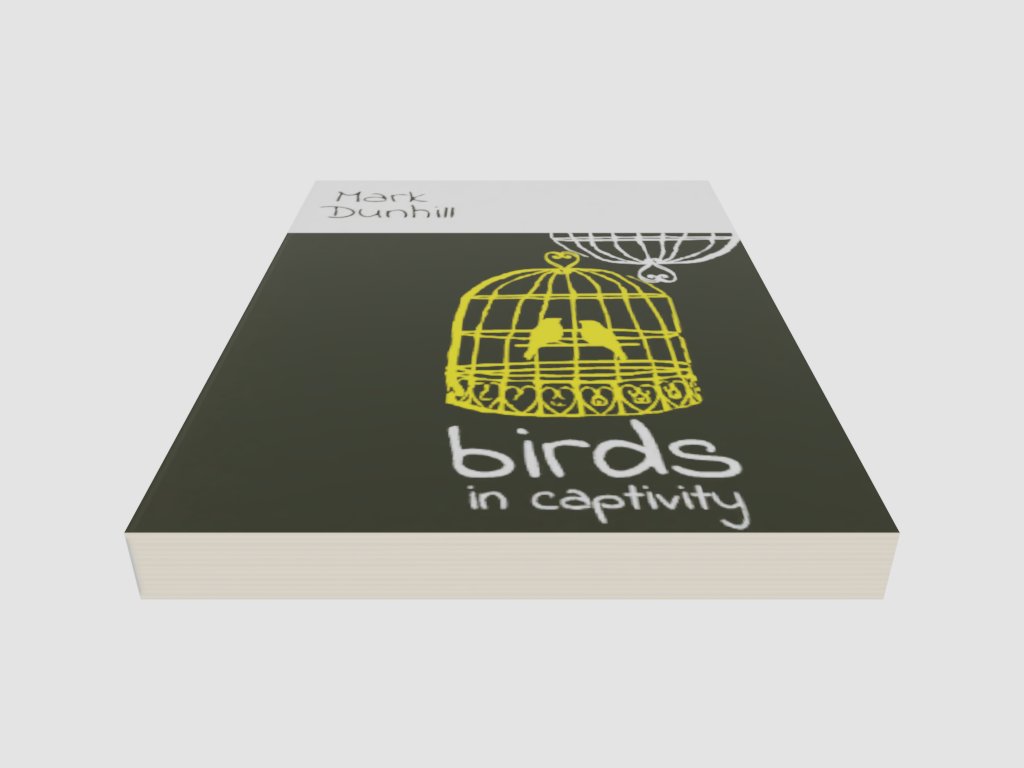}}
	\end{tabular}}}}	
	\caption{
	Manipulable books. They will be auto-scaled to match three pre-defined sizes: small, medium, and large. 
	}
    \label{tab:book}
	\vspace{-.0em}
\end{table*}
\begin{table*}[t!]\small
	\centering
	{\setlength{\tabcolsep}{0pt} 
	\makebox[\linewidth]{\resizebox{1.\linewidth}{!}{%
    \begin{tabular}{@{} 
            m{0.19\linewidth} @{\hspace{0.01\linewidth}}  
            m{0.19\linewidth} @{\hspace{0.01\linewidth}}  
            m{0.19\linewidth} @{\hspace{0.01\linewidth}}  
            m{0.19\linewidth} @{\hspace{0.01\linewidth}} 
            m{0.19\linewidth} 
            @{}}
	\raisebox{-\totalheight}{\includegraphics[width=\linewidth]{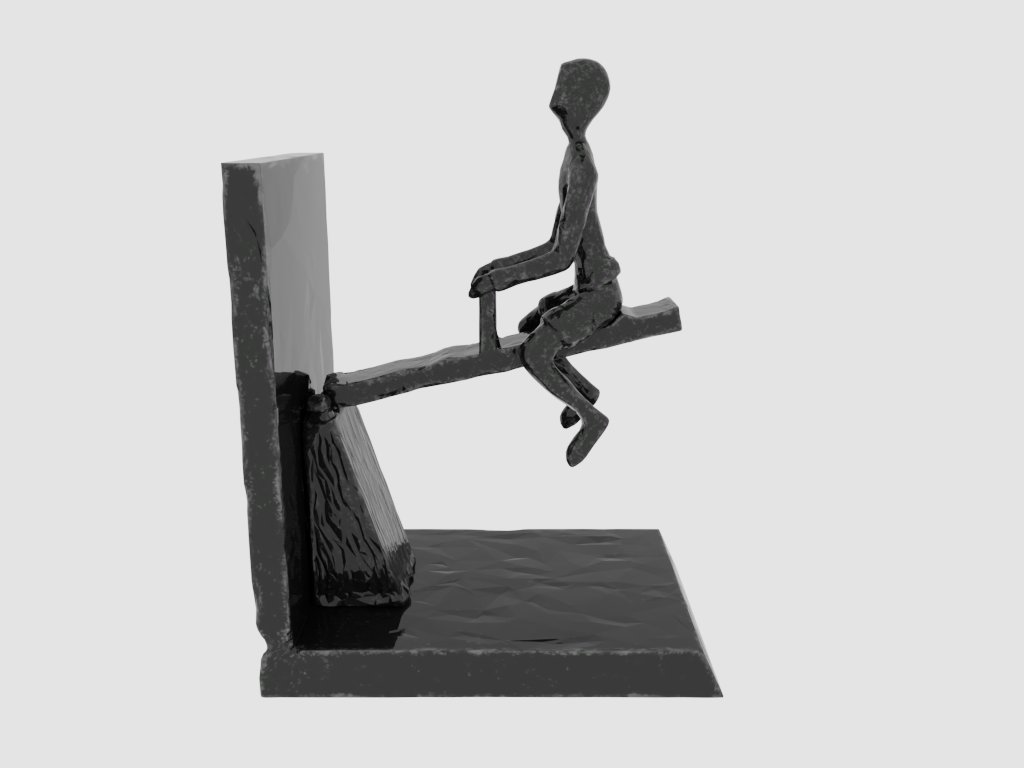}} &
	\raisebox{-\totalheight}{\includegraphics[width=\linewidth]{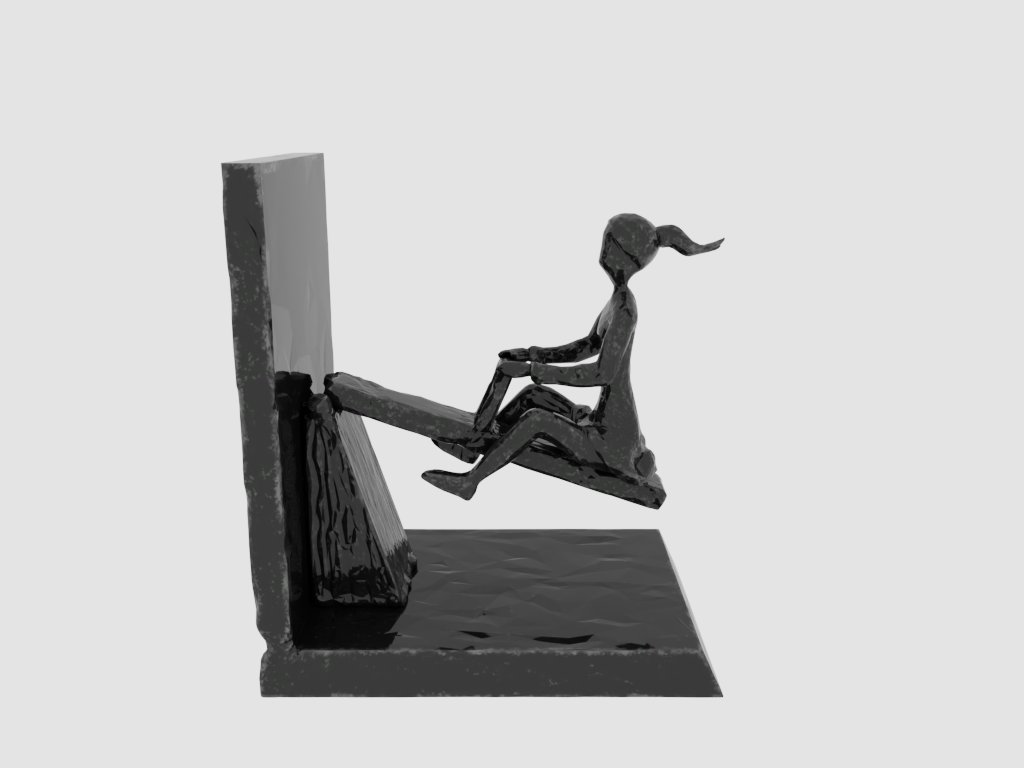}} &
	\raisebox{-\totalheight}{\includegraphics[width=\linewidth]{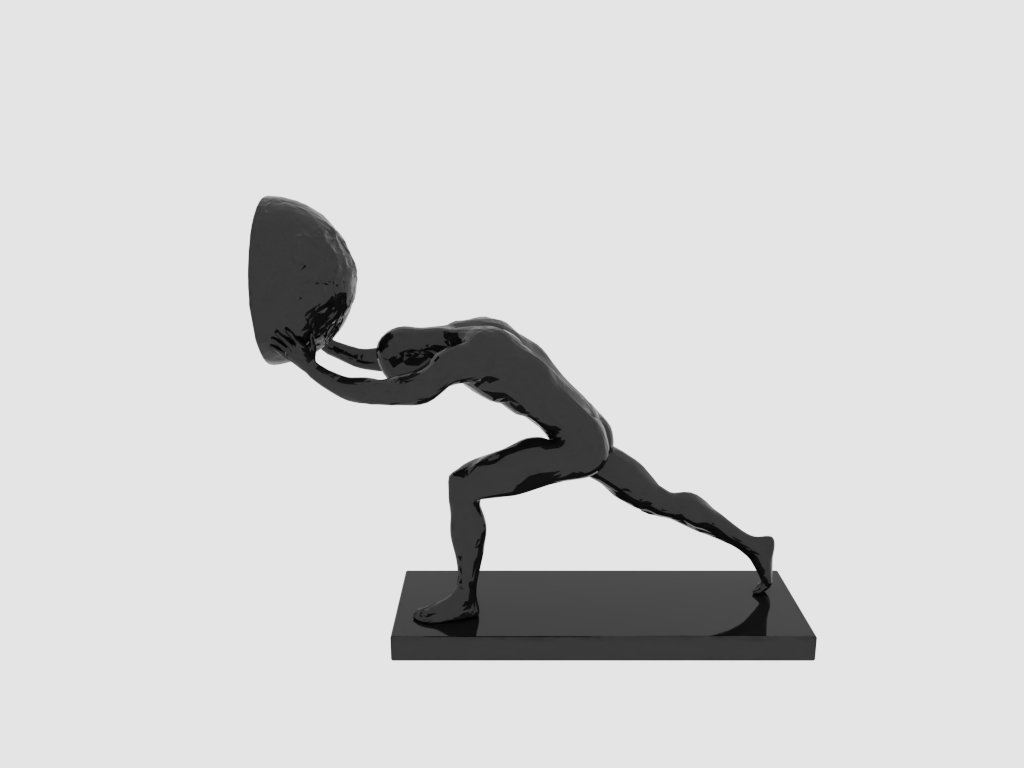}} &
    \raisebox{-\totalheight}{\includegraphics[width=\linewidth]{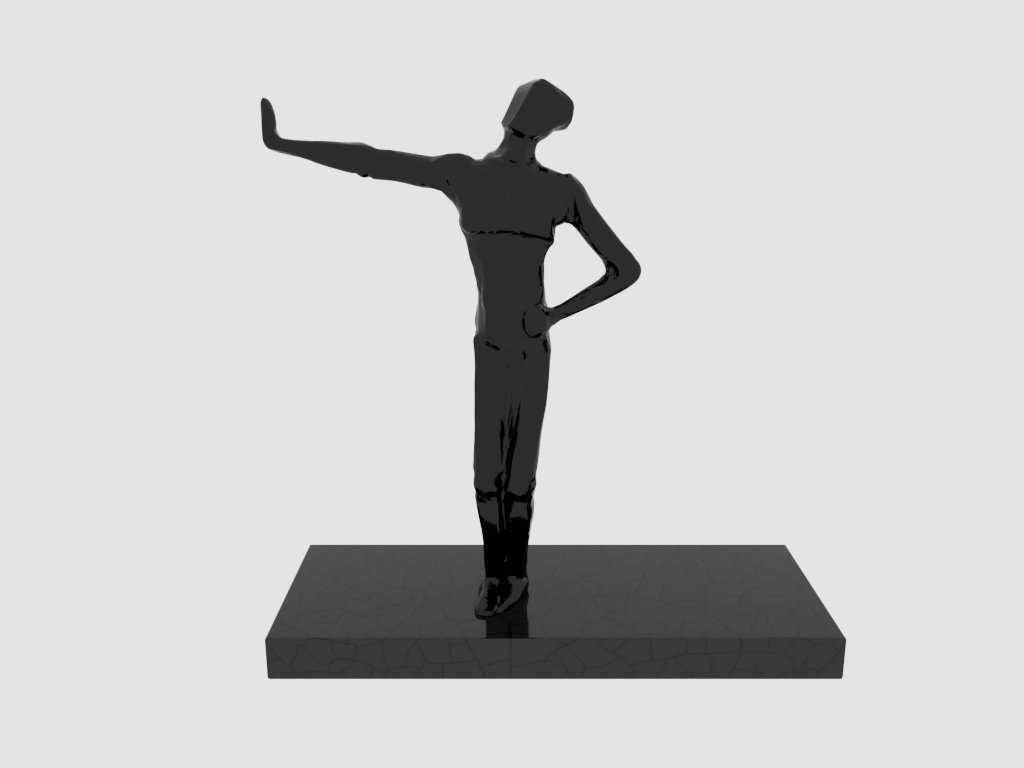}} &
	\raisebox{-\totalheight}{\includegraphics[width=\linewidth]{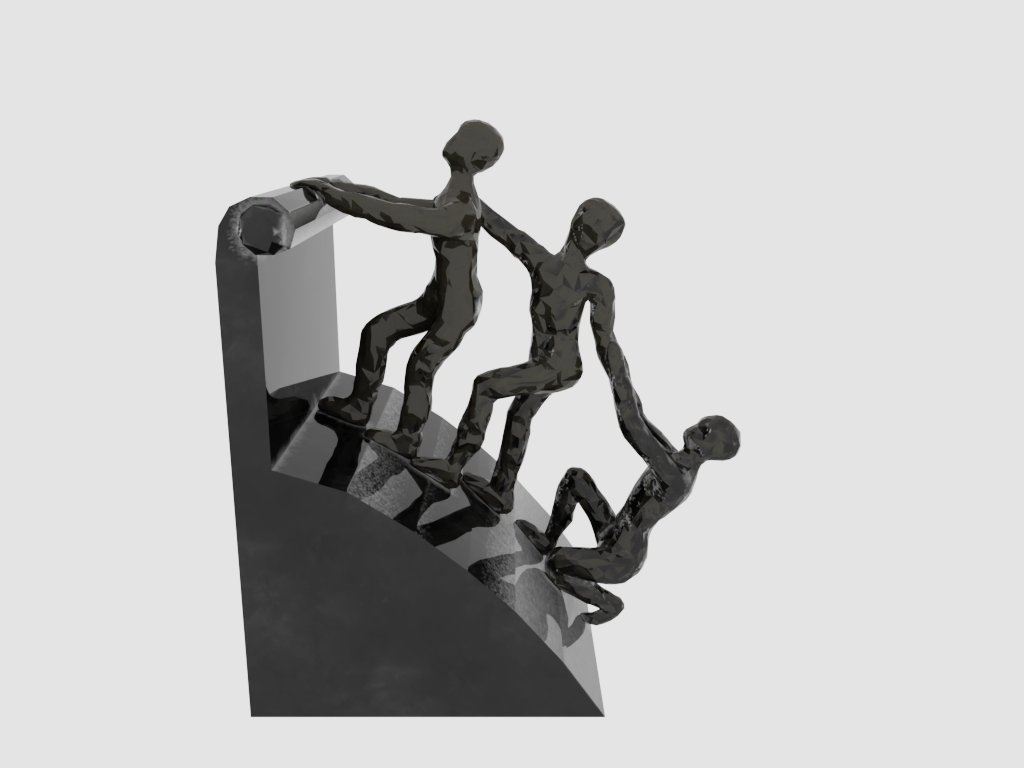}}
	\end{tabular}}}}
	\caption{
	Different bookends are used to help create tilting poses of books.
	}
    \label{tab:bookend}
	\vspace{-.0em}
\end{table*}

\begin{table*}[ht]
    \centering
    \caption{Attributes of \model assets. Oriented assets have an intrinsic frame while non-oriented assets do not.}
    \label{tab:obj_dims}
    \begin{tabular}{l @{\hspace{0.3cm}} ccr @{\hspace{0.3cm}} r @{\hspace{0.3cm}} r}
    \toprule
        Name & Type & Oriented & L (cm) & W (cm) & H (cm) \\
    \midrule
        Alarm clock & near & \checkmark & 7 & 13 & 17 \\
        Armento Rider & near & \checkmark & 24 & 7 & 24 \\
        Bicycle sculpture & near & \checkmark & 21 & 8 & 18 \\
        Picture frame & near & \checkmark & 13 & 22 & 18 \\
        Teddy bear & near & \checkmark  & 20 & 23 & 25 \\
        Newton's cradle & near & \xmark & 10 & 15 & 14 \\
        Geosphere sculpture & near & \xmark & 15 & 15 & 15 \\
        Pillar bookend & near & \xmark & 7 & 16 & 13 \\
        Rubik's cube & near & \xmark & 6 & 6 & 6 \\
        Succulents & near & \xmark & 17 & 15 & 29 \\
        Ceramic jar & near & \xmark & 6 & 6 & 8 \\
        Pagoda statue & near & \xmark & 13 & 14 & 21 \\
        Cheval mirror & distant & \checkmark & 40 & 42 & 43 \\
        Lady with an Ermine & distant & \checkmark & 62 & 59 & 52 \\
        Mona Lisa & distant & \checkmark & 62 & 54 & 52 \\
        Adoration of the Magi & distant & \checkmark & 62 & 79 & 52 \\
        Floor lamp 1 & distant & \xmark & 48 & 48 & 59 \\
        Floor lamp 2 & distant & \xmark & 51 & 51 & 60 \\
        Magnolia sieboldii & distant & \xmark & 61 & 65 & 51 \\
        Philadelphus shrub & distant & \xmark & 63 & 61 & 06 \\
        Juniperus communis & distant & \xmark & 46 & 45 & 34 \\
        Copper scale & distant & \xmark & 52 & 61 & 46 \\
        Decorative disk & distant & \checkmark & 52 & 52 & 43 \\
        Marble bust & distant & \checkmark & 52 & 52 & 53 \\
        Table $1-8$ & table & \checkmark & 60 & 140 & 70 \\
        Shelf 1 & shelf & \checkmark & 45 & 149 & 215 \\
        Shelf 2 & shelf & \checkmark  & 45 & 176 & 190 \\
        Shelf 3 & shelf & \checkmark & 45 & 176 & 190 \\
        Shelf 4 & shelf & \checkmark & 45 & 171 & 190 \\
        Shelf 5 & shelf & \checkmark & 45 & 149 & 190 \\
        Shelf 6 & shelf & \checkmark & 45 & 164 & 187 \\
        Book-small  & book & \xmark & 17.5 -- 18.8 & 10.8 -- 13.0 & 1.5 -- 1.8 \\
        Book-medium & book & \xmark & 21.6 -- 25.0 & 14.0 -- 17.6 & 2.0 -- 2.5 \\
        Book-large  & book & \xmark & 25.4 -- 30.5 & 20.3 -- 24.1 & 3.7 -- 4.0 \\
        \bottomrule
    \end{tabular}
\end{table*}


\end{document}